\documentclass{article}
\usepackage{lmodern}

    \usepackage[preprint]{neurips2025}

\usepackage[utf8]{inputenc} 
\usepackage[T1]{fontenc}    
\usepackage{hyperref}       
\usepackage{url}            
\usepackage{booktabs}       
\usepackage{amsfonts}       
\usepackage{nicefrac}       
\usepackage{microtype}      
\usepackage{xcolor}         

\usepackage[pdftex]{graphicx}
\usepackage{amsmath}
\usepackage{amsthm}
\usepackage{amssymb}
\usepackage{wrapfig}
\usepackage{graphicx}
\usepackage{lipsum} 
\usepackage[normalem]{ulem}
\useunder{\uline}{\ul}{}
\usepackage[noend]{algpseudocode}
\usepackage{algorithm}
\usepackage{multirow}
\usepackage{rotating}
\usepackage[normalem]{ulem}
\useunder{\uline}{\ul}{}
\usepackage{makecell}
\usepackage{pdfpages}

\newtheorem{theorem}{Theorem}

\usepackage{pifont}

\title{OLinear: A Linear Model for Time Series Forecasting in Orthogonally Transformed Domain}

\author{%
  Wenzhen Yue\\
  State Key Laboratory of General AI \\
  Peking University, Beijing 100871 \\
  \texttt{yuewenzhen@stu.pku.edu.cn} \\
  \And
  Yong Liu \\
  School of Software, BNRist\\
  Tsinghua University, Beijing 100084  \\
  \texttt{liuyong21@mails.tsinghua.edu.cn} \\
  \And
  Haoxuan Li \\
  Center for Data Science \\
  Peking University, Beijing 100871 \\
  \texttt{hxli@stu.pku.edu.cn} \\
  \And
  Hao Wang \\
  Department of Control Science and Engineering \\
  Zhejiang University, Hangzhou 310058 \\
  \texttt{haohaow@zju.edu.cn} \\
  \And
  Xianghua Ying\\
  State Key Laboratory of General AI \\
  Peking University, Beijing 100871 \\
  \texttt{xhying@pku.edu.cn} \\
  \And
  Ruohao Guo\\
  State Key Laboratory of General AI \\
  Peking University, Beijing 100871 \\
  \texttt{ruohguo@foxmail.com} \\
  \And
  Bowei Xing\\
  State Key Laboratory of General AI \\
  Peking University, Beijing 100871 \\
  \texttt{xingbowei@pku.edu.cn} \\
  \And
  Ji Shi\\
  State Key Laboratory of General AI \\
  Peking University, Beijing 100871 \\
  \texttt{sjj118@pku.edu.cn} \\
}

\begin{document}

\maketitle

\begin{abstract}

This paper presents \textbf{OLinear}, a \textbf{linear}-based multivariate time series forecasting model that operates in an \textbf{o}rthogonally transformed domain. Recent forecasting models typically adopt the temporal forecast (TF) paradigm, which directly encode and decode time series in the time domain. However, the entangled step-wise dependencies in series data can hinder the performance of TF. To address this, some forecasters conduct encoding and decoding in the transformed domain using fixed, dataset-independent bases (e.g., sine and cosine signals in the Fourier transform). In contrast, we utilize \textbf{OrthoTrans}, a data-adaptive transformation based on an orthogonal matrix that diagonalizes the series' temporal Pearson correlation matrix. This approach enables more effective encoding and decoding in the decorrelated feature domain and can serve as a plug-in module to enhance existing forecasters. To enhance the representation learning for multivariate time series, we introduce a customized linear layer, \textbf{NormLin}, which employs a normalized weight matrix to capture multivariate dependencies. Empirically, the NormLin module shows a surprising performance advantage over multi-head self-attention, while requiring nearly half the FLOPs. Extensive experiments on 24 benchmarks and 140 forecasting tasks demonstrate that OLinear consistently achieves state-of-the-art performance with high efficiency. Notably, as a plug-in replacement for self-attention, the NormLin module consistently enhances Transformer-based forecasters. The code and datasets are available at \url{https://anonymous.4open.science/r/OLinear}.

\end{abstract}

\section{Introduction}

Multivariate time series forecasting is critical in fields such as weather \citep{nature_weather}, transportation \citep{traffic}, energy \citep{informer}, and finance \citep{survey4}. Time series forecasters typically adopt the temporal forecast (TF) paradigm \citep{timemixer++,itransformer,patchtst,timemixer}, which encodes time series into latent representations and decodes them back, all within the time domain. However, this paradigm struggles to fully exploit the forecasting potential  in the presence of entangled intra-series dependencies \citep{frets,freeformer}. To mitigate this issue, recent studies apply Fourier \citep{filternet,frets,freeformer} or wavelet \citep{wavelet_token} transforms to obtain the decorrelated feature sequence and perform encoding and decoding in the transformed domain. Nevertheless, these methods rely on dataset-independent bases, which fail to exploit the dataset-specific temporal correlation information.

In this paper, we introduce \textbf{OrthoTrans}, a dataset-adaptive transformation scheme that constructs an orthogonal basis via eigenvalue decomposition of the temporal Pearson correlation matrix \citep{statistical}. Projecting the series onto this basis obtains the decorrelated feature domain, providing a disentangled input for linear encoding and empirically improving forecasting performance. Notably, OrthoTrans is modular and can be integrated into existing forecasters to enhance their performance. In-depth ablation studies reveal that OrthoTrans promotes representation diversity and increases the rank of attention matrices in Transformer-based models.



As OrthoTrans transforms complex temporal variations into decorrelated features, the representation learning process can be effectively handled by linear layers \citep{frets}. Specifically, we employ a linear-based Cross-Series Learner (CSL) and Intra-Series Learner (ISL) to model multivariate correlations and sequential dynamics, respectively. To motivate our design, we note that in the classic self-attention mechanism, the attention entries are all positive with the row-wise L1 norm fixed as 1. Inspired by this, we design a new \textbf{lin}ear layer in CSL, called \textbf{NormLin}, where the weight matrix entries are made positive via the \texttt{Softplus} function and then row-wise \textbf{norm}alized. Surprisingly, the NormLin module consistently outperforms the self-attention mechanism while improving computational efficiency in the field of time series forecasting. The overall model is referred to as \textbf{OLinear}, and our contributions can be summarized as follows:

\begin{itemize}

\item In contrast to the commonly used TF paradigm, we introduce \textbf{OrthoTrans}, a dataset-adaptive transformation scheme which leverages the orthogonal matrix derived from the eigenvalue decomposition of the temporal Pearson correlation matrix. As a plug-in, it consistently improves the performance of existing forecasters.

\item For better representation learning, we present the \textbf{NormLin} layer, which employs a row-normalized weight matrix to capture  multivariate correlations. Notably, as a plug-in, the NormLin module improves both the accuracy and efficiency of Transformer-based forecasters. It also adapts well to decoder architectures and large-scale time series models.

\item  Extensive experiments on 24 benchmarks and 140 forecasting tasks (covering various datasets and prediction settings) demonstrate that \textbf{OLinear} consistently achieves state-of-the-art performance with competitive computational  efficiency. 

\end{itemize}

\section{Related work}

\subsection{Transformed domain in time series forecasting}

Deep learning–based forecasters typically adopt the TF paradigm \citep{timemixer++,timemixer,Leddam_icml,card,itransformer,patchtst,timesnet,linear}, with the entire process performed in the time domain. However, this paradigm may underperform in the presence of strong intra-series correlations \citep{frets,fredf}. To mitigate this, recent models propose to forecast in the frequency domain \citep{filternet,fits,frets} or the  wavelet domain \citep{wavelet_token} where step-wise correlations are reduced and performance improvements are observed.
FreDF \citep{fredf} incorporates a frequency-domain regularization term into the loss function to address the issue of biased forecast. However, classical Fourier and wavelet transforms employ fixed, dataset-agnostic bases that do not explicitly use the specific statistical characteristics of the dataset. In this work, we leverage orthogonal matrices derived from the temporal Pearson correlation matrix to decorrelate the series data \citep{pca}. This approach provides a more suitable input for linear encoding (Appendix~\ref{orthotrans_imp_rank}) and improves forecasting performance. Specifically, for Transformer-based forecasters, OrthoTrans increases the rank of the attention matrices, thereby enhancing the model’s expressiveness.


\subsection{Time series forecasters}

Deep learning based time series forecasters can generally be categorized into Transformer-based \citep{timemixer++,patchtst,itransformer,card,Leddam_icml}, Linear-based \citep{timemixer,linear,frets,fits,filternet}, TCN-based \citep{timesnet,moderntcn}, RNN-based \citep{rnn,rnn_nips2018}, and GNN-based \citep{crossgnn,fouriergnn} models. Recently, research interests increasingly focus on Transformer-based and linear-based methods, each offering distinct advantages. Transformer-based forecasters typically exhibit strong expressiveness, whereas linear-based models offer better computational efficiency. In this paper, we aim to achieve state-of-the-art performance using an efficient linear-based model. Specifically, we employ the linear-based NormLin module to robustly model inter-variate correlations. Remarkably, the NormLin module consistently outperforms the self-attention mechanism in Transformer-based forecasters, highlighting the strong capability of linear layers for time series forecasting.

\section{Preliminaries} \label{preliminaries}

Real-world time series often exhibit temporal dependencies, where each time step is influenced by its predecessors. In this section, we derive the expected value of a future step given its past, assuming a multivariate Gaussian distribution. Without loss of generality,  we consider a series $\mathbf{x} \in \mathbb{R}^{\mathsf{T}}$, and the following time step $y \in \mathbb{R}$. 
We now present the following theorem.

\begin{theorem}[Expected value of $y$] \label{theorem1}

We assume that $\mathbf{x} \sim \mathcal{N}  \left ( \mu _ {\mathbf{x}}, \Sigma _{\mathbf{x}} \right ) $, $y \sim \mathcal{N} \left ( \mu _y, \sigma  _y^2 \right ) $, and the joint distribution $[\mathbf{x},y] \sim \mathcal{N} \left ( \left [ \mu _{\mathbf{x}}, \mu _y \right ],  \begin{bmatrix}
 \Sigma _{\mathbf{x}} & \Sigma _{\mathbf{x}y} \\
 \Sigma _{\mathbf{x}y}^{\mathsf{T}} & \sigma _y^2
\end{bmatrix}   \right )$. Here, $\mathcal{N}\left ( \cdot , \cdot \right ) $ denotes a Gaussian distribution with the specified mean vector and covariance matrix. $\Sigma _{\mathbf{x}} \in \mathbb{R}^{t\times t}$ is symmetric and positive definite. $\Sigma _{\mathbf{x}y} \in \mathbb{R}^{t\times 1}$ denotes the covariance between $\mathbf{x}$ and $y$. Then, the expected value of $y$ given $\mathbf{x}$ is

\begin{equation}
\mu _{y\mid \mathbf{x} } = \mu _y + \Sigma _{\mathbf{x}y}^{\mathsf{T}}\Sigma _{\mathbf{x}}^{-1} \left ( \mathbf{x}-\mu _{\mathbf{x}} \right ) 
\label{eq1}
\end{equation}

\end{theorem}

The proof of Theorem~\ref{theorem1} is provided in Appendix \ref{proof}. Equation~\ref{eq1} shows that for a temporally correlated series $[\mathbf{x}, y]$, the expected value of $y$ depends not only on its own mean $\mu_y$, but also on its past observations  $\mathbf{x}$.  The second term in Equation~\ref{eq1} introduces additional difficulty to forecasting under the TF paradigm. In contrast, decorrelating the series simplifies the forecasting task. In this work, we introduce OrthoTrans to transform the original series into a decorrelated transformed domain, converting the forecasting problem into an \textit{inter-independent feature prediction} task. Experiments in Section~\ref{exp_base} validate the effectiveness of this transformation strategy and its generality when used as a plug-in for other forecasters.

\begin{figure}[t]
   \centering
   \includegraphics[width=1.0\linewidth]{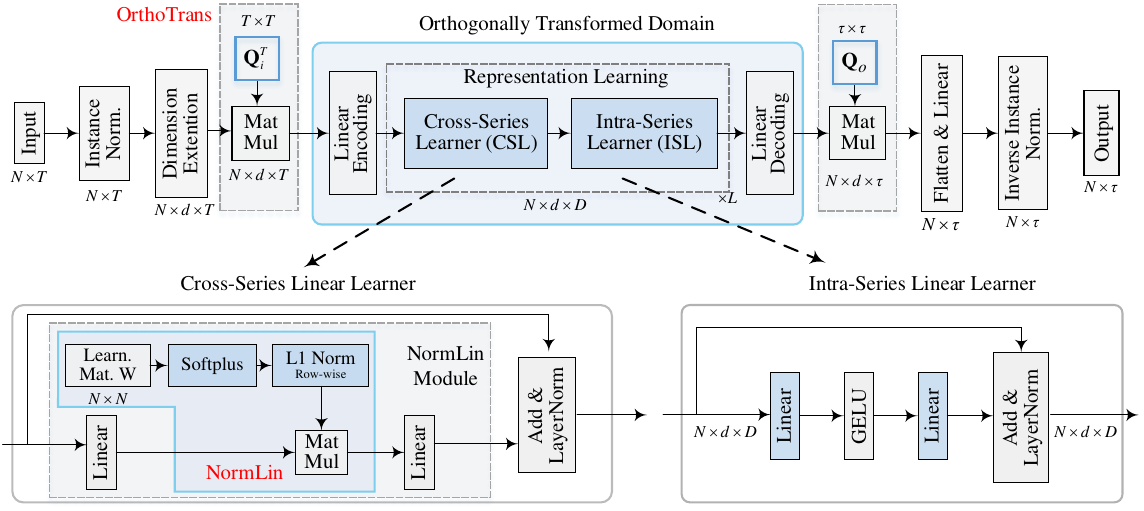}
   \caption{Overall structure of OLinear. The orthogonal matrix derived from the series' temporal Pearson correlation matrix transforms the series into the new feature domain with removed sequential correlations. The cross-series learners and intra-series learners are then employed for robust representation learning. Specifically, we introduce the NormLin layer, whose weight matrix is processed by the \texttt{Softplus} function and row-wise L1 normalization, to capture multivariate correlations.}
   \label{fig_arch}
\end{figure}

\section{Method}
\label{method}

\paragraph{Problem formulation}
For multivariate time series forecasting, given a historical sequence $\mathbf{X} =\left \{ \mathbf{x} _{1}, \cdots, \mathbf{x} _{T}  \right \} \in \mathbb{R}^{N \times T} $ with $T$ time steps and $N$ variates, the task is to  predict the future $\tau$ time steps $\mathbf{Y} =\left \{ \mathbf{x}_{T+1},\cdots, \mathbf{x}_{T+\tau}   \right \} \in \mathbb{R}^{N \times \tau}$. Our goal is to approximate the ground truth $\mathbf{Y}$ as closely as possible  with predictions $\hat{\mathbf{Y}}$.

\subsection{Overall architecture}

As shown in Figure~\ref{fig_arch}, OLinear adopts a simple yet effective architecture. Given the input series $\mathbf{X}$, a RevIN \citep{revin} layer first performs instance normalization to mitigate non-stationarity. A dimension extension module then enhances expressiveness \citep{frets} by computing the outer product with a learnable vector $\phi_d \in \mathbb{R}^d$, where $d$ is the embedding size. Next, the time domain is decorrelated by multiplying with a transposed orthogonal matrix $\mathbf{Q}_i^{\mathsf{T}} \in \mathbb{R}^{T \times T}$, as detailed in Section~\ref{ortho_trans_section}. This process can be formulated as $\mathbf{Z} = \left ( \mathrm{RevIN} _{\mathrm{Norm} } \left ( \mathbf{X} \right ) \otimes \phi _d \right ) \cdot \mathbf{Q }_i^{\mathsf{T}} \in \mathbb{R}^{N\times d\times T}$, where $\otimes$ and $\cdot$ denote the outer product and standard matrix multiplication, respectively. 


We then perform encoding and forecasting in the transformed domain. Specifically, a linear layer first encodes the decorrelated features $\mathbf{Z}$ to the model dimension $D$. The cross-series learner (CSL) and intra-series learner (ISL) subsequently  capture multivariate correlations and model intra-series dynamics, respectively. After passing through $L$ stacked blocks, the representation is decoded to the desired prediction length $\tau$, and then mapped back to the time domain via multiplication with the orthogonal matrix $\mathbf{Q}_o$.
The overall process is summarized as:

\begin{equation}
\begin{aligned}
\tilde{\mathbf{H}}^0  &= \mathrm{LinearEncode} \left ( \mathbf{Z} \right ) \in \mathbb{R}^{N\times d \times  D}, \\
\tilde{\mathbf{H}}^{l+1} &= \mathrm{ISL}\left ( \mathrm{CSL} \left ( \tilde{\mathbf{H}}^{l} \right )  \right ), \quad l=0,\cdots,L-1, \\
\tilde{\mathbf{Y}} &= \mathrm{LinearDecode} \left ( \tilde{\mathbf{H}}^{L} \right ) \cdot \mathbf{Q}_o  \in \mathbb{R}^{N\times d \times  \tau}.
\end{aligned}
\label{eq:mainblock}
\end{equation}

\setlength{\abovecaptionskip}{0pt} 
\setlength{\intextsep}{0pt} 
\begin{wrapfigure}[17]{r}{0.55\textwidth}
  \begin{center}
    \includegraphics[width=0.55\textwidth]{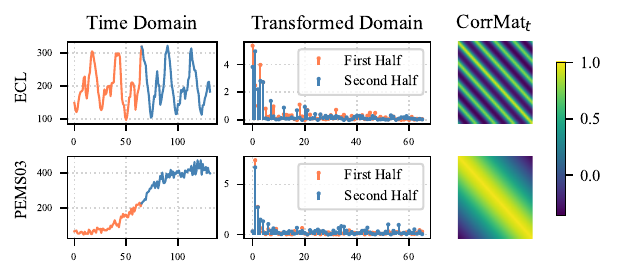}
  \end{center}
  \caption{Comparison between the time and transformed domains.  (1) The series exhibits strong consistency in the transformed domain, which is desirable for forecasting. (2) Sequence correlations are effectively reduced in the transformed domain. (3) Periodicity can be clearly reflected in the correlation matrix. (4) The new domain typically exhibits energy compaction, with  only a few dominant dimensions. More cases are shown in Figure~\ref{fig:time_trans_corr_Q}.}
  \label{fig_time_trans}
\end{wrapfigure}

Finally, the flattened output is mapped to shape $N \times \tau$ via a linear layer, then de-normalized to yield the final prediction: $\hat{\mathbf{Y}}=\mathrm{RevIN} _{\mathrm{DeNorm} }\left ( \mathrm{FlattenLinear}\left ( \tilde{\mathbf{Y}}  \right )  \right )$.


\subsection{Orthogonal transformation (OrthoTrans)} \label{ortho_trans_section}

One effective approach for decorrelating the series is based on the Pearson correlation matrix. Let $\mathbf{X}^{train} \in \mathbb{R}^{N \times M}$ denote the training set, where $M$ is the length of the training series. For each variate $j$, we generate $T$ lagged series with temporal offsets from $0$ to $T-1$: $\mathbf{s}_i^j = \mathbf{X} \left [j,\ i:M-T+i \right ], \  i=0,\cdots T-1,$ where $\mathbf{s}_i^j$ denotes the $i$-th temporally lagged series of variate $j$. We then compute the Pearson correlation matrix $\mathrm{CorrMat} _t^j$ of $\{\mathbf{s}_i^j\}$, whose $(p, q)$-th entry is $\frac{\mathrm{Cov}\left ( \mathbf{s}_p ,\mathbf{s}_q   \right ) }{\sqrt{\mathrm{Var}\left ( \mathbf{s}_p \right ) \cdot \mathrm{Var}\left ( \mathbf{s}_q \right )} }, \  p,q\in \left \{ 0,\cdots, T-1 \right \} $.




Here, $\mathrm{Cov}(\cdot)$ denotes the covariance of two series, and $\mathrm{Var}(\cdot)$ denotes the variance. The final Pearson correlation matrix is then obtained by averaging over all variates: $\mathrm{CorrMat} _t = \frac{1}{N} \sum_{j=0}^{N-1} \mathrm{CorrMat} _t^j$.


Following the above procedure, the resulting $\mathrm{CorrMat}_t$ is symmetric with all diagonal entries equal to 1.  Based on the properties of symmetric matrices, we perform eigenvalue decomposition as $\mathrm{CorrMat}_t = \mathbf{Q}_i \Lambda \mathbf{Q}_i^{\mathsf{T}}$, where $\mathbf{Q}_i$ is an orthogonal matrix \citep{matrix} whose columns are the eigenvectors of $\mathrm{CorrMat}_t$, and $\Lambda$ is a diagonal matrix containing the corresponding eigenvalues.


For the input series $\mathbf{x}$ with temporal correlation matrix $\mathrm{CorrMat}_t$ and unit variance, the covariance matrix of the transformed vector $\mathbf{Q}_i^{\mathsf{T}} \mathbf{x}$ is $\mathrm{CovMat}\left ( \mathbf{Q}_i^{\mathsf{T}} \mathbf{x}  \right ) = \mathbf{Q}_i^{\mathsf{T}} \mathrm{CorrMat}_t\mathbf{Q}_i
=\mathbf{Q}_i^{\mathsf{T}} \mathbf{Q}_i \Lambda \mathbf{Q}_i^{\mathsf{T}} \mathbf{Q}_i = \Lambda$, which is a diagonal matrix. Therefore, the entries of $\mathbf{Q}_i^{\mathsf{T}} \mathbf{x}$ are linearly independent, removing sequential dependencies. Similarly, we can compute $\mathbf{Q}_o \in \mathbb{R}^{\tau \times \tau}$, and recover the temporal correlations by multiplying with $\mathbf{Q}_o$. Note that $\mathbf{Q}_i$ and $\mathbf{Q}_o$ are pre-computed and used throughout training and inference. Ablation studies on these two matrices are discussed in Appendix~\ref{abl_qi_qo}. Moreover, as shown in Appendix~\ref{Q_mat_robust}, OLinear remains robust even when the orthogonal matrices $\mathbf{Q}_i$ and $\mathbf{Q}_o$ are computed with limited training data.


\paragraph{Discussion}
 (1) Mathematically, $\mathbf{Q}_i^{\mathsf{T}} \mathbf{x}$ projects $\mathbf{x}$ onto the eigenvectors of the temporal correlation matrix $\mathrm{CorrMat}_t$. The transformed domain exhibits energy compaction, and \textbf{noise is suppressed} in the primary components \citep{prml_book}. (2) Compared to DFT and wavelet transforms, or to the TF paradigm without any transformation, OrthoTrans produces \textbf{higher-rank attention matrices} for Transformer-based forecasters (see Figures~\ref{fig:orthotrans_improve_rank}– \ref{fig:orthotrans_improve_rank_patchtst}), indicating greater  representation diversity and enhanced model expressiveness. This partly explains why OrthoTrans enhances other forecasters when used as a plug-in module.

\subsection{Representation learning}


\paragraph{CSL and NormLin}


Inspired by the properties of attention matrices in self-attention \citep{transformer}, we impose two constraints on the weight matrix of the linear layer for robust multivariate representation learning: (1) all entries must be positive, and (2) each row must sum to 1. To enforce these constraints, we apply the \texttt{Softplus} function followed by row-wise L1 normalization to the learnable weight matrix $\mathbf{W}$. The resulting layer is referred to as NormLin, and is defined as follows:


\begin{equation}
\mathrm{NormLin}\left ( \mathbf{x}  \right )  = \mathrm{RowNorm} _{\mathrm{L1} }\left ( \mathrm{Softplus} \left ( \mathbf{W}  \right )  \right ) \ \mathbf{x}, \mathbf{W} \in \mathbb{R}^ {N \times N}, \mathbf{x} \in \mathbb{R}^N.
\label{eq:norlin}
\end{equation}

The variants of NormLin are discussed in Appendix~\ref{normlin_abl}. Incorporating the other components in Figure~\ref{fig_arch}(b), the CSL process is formulated as:


\begin{equation}
\mathrm{CSL} \left ( \tilde{\mathbf{H}}   \right ) = \mathrm{LayerNorm}\left ( \tilde{\mathbf{H}} + \mathrm{Linear} \left ( \mathrm{NormLin} \left ( \mathrm{Linear} \left ( \tilde{\mathbf{H}} \right )  \right )  \right )  \right )      ,
\label{eq:csl}
\end{equation}

\begin{wraptable}[10]{r}{0.42\textwidth}
\caption{Comparison of FLOPs and memory usage between the NormLin module and multi-head self-attention (MHSA). Here, $h$ denotes the number of attention heads, typically set to 8.}
 \vspace{4pt}
\label{tab:flops}
\centering
\setlength{\tabcolsep}{1pt}
\renewcommand{\arraystretch}{1.0} 
{\fontsize{7}{9}\selectfont
\begin{tabular}{@{}ccc@{}}
\toprule
Module & NormLin Module & MHSA \\ \midrule
FLOPs  & $ \mathcal{O} \left ( N^2D+2ND^2 \right )$             & $ \mathcal{O} \left ( 2N^2D+4ND^2 \right )$    \\
Memory & $ \mathcal{O} \left ( N^2+ND \right )$             &  $\mathcal{O} \left ( hN^2+ND \right )$    \\ \bottomrule
\end{tabular}
}
\end{wraptable}

where $\mathrm{Linear}(\cdot)$ and $\mathrm{NormLin}(\cdot)$ operate on the sequence and variate dimensions, respectively. We define \textbf{the NormLin module} as $\mathrm{NormLinModule}\left ( \cdot \right ) \triangleq \mathrm{Linear} \left ( \mathrm{NormLin} \left ( \mathrm{Linear} \left ( \cdot \right )  \right )  \right )$. Ablation studies on the two linear layers in $\mathrm{NormLinModule}\left ( \cdot \right )$ are presented in Appendix~\ref{pre_post_lin}.

\paragraph{ISL} We adopt two linear layers separated by the \texttt{GELU} activation function as a powerful predictive representation learner  \citep{rlinear}. It has been well established that MLPs are highly effective for encoding sequential dynamics and decoding future series \citep{linear,tide}. Similar to CSL, residual connections and LayerNorm are also applied. The complete ISL process is defined as:

\begin{equation}
\mathrm{ISL} \left ( \tilde{\mathbf{H}}   \right ) = \mathrm{LayerNorm}\left ( \tilde{\mathbf{H}} + \mathrm{Linear} \left ( \mathrm{GELU} \left ( \mathrm{Linear} \left ( \tilde{\mathbf{H}} \right )  \right )  \right )  \right ).
\label{eq:isl}
\end{equation}

\paragraph{Discussion} 

 Compared to the self-attention mechanism, the NormLin module offers three main advantages:
(1) As shown in Table~\ref{tab:flops}, it reduces \textbf{computational} complexity by half and decreases memory footprint by a factor of $1/h$, where $h$ is the number of attention heads;
(2) The learned weight matrix in NormLin naturally exhibits \textbf{high rank}, in contrast to the low-rank nature of self-attention (see Appendix~\ref{low_rank}). The low-rank issue in self-attention could arise from the sharp value concentration of the Softmax function. A higher-rank weight (or attention) matrix often better preserves the rank of the representation space and thus improves the model’s expressiveness \citep{flattentrans}. 
(3) From the perspective of \textbf{gradient flow}, the NormLin layer provides a more direct backpropagation path for optimizing weight entries. Appendix~\ref{jacob} shows that the Jacobian matrices of $\mathrm{Softmax}(\cdot)$ in self-attention and $\mathrm{Norm}_{\mathrm{L1}}(\mathrm{Softplus}(\cdot))$ in the NormLin layer share a similar structure, but the latter offers greater flexibility. As illustrated in Figure~\ref{fig:jacob_mat}, the Jacobian matrix of NormLin typically contains more large-magnitude entries, indicating stronger and more effective gradient propagation during training.


\section{Experiments}
\label{Experiments}

\paragraph{Datasets and implementation details} 
We extensively evaluate OLinear using 24 diverse real-world datasets: \textbf{ETT} (four subsets), \textbf{Weather}, \textbf{ECL}, \textbf{Traffic}, \textbf{Exchange}, \textbf{Solar-Energy}, \textbf{PEMS} (four subsets), \textbf{ILI}, \textbf{COVID-19}, \textbf{METR-LA}, \textbf{NASDAQ}, \textbf{Wiki}, \textbf{SP500}, \textbf{DowJones}, \textbf{CarSales}, \textbf{Power}, \textbf{Website}, \textbf{Unemp}. The weighted L1 loss function from CARD \citep{card} is adopted. The embedding size $d$ is set as 16. Dataset description and more implementation details are presented in Appendices~\ref{dataset} and \ref{imple_details}, respectively.

\subsection{Forecasting performance}

\paragraph{Baselines} 
We carefully choose 11 well-acknowledged state-of-the-art forecasting models as our baselines, including (1) Linear-based models: TimeMixer \citep{timemixer}, FilterNet \citep{filternet}, FITS \citep{fits}, DLinear \citep{linear}; (2) Transformer-based models: TimeMixer++ \citep{timemixer++}, Leddam \citep{Leddam_icml}, CARD \citep{card}, Fredformer \citep{fredformer}, iTransformer \citep{itransformer}, PatchTST \citep{patchtst}; (3) TCN-based model: TimesNet \citep{timesnet}.

\begin{table}[t]
\caption{Long-term forecasting results with prediction lengths $\tau \in \left \{ 12,24,48,96 \right \}$ for PEMS and $\tau \in \left \{ 96,192,336,720 \right \}$ for others. Lookback horizon $T=96$. Results are averaged over four prediction lengths. \textit{Avg} denotes further averaging over subsets. Full results are shown in Table~\ref{tab:full_long}. }
\label{tab:long_term}
\centering
\setlength{\tabcolsep}{0.9pt}
\renewcommand{\arraystretch}{1.0} 
{\fontsize{7}{9}\selectfont
\begin{tabular}{@{}ccccccccccccccccccccccc@{}}
\toprule
Model       & 
\multicolumn{2}{c}{\begin{tabular}[c]{@{}c@{}}OLinear\\      (Ours) \end{tabular}}                                                     & 
\multicolumn{2}{c}{\begin{tabular}[c]{@{}c@{}}TimeMix.\\      \citeyear{timemixer} \end{tabular}} & 
\multicolumn{2}{c}{\begin{tabular}[c]{@{}c@{}}FilterNet\\      \citeyear{filternet} \end{tabular}} & 
\multicolumn{2}{c}{\begin{tabular}[c]{@{}c@{}}DLinear\\      \citeyear{linear} \end{tabular}} & 
\multicolumn{2}{c}{\begin{tabular}[c]{@{}c@{}}TimeMix.++\\      \citeyear{timemixer++} \end{tabular}} & 
\multicolumn{2}{c}{\begin{tabular}[c]{@{}c@{}}Leddam\\      \citeyear{Leddam_icml} \end{tabular}} & 
\multicolumn{2}{c}{\begin{tabular}[c]{@{}c@{}}CARD\\      \citeyear{card} \end{tabular}} & 
\multicolumn{2}{c}{\begin{tabular}[c]{@{}c@{}}Fredformer\\      \citeyear{fredformer} \end{tabular}} & 
\multicolumn{2}{c}{\begin{tabular}[c]{@{}c@{}}iTrans.\\      \citeyear{itransformer} \end{tabular}} & 
\multicolumn{2}{c}{\begin{tabular}[c]{@{}c@{}}PatchTST\\      \citeyear{patchtst} \end{tabular}} & 
\multicolumn{2}{c}{\begin{tabular}[c]{@{}c@{}}TimesNet\\      \citeyear{timesnet} \end{tabular}} \\ \midrule
Metric       & MSE                                   & MAE                                   & MSE                                    & MAE                                    & MSE                                    & MAE                                    & MSE                                                  & MAE                    & MSE                                     & MAE                                     & MSE                                   & MAE                                  & MSE                                  & MAE                                 & MSE                                     & MAE                                    & MSE                                                  & MAE                    & MSE                            & MAE                                           & MSE                                    & MAE                                   \\ \midrule
ETT(Avg)     & {\color[HTML]{0000FF} {\ul 0.359}}    & {\color[HTML]{FF0000} \textbf{0.376}} & 0.367                                  & 0.388                                  & 0.375                                  & 0.394                                  & 0.442                                                & 0.444                  & {\color[HTML]{FF0000} \textbf{0.349}}     & {\color[HTML]{0000FF} {\ul 0.377}}    & 0.367                                 & 0.387                                & 0.366                 & 0.380                                              & 0.366                                   & 0.385                                  & 0.383                                                & 0.399                  & 0.380                                  & 0.396                                 & 0.391                                  & 0.404                                 \\
ECL          & {\color[HTML]{FF0000} \textbf{0.159}} & {\color[HTML]{FF0000} \textbf{0.248}} & 0.182                                  & 0.273                                  & 0.173                                  & 0.268                                  & 0.212                                                & 0.300                  & {\color[HTML]{0000FF} {\ul 0.165}}      & {\color[HTML]{0000FF} {\ul 0.253}}      & 0.169                                 & 0.263                                & 0.168                                & 0.258                               & 0.176                                   & 0.269                                  & 0.178                                                & 0.270                  & 0.208                          & 0.295                                         & 0.192                                  & 0.295                                 \\
Exchange     & 0.355                                 & {\color[HTML]{0000FF} {\ul 0.399}}    & 0.387                                  & 0.416                                  & 0.388                                  & 0.419                                  & {\color[HTML]{0000FF} {\ul 0.354}}                   & 0.414                  & 0.357                                   & 0.409                                   & {\color[HTML]{0000FF} {\ul 0.354}}    & 0.402                                & 0.362                                & 0.402                               & {\color[HTML]{FF0000} \textbf{0.333}}   & {\color[HTML]{FF0000} \textbf{0.391}}  & 0.360                                                & 0.403                  & 0.367                          & 0.404                                         & 0.416                                  & 0.443                                 \\
Traffic      & 0.451                                 & {\color[HTML]{FF0000} \textbf{0.247}} & 0.485                                  & 0.298                                  & 0.463                                  & 0.310                                  & 0.625                                                & 0.383                  & {\color[HTML]{FF0000} \textbf{0.416}}   & {\color[HTML]{0000FF} {\ul 0.264}}      & 0.467                                 & 0.294                                & 0.453                                & 0.282                               & 0.433                                   & 0.291                                  & {\color[HTML]{0000FF} {\ul 0.428}}                   & 0.282                  & 0.531                          & 0.343                                         & 0.620                                  & 0.336                                 \\
Weather      & {\color[HTML]{0000FF} {\ul 0.237}}    & {\color[HTML]{FF0000} \textbf{0.260}} & 0.240                                  & 0.272                                  & 0.245                                  & 0.272                                  & 0.265                                                & 0.317                  & {\color[HTML]{FF0000} \textbf{0.226}}     & {\color[HTML]{0000FF} {\ul 0.262}}    & 0.242                                 & 0.272                                & 0.239                 & 0.265                                              & 0.246                                   & 0.272                                  & 0.258                                                & 0.279                  & 0.259                                  & 0.281                                 & 0.259                                  & 0.287                                 \\
Solar & {\color[HTML]{0000FF} {\ul 0.215}}    & {\color[HTML]{FF0000} \textbf{0.217}} & 0.216                                  & 0.280                                  & 0.235                                  & 0.266                                  & 0.330                                                & 0.401                  & {\color[HTML]{FF0000} \textbf{0.203}}   & 0.258                                   & 0.230                                 & 0.264                                & 0.237                                & {\color[HTML]{0000FF} {\ul 0.237}}  & 0.226                                   & 0.262                                  & 0.233                                                & 0.262                  & 0.270                          & 0.307                                         & 0.301                                  & 0.319                                 \\
PEMS03       & {\color[HTML]{FF0000} \textbf{0.095}} & {\color[HTML]{FF0000} \textbf{0.199}} & 0.167                                  & 0.267                                  & 0.145                                  & 0.251                                  & 0.278                                                & 0.375                  & 0.165                                   & 0.263                                   & {\color[HTML]{0000FF} {\ul 0.107}}    & {\color[HTML]{0000FF} {\ul 0.210}}   & 0.174                                & 0.275                               & 0.135                                   & 0.243                                  & 0.113                                                & 0.221                  & 0.180                          & 0.291                                         & 0.147                                  & 0.248                                 \\
PEMS04       & {\color[HTML]{FF0000} \textbf{0.091}} & {\color[HTML]{FF0000} \textbf{0.190}} & 0.185                                  & 0.287                                  & 0.146                                  & 0.258                                  & 0.295                                                & 0.388                  & 0.136                                   & 0.251                                   & {\color[HTML]{0000FF} {\ul 0.103}}    & {\color[HTML]{0000FF} {\ul 0.210}}   & 0.206                                & 0.299                               & 0.162                                   & 0.261                                  & 0.111                                                & 0.221                  & 0.195                          & 0.307                                         & 0.129                                  & 0.241                                 \\
PEMS07       & {\color[HTML]{FF0000} \textbf{0.077}} & {\color[HTML]{FF0000} \textbf{0.164}} & 0.181                                  & 0.271                                  & 0.123                                  & 0.229                                  & 0.329                                                & 0.395                  & 0.152                                   & 0.258                                   & {\color[HTML]{0000FF} {\ul 0.084}}    & {\color[HTML]{0000FF} {\ul 0.180}}   & 0.149                                & 0.247                               & 0.121                                   & 0.222                                  & 0.101                                                & 0.204                  & 0.211                          & 0.303                                         & 0.124                                  & 0.225                                 \\
PEMS08       & {\color[HTML]{FF0000} \textbf{0.113}} & {\color[HTML]{FF0000} \textbf{0.194}} & 0.226                                  & 0.299                                  & 0.172                                  & 0.260                                  & 0.379                                                & 0.416                  & 0.200                                   & 0.279                                   & {\color[HTML]{0000FF} {\ul 0.122}}    & {\color[HTML]{0000FF} {\ul 0.211}}   & 0.201                                & 0.280                               & 0.161                                   & 0.250                                  & 0.150                                                & 0.226                  & 0.280                          & 0.321                                         & 0.193                                  & 0.271                                 \\ \midrule
1st Count   & 5                                     & 9                                     & 0                                      & 0                                      & 0                                      & 0                                      & 0                                                    & 0                      & 4                                         & 0                                     & 0                                     & 0                                    & 0                     & 0                                                  & 1                                       & 1                                      & 0                                                    & 0                      & 0                                      & 0                                     & 0                                      & 0                                     \\ \bottomrule
\end{tabular}

}
\end{table}


\begin{table}[t]
\caption{Short-term forecasting results. Two settings are applied:  S1 (Input-12, Predict-$\left \{ 3,6,9,12 \right \}$) and S2 (Input-36, Predict-$\left \{ 24,36,48,60 \right \}$). Average results across eight prediction lengths are reported. Full results are provided in Tables~\ref{tab:full_short_part1} and \ref{tab:full_short_part2}.}
\label{tab:short_term}
\centering
\setlength{\tabcolsep}{0.85pt}
\renewcommand{\arraystretch}{1.0} 
{\fontsize{7}{9}\selectfont
\begin{tabular}{@{}ccccccccccccccccccccccc@{}}
\toprule
Model       & 
\multicolumn{2}{c}{\begin{tabular}[c]{@{}c@{}}OLinear\\      (Ours) \end{tabular}}                                                     & 
\multicolumn{2}{c}{\begin{tabular}[c]{@{}c@{}}TimeMix.\\      \citeyear{timemixer} \end{tabular}} & 
\multicolumn{2}{c}{\begin{tabular}[c]{@{}c@{}}FilterNet\\      \citeyear{filternet} \end{tabular}} & 
\multicolumn{2}{c}{\begin{tabular}[c]{@{}c@{}}DLinear\\      \citeyear{linear} \end{tabular}} & 
\multicolumn{2}{c}{\begin{tabular}[c]{@{}c@{}}TimeMix.++\\      \citeyear{timemixer++} \end{tabular}} & 
\multicolumn{2}{c}{\begin{tabular}[c]{@{}c@{}}Leddam\\      \citeyear{Leddam_icml} \end{tabular}} & 
\multicolumn{2}{c}{\begin{tabular}[c]{@{}c@{}}CARD\\      \citeyear{card} \end{tabular}} & 
\multicolumn{2}{c}{\begin{tabular}[c]{@{}c@{}}Fredformer\\      \citeyear{fredformer} \end{tabular}} & 
\multicolumn{2}{c}{\begin{tabular}[c]{@{}c@{}}iTrans.\\      \citeyear{itransformer} \end{tabular}} & 
\multicolumn{2}{c}{\begin{tabular}[c]{@{}c@{}}PatchTST\\      \citeyear{patchtst} \end{tabular}} & 
\multicolumn{2}{c}{\begin{tabular}[c]{@{}c@{}}TimesNet\\      \citeyear{timesnet} \end{tabular}} \\ \midrule
Metric    & MSE                                       & MAE                                       & MSE                                      & MAE                                  & MSE                                    & MAE                                    & MSE                                     & MAE                                 & MSE                                                     & MAE                     & MSE                                   & MAE                                  & MSE                                & MAE                                   & MSE                                      & MAE                                   & MSE                                   & MAE                                   & MSE                                    & MAE                                   & MSE                                    & MAE                                   \\ \midrule
ILI       & {\color[HTML]{FF0000} \textbf{1.429}}     & {\color[HTML]{FF0000} \textbf{0.690}}     & 1.864                                    & 0.806                                & 1.793                                  & 0.791                                  & 2.742                                   & 1.126                               & 1.805                                                   & 0.793                   & 1.725                                 & 0.777                                & 1.959                              & 0.822                                 & 1.732                                    & 0.797                                 & {\color[HTML]{0000FF} {\ul 1.715}}    & {\color[HTML]{0000FF} {\ul 0.773}}    & 1.905                                  & 0.804                                 & 1.809                                  & 0.807                                 \\
COVID-19  & {\color[HTML]{FF0000} \textbf{5.187}}     & {\color[HTML]{FF0000} \textbf{1.211}}     & 5.919                                    & 1.350                                & 5.607                                  & 1.322                                  & 8.279                                   & 1.601                               & 5.974                                                   & 1.369                   & {\color[HTML]{0000FF} {\ul 5.251}}    & {\color[HTML]{0000FF} {\ul 1.285}}   & 5.536                              & 1.314                                 & 5.279                                    & 1.287                                 & 5.301                                 & 1.293                                 & 5.836                                  & 1.362                                 & 6.106                                  & 1.369                                 \\
METR-LA   & 0.587                                     & {\color[HTML]{FF0000} \textbf{0.311}}     & 0.608                                    & 0.372                                & 0.603                                  & 0.366                                  & {\color[HTML]{0000FF} {\ul 0.580}}      & 0.422                               & {\color[HTML]{FF0000} \textbf{0.567}}                   & 0.363                   & 0.603                                 & 0.367                                & 0.639                              & {\color[HTML]{0000FF} {\ul 0.350}}    & 0.617                                    & 0.369                                 & 0.627                                 & 0.373                                 & 0.614                                  & 0.372                                 & 0.617                                  & 0.370                                 \\
NASDAQ    & {\color[HTML]{0000FF} {\ul 0.121}}        & {\color[HTML]{FF0000} \textbf{0.201}}     & {\color[HTML]{FF0000} \textbf{0.120}}    & {\color[HTML]{0000FF} {\ul 0.204}}   & 0.127                                  & 0.211                                  & 0.150                                   & 0.251                               & 0.125                                                   & 0.210                   & 0.128                                 & 0.211                                & 0.125                              & 0.207                                 & 0.127                                    & 0.210                                 & 0.133                                 & 0.217                                 & 0.128                                  & 0.209                                 & 0.161                                  & 0.247                                 \\
Wiki      & {\color[HTML]{0000FF} {\ul 6.395}}        & {\color[HTML]{FF0000} \textbf{0.415}}     & 6.443                                    & 0.439                                & 6.457                                  & 0.439                                  & 6.420                                   & 0.510                               & 6.430                                                   & 0.443                   & 6.417                                 & 0.433                                & 6.419                              & 0.427                                 & {\color[HTML]{FF0000} \textbf{6.318}}    & 0.429                                 & 6.422                                 & 0.432                                 & 6.368                                  & {\color[HTML]{0000FF} {\ul 0.424}}    & 7.633                                  & 0.572                                 \\
SP500     & {\color[HTML]{FF0000} \textbf{0.146}}     & {\color[HTML]{FF0000} \textbf{0.250}}     & 0.153                                    & 0.265                                & 0.164                                  & 0.279                                  & 0.178                                   & 0.298                               & 0.157                                                   & 0.270                   & 0.163                                 & 0.282                                & {\color[HTML]{0000FF} {\ul 0.147}} & {\color[HTML]{0000FF} {\ul 0.252}}    & 0.167                                    & 0.286                                 & 0.161                                 & 0.279                                 & 0.159                                  & 0.277                                 & 0.150                                  & 0.262                                 \\
DowJones  & {\color[HTML]{FF0000} \textbf{7.686}}     & {\color[HTML]{FF0000} \textbf{0.619}}     & 8.499                                    & 0.633                                & 8.283                                  & 0.633                                  & 7.893                                   & 0.626                               & 8.895                                                   & 0.643                   & 8.257                                 & 0.633                                & {\color[HTML]{0000FF} {\ul 7.699}} & {\color[HTML]{FF0000} \textbf{0.619}} & 8.041                                    & {\color[HTML]{0000FF} {\ul 0.625}}    & 8.177                                 & 0.630                                 & 7.991                                  & 0.626                                 & 10.960                                 & 0.737                                 \\
CarSales  & 0.330                                     & {\color[HTML]{FF0000} \textbf{0.305}}     & 0.333                                    & 0.322                                & 0.328                                  & 0.319                                  & 0.387                                   & 0.376                               & 0.337                                                   & 0.321                   & 0.335                                 & 0.322                                & 0.347                              & 0.324                                 & 0.335                                    & 0.325                                 & {\color[HTML]{FF0000} \textbf{0.311}} & {\color[HTML]{0000FF} {\ul 0.307}}    & {\color[HTML]{0000FF} {\ul 0.327}}     & 0.318                                 & 0.334                                  & 0.328                                 \\
Power     & {\color[HTML]{0000FF} {\ul 1.248}}        & {\color[HTML]{FF0000} \textbf{0.835}}     & {\color[HTML]{FF0000} \textbf{1.234}}    & {\color[HTML]{0000FF} {\ul 0.840}}   & 1.309                                  & 0.870                                  & 1.278                                   & 0.870                               & {\color[HTML]{FF0000} \textbf{1.234}}                   & 0.841                   & 1.295                                 & 0.868                                & 1.288                              & 0.847                                 & 1.302                                    & 0.870                                 & 1.324                                 & 0.874                                 & 1.311                                  & 0.873                                 & 1.317                                  & 0.871                                 \\
Website   & {\color[HTML]{0000FF} {\ul 0.225}}        & {\color[HTML]{0000FF} {\ul 0.311}}        & 0.279                                    & 0.358                                & 0.297                                  & 0.367                                  & 0.302                                   & 0.389                               & 0.260                                                   & 0.344                   & 0.264                                 & 0.348                                & 0.303                              & 0.366                                 & 0.266                                    & 0.351                                 & {\color[HTML]{FF0000} \textbf{0.179}} & {\color[HTML]{FF0000} \textbf{0.297}} & 0.284                                  & 0.362                                 & 0.251                                  & 0.341                                 \\
Unemp     & {\color[HTML]{0000FF} {\ul 0.729}}        & {\color[HTML]{FF0000} \textbf{0.461}}     & 1.581                                    & 0.708                                & 1.286                                  & 0.627                                  & {\color[HTML]{FF0000} \textbf{0.565}}   & {\color[HTML]{0000FF} {\ul 0.509}}  & 1.506                                                   & 0.678                   & 1.502                                 & 0.689                                & 1.163                              & 0.596                                 & 2.048                                    & 0.789                                 & 1.408                                 & 0.666                                 & 1.237                                  & 0.624                                 & 2.328                                  & 0.852                                 \\ \midrule
1st Count & 4                                         & 10                                        & 2                                        & 0                                    & 0                                      & 0                                      & 1                                       & 0                                   & 2                                                       & 0                       & 0                                     & 0                                    & 0                                  & 1                                     & 1                                        & 0                                     & 2                                     & 1                                     & 0                                      & 0                                     & 0                                      & 0                                     \\ \bottomrule
\end{tabular}
}
\end{table}

\paragraph{Main results}

\begin{wrapfigure}[12]{r}{0.5\textwidth}
  \begin{center}
    \includegraphics[width=0.5\textwidth]{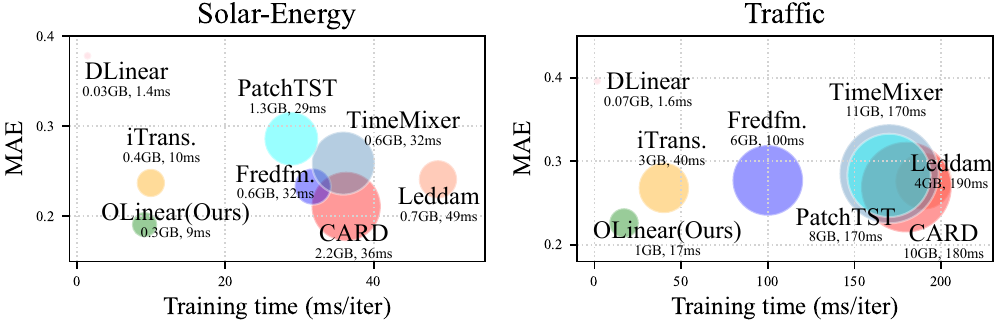}
  \end{center}
   \caption{Model efficiency comparison. Bubble areas represent GPU memory usage during training, scaled independently in the two subfigures. Resource footprint data are from Table~\ref{tab:GPU}.}
  \label{fig:gpu_main_text}
\end{wrapfigure}

Comprehensive long-term and short-term forecasting results are presented in Tables~\ref{tab:long_term} and \ref{tab:short_term}, respectively, with the best results highlighted in {\color[HTML]{FF0000} \textbf{bold}} and the second-best {\color[HTML]{0000FF} {\ul underlined}}. Lower MSE/MAE values indicate more accurate predictions. Across a wide range of benchmarks, OLinear consistently outperforms state-of-the-art Transformer-based and linear-based forecasters. Notably, these gains are achieved with high computational efficiency. (Figure~\ref{fig:gpu_main_text} and Table~\ref{tab:GPU}).

\setlength{\abovecaptionskip}{0pt} 
\setlength{\intextsep}{0pt} 

We attribute this superior performance to the adopted OrthoTrans and NormLin modules. The effectiveness of the simpler NormLin module challenges the necessity of the widely adopted multi-head self-attention mechanism, which has been a dominant design in prior works. To further validate robustness, we evaluate OLinear under varying lookback lengths (Table~\ref{tab:best_var_lookback}), where it consistently outperforms existing state-of-the-art methods.

\begin{figure}[t]
   \centering
   \includegraphics[width=1.0\linewidth]{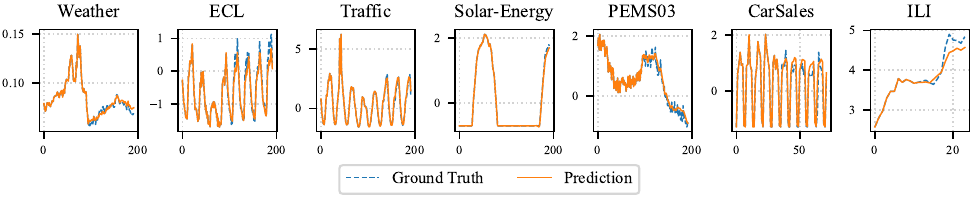}
   \caption{Visualization of the forecasting results of OLinear, demonstrating great accuracy.}
   \label{fig_visual}
\end{figure}

Figure~\ref{fig_visual} shows the prediction visualizations of OLinear. Moreover, Table~\ref{tab:robust_compare} demonstrates that OLinear exhibits greater robustness to random seeds compared to state-of-the-art Transformer-based forecasters such as TimeMixer++ and iTransformer.

\begin{table}[t]
\caption{Ablation study on transformation bases. \textit{S1} and \textit{S2} represent `Input-12, Predict-$\{3,6,9,12\}$' and `Input-36, Predict-$\{24,36,48,60\}$', respectively. \textit{Wavelet1} and \textit{Wavelet2} use the Haar and discrete Meyer wavelets. \textit{Identity} denotes no transformation. Full results are in Table~\ref{tab:base_full}.}

\label{tab:base}
\centering
\setlength{\tabcolsep}{3.1pt}
\renewcommand{\arraystretch}{1.0} 
{\fontsize{8}{9}\selectfont
\begin{tabular}{@{}ccccccccccccccc@{}}
\toprule
Dataset   & \multicolumn{2}{c}{ECL}                                                       & \multicolumn{2}{c}{Solar-Energy}                                              & \multicolumn{2}{c}{PEMS03}                                                    & \multicolumn{2}{c}{Power (S2)}                                                & \multicolumn{2}{c}{ILI (S1)}                                                  & \multicolumn{2}{c}{COVID (S2)}                                             & \multicolumn{2}{c}{METR-LA (S2)}                                              \\ \midrule
Metric    & MSE                                   & MAE                                   & MSE                                   & MAE                                   & MSE                                   & MAE                                   & MSE                                   & MAE                                   & MSE                                   & MAE                                   & MSE                                   & MAE                                   & MSE                                   & MAE                                   \\ \midrule
Ours      & {\color[HTML]{FF0000} \textbf{0.159}} & {\color[HTML]{FF0000} \textbf{0.248}} & {\color[HTML]{FF0000} \textbf{0.215}} & {\color[HTML]{FF0000} \textbf{0.217}} & {\color[HTML]{FF0000} \textbf{0.095}} & {\color[HTML]{FF0000} \textbf{0.199}} & {\color[HTML]{FF0000} \textbf{1.487}} & {\color[HTML]{FF0000} \textbf{0.922}} & {\color[HTML]{FF0000} \textbf{1.094}} & {\color[HTML]{FF0000} \textbf{0.578}} & {\color[HTML]{FF0000} \textbf{8.467}} & {\color[HTML]{FF0000} \textbf{1.754}} & {\color[HTML]{FF0000} \textbf{0.838}} & {\color[HTML]{FF0000} \textbf{0.402}} \\
Fourier   & 0.161                                 & 0.250                                 & {\color[HTML]{0000FF} {\ul 0.219}}    & {\color[HTML]{0000FF} {\ul 0.219}}    & {\color[HTML]{0000FF} {\ul 0.101}}    & {\color[HTML]{0000FF} {\ul 0.204}}    & 1.614                                 & 0.967                                 & 1.268                                 & 0.584                                 & 9.165                                 & 1.839                                 & 0.843                                 & {\color[HTML]{0000FF} {\ul 0.403}}    \\
Wavelet1  & {\color[HTML]{0000FF} {\ul 0.160}}    & {\color[HTML]{0000FF} {\ul 0.249}}    & 0.221                                 & 0.221                                 & 0.107                                 & 0.210                                 & 1.663                                 & 0.987                                 & {\color[HTML]{0000FF} {\ul 0.116}}                                & {\color[HTML]{0000FF} {\ul 0.580}}    & 8.666                                 & 1.799                                 & {\color[HTML]{0000FF} {\ul 0.840}}    & 0.404                                 \\
Wavelet2  & 0.162                                 & 0.251                                 & 0.226                                 & 0.224                                 & 0.108                                 & 0.210                                 & 1.664                                 & 0.987                                 & 1.177                                 & 0.594                                 & 8.949                                 & 1.840                                 & 0.843                                 & 0.406                                 \\
Chebyshev & 0.218                                 & 0.295                                 & 0.226                                 & 0.226                                 & 0.105                                 & 0.207                                 & 1.570                                 & 0.965                                 & 1.217                                 & 0.597                                 & 9.330                                 & 1.875                                 & 0.854                                 & 0.407                                 \\
Laguerre  & 0.167                                 & 0.255                                 & 0.233                                 & 0.230                                 & 0.111                                 & 0.214                                 & 1.659                                 & 0.984                                 & 1.353                                 & 0.651                                 & 9.302                                 & 1.890                                 & 0.868                                 & 0.420                                 \\
Legendre  & 0.161                                 & 0.250                                 & 0.243                                 & 0.235                                 & 0.109                                 & 0.213                                 & 1.685                                 & 0.995                                 & 1.177                                 & 0.603                                 & {\color[HTML]{0000FF} {\ul 8.550}}    & {\color[HTML]{0000FF} {\ul 1.798}}    & 0.841                                 & 0.404                                 \\
Identity  & 0.163                                 & 0.252                                 & 0.227                                 & 0.225                                 & 0.106                                 & 0.209                                 & {\color[HTML]{0000FF} {\ul 1.542}}    & {\color[HTML]{0000FF} {\ul 0.945}}    & 1.153    & 0.587                                 & 8.856                                 & 1.819                                 & 0.848                                 & 0.408                                 \\ \bottomrule
\end{tabular}
}
\end{table}

\subsection{Model analysis}

\paragraph{Transformation bases} \label{exp_base}

We replace OrthoTrans with several commonly used bases, including the Fourier basis, two wavelet bases (Haar and discrete Meyer wavelets), and three polynomial bases. As shown in Table~\ref{tab:base}, our method consistently outperforms all alternatives. Specifically, it achieves a 5.5\% reduction in MSE over the Fourier basis, and a 10.0\% improvement over the no-transformation baseline (i.e., the TF paradigm) on PEMS03.


\paragraph{OrthoTrans as a plug-in}
We further integrate OrthoTrans into three classic forecasters: iTransformer, PatchTST, and RLinear \citep{rlinear}. As shown in Table~\ref{tab:base_iTrans}, OrthoTrans yields average MSE improvements of 5.1\% and 10.1\% for iTransformer and PatchTST, respectively, highlighting the benefit of incorporating dataset-specific statistical information into the model design. This improvement can be attributed to the increased attention matrix rank introduced by OrthoTrans (see Appendix~\ref{orthotrans_imp_rank}), which indicates enlarged representation space and enhanced model capacity~\citep{flattentrans}.

\begin{wraptable}[14]{r}{0.5\textwidth}
\caption{Applying OrthoTrans (O.Trans) to iTransformer, PatchTST and RLinear. Average MSEs are reported. Full results are  in Table~\ref{tab:base_iTrans_full}.}
 \vspace{4pt}
\label{tab:base_iTrans}
\centering
\setlength{\tabcolsep}{3pt}
\renewcommand{\arraystretch}{1.0} 
{\fontsize{7}{9}\selectfont
\begin{tabular}{@{}ccccccc@{}}
\toprule
\multirow{2}{*}{Model}                  & \multicolumn{2}{c}{\begin{tabular}[c]{@{}c@{}}iTrans.\\  \citeyear{itransformer} \end{tabular}} & \multicolumn{2}{c}{\begin{tabular}[c]{@{}c@{}}PatchTST\\      \citeyear{patchtst} \end{tabular}} & \multicolumn{2}{c}{\begin{tabular}[c]{@{}c@{}}RLinear\\    \citeyear{rlinear} \end{tabular}} \\ \cmidrule(l){2-7} 
 & Van.                                  & +O.Trans                              & Van.                   & +O.Trans                                              & Van.                  & +O.Trans                                              \\ \midrule
ETTm1                    & 0.407 & {\color[HTML]{FF0000} \textbf{0.404}}                                & 0.387                  & {\color[HTML]{FF0000} \textbf{0.384}}                 & 0.414                 & {\color[HTML]{FF0000} \textbf{0.408}}                 \\
ECL                      & 0.178                                 & {\color[HTML]{FF0000} \textbf{0.171}} & 0.208                  & {\color[HTML]{FF0000} \textbf{0.181}}                 & 0.219                 & {\color[HTML]{FF0000} \textbf{0.214}}                 \\
PEMS03                   & 0.113                                 & {\color[HTML]{FF0000} \textbf{0.103}} & 0.180                  & {\color[HTML]{FF0000} \textbf{0.163}}                 & 0.495                 & {\color[HTML]{FF0000} \textbf{0.477}}                 \\
PEMS07                   & 0.101                                 & {\color[HTML]{FF0000} \textbf{0.085}} & 0.211                  & {\color[HTML]{FF0000} \textbf{0.147}}                 & 0.504                 & {\color[HTML]{FF0000} \textbf{0.485}}                 \\
Solar                    & 0.233                                 & {\color[HTML]{FF0000} \textbf{0.228}} & 0.270                  & {\color[HTML]{FF0000} \textbf{0.239}}                 & 0.369                 & {\color[HTML]{FF0000} \textbf{0.354}}                 \\
Weather                  & 0.258                                 & {\color[HTML]{FF0000} \textbf{0.252}} & 0.259                  & {\color[HTML]{FF0000} \textbf{0.246}}                 & 0.272                 & {\color[HTML]{FF0000} \textbf{0.269}}                 \\
METR-LA                  & 0.338                                 & {\color[HTML]{FF0000} \textbf{0.329}} & 0.335                  & {\color[HTML]{FF0000} \textbf{0.333}}                 & 0.342                 & {\color[HTML]{FF0000} \textbf{0.341}}                 \\ \bottomrule
\end{tabular}
}
\end{wraptable}

\paragraph{Representation learning}

To validate the rationality of our CSL and ISL designs, we conduct ablation studies by replacing or removing their core components—NormLin and (standard) linear layers. As shown in Table~\ref{tab:var_temp}, our design—applying NormLin along the variate dimension and standard linear layers along the temporal dimension—consistently achieves the best performance.
Notably, applying NormLin along the \textit{variate} dimension consistently outperforms its \textbf{self-attention} counterpart (last row), with reduced computational cost.  Furthermore, removing the ISL module (third-last row) results in a 6.2\% performance decline, highlighting the importance of updating temporal representations. Interestingly, on small-scale datasets with fewer variates (e.g., NASDAQ and ILI), the model with only temporal linear layers (third row) exhibits competitive performance, implying that NormLin is more beneficial when handling a larger number of variates.

%


\vspace{10pt}
\begin{table}[h]
\caption{Ablations on the CSL and ISL design. `NormLin (Temporal)' replaces the linear layer in Eq.~\ref{eq:isl} with the NormLin layer, and `Linear (Variate)' replaces the NormLin layer in Eq.~\ref{eq:csl} with a standard linear layer. `Attn.' denotes the self-attention mechanism. Results are averaged over four prediction lengths. Full results are shown in Table~\ref{tab:var_temp_full}.}
\label{tab:var_temp}
\centering
\setlength{\tabcolsep}{2.5pt}
\renewcommand{\arraystretch}{1.0} 
{\fontsize{7}{8}\selectfont
\begin{tabular}{@{}cccccccccccccccccc@{}}
\toprule
                          &                         & \multicolumn{2}{c}{ECL}                                                       & \multicolumn{2}{c}{Traffic}                                                   & \multicolumn{2}{c}{Solar}                                                     & \multicolumn{2}{c}{PEMS03}                                                    & \multicolumn{2}{c}{Weather}                                                   & \multicolumn{2}{c}{ETTm1}                                                     & \multicolumn{2}{c}{NASDAQ (S1)}                                               & \multicolumn{2}{c}{ILI (S2)}                                                  \\ \cmidrule(l){3-18} 
\multirow{-2}{*}{Variate} & \multirow{-2}{*}{Temp.} & MSE                                   & MAE                                   & MSE                                   & MAE                                   & MSE                                   & MAE                                   & MSE                                   & MAE                                   & MSE                                   & MAE                                   & MSE                                   & MAE                                   & MSE                                   & MAE                                   & MSE                                   & MAE                                   \\ \midrule
NormLin                   & Linear                  & {\color[HTML]{FF0000} \textbf{0.159}} & {\color[HTML]{FF0000} \textbf{0.248}} & {\color[HTML]{FF0000} \textbf{0.451}} & {\color[HTML]{FF0000} \textbf{0.247}} & {\color[HTML]{FF0000} \textbf{0.215}} & {\color[HTML]{FF0000} \textbf{0.217}} & {\color[HTML]{FF0000} \textbf{0.095}} & {\color[HTML]{FF0000} \textbf{0.199}} & {\color[HTML]{FF0000} \textbf{0.237}} & {\color[HTML]{FF0000} \textbf{0.260}} & {\color[HTML]{FF0000} \textbf{0.374}} & {\color[HTML]{FF0000} \textbf{0.377}} & {\color[HTML]{0000FF} {\ul 0.055}}  & {\color[HTML]{0000FF} {\ul 0.125}}  & {\color[HTML]{FF0000} \textbf{1.764}} & {\color[HTML]{FF0000} \textbf{0.802}} \\
Linear                    & Linear                  & 0.178                                 & 0.272                                 & 0.606                                 & 0.320                                 & 0.246                                 & 0.238                                 & 0.121                                 & 0.226                                 & {\color[HTML]{0000FF} {\ul 0.238}}    & {\color[HTML]{0000FF} {\ul 0.261}}    & {\color[HTML]{0000FF} {\ul 0.377}}    & 0.380                                 & 0.057                                 & 0.132                                 & 1.938                                 & 0.837                                 \\
w/o                       & Linear                  & 0.178                                 & 0.259                                 & 0.482                                 & 0.257                                 & 0.241                                 & 0.232                                 & 0.147                                 & 0.234                                 & 0.247                                 & 0.266                                 & 0.378                                 & {\color[HTML]{0000FF} {\ul 0.379}}    & {\color[HTML]{FF0000} \textbf{0.054}}    & {\color[HTML]{FF0000} \textbf{0.124}}    & {\color[HTML]{0000FF} {\ul 1.864}}    & {\color[HTML]{0000FF} {\ul 0.823}}    \\ \midrule
NormLin                   & NormLin                 & 0.169                                 & 0.257                                 & 0.460    & 0.275                                 & 0.252                                 & 0.240                                 & 0.112                                 & 0.214                                 & 0.239                                 & {\color[HTML]{0000FF} {\ul 0.261}}                                & 0.381                                 & 0.382                                 & {\color[HTML]{0000FF} {\ul 0.055}}                                 & 0.126                                 & 1.947                                 & 0.836                                 \\
Linear                    & NormLin                 & 0.183                                 & 0.276                                 & 0.578                                 & 0.339                                 & 0.262                                 & 0.254                                 & 0.143                                 & 0.246                                 & 0.240                                 & {\color[HTML]{0000FF} {\ul 0.261}}                                  & 0.383                                 & 0.384                                 & 0.057                                 & 0.133                                 & 2.037                                 & 0.867                                 \\
w/o                       & NormLin                 & 0.185                                 & 0.266                                 & 0.493                                 & 0.290                                 & 0.283                                 & 0.262                                 & 0.182                                 & 0.269                                 & 0.246                                 & 0.265                                 & 0.384                                 & 0.385                                 & {\color[HTML]{0000FF} {\ul 0.055}}                                 & {\color[HTML]{0000FF} {\ul 0.125}}                                 & 2.093                                 & 0.874                                 \\ \midrule
NormLin                   & w/o                     & 0.169                                 & 0.257                                 & 0.460    & 0.275                                 & 0.253                                 & 0.241                                 & 0.114                                 & 0.215                                 & 0.239                                 & {\color[HTML]{0000FF} {\ul 0.261}}                                 & 0.380                                 & 0.382                                 & {\color[HTML]{0000FF} {\ul 0.055}}                                & 0.126                                 & 1.940                                 & 0.837                                 \\
Linear                    & w/o                     & 0.183                                 & 0.276                                 & 0.591                                 & 0.341                                 & 0.262                                 & 0.254                                 & 0.142                                 & 0.246                                 & 0.240                                 & 0.262                                 & 0.384                                 & 0.384                                 & 0.057                                 & 0.132                                 & 2.073                                 & 0.874                                 \\ \midrule
Attn.                     & Linear                  & {\color[HTML]{0000FF} {\ul 0.166}}    & {\color[HTML]{0000FF} {\ul 0.255}}    & {\color[HTML]{0000FF} {\ul 0.457}}                                & {\color[HTML]{0000FF} {\ul 0.251}}    & {\color[HTML]{0000FF} {\ul 0.220}}    & {\color[HTML]{0000FF} {\ul 0.221}}    & {\color[HTML]{0000FF} {\ul 0.097}}    & {\color[HTML]{0000FF} {\ul 0.202}}    & 0.244                                 & 0.265                                 & 0.391                                 & 0.389                                 & 0.056                                 & 0.126                                 & 2.022                                 & 0.847                                 \\ \bottomrule
\end{tabular}
}
\end{table}

\paragraph{OLinear-C} \label{OLinear-c}

\begin{wrapfigure}[12]{r}{0.4\textwidth}
  \begin{center}
    \includegraphics[width=0.4\textwidth]{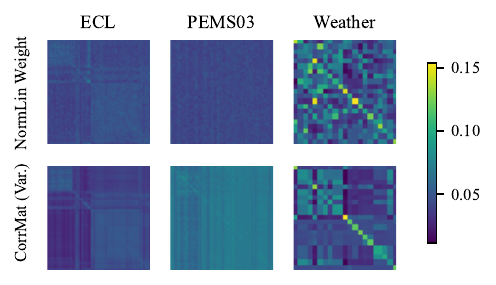}
  \end{center}
   \caption{The learned NormLin weights exhibit similarity to the multivariate correlation matrix (after \texttt{Softmax}).}
  \label{fig:norm_weight_vs_corr}
\end{wrapfigure}

As shown in Figure~\ref{fig:norm_weight_vs_corr}, the learned weight matrix resemble $\mathrm{Softmax} \left ( \mathrm{CorrMat} _v \right )$, where $\mathrm{CorrMat} _v$ is the Pearson correlation matrix across variates. Motivated by this, we replace the learnable weights in NormLin with the pre-computed $\mathrm{Softmax} \left ( \mathrm{CorrMat} _v \right )$, resulting in a simplified variant: $\mathrm{NormLin}_c\left ( \mathbf{x}  \right )  = \mathrm{Softmax} \left ( \mathrm{CorrMat} _v \right ) \  \mathbf{x} $. We refer to the model with this NormLin variant as \textbf{OLinear-C}.

\begin{table}[t]
\caption{Performance comparison of OLinear and OLinear-C. Full results are shown in Table~\ref{tab:ortho-c-full}. }
\label{tab:OLinear-c_short}
\centering
\setlength{\tabcolsep}{2.2pt}
\renewcommand{\arraystretch}{1.0} 
{\fontsize{7}{9}\selectfont
\begin{tabular}{@{}ccccccccccccccccccc@{}}
\toprule
Dataset       & \multicolumn{2}{c}{ECL}                                                       & \multicolumn{2}{c}{Traffic}                                                   & \multicolumn{2}{c}{ETT}                                                       & \multicolumn{2}{c}{Solar}                                                     & \multicolumn{2}{c}{PEMS}                                                      & \multicolumn{2}{c}{CarSales}                                                  & \multicolumn{2}{c}{ILI}                                                       & \multicolumn{2}{c}{COVID-19}                                                  & \multicolumn{2}{c}{Unemp}                                                     \\ \midrule
Metric        & MSE                                   & MAE                                   & MSE                                   & MAE                                   & MSE                                   & MAE                                   & MSE                                   & MAE                                   & MSE                                   & MAE                                   & MSE                                   & MAE                                   & MSE                                   & MAE                                   & MSE                                   & MAE                                   & MSE                                   & MAE                                   \\ \midrule
OLinear   & {\color[HTML]{FF0000} \textbf{0.159}} & {\color[HTML]{FF0000} \textbf{0.248}} & {\color[HTML]{FF0000} \textbf{0.451}} & {\color[HTML]{FF0000} \textbf{0.247}} & {\color[HTML]{FF0000} \textbf{0.359}} & {\color[HTML]{FF0000} \textbf{0.376}} & {\color[HTML]{FF0000} \textbf{0.215}} & {\color[HTML]{FF0000} \textbf{0.217}} & {\color[HTML]{FF0000} \textbf{0.094}} & {\color[HTML]{FF0000} \textbf{0.187}} & {\color[HTML]{FF0000} \textbf{0.330}} & {\color[HTML]{FF0000} \textbf{0.305}} & {\color[HTML]{FF0000} \textbf{1.429}} & {\color[HTML]{FF0000} \textbf{0.690}} & {\color[HTML]{FF0000} \textbf{5.187}} & {\color[HTML]{FF0000} \textbf{1.211}} & {\color[HTML]{FF0000} \textbf{0.729}} & {\color[HTML]{FF0000} \textbf{0.461}} \\
OLinear-C & 0.161                                 & 0.249                                 & {\color[HTML]{FF0000} \textbf{0.451}} & {\color[HTML]{FF0000} \textbf{0.247}} & {\color[HTML]{FF0000} \textbf{0.359}} & {\color[HTML]{FF0000} \textbf{0.376}} & {\color[HTML]{FF0000} \textbf{0.215}} & {\color[HTML]{FF0000} \textbf{0.217}} & {\color[HTML]{FF0000} \textbf{0.094}} & {\color[HTML]{FF0000} \textbf{0.187}} & {\color[HTML]{FF0000} \textbf{0.330}} & {\color[HTML]{FF0000} \textbf{0.305}} & 1.463                                 & 0.698                                 & 5.346                                 & 1.247                                 & 0.766                                 & 0.474                                 \\ \bottomrule
\end{tabular}
}
\end{table}

\setlength{\abovecaptionskip}{0pt} 
\setlength{\intextsep}{0pt} 

Table~\ref{tab:OLinear-c_short} shows that OLinear-C performs comparably to OLinear with reduced learnable parameters. Comprehensive evaluation of OLinear-C and the variants of $\mathrm{NormLin}_c$ are discussed in Appendix~\ref{append_orthoc}.

\subsection{Generality and scalability of the NormLin module}

\begin{table}[t]
  \centering
  \begin{minipage}[t]{0.48\textwidth}
    \centering
    \caption{Comparison of the NormLin module with state-of-the-art attention variants. Average MSEs are reported. Full results are in Table~\ref{tab:normlin_vs_attn_full}.}
    \label{tab:normlin_vs_attn}
    \setlength{\tabcolsep}{1pt}
    \renewcommand{\arraystretch}{1.0} 
    {\fontsize{7}{9}\selectfont
    \begin{tabular}{@{}cccccccc@{}}
\toprule
Dataset & \begin{tabular}[c]{@{}c@{}}NormLin\\      (Ours)\end{tabular} & \begin{tabular}[c]{@{}c@{}}Trans.\\      \citeyear{transformer} \end{tabular} & 
\begin{tabular}[c]{@{}c@{}}Refm.\\ \citeyear{reformer} \end{tabular} & 
\begin{tabular}[c]{@{}c@{}}Flowfm.\\   \citeyear{flowformer}   \end{tabular} & 
\begin{tabular}[c]{@{}c@{}}Flatten\\  \citeyear{flattentrans} \end{tabular} & 
\begin{tabular}[c]{@{}c@{}}Mamba\\ \citeyear{mamba}  \end{tabular} & \begin{tabular}[c]{@{}c@{}}E.Attn.\\   \citeyear{freeformer} \end{tabular} \\ \midrule
ECL     & {\color[HTML]{FF0000} \textbf{0.159}}                         & 0.166                                                    & 0.167                                                   & 0.165                                                     & {\color[HTML]{0000FF} {\ul 0.164}}                        & 0.176                                                   & {\color[HTML]{FF0000} \textbf{0.159}}                       \\
Traffic & {\color[HTML]{0000FF} {\ul 0.451}}                            & 0.457                                                    & 0.459                                                   & 0.460                                                     & 0.464                                                     & 0.456                                                   & {\color[HTML]{FF0000} \textbf{0.439}}                       \\
PEMS03  & {\color[HTML]{FF0000} \textbf{0.095}}                         & 0.097                                                    & {\color[HTML]{0000FF} {\ul 0.096}}                      & 0.099                                                     & 0.101                                                     & 0.104                                                   & 0.097                                                       \\
Weather & {\color[HTML]{FF0000} \textbf{0.237}}                         & 0.244                                                    & {\color[HTML]{0000FF} {\ul 0.241}}                      & 0.242                                                     & 0.246                                                     & 0.242                                                   & {\color[HTML]{0000FF} {\ul 0.241}}                          \\
Solar   & {\color[HTML]{FF0000} \textbf{0.215}}                         & 0.223                                                    & {\color[HTML]{0000FF} {\ul 0.216}}                      & 0.222                                                     & 0.231                                                     & 0.228                                                   & 0.217                                                       \\
ILI     & {\color[HTML]{FF0000} \textbf{1.764}}                         & 2.022                                                    & {\color[HTML]{0000FF} {\ul 1.821}}                      & 1.881                                                     & 2.134                                                     & 1.950                                                   & 1.878                                                       \\
NASDAQ  & {\color[HTML]{FF0000} \textbf{0.055}}                         & {\color[HTML]{0000FF} {\ul 0.056}}                       & {\color[HTML]{FF0000} \textbf{0.055}}                   & {\color[HTML]{FF0000} \textbf{0.055}}                     & 0.057                                                     & {\color[HTML]{FF0000} \textbf{0.055}}                   & {\color[HTML]{FF0000} \textbf{0.055}}                       \\ \bottomrule
\end{tabular}
    }
  \end{minipage}
  \hfill
  \begin{minipage}[t]{0.48\textwidth}
    \centering
    \caption{Applying NormLin (N.Lin) to state-of-the-art Transformer-based forecasters. Average MSEs are reported. Full results are in Table~\ref{tab:normlin_itrans_full}.}
    \label{tab:normlin_itrans}
    \setlength{\tabcolsep}{1.5pt}
    \renewcommand{\arraystretch}{0.93} 
    {\fontsize{7}{9}\selectfont
    \begin{tabular}{@{}ccccccccc@{}}
\toprule
                          & \multicolumn{2}{c}{iTrans.}                   & \multicolumn{2}{c}{PatchTST}                  & \multicolumn{2}{c}{Leddam}                                                    & \multicolumn{2}{c}{Fredformer}                                                \\ \cmidrule(l){2-9} 
\multirow{-2}{*}{Dataset} & Van.  & N.Lin                                 & Van.  & N.Lin                                 & Van.                                  & N.Lin                                 & Van.                                  & N.Lin                                 \\ \midrule
ETTm1                     & 0.407 & {\color[HTML]{FF0000} \textbf{0.388}} & 0.387 & {\color[HTML]{FF0000} \textbf{0.379}} & 0.386                                 & {\color[HTML]{FF0000} \textbf{0.381}} & 0.384                                 & {\color[HTML]{FF0000} \textbf{0.381}} \\
ECL                       & 0.178 & {\color[HTML]{FF0000} \textbf{0.166}} & 0.208 & {\color[HTML]{FF0000} \textbf{0.181}} & 0.169                                 & {\color[HTML]{FF0000} \textbf{0.165}} & 0.176                                 & {\color[HTML]{FF0000} \textbf{0.169}} \\
PEMS03                    & 0.113 & {\color[HTML]{FF0000} \textbf{0.102}} & 0.180 & {\color[HTML]{FF0000} \textbf{0.146}} & 0.107                                 & {\color[HTML]{FF0000} \textbf{0.103}} & 0.134                                 & {\color[HTML]{FF0000} \textbf{0.108}} \\
PEMS07                    & 0.101 & {\color[HTML]{FF0000} \textbf{0.086}} & 0.211 & {\color[HTML]{FF0000} \textbf{0.168}} & 0.084                                 & {\color[HTML]{FF0000} \textbf{0.082}} & 0.121                                 & {\color[HTML]{FF0000} \textbf{0.096}} \\
Solar                     & 0.233 & {\color[HTML]{FF0000} \textbf{0.226}} & 0.270 & {\color[HTML]{FF0000} \textbf{0.237}} & 0.230                                 & {\color[HTML]{FF0000} \textbf{0.222}} & {\color[HTML]{FF0000} \textbf{0.226}} & {\color[HTML]{FF0000} \textbf{0.226}}                                \\
Weather                   & 0.258 & {\color[HTML]{FF0000} \textbf{0.245}} & 0.259 & {\color[HTML]{FF0000} \textbf{0.245}} & {\color[HTML]{FF0000} \textbf{0.242}} & {\color[HTML]{FF0000} \textbf{0.242}} & 0.246                                 & {\color[HTML]{FF0000} \textbf{0.240}} \\
METR-LA                   & 0.338 & {\color[HTML]{FF0000} \textbf{0.328}} & 0.335 & {\color[HTML]{FF0000} \textbf{0.333}} & 0.327                                 & {\color[HTML]{FF0000} \textbf{0.320}} & 0.336                                 & {\color[HTML]{FF0000} \textbf{0.329}} \\ \bottomrule
\end{tabular}
    }
  \end{minipage}
\end{table}

In our CSL module, we employ the NormLin module—comprising the NormLin layer and its associated pre- and post-linear layers (Equation~\ref{eq:csl})—to capture multivariate correlations.
Despite its simple architecture, the NormLin module demonstrates strong capability, generality, and scalability. Specifically, it consistently outperforms the classic multi-head self-attention mechanism and its variants, offering a compelling alternative for token dependency modeling.

\paragraph{Comparison with attention mechanisms}
To assess its effectiveness, we compare NormLin with classic attention variants, such as Reformer \citep{reformer}, Flowformer \citep{flowformer}, FLatten \citep{flashattention}, and Mamba \citep{mamba}. As shown in Table~\ref{tab:normlin_vs_attn}, NormLin consistently outperforms these methods, demonstrating that the simple normalized weight matrix performs better than the complex query-key interaction mechanism for time series forecasting.

\begin{wraptable}[13]{r}{0.45\textwidth}
\caption{Applying NormLin to the large time series model Timer. For fine-tuning, 5\% samples of trainset are used. MSEs are reported.}
 \vspace{4pt}
\label{tab:timer_normlin}
\centering
\setlength{\tabcolsep}{6pt}
\renewcommand{\arraystretch}{1.0} 
{\fontsize{8}{9}\selectfont
\begin{tabular}{@{}ccccc@{}}
\toprule
\multirow{2}{*}{Dataset}      & \multicolumn{2}{c}{Fine-tuning}               & \multicolumn{2}{c}{Zero-shot}                                                 \\ \cmidrule(l){2-5} 
\ & Van.  & NormLin                               & Van.                                  & NormLin                               \\ \midrule
ETTh1                     & 0.362 & {\color[HTML]{FF0000} \textbf{0.360}} & 0.438                                 & {\color[HTML]{FF0000} \textbf{0.404}} \\
ETTh2                     & 0.280 & {\color[HTML]{FF0000} \textbf{0.269}} & 0.314                                 & {\color[HTML]{FF0000} \textbf{0.274}} \\
ETTm1                     & 0.321 & {\color[HTML]{FF0000} \textbf{0.309}} & 0.690                                 & {\color[HTML]{FF0000} \textbf{0.632}} \\
ETTm2                     & 0.176 & {\color[HTML]{FF0000} \textbf{0.172}} & 0.213                                 & {\color[HTML]{FF0000} \textbf{0.212}} \\
ECL                       & 0.132 & {\color[HTML]{FF0000} \textbf{0.130}} & 0.192                                 & {\color[HTML]{FF0000} \textbf{0.183}} \\
Traffic                   & 0.361 & {\color[HTML]{FF0000} \textbf{0.353}} & {\color[HTML]{FF0000} \textbf{0.458}} & 0.462                                 \\
Weather                   & 0.151 & {\color[HTML]{FF0000} \textbf{0.149}} & 0.181                                 & {\color[HTML]{FF0000} \textbf{0.174}} \\ \bottomrule
\end{tabular}
}
\end{wraptable}

\paragraph{NormLin as a plug-in}
To demonstrate the generalizability of the NormLin module, we replace self-attention with NormLin in Transformer-based forecasters. As shown in Table~\ref{tab:normlin_itrans}, this substitution leads to notable MSE improvements—6.7\% for iTransformer and 10.3\% for PatchTST—validating its plug-and-play effectiveness. These results also validate NormLin's capability to model dependencies across multiple token types (i.e., variate, temporal, and frequency-domain), highlighting  its potential as a universal token dependency learner for time series forecasting. Furthermore, the NormLin module consistently improves both training and inference efficiency across these models (see Table~\ref{tab:GPU_normlin_plugin_compare}); for example, it boosts iTransformer's inference efficiency by an average of 53\%.

\paragraph{Scalability of the NormLin module}

Decoder-only Transformers have become the de facto architecture choice for large time series models \citep{timer,chronos,timesfm}.
We take Timer \citep{timer} as a representative example  and replace self-attention in Timer with NormLin.
To align with Timer’s decoder-only structure, we apply a causal mask by zeroing out the upper triangular part of NormLin’s weight matrix.
The modified model is pre-trained on the UTSD dataset \citep{timer}, which spans seven domains and contains up to 1 billion time points.
As shown in Table~\ref{tab:timer_normlin}, NormLin improves performance in both zero-shot and fine-tuning scenarios.
For example, the zero-shot MSE on ETTh2 and ETTm1 is reduced by 12.7\% and 8.4\%, respectively.
These results demonstrate that the NormLin module adapts well to decoder-only architectures and scales effectively to large-scale pre-training scenarios.

\section{Conclusion}

In this work, we present OLinear, a simple yet effective linear-based forecaster that achieves state-of-the-art performance,  built on two core components: (1) OrthoTrans, an orthogonal transformation that decorrelates temporal dependencies to facilitate better encoding and forecasting, and (2) the NormLin module, a powerful and general-purpose token dependency learner. Notably, both modules consistently improve existing forecasters when used as plug-ins. In future work, we plan to apply the NormLin module to large time series models and broader time series analysis tasks.




\nocite{linear_softmax,freeformer}
\clearpage
\bibliographystyle{unsrtnat}
\bibliography{neurips2025}

\begin{thebibliography}{52}
\providecommand{\natexlab}[1]{#1}
\providecommand{\url}[1]{\texttt{#1}}
\expandafter\ifx\csname urlstyle\endcsname\relax
  \providecommand{\doi}[1]{doi: #1}\else
  \providecommand{\doi}{doi: \begingroup \urlstyle{rm}\Url}\fi

\bibitem[Wu et~al.(2023{\natexlab{a}})Wu, Zhou, Long, and Wang]{nature_weather}
Haixu Wu, Hang Zhou, Mingsheng Long, and Jianmin Wang.
\newblock Interpretable weather forecasting for worldwide stations with a unified deep model.
\newblock \emph{Nature Machine Intelligence}, 5\penalty0 (6):\penalty0 602--611, 2023{\natexlab{a}}.

\bibitem[Ma et~al.(2021)Ma, Dai, and Zhou]{traffic}
Changxi Ma, Guowen Dai, and Jibiao Zhou.
\newblock Short-term traffic flow prediction for urban road sections based on time series analysis and lstm\_bilstm method.
\newblock \emph{IEEE Transactions on Intelligent Transportation Systems}, 23\penalty0 (6):\penalty0 5615--5624, 2021.

\bibitem[Zhou et~al.(2021)Zhou, Zhang, Peng, Zhang, Li, Xiong, and Zhang]{informer}
Haoyi Zhou, Shanghang Zhang, Jieqi Peng, Shuai Zhang, Jianxin Li, Hui Xiong, and Wancai Zhang.
\newblock Informer: Beyond efficient transformer for long sequence time-series forecasting.
\newblock In \emph{AAAI}, volume~35, pages 11106--11115, 2021.

\bibitem[Chen et~al.(2023)Chen, Ma, Li, Wang, and Li]{survey4}
Zonglei Chen, Minbo Ma, Tianrui Li, Hongjun Wang, and Chongshou Li.
\newblock Long sequence time-series forecasting with deep learning: A survey.
\newblock \emph{Information Fusion}, 97:\penalty0 101819, 2023.

\bibitem[Wang et~al.(2025{\natexlab{a}})Wang, Li, Shi, Ye, Mo, Lin, Ju, Chu, and Jin]{timemixer++}
Shiyu Wang, Jiawei Li, Xiaoming Shi, Zhou Ye, Baichuan Mo, Wenze Lin, Shengtong Ju, Zhixuan Chu, and Ming Jin.
\newblock Timemixer++: A general time series pattern machine for universal predictive analysis.
\newblock In \emph{ICLR}, 2025{\natexlab{a}}.

\bibitem[Liu et~al.(2024{\natexlab{a}})Liu, Hu, Zhang, Wu, Wang, Ma, and Long]{itransformer}
Yong Liu, Tengge Hu, Haoran Zhang, Haixu Wu, Shiyu Wang, Lintao Ma, and Mingsheng Long.
\newblock itransformer: Inverted transformers are effective for time series forecasting.
\newblock In \emph{ICLR}, 2024{\natexlab{a}}.

\bibitem[Nie et~al.(2023)Nie, Nguyen, Sinthong, and Kalagnanam]{patchtst}
Yuqi Nie, Nam~H. Nguyen, Phanwadee Sinthong, and Jayant Kalagnanam.
\newblock A time series is worth 64 words: Long-term forecasting with transformers.
\newblock \emph{ICLR}, 2023.

\bibitem[Wang et~al.(2024{\natexlab{a}})Wang, Wu, Shi, Hu, Luo, Ma, Zhang, and Zhou]{timemixer}
Shiyu Wang, Haixu Wu, Xiaoming Shi, Tengge Hu, Huakun Luo, Lintao Ma, James~Y. Zhang, and Jun Zhou.
\newblock Timemixer: Decomposable multiscale mixing for time series forecasting.
\newblock In \emph{ICLR}, 2024{\natexlab{a}}.

\bibitem[Yi et~al.(2023)Yi, Zhang, Fan, Wang, Wang, He, An, Lian, Cao, and Niu]{frets}
Kun Yi, Qi~Zhang, Wei Fan, Shoujin Wang, Pengyang Wang, Hui He, Ning An, Defu Lian, Longbing Cao, and Zhendong Niu.
\newblock Frequency-domain mlps are more effective learners in time series forecasting.
\newblock In \emph{NeurIPS}, 2023.

\bibitem[Yue et~al.(2025)Yue, Liu, Ying, Xing, Guo, and Shi]{freeformer}
Wenzhen Yue, Yong Liu, Xianghua Ying, Bowei Xing, Ruohao Guo, and Ji~Shi.
\newblock Freeformer: Frequency enhanced transformer for multivariate time series forecasting.
\newblock \emph{arXiv preprint arXiv:2501.13989}, 2025.

\bibitem[Yi et~al.(2024{\natexlab{a}})Yi, Fei, Zhang, He, Hao, Lian, and Fan]{filternet}
Kun Yi, Jingru Fei, Qi~Zhang, Hui He, Shufeng Hao, Defu Lian, and Wei Fan.
\newblock Filternet: Harnessing frequency filters for time series forecasting.
\newblock \emph{Advances in Neural Information Processing Systems}, 37:\penalty0 55115--55140, 2024{\natexlab{a}}.

\bibitem[Masserano et~al.(2024)Masserano, Ansari, Han, Zhang, Faloutsos, Mahoney, Wilson, Park, Rangapuram, Maddix, et~al.]{wavelet_token}
Luca Masserano, Abdul~Fatir Ansari, Boran Han, Xiyuan Zhang, Christos Faloutsos, Michael~W Mahoney, Andrew~Gordon Wilson, Youngsuk Park, Syama Rangapuram, Danielle~C Maddix, et~al.
\newblock Enhancing foundation models for time series forecasting via wavelet-based tokenization.
\newblock \emph{arXiv preprint arXiv:2412.05244}, 2024.

\bibitem[Gray and Davisson(2004)]{statistical}
Robert~M Gray and Lee~D Davisson.
\newblock \emph{An introduction to statistical signal processing}.
\newblock Cambridge University Press, 2004.

\bibitem[Yu et~al.(2024)Yu, Zou, Hu, Aviles-Rivero, Qin, and Wang]{Leddam_icml}
Guoqi Yu, Jing Zou, Xiaowei Hu, Angelica~I Aviles-Rivero, Jing Qin, and Shujun Wang.
\newblock Revitalizing multivariate time series forecasting: Learnable decomposition with inter-series dependencies and intra-series variations modeling.
\newblock In \emph{ICML}, 2024.

\bibitem[Wang et~al.(2024{\natexlab{b}})Wang, Zhou, Wen, Gao, Ding, and Jin]{card}
Xue Wang, Tian Zhou, Qingsong Wen, Jinyang Gao, Bolin Ding, and Rong Jin.
\newblock Card: Channel aligned robust blend transformer for time series forecasting.
\newblock In \emph{ICLR}, 2024{\natexlab{b}}.

\bibitem[Wu et~al.(2023{\natexlab{b}})Wu, Hu, Liu, Zhou, Wang, and Long]{timesnet}
Haixu Wu, Tengge Hu, Yong Liu, Hang Zhou, Jianmin Wang, and Mingsheng Long.
\newblock Timesnet: Temporal 2d-variation modeling for general time series analysis.
\newblock In \emph{ICLR}, 2023{\natexlab{b}}.

\bibitem[Zeng et~al.(2023)Zeng, Chen, Zhang, and Xu]{linear}
Ailing Zeng, Muxi Chen, Lei Zhang, and Qiang Xu.
\newblock Are transformers effective for time series forecasting?
\newblock In \emph{AAAI}, volume~37, pages 11121--11128, 2023.

\bibitem[Wang et~al.(2025{\natexlab{b}})Wang, Pan, Shen, Chen, Yang, Yang, Zhang, Liu, Li, and Tao]{fredf}
Hao Wang, Lichen Pan, Yuan Shen, Zhichao Chen, Degui Yang, Yifei Yang, Sen Zhang, Xinggao Liu, Haoxuan Li, and Dacheng Tao.
\newblock Fredf: Learning to forecast in the frequency domain.
\newblock In \emph{ICLR}, 2025{\natexlab{b}}.

\bibitem[Xu et~al.(2024)Xu, Zeng, and Xu]{fits}
Zhijian Xu, Ailing Zeng, and Qiang Xu.
\newblock Fits: Modeling time series with $10 k$ parameters.
\newblock In \emph{ICLR}, 2024.

\bibitem[Jolliffe(2002)]{pca}
Ian~T Jolliffe.
\newblock \emph{Principal component analysis for special types of data}.
\newblock Springer, 2002.

\bibitem[Luo and Wang(2024)]{moderntcn}
Donghao Luo and Xue Wang.
\newblock Moderntcn: A modern pure convolution structure for general time series analysis.
\newblock In \emph{ICLR}, 2024.

\bibitem[Lai et~al.(2018)Lai, Chang, Yang, and Liu]{rnn}
Guokun Lai, Wei-Cheng Chang, Yiming Yang, and Hanxiao Liu.
\newblock Modeling long-and short-term temporal patterns with deep neural networks.
\newblock In \emph{SIGIR}, pages 95--104, 2018.

\bibitem[Rangapuram et~al.(2018)Rangapuram, Seeger, Gasthaus, Stella, Wang, and Januschowski]{rnn_nips2018}
Syama~Sundar Rangapuram, Matthias~W Seeger, Jan Gasthaus, Lorenzo Stella, Yuyang Wang, and Tim Januschowski.
\newblock Deep state space models for time series forecasting.
\newblock \emph{NeurIPS}, 31, 2018.

\bibitem[Huang et~al.(2023)Huang, Shen, Zhang, Ding, Wang, Zhou, and Wang]{crossgnn}
Qihe Huang, Lei Shen, Ruixin Zhang, Shouhong Ding, Binwu Wang, Zhengyang Zhou, and Yang Wang.
\newblock Crossgnn: Confronting noisy multivariate time series via cross interaction refinement.
\newblock \emph{NeurIPS}, 36:\penalty0 46885--46902, 2023.

\bibitem[Yi et~al.(2024{\natexlab{b}})Yi, Zhang, Fan, He, Hu, Wang, An, Cao, and Niu]{fouriergnn}
Kun Yi, Qi~Zhang, Wei Fan, Hui He, Liang Hu, Pengyang Wang, Ning An, Longbing Cao, and Zhendong Niu.
\newblock Fouriergnn: Rethinking multivariate time series forecasting from a pure graph perspective.
\newblock \emph{Advances in Neural Information Processing Systems}, 36, 2024{\natexlab{b}}.

\bibitem[Kim et~al.(2021)Kim, Kim, Tae, Park, Choi, and Choo]{revin}
Taesung Kim, Jinhee Kim, Yunwon Tae, Cheonbok Park, Jang-Ho Choi, and Jaegul Choo.
\newblock Reversible instance normalization for accurate time-series forecasting against distribution shift.
\newblock In \emph{ICLR}, 2021.

\bibitem[Horn and Johnson(2012)]{matrix}
Roger~A Horn and Charles~R Johnson.
\newblock \emph{Matrix analysis}.
\newblock Cambridge university press, 2012.

\bibitem[Bishop and Nasrabadi(2006)]{prml_book}
Christopher~M Bishop and Nasser~M Nasrabadi.
\newblock \emph{Pattern recognition and machine learning}, volume~4.
\newblock Springer, 2006.

\bibitem[Vaswani et~al.(2017)Vaswani, Shazeer, Parmar, Uszkoreit, Jones, Gomez, Kaiser, and Polosukhin]{transformer}
Ashish Vaswani, Noam Shazeer, Niki Parmar, Jakob Uszkoreit, Llion Jones, Aidan~N Gomez, {\L}ukasz Kaiser, and Illia Polosukhin.
\newblock Attention is all you need.
\newblock \emph{NIPS}, 30, 2017.

\bibitem[Li et~al.(2023)Li, Qi, Li, and Xu]{rlinear}
Zhe Li, Shiyi Qi, Yiduo Li, and Zenglin Xu.
\newblock Revisiting long-term time series forecasting: An investigation on linear mapping.
\newblock \emph{arXiv preprint arXiv:2305.10721}, 2023.

\bibitem[Das et~al.(2023)Das, Kong, Leach, Sen, and Yu]{tide}
Abhimanyu Das, Weihao Kong, Andrew Leach, Rajat Sen, and Rose Yu.
\newblock Long-term forecasting with tide: Time-series dense encoder.
\newblock \emph{arXiv preprint arXiv:2304.08424}, 2023.

\bibitem[Han et~al.(2023)Han, Pan, Han, Song, and Huang]{flattentrans}
Dongchen Han, Xuran Pan, Yizeng Han, Shiji Song, and Gao Huang.
\newblock Flatten transformer: Vision transformer using focused linear attention.
\newblock In \emph{ICCV}, pages 5961--5971, 2023.

\bibitem[Piao et~al.(2024)Piao, Chen, Murayama, Matsubara, and Sakurai]{fredformer}
Xihao Piao, Zheng Chen, Taichi Murayama, Yasuko Matsubara, and Yasushi Sakurai.
\newblock Fredformer: Frequency debiased transformer for time series forecasting.
\newblock In \emph{SIGKDD}, 2024.

\bibitem[Kitaev et~al.(2020)Kitaev, Kaiser, and Levskaya]{reformer}
Nikita Kitaev, {\L}ukasz Kaiser, and Anselm Levskaya.
\newblock Reformer: The efficient transformer.
\newblock \emph{ICLR}, 2020.

\bibitem[Wu et~al.(2022)Wu, Wu, Xu, Wang, and Long]{flowformer}
Haixu Wu, Jialong Wu, Jiehui Xu, Jianmin Wang, and Mingsheng Long.
\newblock Flowformer: Linearizing transformers with conservation flows.
\newblock In \emph{ICML}, 2022.

\bibitem[Gu and Dao(2023)]{mamba}
Albert Gu and Tri Dao.
\newblock Mamba: Linear-time sequence modeling with selective state spaces.
\newblock \emph{arXiv preprint arXiv:2312.00752}, 2023.

\bibitem[Dao et~al.(2022)Dao, Fu, Ermon, Rudra, and R{\'e}]{flashattention}
Tri Dao, Dan Fu, Stefano Ermon, Atri Rudra, and Christopher R{\'e}.
\newblock Flashattention: Fast and memory-efficient exact attention with io-awareness.
\newblock \emph{NeurIPS}, 35:\penalty0 16344--16359, 2022.

\bibitem[Liu et~al.(2024{\natexlab{b}})Liu, Zhang, Li, Huang, Wang, and Long]{timer}
Yong Liu, Haoran Zhang, Chenyu Li, Xiangdong Huang, Jianmin Wang, and Mingsheng Long.
\newblock Timer: Transformers for time series analysis at scale.
\newblock In \emph{ICML}, 2024{\natexlab{b}}.

\bibitem[Ansari et~al.(2024)Ansari, Stella, Turkmen, Zhang, Mercado, Shen, Shchur, Rangapuram, Arango, Kapoor, et~al.]{chronos}
Abdul~Fatir Ansari, Lorenzo Stella, Caner Turkmen, Xiyuan Zhang, Pedro Mercado, Huibin Shen, Oleksandr Shchur, Syama~Sundar Rangapuram, Sebastian~Pineda Arango, Shubham Kapoor, et~al.
\newblock Chronos: Learning the language of time series.
\newblock \emph{arXiv preprint arXiv:2403.07815}, 2024.

\bibitem[Das et~al.(2024)Das, Kong, Sen, and Zhou]{timesfm}
Abhimanyu Das, Weihao Kong, Rajat Sen, and Yichen Zhou.
\newblock A decoder-only foundation model for time-series forecasting.
\newblock In \emph{Forty-first International Conference on Machine Learning}, 2024.

\bibitem[Yue et~al.(2024)Yue, Ying, Guo, Chen, Zhu, Shi, Xing, and Chen]{linear_softmax}
Wenzhen Yue, Xianghua Ying, Ruohao Guo, Dongdong Chen, Yuqing Zhu, Ji~Shi, Bowei Xing, and Taiyan Chen.
\newblock Sub-adjacent transformer: Improving time series anomaly detection with reconstruction error from sub-adjacent neighborhoods.
\newblock In \emph{IJCAI}, 2024.

\bibitem[Surya~Duvvuri and Dhillon(2024)]{laser}
Sai Surya~Duvvuri and Inderjit~S Dhillon.
\newblock Laser: Attention with exponential transformation.
\newblock \emph{arXiv e-prints}, pages arXiv--2411, 2024.

\bibitem[Wu et~al.(2021)Wu, Xu, Wang, and Long]{autoformer}
Haixu Wu, Jiehui Xu, Jianmin Wang, and Mingsheng Long.
\newblock Autoformer: Decomposition transformers with auto-correlation for long-term series forecasting.
\newblock \emph{NeurIPS}, 34:\penalty0 22419--22430, 2021.

\bibitem[Liu et~al.(2022)Liu, Zeng, Chen, Xu, Lai, Ma, and Xu]{scinet}
Minhao Liu, Ailing Zeng, Muxi Chen, Zhijian Xu, Qiuxia Lai, Lingna Ma, and Qiang Xu.
\newblock Scinet: Time series modeling and forecasting with sample convolution and interaction.
\newblock \emph{NeurIPS}, 35:\penalty0 5816--5828, 2022.

\bibitem[Chen et~al.(2022)Chen, Segovia-Dominguez, Coskunuzer, and Gel]{chen2022tamp}
Yuzhou Chen, Ignacio Segovia-Dominguez, Baris Coskunuzer, and Yulia Gel.
\newblock Tamp-s2gcnets: coupling time-aware multipersistence knowledge representation with spatio-supra graph convolutional networks for time-series forecasting.
\newblock In \emph{ICLR}, 2022.

\bibitem[Kingma and Ba(2015)]{adam_opt}
Diederik~P. Kingma and Jimmy Ba.
\newblock Adam: A method for stochastic optimization.
\newblock In \emph{ICLR}, 2015.

\bibitem[Paszke et~al.(2019)Paszke, Gross, Massa, Lerer, Bradbury, Chanan, Killeen, Lin, Gimelshein, Antiga, et~al.]{pytorch}
Adam Paszke, Sam Gross, Francisco Massa, Adam Lerer, James Bradbury, Gregory Chanan, Trevor Killeen, Zeming Lin, Natalia Gimelshein, Luca Antiga, et~al.
\newblock Pytorch: An imperative style, high-performance deep learning library.
\newblock \emph{Advances in neural information processing systems}, 32, 2019.

\bibitem[Brown et~al.(2020)Brown, Mann, Ryder, Subbiah, Kaplan, Dhariwal, Neelakantan, Shyam, Sastry, Askell, et~al.]{gpt3_2020}
Tom Brown, Benjamin Mann, Nick Ryder, Melanie Subbiah, Jared~D Kaplan, Prafulla Dhariwal, Arvind Neelakantan, Pranav Shyam, Girish Sastry, Amanda Askell, et~al.
\newblock Language models are few-shot learners.
\newblock \emph{NIPS}, 33:\penalty0 1877--1901, 2020.

\bibitem[Zhang and Yan(2023)]{crossformer}
Yunhao Zhang and Junchi Yan.
\newblock Crossformer: Transformer utilizing cross-dimension dependency for multivariate time series forecasting.
\newblock In \emph{ICLR}, 2023.

\bibitem[Zhou et~al.(2022)Zhou, Ma, Wen, Wang, Sun, and Jin]{fedformer}
Tian Zhou, Ziqing Ma, Qingsong Wen, Xue Wang, Liang Sun, and Rong Jin.
\newblock Fedformer: Frequency enhanced decomposed transformer for long-term series forecasting.
\newblock In \emph{ICML}, pages 27268--27286. PMLR, 2022.

\bibitem[Gruver et~al.(2023)Gruver, Finzi, Qiu, and Wilson]{LLMtime}
Nate Gruver, Marc Finzi, Shikai Qiu, and Andrew~G Wilson.
\newblock Large language models are zero-shot time series forecasters.
\newblock \emph{Advances in Neural Information Processing Systems}, 36:\penalty0 19622--19635, 2023.

\bibitem[Xiong et~al.(2021)Xiong, Zeng, Chakraborty, Tan, Fung, Li, and Singh]{nystromformer}
Yunyang Xiong, Zhanpeng Zeng, Rudrasis Chakraborty, Mingxing Tan, Glenn Fung, Yin Li, and Vikas Singh.
\newblock Nystr{\"o}mformer: A nystr{\"o}m-based algorithm for approximating self-attention.
\newblock In \emph{Proceedings of the AAAI conference on artificial intelligence}, volume~35, pages 14138--14148, 2021.

\end{thebibliography}

\clearpage

\appendix


\section{Proof of Theorem 1} \label{proof}

\begin{proof}

For clarity, we denote $\mathbf{z} \triangleq \begin{bmatrix} \mathbf{x} \\y\end{bmatrix} \in \mathbb{R}^{t+1} $, $\mu_z \triangleq \begin{bmatrix} \mu_{\mathbf{x}} \\ \mu_y \end{bmatrix} \in \mathbb{R}^{t+1} $, and $\Sigma \triangleq \begin{bmatrix}
 \Sigma _{\mathbf{x}} & \Sigma _{\mathbf{x}y} \\
 \Sigma _{\mathbf{x}y}^{\mathsf{T}} & \sigma _y^2
\end{bmatrix}$. According to the definition, the probability density function of $\mathrm{z}$, i.e, the joint density of $\mathbf{x}$ and $y$,  is

\begin{equation}
p(\mathbf{z}) = p(\mathbf{x}, y)=\frac{1}{(2\pi)^{(t+1)/2} |\Sigma|^{1/2}} \exp\left( -\frac{1}{2} (\mathbf{z} - \mu_z)^{\mathsf{T}} \Sigma^{-1} (\mathbf{z} - \mu_z) \right),
\label{eq:joint_pdf}
\end{equation}

where $|\Sigma|$ denotes the determinant of $\Sigma$. In the following, we ignore the constant coefficient and focus on the exponential term.

Using the standard block matrix inverse formula \citep{matrix}, we have 

\begin{equation}
\Sigma^{-1} =
\begin{bmatrix}
\Sigma_{\mathbf{x}}^{-1} + d^{-1} \Sigma_{\mathbf{x}}^{-1} \Sigma_{\mathbf{x}y}  \Sigma_{\mathbf{x}y}^{\mathsf{T}} \Sigma_{\mathbf{x}}^{-1} & -d^{-1} \Sigma_{\mathbf{x}}^{-1} \Sigma_{\mathbf{x}y}  \\
- d^{-1} \Sigma_{\mathbf{x}y}^{\mathsf{T}} \Sigma_{\mathbf{x}}^{-1} & d^{-1}
\end{bmatrix}  \triangleq \begin{bmatrix}
\mathbf{A} & -\mathbf{b} \\
-\mathbf{b}^{\mathsf{T}} & d^{-1}
\end{bmatrix},
\label{eq:block_inverse}
\end{equation}

where the scalar $d = \sigma_y^2 - \Sigma_{\mathbf{x}y}^{\mathsf{T}} \Sigma_{\mathbf{x}}^{-1} \Sigma_{\mathbf{x}y}$ is the Schur complement \citep{matrix}. Letting  $\mathbf{b} \in \mathbb{R}^{t\times 1}$ denote $d^{-1} \Sigma_{\mathbf{x}}^{-1} \Sigma_{\mathbf{x}y}$, the matrix $\mathbf{A}$ becomes $\Sigma_{\mathbf{x}}^{-1} + d \cdot \mathbf{b}\mathbf{b}^{\mathsf{T}}$. 

For notational simplicity, we define $\mathbf{\alpha} \triangleq  \mathbf{x} - \mu_{\mathbf{x}} \in \mathbb{R}^{t\times 1}$ and $\beta \triangleq  y - \mu_y \in \mathbb{R}$. Then, the exponent in Equation~\ref{eq:joint_pdf} (ignoring the constant factor $-1/2$) becomes

\begin{equation}
\begin{aligned}
(\mathbf{z} - \mu_z)^{\mathsf{T}} \Sigma^{-1} (\mathbf{z} - \mu_z) &= \begin{bmatrix} \alpha ^{\mathsf{T}} & \beta \end{bmatrix} \begin{bmatrix}
\Sigma_{\mathbf{x}}^{-1} + d \cdot \mathbf{b}\mathbf{b}^{\mathsf{T}}  & -\mathbf{b} \\
 -\mathbf{b}^{\mathsf{T}} & d^{-1}
\end{bmatrix} \begin{bmatrix} \alpha  \\ \beta \end{bmatrix} \\
&=\alpha ^{\mathsf{T}}\Sigma_{\mathbf{x}}^{-1}\alpha + d \cdot \alpha ^{\mathsf{T}} \mathbf{b}\mathbf{b}^{\mathsf{T}} - 
2\beta \alpha ^{\mathsf{T}} \mathbf{b} + d^{-1}\beta ^2 \\
&=\alpha ^{\mathsf{T}}\Sigma_{\mathbf{x}}^{-1}\alpha + d^{-1}\left ( \beta ^2 -2d \cdot \alpha ^{\mathsf{T}} \mathbf{b} \cdot \beta 
+ d^2 \cdot \alpha ^{\mathsf{T}} \mathbf{b}\mathbf{b}^{\mathsf{T}}\alpha  \right )  \\  
&=\alpha ^{\mathsf{T}}\Sigma_{\mathbf{x}}^{-1}\alpha + d^{-1} \left ( \beta - d \alpha ^{\mathsf{T}} \mathbf{b} \right ) ^2.
\end{aligned}
\label{eq:myexponent}
\end{equation}

Since  $p(\mathbf{x}) \propto \exp\left( -\frac{1}{2} \alpha^{\mathsf{T}} \Sigma_{\mathbf{x}}^{-1} \alpha \right)$, it follows that

\begin{equation}
\begin{aligned}
p(y \mid \mathbf{x}) &= \frac{p(\mathbf{x}, y)}{p(\mathbf{x})}  \\
& \propto \exp\left( -\frac{1}{2} \left( \alpha ^{\mathsf{T}}\Sigma_{\mathbf{x}}^{-1}\alpha + d^{-1} \left ( \beta - d \alpha ^{\mathsf{T}} \mathbf{b} \right ) ^2 -\alpha ^{\mathsf{T}}\Sigma_x^{-1}\alpha \right) \right) \\
& = \exp\left( - \frac{\left ( \beta - d \alpha ^{\mathsf{T}} \mathbf{b} \right ) ^2 }{2d}   \right) \\
& = \exp\left( - \frac{\left ( y - \mu _y - d \alpha ^{\mathsf{T}} \mathbf{b} \right ) ^2 }{2d}   \right).
\end{aligned}
\label{eq:conditionpdf}
\end{equation}

Therefore, the conditional density of $y$ give $\mathbf{x}$ is also a Gaussian distribution, whose mean is

\begin{equation}
\begin{aligned}
\mu _y + d \alpha ^{\mathsf{T}} \mathbf{b} & = \mu _y + d \mathbf{b}^{\mathsf{T}} \alpha   \\
&= \mu _y + d \cdot d^{-1} \Sigma_{\mathbf{x}y}^{\mathsf{T}} \Sigma_{\mathbf{x}}^{-1} \left ( \mathbf{x}-\mu _x  \right )  \\
&= \mu _y +  \Sigma_{\mathbf{x}y}^{\mathsf{T}} \Sigma_{\mathbf{x}}^{-1} \left ( \mathbf{x}-\mu _{\mathbf{x}}  \right ),
\end{aligned}
\label{eq:mean}
\end{equation}

and the variance is $d = \sigma_y^2 - \Sigma_{\mathbf{x}y}^{\mathsf{T}} \Sigma_{\mathbf{x}}^{-1} \Sigma_{\mathbf{x}y}$. This completes the proof.

\end{proof}

\section{Jacobian matrix comparison of self-attention and NormLin} \label{jacob}

In this section, we analyze the gradients of the non-linear transformations in the self-attention mechanism and the NormLin module, focusing on the attention/weight rows, denoted as $\mathbf{a} \in \mathbb{R}^N$.

\subsection{Jacobian matrix of Softmax in self-attention}

Let $\mathbf{c} \triangleq \mathrm{Softmax}(\mathbf{a}) \in \mathbb{R}^N$. We first compute the partial derivatives element-wise, and then rewrite the results in matrix form. Based on the definition of $\mathrm{Softmax}$, the $i$-th element of $\mathbf{c}$ is 
$\mathbf{c}_i = \frac{e^{\mathbf{a}_i}}{\sum_{k=1}^{N}e^{\mathbf{a}_k}}$. The partial derivative of $\mathbf{c}_i$ with respect to $\mathbf{a}_i$ is:

\begin{equation}
\begin{aligned}
\frac{\partial \mathbf{c} _i}{\partial \mathbf{a} _i} = \frac{e^{\mathbf{a}_i} }{\sum_{k  = 1}^{N}e^{\mathbf{a}_k} }-\frac{e^{\mathbf{a}_i}\cdot e^{\mathbf{a}_i}}{(\sum_{k  = 1}^{N}e^{\mathbf{a}_k} )^2}  = \mathbf{c} _i - \mathbf{c} _i^2.
\end{aligned} 
\label{eq20}
\end{equation}

For $j \neq i$, the partial derivative of $\mathbf{c} _i$ with respect to $\mathbf{a}_j$  is:

\begin{equation}
\begin{aligned}
\frac{\partial \mathbf{c} _i}{\partial \mathbf{a} _j} = -\frac{e^{\mathbf{a}_i}\cdot e^{\mathbf{a}_j}}{(\sum_{k = 1}^{N}e^{\mathbf{a}_k} )^2}  = -\mathbf{c}_i \mathbf{c} _j
\end{aligned} 
\label{eq21}
\end{equation}

Combining Equations~\eqref{eq20} and~\eqref{eq21}, we obtain:

\begin{equation}
\begin{aligned}
\frac{\partial \mathbf{c}_i }{\partial \mathbf{a} }  = \mathbf{c}_i [\delta_{1i}, \delta_{2i}, \ldots, \delta_{Ni}]  - \mathbf{c}_i \mathbf{c}^{\mathsf{T}}, 
\end{aligned} 
\label{eq22}
\end{equation}

where $\delta_{ji} = 1$ if $j = i$, otherwise 0. Therefore, the Jacobian matrix $\frac{\partial \mathbf{c} }{\partial \mathbf{a} }$ can be written as

\begin{equation}
\frac{\partial \mathbf{c} }{\partial \mathbf{a} }  = \mathrm{Diag} (\mathbf{c}) - \mathbf{c}\mathbf{c}^{\mathsf{T}},
\label{eq:softmax_jacob}
\end{equation}

where $\mathrm{Diag} (\mathbf{c})$ is the diagonal matrix with $\mathbf{c}$ as its diagonal. We can observe that the Jacobian matrix is a function of $\mathbf{c}$. Since the Softmax function could cause sharp value concentration, the Jacobian matrix may exhibit sparsity, with most entries being close to zero \citep{laser}.

\subsection{Jacobian matrix of the NormLin layer}

Let $\mathbf{c} \triangleq \mathrm{Norm}_{\mathrm{L1} }\left ( \mathrm{Softplus} \left ( \mathbf{a}  \right )  \right )  \in \mathbb{R}^N$, where $\mathrm{Norm}_{\mathrm{L1} } (\cdot)$ denotes the L1 normalization. For notational simplicity, let $\mathbf{b} \triangleq \mathrm{Softplus}(\mathbf{a})$.
We first analyze $\frac{\partial \mathbf{c} }{\partial \mathbf{b} }$. Since all entries of $\mathbf{b}$ are positive, we have:

\begin{equation}
\mathbf{c}_i = \frac{\mathbf{b}_i}{\sum_{k=1}^{N}\mathbf{b}_k}.
\label{eq24}
\end{equation}

Therefore, it follows that

\begin{equation}
\begin{aligned}
\frac{\partial \mathbf{c}_i}{\partial \mathbf{b}_i} = \frac{1}{\sum_{k=1}^{N}\mathbf{b}_k} -\frac{\mathbf{b}_i}{(\sum_{k=1}^{N}\mathbf{b}_k)^2} =\frac{1}{\left \| \mathbf{b} \right \|_1^2 } \left ( \left \| \mathbf{b} \right \|_1- \mathbf{b}_i \right ).
\end{aligned}
\label{eq25}
\end{equation}

For $j \neq i$, the partial derivative $\frac{\partial \mathbf{c}_i}{\partial \mathbf{b}_j}$ is:

\begin{equation}
\begin{aligned}
\frac{\partial \mathbf{c}_i}{\partial \mathbf{b}_j}   = 
 -\frac{\mathbf{b}_i}{(\sum_{k=1}^{N}\mathbf{b}_k)^2} 
 = -\frac{1}{\left \| \mathbf{b} \right \|_1^2 }  \mathbf{b}_i.
\end{aligned}
\label{eq26}
\end{equation}

Combining Equations~\eqref{eq25} and~\eqref{eq26}, we obtain the Jacobian matrix:

\begin{equation}
\begin{aligned}
\frac{\partial \mathbf{c}}{\partial \mathbf{b}} 
 = \frac{1}{\left \| \mathbf{b} \right \|_1^2 } 
\left ( \left \| \mathbf{b} \right \|_1 \mathbf{I}- \mathbf{b} \mathbf{1}^{\mathsf{T}}  \right ), 
\end{aligned}
\label{eq27}
\end{equation}

where $\mathbf{1} \in \mathbb{R}^N$ is the all-ones vector.

For the derivative of $\mathbf{b}_i$ with respect to $\mathbf{a}_j$ with $1 \le i,j \le N$, we have

\begin{equation}
\begin{aligned}
\frac{\partial \mathbf{b}_i}{\partial \mathbf{a}_j} & = \delta_{ij} \cdot \frac{\partial \mathbf{b}_i}{\partial \mathbf{a}_i}  = \delta_{ij} \cdot \frac{\partial }{\partial \mathbf{a}_i} \mathrm{In}(1+e^{\mathbf{a}_i}) \\
&= \delta_{ij} \cdot \frac{e^{\mathbf{a}_i}}{1+e^{\mathbf{a}_i}} = \delta_{ij} \cdot \mathrm{Sigmoid} (\mathbf{a}_i).
\end{aligned}
\label{eq28}
\end{equation}

Therefore, $\frac{\partial \mathbf{b}}{\partial \mathbf{a}}$ can be written as

\begin{equation}
\begin{aligned}
\frac{\partial \mathbf{b}}{\partial \mathbf{a}}  = \mathrm{Diag}  \left ( \mathrm{Sigmoid} (\mathbf{a}) \right ) \triangleq \mathrm{Diag}  \left ( \tilde{\mathbf{b}} \right ),
\end{aligned}
\label{eq29}
\end{equation}

where $\mathrm{Sigmoid} (\cdot)$ operates element-wise, and $\tilde{\mathbf{b}} \triangleq \mathrm{Sigmoid} (\mathbf{a})$ is defined for clarity.  
Using the chain rule, the Jacobian matrix of $\mathbf{c}$ with respect to $\mathbf{a}$ can be derived as follows:

\begin{equation}
\begin{aligned}
\frac{\partial \mathbf{c}}{\partial \mathbf{a}} & = \frac{\partial \mathbf{c}}{\partial \mathbf{b}} \cdot
\frac{\partial \mathbf{b}}{\partial \mathbf{a}}  = \frac{1}{\left \| \mathbf{b} \right \|_1^2 } 
\left ( \left \| \mathbf{b} \right \|_1 \mathbf{I}- \mathbf{b} \mathbf{1}^{\mathsf{T}}  \right )
\mathrm{Diag}  \left ( \tilde{\mathbf{b}} \right ) \\
& =\frac{1}{\left \| \mathbf{b} \right \|_1^2 } 
\left ( \left \| \mathbf{b} \right \|_1 \mathrm{Diag}  \left ( \tilde{\mathbf{b}} \right ) - \mathbf{b} \tilde{\mathbf{b}}^{\mathsf{T}} \right ) \\
&=\frac{1}{\left \| \mathbf{b} \right \|_1 } 
\left (  \mathrm{Diag}  \left ( \tilde{\mathbf{b}} \right ) - \bar{\mathbf{b}} \tilde{\mathbf{b}}^{\mathsf{T}} \right ),
\end{aligned}
\label{eq:normlin_jacob}
\end{equation}

where we use the fact that $\mathbf{1}^{\mathsf{T}} \mathrm{Diag}  \left ( \tilde{\mathbf{b}} \right ) = \tilde{\mathbf{b}}^{\mathsf{T}} $, and $\bar{\mathbf{b}} = \frac{\mathbf{b}}{\left \| \mathbf{b} \right \|_1 }$ is the normalized $\mathbf{b}$. 

\begin{figure}[t]
   \centering
   \includegraphics[width=1.0\linewidth]{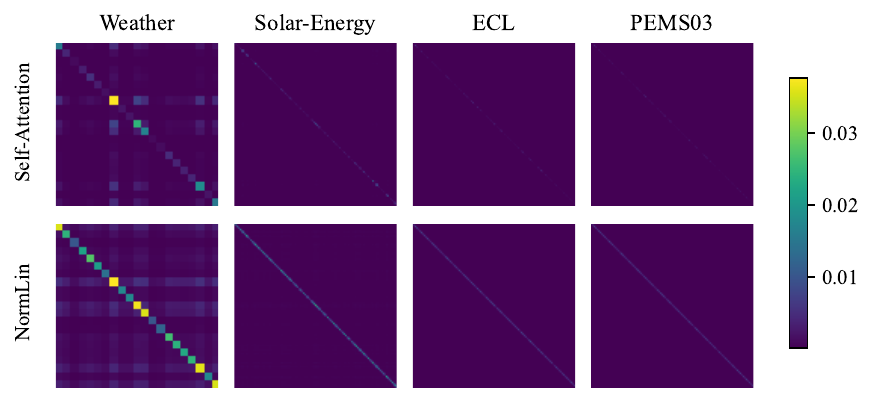}
   \caption{Comparison of Jacobian matrices of $\mathrm{Softmax} \left ( \cdot  \right )$ in self-attention and  $\mathrm{Norm}_{\mathrm{L1} }\left ( \mathrm{Softplus} \left ( \cdot  \right )  \right )$ in the NormLin layer. The absolute values of the matrix entries are visualized. NormLin and self-attention exhibit similar patterns; however, NormLin generally yields diagonal entries of higher magnitude, suggesting more effective gradient propagation. The same input vector is used for both to ensure a fair comparison.}
   \label{fig:jacob_mat}
\end{figure}

Equations~\eqref{eq:normlin_jacob} and~\eqref{eq:softmax_jacob} share similar structures, with the former introducing an additional learnable scaling factor $\frac{1}{\left \| \mathbf{b} \right \|_1}$, which offers greater flexibility.
The entries of $\tilde{\mathbf{b}} = \mathrm{Sigmoid} (\mathbf{a})$ in Equation~\eqref{eq:normlin_jacob} are generally larger than those of $\mathbf{c} = \mathrm{Softmax}(\mathbf{a})$ in Equation~\eqref{eq:softmax_jacob}, particularly when $\mathbf{a}$ contains small values near zero. This property may contribute to stronger gradients in NormLin.
Figure~\ref{fig:jacob_mat} illustrates the Jacobian matrices of  the self-attention mechanism and the NormLin layer under the same input $\mathbf{a}$, highlighting that NormLin tends to produce stronger gradient values.


\section{Dataset description} \label{dataset}

\begin{table}[p]
\caption{Detailed dataset descriptions and statistics. \textit{Dim} denotes the number of variates for each dataset. \textit{Frequency} refers to the time interval between consecutive steps. \textit{Split} indicates the (Train, Validation, Test) ratio. \textit{Prediction len.} represents the prediction lengths. For long-term forecasting, the input length is fixed at 96. For short-term forecasting, we adopt two settings: S1 (Input-12, Predict-$\{3,6,9,12\}$) and S2 (Input-36, Predict-$\{24,36,48,60\}$). In total, 140 prediction tasks across various datasets and prediction length settings are evaluated in this work.}

\label{tab:dataset}
\centering
\setlength{\tabcolsep}{6pt}
\begin{tabular}{@{}lcccccc@{}}
\toprule
Dataset                  & Dim          & Frequency          & Total len.           & Split & Prediction len.                                                           & Information             \\ \midrule
ETTh1, ETTh2              & 7                   & Hourly                 & 17420                & 6:2:2         & \{96,192,336,720\}                                                          & Electricity             \\ \midrule
ETTm1, ETTm2              & 7                   & 15 mins                & 69680                & 6:2:2         & \{96,192,336,720\}                                                          & Electricity             \\ \midrule
Weather                   & 21                  & 10 mins                & 52696                & 7:1:2         & \{96,192,336,720\}                                                          & Weather                 \\ \midrule
ECL                       & 321                 & Hourly                 & 26304                & 7:1:2         & \{96,192,336,720\}                                                          & Electricity             \\ \midrule
Traffic                   & 862                 & Hourly                 & 17544                & 7:1:2         & \{96,192,336,720\}                                                          & Transportation          \\ \midrule
Exchange                  & 8                   & Daily                  & 7588                 & 7:1:2         & \{96,192,336,720\}                                                          & Economy                 \\ \midrule
Solar-Energy              & 137                 & 10 mins                & 52560                & 7:1:2         & \{96,192,336,720\}                                                          & Energy                  \\ \midrule
PEMS03                    & 358                 & 5 mins                 & 26209                & 6:2:2         & \{12,24,48,96\}                                                             & Transportation          \\ \midrule
PEMS04                    & 307                 & 5 mins                 & 16992                & 6:2:2         & \{12,24,48,96\}                                                             & Transportation          \\ \midrule
PEMS07                    & 883                 & 5 mins                 & 28224                & 6:2:2         & \{12,24,48,96\}                                                             & Transportation          \\ \midrule
PEMS08                    & 170                 & 5 mins                 & 17856                & 6:2:2         & \{12,24,48,96\}                                                             & Transportation          \\ \midrule
ILI                       & 7                   & Weekly                 & 966                  & 7:1:2         & \begin{tabular}[c]{@{}c@{}}\{3,6,9,12\}\\      \{24,36,48,60\}\end{tabular} & Health                  \\ \midrule
\multirow{2}{*}{COVID-19} & \multirow{2}{*}{55} & \multirow{2}{*}{Daily} & \multirow{2}{*}{335} & 7:1:2         & \{3,6,9,12\}                                                                & \multirow{2}{*}{Health} \\ \cmidrule(lr){5-6}
                          &                     &                        &                      & 6:2:2         & \{24,36,48,60\}                                                             &                         \\ \midrule
METR-LA                   & 207                 & 5 mins                 & 34272                & 7:1:2         & \begin{tabular}[c]{@{}c@{}}\{3,6,9,12\}\\      \{24,36,48,60\}\end{tabular} & Transportation          \\ \midrule
NASDAQ                    & 12                  & Daily                  & 3914                 & 7:1:2         & \begin{tabular}[c]{@{}c@{}}\{3,6,9,12\}\\      \{24,36,48,60\}\end{tabular} & Finance                 \\ \midrule
Wiki                      & 99                  & Daily                  & 730                  & 7:1:2         & \begin{tabular}[c]{@{}c@{}}\{3,6,9,12\}\\      \{24,36,48,60\}\end{tabular} & Web                     \\ \midrule
SP500                     & 5                   & Daily                  & 8077                 & 7:1:2         & \begin{tabular}[c]{@{}c@{}}\{3,6,9,12\}\\      \{24,36,48,60\}\end{tabular} & Finance                 \\ \midrule
DowJones                  & 27                  & Daily                  & 6577                 & 7:1:2         & \begin{tabular}[c]{@{}c@{}}\{3,6,9,12\}\\      \{24,36,48,60\}\end{tabular} & Finance                 \\ \midrule
CarSales                   & 10                  & Daily                  & 6728                 & 7:1:2         & \begin{tabular}[c]{@{}c@{}}\{3,6,9,12\}\\      \{24,36,48,60\}\end{tabular} & Market                    \\ \midrule
Power                     & 2                   & Daily                  & 1186                 & 7:1:2         & \begin{tabular}[c]{@{}c@{}}\{3,6,9,12\}\\      \{24,36,48,60\}\end{tabular} & Energy                  \\ \midrule
Website                   & 4                   & Daily                  & 2167                 & 7:1:2         & \begin{tabular}[c]{@{}c@{}}\{3,6,9,12\}\\      \{24,36,48,60\}\end{tabular} & Web                     \\ \midrule
Unemp                        & 53                  & Monthly                & 531                  & 6:2:2         & \begin{tabular}[c]{@{}c@{}}\{3,6,9,12\}\\      \{24,36,48,60\}\end{tabular} & Society                 \\ \bottomrule
\end{tabular}
\end{table}

In this work, the following real-world datasets are used for performance evaluation. The dataset details are presented in Table~\ref{tab:dataset}.

\begin{itemize}

\item \textbf{ETT} datasets \citep{informer} record seven channels related to electricity transformers from July 2016 to July 2018. It contains four datasets: ETTh1 and ETTh2, with hourly recordings, and ETTm1 and ETTm2, with 15-minute recordings.

\item \textbf{Weather} \citep{autoformer} contains 21 meteorological variables (e.g., air temperature, humidity) recorded every 10 minutes in 2020.

\item \textbf{ECL} \citep{autoformer} records hourly electricity consumption of 321 consumers from July 2016 to July 2019.

\item \textbf{Traffic} \citep{autoformer} includes hourly road occupancy rates from 862 sensors in the Bay Area from January 2015 to December 2016.

\item \textbf{Exchange} \citep{autoformer} collects daily exchange rates for eight countries from January 1990 to October 2010.

\item \textbf{Solar-Energy} \citep{rnn} records the solar power output every 10 minutes from 137 photovoltaic plants in 2006.

\item \textbf{PEMS} \citep{scinet} provides public traffic sensor  data from California, collected every 5 minutes. We use its four subsets (PEMS03, PEMS04, PEMS07, PEMS08) in this study.

\item \textbf{ILI} \footnote{\url{https://gis.cdc.gov/grasp/fluview/fluportaldashboard.html}} contains weekly records of influenza-like illness patient counts provided by the U.S. CDC from 2002 to 2021.

\item \textbf{COVID-19} \citep{chen2022tamp} includes daily records of COVID-19 hospitalizations in California in 2020, provided by Johns Hopkins University.

\item \textbf{METR-LA} \footnote{\url{https://github.com/liyaguang/DCRNN}} collects traffic network data in Los Angeles every 5 minutes from March to June 2012. A total of 207 channels are included.

\item \textbf{NASDAQ} \footnote{\url{https://www.kaggle.com/datasets/sai14karthik/nasdq-dataset}} includes daily NASDAQ index and key economic indicators (e.g., interest rate and gold price) from 2010 to 2024.

\item \textbf{Wiki} \footnote{\url{https://www.kaggle.com/datasets/sandeshbhat/wikipedia-web-traffic-201819}} records daily page view counts for Wikipedia articles over two years (2018–2019). The first 99 channels are used in this study.

\item \textbf{SP500} records daily SP500 index data (e.g., opening price, closing price, and trading volume) from January 1993 to February 2025.

\item \textbf{DowJones}  collects daily stock prices of 27 Dow Jones Industrial Average (DJIA) component companies from January 1999 to March 2025.

\item \textbf{CarSales} collects daily sales of 10 vehicle brands (e.g., Toyota, Honda) in the U.S. from January 2005 to June 2023. The data are compiled from the \textit{Vehicles Sales} dataset \footnote{\url{https://www.kaggle.com/datasets/crisbam/vehicles-sales/data}} on Kaggle.

\item \textbf{Power} contains daily wind and solar energy production (in MW) records for the French grid from April 2020 to June 2023. The data are compiled from the \textit{Wind \& Solar Daily Power Production} dataset \footnote{\url{https://www.kaggle.com/datasets/henriupton/wind-solar-electricity-production}} on Kaggle.

\item \textbf{Website} \footnote{\url{https://www.kaggle.com/datasets/bobnau/daily-website-visitors}} contains six years of daily visit data (e.g., first-time and returning visits) to an academic website, spanning from September 2014 to August 2020.

\item \textbf{Unemp} contains monthly unemployment figures for 50 U.S. states and three other territories from January 1976 to March 2020, sourced from the official website of the U.S. Bureau of Labor Statistics\footnote{\url{https://www.bls.gov/web/laus.supp.toc.htm}}.

\end{itemize}

\section{Implementation details}  \label{imple_details}

OLinear is optimized using the ADAM optimizer \citep{adam_opt}, with the initial learning rate selected from $\left \{ 10^{-4}, 2\times 10^{-4}, 5\times 10^{-4} \right \}$. The model dimension $D$ is chosen from $\left \{ 128, 256, 512 \right \}$, while the embedding size $d$ is set to 16. The batch size is selected from $\left \{ 4, 8, 16, 32 \right \}$  depending on the dataset scale. The block number $L$ is chosen from $\left \{ 1, 2, 3 \right \}$. Training is performed for up to 50 epochs with early stopping, which halts training if the validation performance does not improve for 10 consecutive epochs. We adopt the weighted L1 loss function following CARD \citep{card}. The experiments are implemented in PyTorch \citep{pytorch} and conducted on an NVIDIA GPU with 24 GB of memory. Hyperparameter sensitivity is discussed in Appendix~\ref{hyperpara}.
For baseline models, we use the reported values from the original papers when available; otherwise, we produce the results using the official code. For the model FilterNet \citep{filternet}, TexFilter is adopted in this work. The code and datasets are available at the following anonymous repository: \url{https://anonymous.4open.science/r/OLinear}.

\section{Showcases}

\subsection{Time domain and transformed domain} \label{append_time_trans}

Figure~\ref{fig:norm_weight_vs_corr} illustrates the temporal domain and its corresponding transformed domain. In the transformed domain, the correlations along the sequence are effectively suppressed. Adjacent time series exhibit strong consistency in this new domain, which is desirable for forecasting tasks. The orthogonal transformation corresponds to projecting the time series onto the eigenvectors of the temporal Pearson correlation matrix $\mathrm{CorrMat}_t$. The eigenvalues of this symmetric positive semi-definite matrix $\mathrm{CorrMat}_t$ typically decay rapidly \citep{pca}, with only a few being dominant. Consequently, the transformed series exhibits sparsity, with most of the energy concentrated in just a few dimensions. Moreover, since noise tends to be evenly distributed across the transformed dimensions, the signal-to-noise ratio (SNR) in the leading components is improved \citep{prml_book}.

\vspace{15pt}
\begin{figure}[t]
   \centering
   \includegraphics[width=1.0\linewidth]{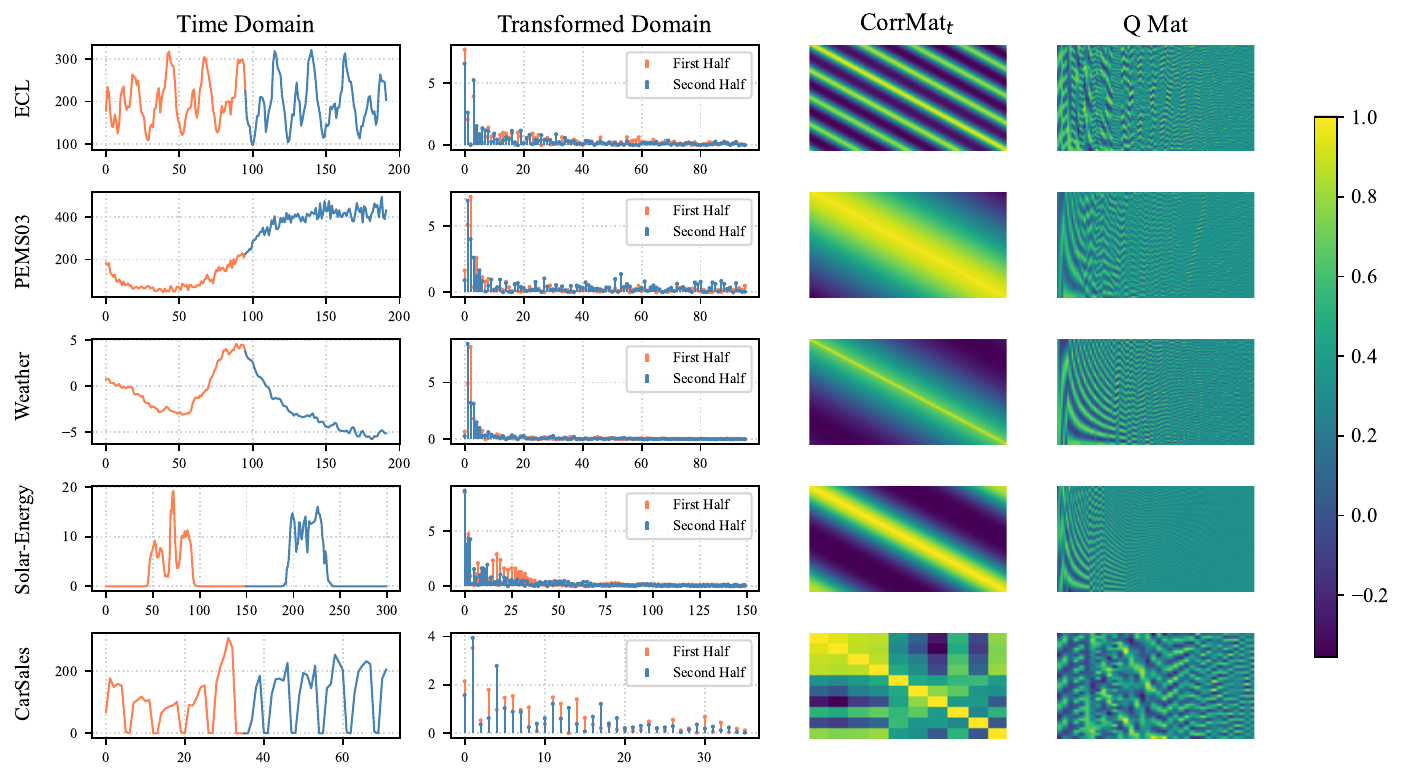}
\caption{Visualization of temporal series, transformed series, the temporal Pearson correlation matrix, and its corresponding orthogonal matrix. The series are normalized prior to transformation, and the absolute values in the transformed domain are shown for clarity. The X-axis in the first and second columns represents the temporal and feature dimensions, respectively.
Adjacent series exhibit consistency in the transformed domain, with sparsity clearly observable. Temporal Pearson correlation matrices on more datasets are presented in Figure~\ref{fig:temp_corr_mat_on_datasets}.}
   \label{fig:time_trans_corr_Q}
\end{figure}

\begin{figure}[b]
   \centering
   \includegraphics[width=0.7\linewidth]{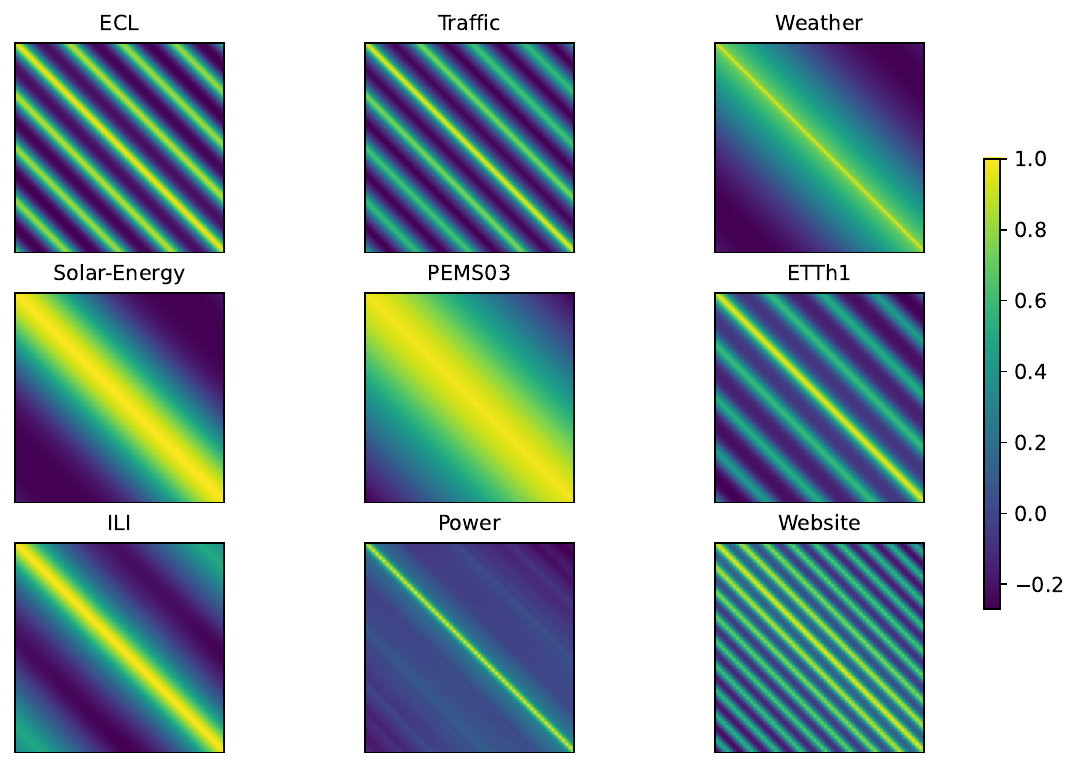}
   \vspace{10pt}
   \caption{Temporal Pearson correlation matrices on various datasets. Periodic patterns are clearly reflected in the correlation matrices. We incorporate such \textit{a priori} statistical information into the model design, leading to improved forecasting performance.}
   \label{fig:temp_corr_mat_on_datasets}
\end{figure}

\subsection{Low-rank attention matrix and high-rank NormLin weight matrix} \label{low_rank}

\begin{figure}[t]
   \centering
   \includegraphics[width=1.0\linewidth]{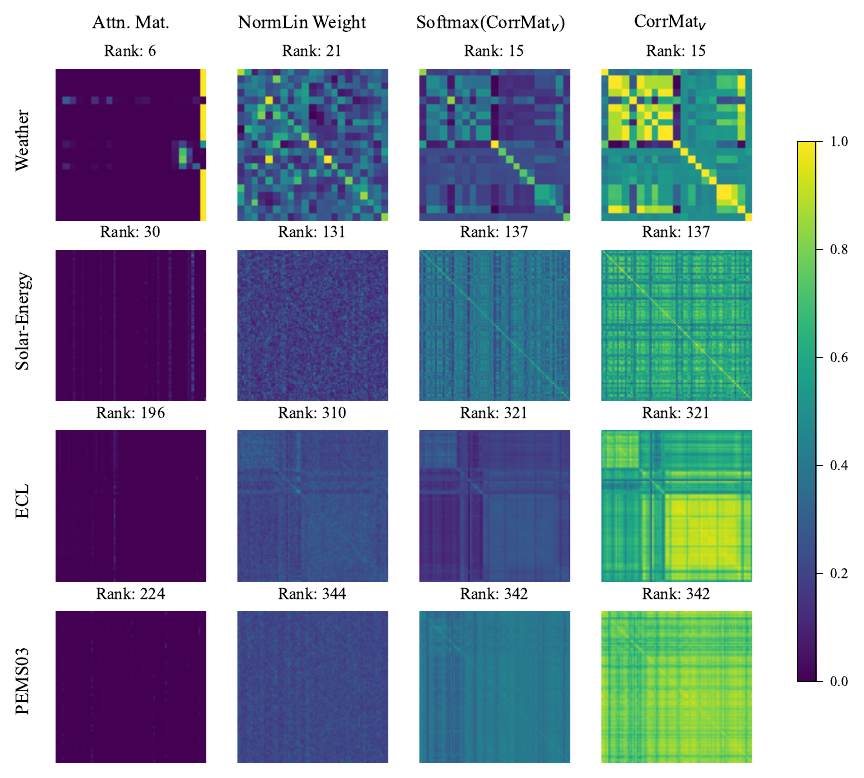}
   \caption{Illustration of attention/weight matrices from self-attention, NormLin, and the across-variate Pearson correlation matrices (with and without Softmax). When applied in OLinear, the self-attention mechanism results in low-rank attention matrices, whereas NormLin produces high-rank weight matrices that better preserve representation diversity. Moreover, the learned NormLin weights resemble the across-variate correlation matrices, suggesting effective modeling of multivariate relationships.}
   \label{fig:attn_normlin_corr_corr}
\end{figure}

We replace the NormLin module in OLinear with a standard self-attention mechanism and observe the typical low-rank property in the resulting attention matrices, as shown in Figure~\ref{fig:attn_normlin_corr_corr}. This phenomenon can be attributed to the sparsity induced by the transformed domain and the sharp focus introduced by the Softmax operation. The low-rank attention matrix could limit the expressive capacity of the model. In contrast, the weight matrices in NormLin exhibit higher rank, better preserving representation diversity.

Furthermore, the learned NormLin weights closely resemble the across-variate Pearson correlation matrix $\mathrm{CorrMat}_v$, suggesting that the NormLin layer effectively captures multivariate correlations. By directly replacing the learnable weight matrix with  $\mathrm{Softmax}(\mathrm{CorrMat}_v)$, we obtain \textbf{OLinear-C}, which also achieves competitive performance, as demonstrated in Appendix~\ref{append_orthoc}.

\subsection{OrthoTrans enhances attention matrix rank} \label{orthotrans_imp_rank}

\begin{figure}[t]
   \centering
   \includegraphics[width=1.0\linewidth]{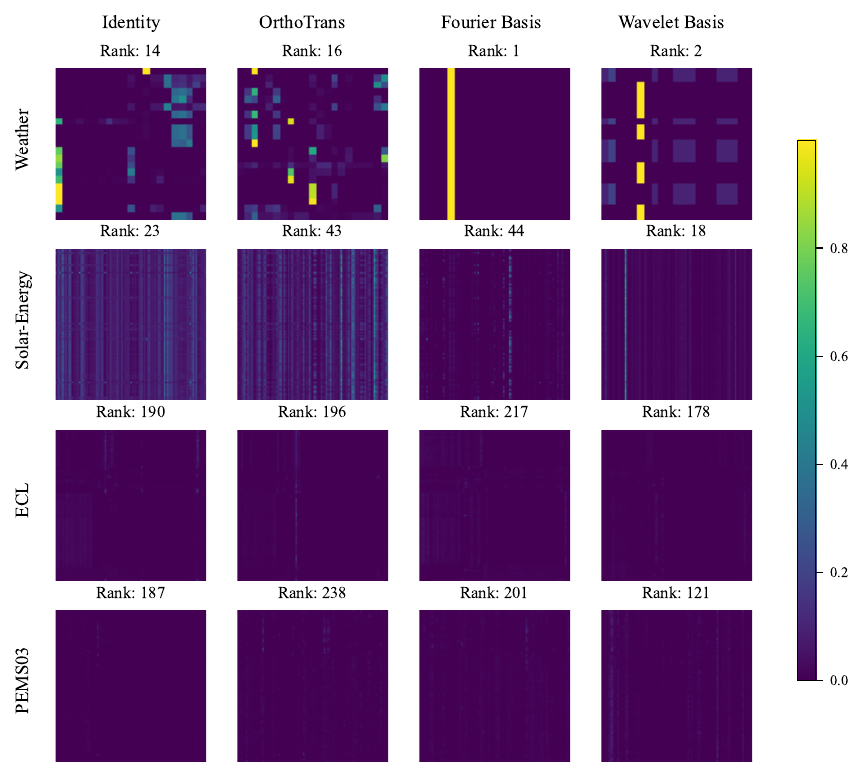}
   \caption{Comparison of attention matrices under various transformation bases for OLinear with the NormLin module replaced by self-attention (\textbf{Example 1}). For a fair comparison, the \textbf{same} input series is used for different bases on each dataset. `Identity' denotes no transformation, and the Haar wavelet is used as the wavelet basis.  In general, OrthoTrans yields higher-rank attention matrices and thus better preserves the representation space.}
   
   \label{fig:orthotrans_improve_rank}
\end{figure}

\begin{figure}[t]
   \centering
   \includegraphics[width=1.0\linewidth]{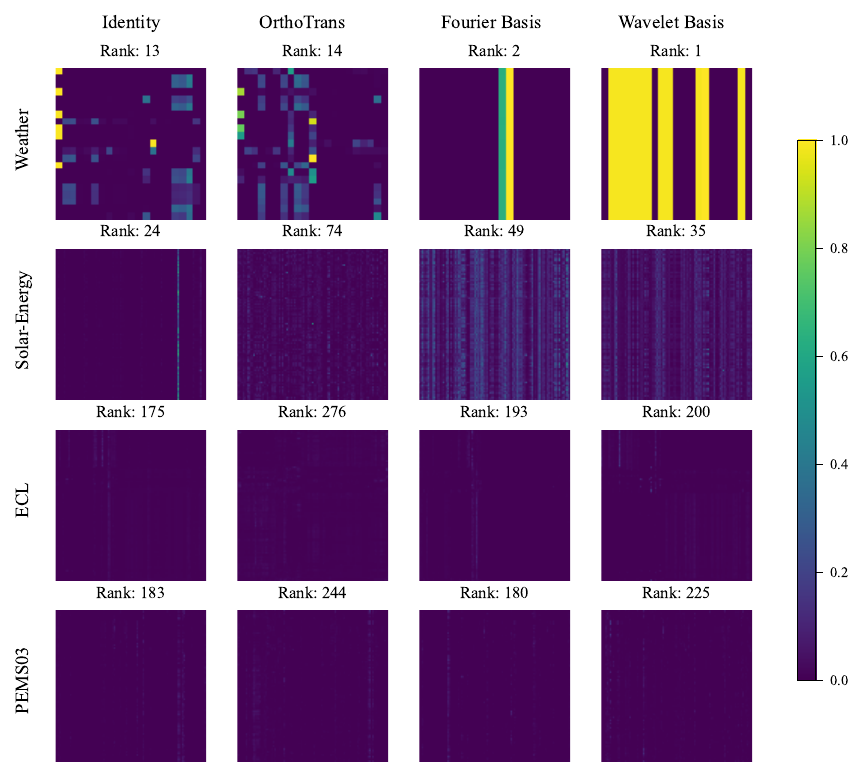}
   \caption{Comparison of attention matrices under various transformation bases for OLinear with the NormLin module replaced by self-attention (\textbf{Example 2}). For a fair comparison, the \textbf{same} input series is used across bases on each dataset. `Identity' denotes no transformation, and the Haar wavelet is used as the wavelet basis.  In general, OrthoTrans yields higher-rank attention matrices and thus better preserves the representation space.}
   \label{fig:orthotrans_improve_rank2}
\end{figure}

\begin{figure}[t]
   \centering
   \includegraphics[width=1.0\linewidth]{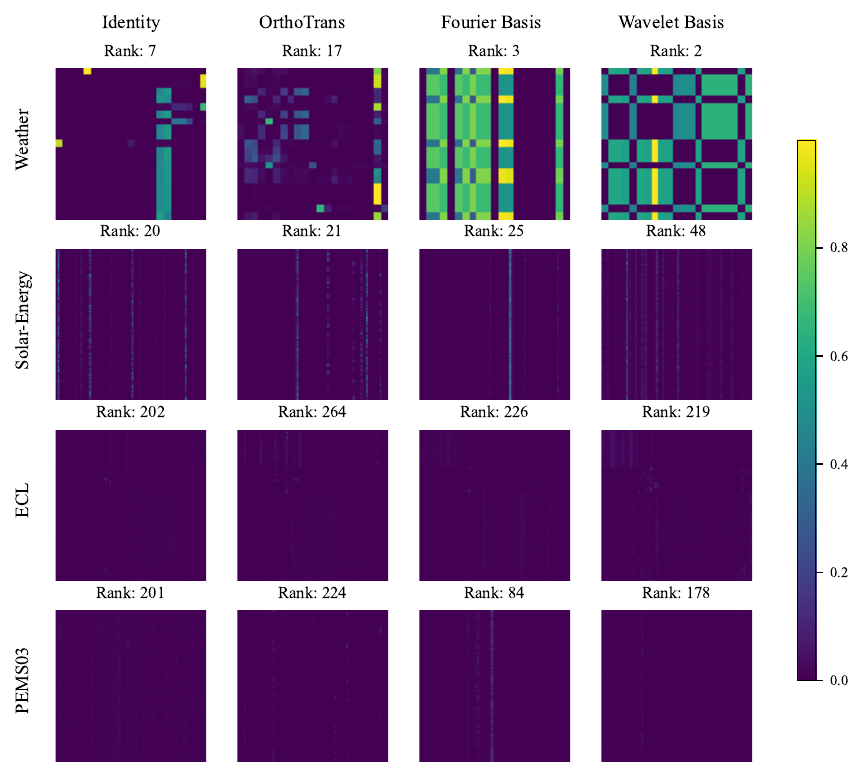}
   \caption{Comparison of attention matrices under various transformation bases for OLinear with the NormLin module replaced by self-attention (\textbf{Example 3}). For a fair comparison, the \textbf{same} input series is used across bases on each dataset. `Identity' denotes no transformation, and the Haar wavelet is used as the wavelet basis.  In general, OrthoTrans yields higher-rank attention matrices and thus better preserves the representation space.}
   \label{fig:orthotrans_improve_rank3}
\end{figure}

\begin{figure}[t]
   \centering
   \includegraphics[width=1.0\linewidth]{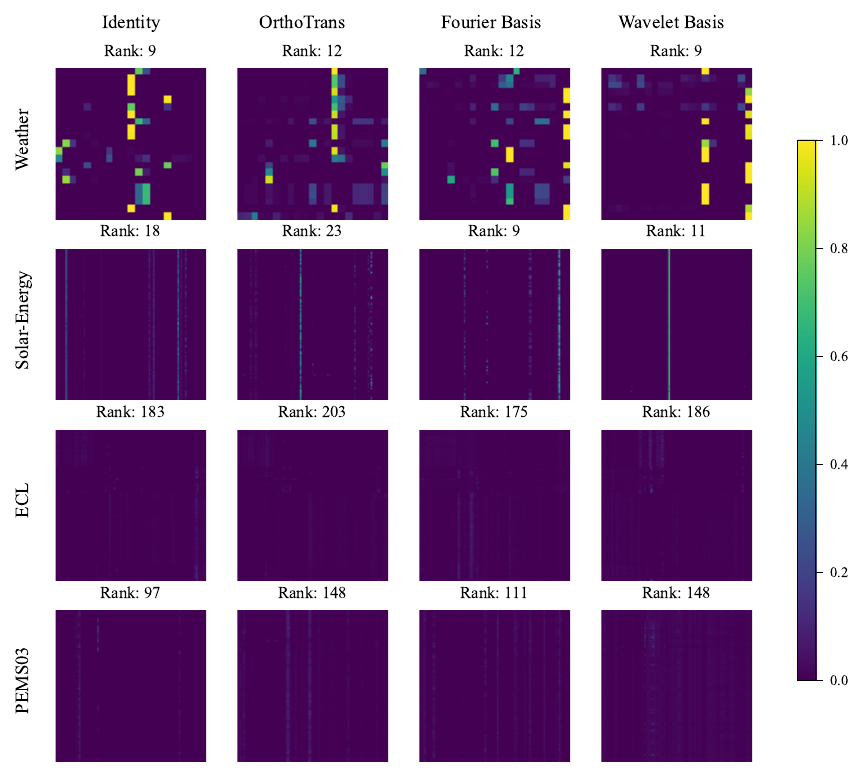}
   \caption{Comparison of attention matrices under various transformation bases for \textbf{iTransformer}. For a fair comparison, the \textbf{same} input series is used across bases on each dataset. `Identity' denotes no transformation, and the Haar wavelet is used as the wavelet basis.  In general, OrthoTrans yields higher-rank attention matrices and thus better preserves the representation space.}
   \label{fig:orthotrans_improve_rank_itrans}
\end{figure}

\begin{figure}[t]
   \centering
   \includegraphics[width=1.0\linewidth]{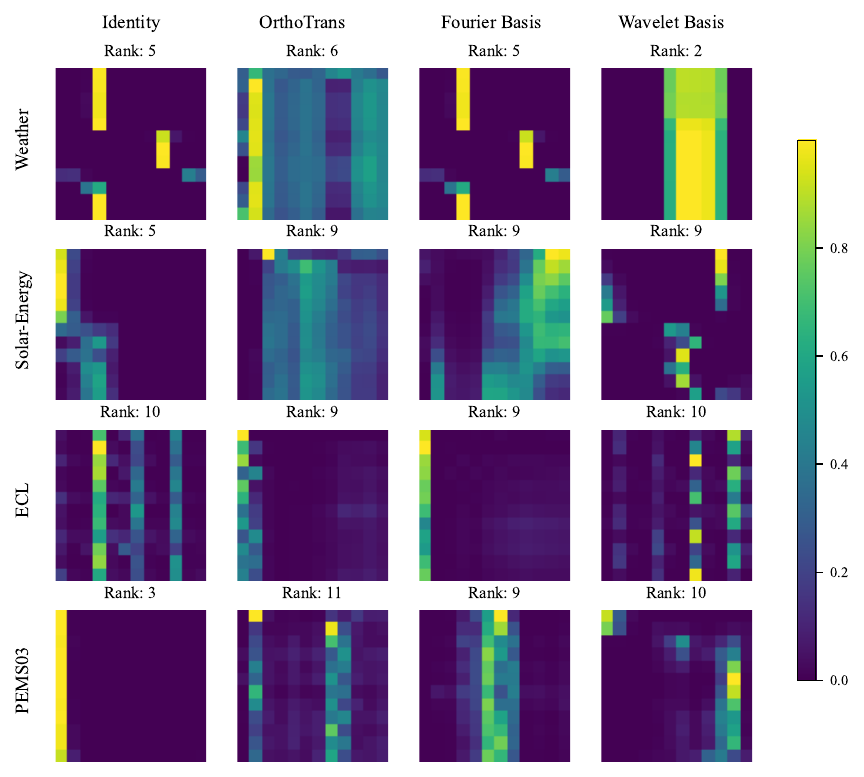}
   \caption{Comparison of attention matrices under various transformation bases for \textbf{PatchTST}. For a fair comparison, the \textbf{same} input series is used across all bases for each dataset. `Identity' denotes no transformation, and the Haar wavelet is used as the wavelet basis.  In general, OrthoTrans yields higher-rank attention matrices and thus better preserves the representation space.}
   \label{fig:orthotrans_improve_rank_patchtst}
\end{figure}

Figures~\ref{fig:orthotrans_improve_rank}–\ref{fig:orthotrans_improve_rank3} visualize the attention matrices of OLinear (with the NormLin module replaced by self-attention) under different transformation bases. Similar results for iTransformer and PatchTST are presented in Figures~\ref{fig:orthotrans_improve_rank_itrans} and \ref{fig:orthotrans_improve_rank_patchtst}, respectively.
As shown, OrthoTrans typically yields higher-rank attention matrices compared to DFT, wavelet transforms, or no transformation. As stated earlier, higher-rank attention matrices better preserve the representation space and can potentially enhance the model's expressive capacity \citep{flattentrans}. This may explain why integrating OrthoTrans as a plug-in consistently improves the performance of Transformer-based forecasters (as shown in Table~\ref{tab:base_iTrans}).

\section{Robustness under various random seeds} \label{robust_appendix}

Table~\ref{tab:robust} presents the standard deviations for different datasets and prediction lengths using seven random seeds. OLinear demonstrates strong robustness across independent runs. Furthermore, as shown in Table~\ref{tab:robust_compare}, our model exhibits better robustness than state-of-the-art Transformer-based models, TimeMixer++ and iTransformer, as measured by the 99\% confidence intervals. 

Note that averaging over $\mathcal{M}$ prediction lengths reduces the standard deviation by a factor of $\frac{1}{\sqrt{\mathcal{M}}}$, making the results more robust to the choice of random seeds. Therefore, we prefer to report the average results in this work to mitigate the influence of randomness.

\begin{table}[t]
\caption{Robustness of OLinear performance. Standard deviations are calculated over seven random seeds. \textit{S1} and \textit{S2} correspond to  $\mathrm{Input} -12, \mathrm{Predict} -\left \{ 3, 6, 9, 12\right \}$ and $\mathrm{Input} -36, \mathrm{Predict} -\left \{ 24, 36, 48, 60\right \}$, respectively. }
\label{tab:robust}
\centering
\setlength{\tabcolsep}{3pt}
\renewcommand{\arraystretch}{1.0} 
{\fontsize{9}{10}\selectfont
\begin{tabular}{@{}cccccccccc@{}}
\toprule
\multicolumn{2}{c}{Dataset}    & \multicolumn{2}{c}{ECL}           & \multicolumn{2}{c}{Traffic}    & \multicolumn{2}{c}{ETTm1}         & \multicolumn{2}{c}{Solar-Energy} \\ \midrule
\multicolumn{2}{c}{Metric}     & MSE              & MAE            & MSE            & MAE           & MSE             & MAE             & MSE             & MAE            \\ \midrule
\multirow{4}{*}{\rotatebox[origin=c]{90}{Horizon}}  & 96  & 0.131±4e-4       & 0.221±4e-4     & 0.398±3e-3     & 0.226±2e-4    & 0.302±7e-4      & 0.334±3e-4      & 0.179±6e-4      & 0.191±5e-4     \\
                         & 192 & 0.150±1e-3       & 0.238±1e-3     & 0.439±3e-3     & 0.241±4e-4    & 0.357±9e-4      & 0.363±5e-4      & 0.209±8e-4      & 0.213±2e-4     \\
                         & 336 & 0.165±1e-3       & 0.254±1e-3     & 0.464±4e-3     & 0.250±3e-4    & 0.387±2e-3      & 0.385±5e-4      & 0.231±8e-4      & 0.229±7e-5     \\
                         & 720 & 0.191±2e-3       & 0.279±2e-3     & 0.502±4e-3     & 0.270±4e-4    & 0.452±1e-3      & 0.426±6e-4      & 0.241±1e-3      & 0.236±4e-4     \\ \midrule
\multicolumn{2}{c}{Dataset}    & \multicolumn{2}{c}{Weather}       & \multicolumn{2}{c}{PEMS03}     & \multicolumn{2}{c}{NASDAQ (S2)}   & \multicolumn{2}{c}{Wiki (S1)}    \\ \midrule
\multicolumn{2}{c}{Metric}     & MSE              & MAE            & MSE            & MAE           & MSE             & MAE             & MSE             & MAE            \\ \midrule
\multirow{4}{*}{\rotatebox[origin=c]{90}{Horizon}} & H1  & 0.153±1e-3       & 0.190±1e-3     & 0.060±3e-4     & 0.159±3e-4    & 0.121±1e-3      & 0.216±9e-4      & 6.161±1e-2      & 0.368±7e-4     \\
                         & H2  & 0.200±2e-3       & 0.235±2e-3     & 0.078±6e-4     & 0.179±5e-4    & 0.163±7e-4      & 0.261±9e-4      & 6.453±9e-3      & 0.385±1e-3     \\
                         & H3  & 0.258±3e-3       & 0.280±2e-3     & 0.104±6e-4     & 0.210±6e-4    & 0.205±2e-3      & 0.296±2e-3      & 6.666±6e-3      & 0.398±1e-3     \\
                         & H4  & 0.337±4e-3       & 0.333±2e-3     & 0.140±2e-3     & 0.247±1e-3    & 0.259±2e-3      & 0.336±2e-3      & 6.834±4e-3      & 0.406±4e-4     \\ \midrule
\multicolumn{2}{c}{Dataset}    & \multicolumn{2}{c}{DowJones (S2)} & \multicolumn{2}{c}{SP500 (S2)} & \multicolumn{2}{c}{CarSales (S1)} & \multicolumn{2}{c}{Power (S1)}   \\ \midrule
\multicolumn{2}{c}{Metric}     & MSE              & MAE            & MSE            & MAE           & MSE             & MAE             & MSE             & MAE            \\ \midrule
\multirow{4}{*}{\rotatebox[origin=c]{90}{Horizon}} & H1  & 7.432±3e-2       & 0.664±9e-4     & 0.155±2e-3     & 0.271±2e-3    & 0.303±2e-3      & 0.277±1e-3      & 0.864±7e-3      & 0.688±4e-3     \\
                         & H2  & 10.848±7e-2      & 0.799±1e-3     & 0.209±2e-3     & 0.317±2e-3    & 0.315±2e-3      & 0.285±2e-3      & 0.991±7e-3      & 0.742±3e-3     \\
                         & H3  & 14.045±1e-1      & 0.914±1e-3     & 0.258±2e-3     & 0.358±1e-3    & 0.327±8e-4      & 0.293±8e-4      & 1.062±1e-2      & 0.770±4e-3     \\
                         & H4  & 16.959±8e-2      & 1.017±3e-3     & 0.305±3e-3     & 0.387±2e-3    & 0.336±5e-4      & 0.301±4e-4      & 1.119±2e-2      & 0.789±6e-3     \\ \bottomrule
\end{tabular}
}
\end{table}

\begin{table}[t]
\caption{99\% confidence intervals of the average performance across four prediction lengths, computed using three times the standard deviation from seven random seeds. The smallest standard deviations are highlighted in {\color[HTML]{FF0000} \textbf{bold}}. \textit{OLinear-C} denotes the variant of OLinear where the weight matrix in NormLin is set to $\mathrm{Softmax} \left ( \mathrm{CorrMat} _v \right )$, with $\mathrm{CorrMat} _v$ being the multivariate correlation matrix.}
\label{tab:robust_compare}
\centering
\setlength{\tabcolsep}{2.5pt}
\renewcommand{\arraystretch}{1.2} 
{\fontsize{8}{10}\selectfont
\begin{tabular}{@{}ccccccccc@{}}
\toprule
Model        & \multicolumn{2}{c}{\begin{tabular}[c]{@{}c@{}}OLinear\\ (Ours)\end{tabular}} & \multicolumn{2}{c}{\begin{tabular}[c]{@{}c@{}}OLinear-C\\ (Ours)\end{tabular}} & 
\multicolumn{2}{c}{\begin{tabular}[c]{@{}c@{}}TimeMixer++\\ \citeyear{timemixer++} \end{tabular}} 
& \multicolumn{2}{c}{\begin{tabular}[c]{@{}c@{}}iTransformer\\ \citeyear{itransformer} \end{tabular}} \\ \midrule
Metric      & MSE                                     & MAE                                    & MSE                                      & MAE                                     & MSE                                   & MAE                                  & MSE                                   & MAE                                   \\ \midrule
Weather      & 0.237±0.006                             & 0.260±{\color[HTML]{FF0000} \textbf{0.003}}                            & 0.238±{\color[HTML]{FF0000} \textbf{0.005}}                              & 0.259±0.004                             & 0.226±0.008                           & 0.262±0.007                          & 0.258±0.009                           & 0.278±0.006                           \\
Solar  & 0.215±{\color[HTML]{FF0000} \textbf{0.001}}                              & 0.217±0.001                            & 0.215±{\color[HTML]{FF0000} \textbf{0.001}}                               & 0.217±{\color[HTML]{FF0000} \textbf{3e-4}}                              & 0.203±{\color[HTML]{FF0000} \textbf{0.001}}                           & 0.238±0.010                          & 0.233±0.009                           & 0.262±0.007                           \\
ECL          & 0.159±{\color[HTML]{FF0000} \textbf{0.001}}                             & 0.248±0.002                            & 0.161±0.006                              & 0.249±0.005                             & 0.165±0.011                           & 0.253±{\color[HTML]{FF0000} \textbf{0.001}}                          & 0.178±0.002                           & 0.270±0.005                           \\
Traffic      & 0.451±{\color[HTML]{FF0000} \textbf{0.003}}                             & 0.247±{\color[HTML]{FF0000} \textbf{0.001}}                            & 0.451±0.006                              & 0.247±{\color[HTML]{FF0000} \textbf{0.001}}                            & 0.416±0.015                           & 0.264±0.013                          & 0.428±0.008                           & 0.282±0.002                           \\
ETTh1        & 0.424±0.003                             & 0.424±0.002                            & 0.424±{\color[HTML]{FF0000} \textbf{0.002}}                              & 0.424±{\color[HTML]{FF0000} \textbf{0.001}}                             & 0.419±0.011                           & 0.432±0.015                          & 0.454±0.004                           & 0.447±0.007                           \\
ETTh2        & 0.367±{\color[HTML]{FF0000} \textbf{0.002}}                             & 0.388±0.002                            & 0.368±{\color[HTML]{FF0000} \textbf{0.002}}                              & 0.389±{\color[HTML]{FF0000} \textbf{0.001}}                             & 0.339±0.009                           & 0.380±0.002                          & 0.383±0.004                           & 0.407±0.007                           \\
ETTm1        & 0.374±{\color[HTML]{FF0000} \textbf{0.001}}                             & 0.377±{\color[HTML]{FF0000} \textbf{0.001}}                            & 0.375±{\color[HTML]{FF0000} \textbf{0.001}}                              & 0.378±{\color[HTML]{FF0000} \textbf{0.001}}                             & 0.369±0.005                           & 0.378±0.007                          & 0.407±0.004                           & 0.410±0.009                           \\
ETTm2        & 0.270±{\color[HTML]{FF0000} \textbf{3e-4}}                              & 0.313±{\color[HTML]{FF0000} \textbf{2e-4}}                             & 0.270±4e-4                               & 0.313±{\color[HTML]{FF0000} \textbf{2e-4}}                              & 0.269±0.002                           & 0.320±0.012                          & 0.288±0.010                           & 0.332±0.003                           \\ \bottomrule
\end{tabular}
}
\end{table}

\section{Full results}  \label{full_results}

Table~\ref{tab:tables_paper_appendix} lists the simplified tables from the main text and their full versions in the appendix.

\begin{table}[ht]
\caption{Overview of tables in the main paper and their full versions in the appendix.}
\label{tab:tables_paper_appendix}
\centering
\setlength{\tabcolsep}{30pt}
\renewcommand{\arraystretch}{1.2} 
{\fontsize{9}{10}\selectfont
\begin{tabular}{@{}ccc@{}}
\toprule
Tables in   paper & Tables in Appendix & Content               \\ \midrule
Table~\ref{tab:long_term}           & Table~\ref{tab:full_long}            & Long-term forecasting \\
Table ~\ref{tab:short_term}           & Tables~\ref{tab:full_short_part1} and \ref{tab:full_short_part2}            & Short-term forecasting \\
Table~\ref{tab:base}           & Table~\ref{tab:base_full}            & Ablation studies on various bases \\

Table~\ref{tab:base_iTrans}          & Table~\ref{tab:base_iTrans_full}            & OrthoTrans as a plug-in \\
Table~\ref{tab:OLinear-c_short}           & Table~\ref{tab:ortho-c-full}            & Performance of OLinear-C \\ 

Table~\ref{tab:var_temp}           & Table~\ref{tab:var_temp_full}            & Ablation studies of OLinear \\ 
Table~\ref{tab:normlin_vs_attn}           & Table~\ref{tab:normlin_vs_attn_full}            & NormLin vs self-attention and its variants \\ 
Table~\ref{tab:normlin_itrans}           & Table~\ref{tab:normlin_itrans_full}            & NormLin  as a plug-in \\ 

\bottomrule
\end{tabular}
}
\end{table}

\begin{table}[p]
\caption{Full results for the long-term forecasting task. The lookback length $T$ is set to 96 for all baselines. \textit{Avg} means the average results across four prediction lengths. This table presents the full version of Table~\ref{tab:long_term}.}
\label{tab:full_long}
\centering
\setlength{\tabcolsep}{2pt}
\renewcommand{\arraystretch}{0.9} 
{\fontsize{5}{7}\selectfont
\begin{tabular}{@{}cccccccccccccccccccccccccc@{}}
\toprule
\multicolumn{2}{c}{Category}         & \multicolumn{10}{c}{Linear-Based}                                                                                                                                                                                                                                                                                                                                                                              & \multicolumn{12}{c}{Transformer-Based}                                                                                                                                                                                                                                                                                                                                                                                                                                                               & \multicolumn{2}{c}{TCN-Based}                                                  \\ \midrule
\multicolumn{2}{c}{Model}           & \multicolumn{2}{c}{\begin{tabular}[c]{@{}c@{}}OLinear\\    (Ours) \end{tabular}}                                                      & 
\multicolumn{2}{c}{\begin{tabular}[c]{@{}c@{}}TimeMixer\\    \citeyear{timemixer}  \end{tabular}} & 
\multicolumn{2}{c}{\begin{tabular}[c]{@{}c@{}}FilterNet\\ \citeyear{filternet} \end{tabular}} & 
\multicolumn{2}{c}{\begin{tabular}[c]{@{}c@{}}FITS\\      \citeyear{fits} \end{tabular}} & 
\multicolumn{2}{c}{\begin{tabular}[c]{@{}c@{}}DLinear\\      \citeyear{linear} \end{tabular}} & 
\multicolumn{2}{c}{\begin{tabular}[c]{@{}c@{}}TimeMixer++\\      \citeyear{timemixer++} \end{tabular}} & 
\multicolumn{2}{c}{\begin{tabular}[c]{@{}c@{}}Leddam\\      \citeyear{Leddam_icml} \end{tabular}} & 
\multicolumn{2}{c}{\begin{tabular}[c]{@{}c@{}}CARD\\      \citeyear{card} \end{tabular}}    & 
\multicolumn{2}{c}{\begin{tabular}[c]{@{}c@{}}Fredformer\\      \citeyear{fredformer} \end{tabular}} & 
\multicolumn{2}{c}{\begin{tabular}[c]{@{}c@{}}iTrans.\\      \citeyear{itransformer} \end{tabular}} & 
\multicolumn{2}{c}{\begin{tabular}[c]{@{}c@{}}PatchTST\\      \citeyear{patchtst} \end{tabular}} & 
\multicolumn{2}{c}{\begin{tabular}[c]{@{}c@{}}TimesNet\\  \citeyear{timesnet}  \end{tabular}} \\ \midrule
\multicolumn{2}{c}{Metric}           & MSE                                   & MAE                                   & MSE                                    & MAE                                    & MSE                                    & MAE                                    & MSE                                                & MAE                   & MSE                                                  & MAE                    & MSE                                     & MAE                                     & MSE                                    & MAE                                 & MSE                                   & MAE                                   & MSE                                     & MAE                                    & MSE                                                  & MAE                    & MSE                                    & MAE                                   & MSE                                    & MAE                                   \\ \midrule
\multirow{5}{*}{\rotatebox[origin=c]{90}{ETTm1}}                      & 96  & {\color[HTML]{FF0000} \textbf{0.302}} & {\color[HTML]{FF0000} \textbf{0.334}} & 0.320                                                 & 0.357                   & 0.321                                  & 0.361                                  & 0.353                                              & 0.375                 & 0.345                                                & 0.372                  & {\color[HTML]{0000FF} {\ul 0.310}}      & {\color[HTML]{FF0000} \textbf{0.334}}   & 0.319                                  & 0.359                               & 0.316                                 & {\color[HTML]{0000FF} {\ul 0.347}}    & 0.326                                   & 0.361                                  & 0.334                                                & 0.368                  & 0.329                                  & 0.367                                 & 0.338                                  & 0.375                                 \\
                               & 192 & {\color[HTML]{0000FF} {\ul 0.357}}    & {\color[HTML]{0000FF} {\ul 0.363}}    & 0.361                                                 & 0.381                   & 0.367                                  & 0.387                                  & 0.486                                              & 0.445                 & 0.380                                                & 0.389                  & {\color[HTML]{FF0000} \textbf{0.348}}   & {\color[HTML]{FF0000} \textbf{0.362}}   & 0.369                                  & 0.383                               & 0.363                                 & 0.370                                 & 0.363                                   & 0.380                                  & 0.377                                                & 0.391                  & 0.367                                  & 0.385                                 & 0.374                                  & 0.387                                 \\
                               & 336 & {\color[HTML]{0000FF} {\ul 0.387}}    & {\color[HTML]{FF0000} \textbf{0.385}} & 0.390                                                 & 0.404                   & 0.401                                  & 0.409                                  & 0.531                                              & 0.475                 & 0.413                                                & 0.413                  & {\color[HTML]{FF0000} \textbf{0.376}}   & 0.391                                   & 0.394                                  & 0.402                               & 0.392                                 & {\color[HTML]{0000FF} {\ul 0.390}}    & 0.395                                   & 0.403                                  & 0.426                                                & 0.420                  & 0.399                                  & 0.410                                 & 0.410                                  & 0.411                                 \\
                               & 720 & {\color[HTML]{0000FF} {\ul 0.452}}    & 0.426                                 & 0.454                                                 & 0.441                   & 0.477                                  & 0.448                                  & 0.600                                              & 0.513                 & 0.474                                                & 0.453                  & {\color[HTML]{FF0000} \textbf{0.440}}   & {\color[HTML]{FF0000} \textbf{0.423}}   & 0.460                                  & 0.442                               & 0.458                                 & {\color[HTML]{0000FF} {\ul 0.425}}    & 0.453                                   & 0.438                                  & 0.491                                                & 0.459                  & 0.454                                  & 0.439                                 & 0.478                                  & 0.450                                 \\ \cmidrule(l){2-26} 
      & Avg & {\color[HTML]{0000FF} {\ul 0.374}}    & {\color[HTML]{FF0000} \textbf{0.377}} & 0.381                                                 & 0.395                   & 0.392                                  & 0.401                                  & 0.493                                              & 0.452                 & 0.403                                                & 0.407                  & {\color[HTML]{FF0000} \textbf{0.369}}   & {\color[HTML]{0000FF} {\ul 0.378}}      & 0.386                                  & 0.397                               & 0.383                                 & 0.384                                 & 0.384                                   & 0.395                                  & 0.407                                                & 0.410                  & 0.387                                  & 0.400                                 & 0.400                                  & 0.406                                 \\ \midrule
\multirow{5}{*}{\rotatebox[origin=c]{90}{ETTm2}}    & 96  & {\color[HTML]{FF0000} \textbf{0.169}} & 0.249                                 & 0.175                                                 & 0.258                   & 0.175                                  & 0.258                                  & 0.182                                              & 0.266                 & 0.193                                                & 0.292                  & {\color[HTML]{0000FF} {\ul 0.170}}      & {\color[HTML]{FF0000} \textbf{0.245}}   & 0.176                                  & 0.257                               & {\color[HTML]{FF0000} \textbf{0.169}} & {\color[HTML]{0000FF} {\ul 0.248}}    & 0.177                                   & 0.259                                  & 0.180                                                & 0.264                  & 0.175                                  & 0.259                                 & 0.187                                  & 0.267                                 \\
                               & 192 & {\color[HTML]{0000FF} {\ul 0.232}}    & {\color[HTML]{FF0000} \textbf{0.290}} & 0.237                                                 & 0.299                   & 0.240                                  & 0.301                                  & 0.253                                              & 0.312                 & 0.284                                                & 0.362                  & {\color[HTML]{FF0000} \textbf{0.229}}   & {\color[HTML]{0000FF} {\ul 0.291}}      & 0.243                                  & 0.303                               & 0.234                                 & 0.292                                 & 0.243                                   & 0.301                                  & 0.250                                                & 0.309                  & 0.241                                  & 0.302                                 & 0.249                                  & 0.309                                 \\
                               & 336 & {\color[HTML]{FF0000} \textbf{0.291}} & {\color[HTML]{FF0000} \textbf{0.328}} & 0.298                                                 & 0.340                   & 0.311                                  & 0.347                                  & 0.313                                              & 0.349                 & 0.369                                                & 0.427                  & 0.303                                   & 0.343                                   & 0.303                                  & 0.341                               & {\color[HTML]{0000FF} {\ul 0.294}}    & {\color[HTML]{0000FF} {\ul 0.339}}    & 0.302                                   & 0.340                                  & 0.311                                                & 0.348                  & 0.305                                  & 0.343                                 & 0.321                                  & 0.351                                 \\
                               & 720 & {\color[HTML]{0000FF} {\ul 0.389}}    & {\color[HTML]{FF0000} \textbf{0.387}} & 0.391                                                 & 0.396                   & 0.414                                  & 0.405                                  & 0.416                                              & 0.406                 & 0.554                                                & 0.522                  & {\color[HTML]{FF0000} \textbf{0.373}}   & 0.399                                   & 0.400                                  & 0.398                               & 0.390                                 & {\color[HTML]{0000FF} {\ul 0.388}}    & 0.397                                   & 0.396                                  & 0.412                                                & 0.407                  & 0.402                                  & 0.400                                 & 0.408                                  & 0.403                                 \\ \cmidrule(l){2-26} 
       & Avg & {\color[HTML]{0000FF} {\ul 0.270}}    & {\color[HTML]{FF0000} \textbf{0.313}} & 0.275                                                 & 0.323                   & 0.285                                  & 0.328                                  & 0.291                                              & 0.333                 & 0.350                                                & 0.401                  & {\color[HTML]{FF0000} \textbf{0.269}}   & 0.320                                   & 0.281                                  & 0.325                               & 0.272                                 & {\color[HTML]{0000FF} {\ul 0.317}}    & 0.279                                   & 0.324                                  & 0.288                                                & 0.332                  & 0.281                                  & 0.326                                 & 0.291                                  & 0.333                                 \\ \midrule
\multirow{5}{*}{\rotatebox[origin=c]{90}{ETTh1}}                        & 96  & {\color[HTML]{FF0000} \textbf{0.360}} & {\color[HTML]{FF0000} \textbf{0.382}} & 0.375                                                 & 0.400                   & 0.382                                  & 0.402                                  & 0.385                                              & 0.394                 & 0.386                                                & 0.400                  & {\color[HTML]{0000FF} {\ul 0.361}}      & 0.403                                   & 0.377                                  & 0.394                               & 0.383                                 & {\color[HTML]{0000FF} {\ul 0.391}}    & 0.373                                   & 0.392                                  & 0.386                                                & 0.405                  & 0.414                                  & 0.419                                 & 0.384                                  & 0.402                                 \\
                               & 192 & {\color[HTML]{FF0000} \textbf{0.416}} & {\color[HTML]{FF0000} \textbf{0.414}} & 0.429                                                 & 0.421                   & 0.430                                  & 0.429                                  & 0.434                                              & 0.422                 & 0.437                                                & 0.432                  & {\color[HTML]{FF0000} \textbf{0.416}}   & 0.441                                   & {\color[HTML]{0000FF} {\ul 0.424}}     & 0.422                               & 0.435                                 & {\color[HTML]{0000FF} {\ul 0.420}}    & 0.433                                   & {\color[HTML]{0000FF} {\ul 0.420}}     & 0.441                                                & 0.436                  & 0.460                                  & 0.445                                 & 0.436                                  & 0.429                                 \\
                               & 336 & {\color[HTML]{0000FF} {\ul 0.457}}    & 0.438                                 & 0.484                                                 & 0.458                   & 0.472                                  & 0.451                                  & 0.476                                              & 0.444                 & 0.481                                                & 0.459                  & {\color[HTML]{FF0000} \textbf{0.430}}   & {\color[HTML]{FF0000} \textbf{0.434}}   & 0.459                                  & 0.442                               & 0.479                                 & 0.442                                 & 0.470                                   & {\color[HTML]{0000FF} {\ul 0.437}}     & 0.487                                                & 0.458                  & 0.501                                  & 0.466                                 & 0.491                                  & 0.469                                 \\
                               & 720 & {\color[HTML]{FF0000} \textbf{0.463}} & 0.462                                 & 0.498                                                 & 0.482                   & 0.481                                  & 0.473                                  & {\color[HTML]{0000FF} {\ul 0.465}}                 & 0.462                 & 0.519                                                & 0.516                  & 0.467                                   & {\color[HTML]{FF0000} \textbf{0.451}}   & {\color[HTML]{FF0000} \textbf{0.463}}  & 0.459                               & 0.471                                 & 0.461                                 & 0.467                                   & {\color[HTML]{0000FF} {\ul 0.456}}     & 0.503                                                & 0.491                  & 0.500                                  & 0.488                                 & 0.521                                  & 0.500                                 \\ \cmidrule(l){2-26} 
      & Avg & {\color[HTML]{0000FF} {\ul 0.424}}    & {\color[HTML]{FF0000} \textbf{0.424}} & 0.447                                                 & 0.440                   & 0.441                                  & 0.439                                  & 0.440                                              & 0.431                 & 0.456                                                & 0.452                  & {\color[HTML]{FF0000} \textbf{0.419}}   & 0.432                                   & 0.431                                  & 0.429                               & 0.442                                 & 0.429                                 & 0.435                                   & {\color[HTML]{0000FF} {\ul 0.426}}     & 0.454                                                & 0.447                  & 0.469                                  & 0.454                                 & 0.458                                  & 0.450                                 \\ \midrule
\multirow{5}{*}{\rotatebox[origin=c]{90}{ETTh2}}   & 96  & 0.284                                 & {\color[HTML]{0000FF} {\ul 0.329}}    & 0.289                                                 & 0.341                   & 0.293                                  & 0.343                                  & 0.292                                              & 0.340                 & 0.333                                                & 0.387                  & {\color[HTML]{FF0000} \textbf{0.276}}   & {\color[HTML]{FF0000} \textbf{0.328}}   & 0.292                                  & 0.343                               & {\color[HTML]{0000FF} {\ul 0.281}}    & 0.330                                 & 0.293                                   & 0.342                                  & 0.297                                                & 0.349                  & 0.292                                  & 0.342                                 & 0.340                                  & 0.374                                 \\
                               & 192 & {\color[HTML]{0000FF} {\ul 0.360}}    & {\color[HTML]{FF0000} \textbf{0.379}} & 0.372                                                 & 0.392                   & 0.374                                  & 0.396                                  & 0.377                                              & 0.391                 & 0.477                                                & 0.476                  & {\color[HTML]{FF0000} \textbf{0.342}}   & {\color[HTML]{FF0000} \textbf{0.379}}   & 0.367                                  & 0.389                               & 0.363                                 & {\color[HTML]{0000FF} {\ul 0.381}}    & 0.371                                   & 0.389                                  & 0.380                                                & 0.400                  & 0.387                                  & 0.400                                 & 0.402                                  & 0.414                                 \\
                               & 336 & 0.409                                 & 0.415                                 & 0.386                                                 & 0.414                   & 0.417                                  & 0.430                                  & 0.416                                              & 0.425                 & 0.594                                                & 0.541                  & {\color[HTML]{FF0000} \textbf{0.346}}   & {\color[HTML]{FF0000} \textbf{0.398}}   & 0.412                                  & 0.424                               & 0.411                                 & 0.418                                 & {\color[HTML]{0000FF} {\ul 0.382}}      & {\color[HTML]{0000FF} {\ul 0.409}}     & 0.428                                                & 0.432                  & 0.426                                  & 0.433                                 & 0.452                                  & 0.452                                 \\
                               & 720 & 0.415                                 & {\color[HTML]{0000FF} {\ul 0.431}}    & {\color[HTML]{0000FF} {\ul 0.412}}                    & 0.434                   & 0.449                                  & 0.460                                  & 0.418                                              & 0.437                 & 0.831                                                & 0.657                  & {\color[HTML]{FF0000} \textbf{0.392}}   & {\color[HTML]{FF0000} \textbf{0.415}}   & 0.419                                  & 0.438                               & 0.416                                 & 0.431                                 & 0.415                                   & 0.434                                  & 0.427                                                & 0.445                  & 0.431                                  & 0.446                                 & 0.462                                  & 0.468                                 \\ \cmidrule(l){2-26} 
      & Avg & 0.367                                 & {\color[HTML]{0000FF} {\ul 0.388}}    & {\color[HTML]{0000FF} {\ul 0.365}}                    & 0.395                   & 0.383                                  & 0.407                                  & 0.376                                              & 0.398                 & 0.559                                                & 0.515                  & {\color[HTML]{FF0000} \textbf{0.339}}   & {\color[HTML]{FF0000} \textbf{0.380}}   & 0.373                                  & 0.399                               & 0.368                                 & 0.390                                 & {\color[HTML]{0000FF} {\ul 0.365}}      & 0.393                                  & 0.383                                                & 0.407                  & 0.384                                  & 0.405                                 & 0.414                                  & 0.427                                 \\ \midrule
\multirow{5}{*}{\rotatebox[origin=c]{90}{ECL}}   & 96  & {\color[HTML]{FF0000} \textbf{0.131}} & {\color[HTML]{FF0000} \textbf{0.221}} & 0.153                                  & 0.247                                  & 0.147                                  & 0.245                                  & 0.198                                              & 0.274                 & 0.197                                                & 0.282                  & {\color[HTML]{0000FF} {\ul 0.135}}      & {\color[HTML]{0000FF} {\ul 0.222}}      & 0.141                                  & 0.235                               & 0.141                                 & 0.233                                 & 0.147                                   & 0.241                                  & 0.148                                                & 0.240                  & 0.161                                  & 0.250                                 & 0.168                                  & 0.272                                 \\
                               & 192 & {\color[HTML]{0000FF} {\ul 0.150}}    & {\color[HTML]{0000FF} {\ul 0.238}}    & 0.166                                  & 0.256                                  & 0.160                                  & 0.250                                  & 0.363                                              & 0.422                 & 0.196                                                & 0.285                  & {\color[HTML]{FF0000} \textbf{0.147}}   & {\color[HTML]{FF0000} \textbf{0.235}}   & 0.159                                  & 0.252                               & 0.160                                 & 0.250                                 & 0.165                                   & 0.258                                  & 0.162                                                & 0.253                  & 0.199                                  & 0.289                                 & 0.184                                  & 0.289                                 \\
                               & 336 & {\color[HTML]{0000FF} {\ul 0.165}}    & {\color[HTML]{0000FF} {\ul 0.254}}    & 0.185                                  & 0.277                                  & 0.173                                  & 0.267                                  & 0.444                                              & 0.490                 & 0.209                                                & 0.301                  & {\color[HTML]{FF0000} \textbf{0.164}}   & {\color[HTML]{FF0000} \textbf{0.245}}   & 0.173                                  & 0.268                               & 0.173                                 & 0.263                                 & 0.177                                   & 0.273                                  & 0.178                                                & 0.269                  & 0.215                                  & 0.305                                 & 0.198                                  & 0.300                                 \\
                               & 720 & {\color[HTML]{FF0000} \textbf{0.191}} & {\color[HTML]{FF0000} \textbf{0.279}} & 0.225                                  & 0.310                                  & 0.210                                  & 0.309                                  & 0.532                                              & 0.551                 & 0.245                                                & 0.333                  & 0.212                                   & 0.310                                   & 0.201                                  & 0.295                               & {\color[HTML]{0000FF} {\ul 0.197}}    & {\color[HTML]{0000FF} {\ul 0.284}}    & 0.213                                   & 0.304                                  & 0.225                                                & 0.317                  & 0.256                                  & 0.337                                 & 0.220                                  & 0.320                                 \\ \cmidrule(l){2-26} 
         & Avg & {\color[HTML]{FF0000} \textbf{0.159}} & {\color[HTML]{FF0000} \textbf{0.248}} & 0.182                                  & 0.273                                  & 0.173                                  & 0.268                                  & 0.384                                              & 0.434                 & 0.212                                                & 0.300                  & {\color[HTML]{0000FF} {\ul 0.165}}      & {\color[HTML]{0000FF} {\ul 0.253}}      & 0.169                                  & 0.263                               & 0.168                                 & 0.258                                 & 0.176                                   & 0.269                                  & 0.178                                                & 0.270                  & 0.208                                  & 0.295                                 & 0.192                                  & 0.295                                 \\ \midrule
\multirow{5}{*}{\rotatebox[origin=c]{90}{Exchange}}      & 96  & {\color[HTML]{FF0000} \textbf{0.082}} & {\color[HTML]{FF0000} \textbf{0.200}} & 0.086                                  & 0.205                                  & 0.091                                  & 0.211                                  & 0.087                                              & 0.208                 & 0.088                                                & 0.218                  & 0.085                                   & 0.214                                   & 0.086                                  & 0.207                               & {\color[HTML]{0000FF} {\ul 0.084}}    & {\color[HTML]{0000FF} {\ul 0.202}}    & {\color[HTML]{0000FF} {\ul 0.084}}      & {\color[HTML]{0000FF} {\ul 0.202}}     & 0.086                                                & 0.206                  & 0.088                                  & 0.205                                 & 0.107                                  & 0.234                                 \\
                               & 192 & {\color[HTML]{FF0000} \textbf{0.171}} & {\color[HTML]{FF0000} \textbf{0.293}} & 0.193                                  & 0.312                                  & 0.186                                  & 0.305                                  & 0.185                                              & 0.306                 & 0.176                                                & 0.315                  & {\color[HTML]{0000FF} {\ul 0.175}}      & 0.313                                   & {\color[HTML]{0000FF} {\ul 0.175}}     & 0.301                               & 0.179                                 & {\color[HTML]{0000FF} {\ul 0.298}}    & 0.178                                   & 0.302                                  & 0.177                                                & 0.299                  & 0.176                                  & 0.299                                 & 0.226                                  & 0.344                                 \\
                               & 336 & 0.331                                 & 0.414                                 & 0.356                                  & 0.433                                  & 0.380                                  & 0.449                                  & 0.342                                              & 0.425                 & {\color[HTML]{0000FF} {\ul 0.313}}                   & 0.427                  & 0.316                                   & 0.420                                   & 0.325                                  & 0.415                               & 0.333                                 & 0.418                                 & 0.319                                   & {\color[HTML]{0000FF} {\ul 0.408}}     & 0.331                                                & 0.417                  & {\color[HTML]{FF0000} \textbf{0.301}}  & {\color[HTML]{FF0000} \textbf{0.397}} & 0.367                                  & 0.448                                 \\
                               & 720 & 0.837                                 & 0.688                                 & 0.912                                  & 0.712                                  & 0.896                                  & 0.712                                  & 0.846                                              & 0.694                 & 0.839                                                & 0.695                  & 0.851                                   & 0.689                                   & {\color[HTML]{0000FF} {\ul 0.831}}     & {\color[HTML]{0000FF} {\ul 0.686}}  & 0.851                                 & 0.691                                 & {\color[HTML]{FF0000} \textbf{0.749}}   & {\color[HTML]{FF0000} \textbf{0.651}}  & 0.847                                                & 0.691                  & 0.901                                  & 0.714                                 & 0.964                                  & 0.746                                 \\ \cmidrule(l){2-26} 
  & Avg & 0.355                                 & {\color[HTML]{0000FF} {\ul 0.399}}    & 0.387                                  & 0.416                                  & 0.388                                  & 0.419                                  & 0.365                                              & 0.408                 & {\color[HTML]{0000FF} {\ul 0.354}}                   & 0.414                  & 0.357                                   & 0.409                                   & {\color[HTML]{0000FF} {\ul 0.354}}     & 0.402                               & 0.362                                 & 0.402                                 & {\color[HTML]{FF0000} \textbf{0.333}}   & {\color[HTML]{FF0000} \textbf{0.391}}  & 0.360                                                & 0.403                  & 0.367                                  & 0.404                                 & 0.416                                  & 0.443                                 \\ \midrule
\multirow{5}{*}{\rotatebox[origin=c]{90}{Traffic}}    & 96  & 0.398                                 & {\color[HTML]{FF0000} \textbf{0.226}} & 0.462                                  & 0.285                                  & 0.430                                  & 0.294                                  & 0.601                                              & 0.361                 & 0.650                                                & 0.396                  & {\color[HTML]{FF0000} \textbf{0.392}}   & {\color[HTML]{0000FF} {\ul 0.253}}      & 0.426                                  & 0.276                               & 0.419                                 & 0.269                                 & 0.406                                   & 0.277                                  & {\color[HTML]{0000FF} {\ul 0.395}}                   & 0.268                  & 0.446                                  & 0.283                                 & 0.593                                  & 0.321                                 \\
                               & 192 & 0.439                                 & {\color[HTML]{FF0000} \textbf{0.241}} & 0.473                                  & 0.296                                  & 0.452                                  & 0.307                                  & 0.603                                              & 0.365                 & 0.598                                                & 0.370                  & {\color[HTML]{FF0000} \textbf{0.402}}   & {\color[HTML]{0000FF} {\ul 0.258}}      & 0.458                                  & 0.289                               & 0.443                                 & 0.276                                 & 0.426                                   & 0.290                                  & {\color[HTML]{0000FF} {\ul 0.417}}                   & 0.276                  & 0.540                                  & 0.354                                 & 0.617                                  & 0.336                                 \\
                               & 336 & 0.464                                 & {\color[HTML]{FF0000} \textbf{0.250}} & 0.498                                  & 0.296                                  & 0.470                                  & 0.316                                  & 0.609                                              & 0.366                 & 0.605                                                & 0.373                  & {\color[HTML]{FF0000} \textbf{0.428}}   & {\color[HTML]{0000FF} {\ul 0.263}}      & 0.486                                  & 0.297                               & 0.460                                 & 0.283                                 & 0.437                                   & 0.292                                  & {\color[HTML]{0000FF} {\ul 0.433}}                   & 0.283                  & 0.551                                  & 0.358                                 & 0.629                                  & 0.336                                 \\
                               & 720 & 0.502                                 & {\color[HTML]{FF0000} \textbf{0.270}} & 0.506                                  & 0.313                                  & 0.498                                  & 0.323                                  & 0.648                                              & 0.387                 & 0.645                                                & 0.394                  & {\color[HTML]{FF0000} \textbf{0.441}}   & {\color[HTML]{0000FF} {\ul 0.282}}      & 0.498                                  & 0.313                               & 0.490                                 & 0.299                                 & {\color[HTML]{0000FF} {\ul 0.462}}      & 0.305                                  & 0.467                                                & 0.302                  & 0.586                                  & 0.375                                 & 0.640                                  & 0.350                                 \\ \cmidrule(l){2-26} 
      & Avg & 0.451                                 & {\color[HTML]{FF0000} \textbf{0.247}} & 0.485                                  & 0.298                                  & 0.463                                  & 0.310                                  & 0.615                                              & 0.370                 & 0.625                                                & 0.383                  & {\color[HTML]{FF0000} \textbf{0.416}}   & {\color[HTML]{0000FF} {\ul 0.264}}      & 0.467                                  & 0.294                               & 0.453                                 & 0.282                                 & 0.433                                   & 0.291                                  & {\color[HTML]{0000FF} {\ul 0.428}}                   & 0.282                  & 0.531                                  & 0.343                                 & 0.620                                  & 0.336                                 \\ \midrule
\multirow{5}{*}{\rotatebox[origin=c]{90}{Weather}}    & 96  & {\color[HTML]{0000FF} {\ul 0.153}}    & {\color[HTML]{0000FF} {\ul 0.190}}    & 0.163                                                 & 0.209                   & 0.162                                  & 0.207                                  & 0.196                                              & 0.236                 & 0.196                                                & 0.255                  & 0.155                                   & 0.205                                   & 0.156                                  & 0.202                               & {\color[HTML]{FF0000} \textbf{0.150}} & {\color[HTML]{FF0000} \textbf{0.188}} & 0.163                                   & 0.207                                  & 0.174                                                & 0.214                  & 0.177                                  & 0.218                                 & 0.172                                  & 0.220                                 \\
                               & 192 & {\color[HTML]{FF0000} \textbf{0.200}} & {\color[HTML]{FF0000} \textbf{0.235}} & 0.208                                                 & 0.250                   & 0.210                                  & 0.250                                  & 0.240                                              & 0.271                 & 0.237                                                & 0.296                  & {\color[HTML]{0000FF} {\ul 0.201}}      & 0.245                                   & 0.207                                  & 0.250                               & 0.202                                 & {\color[HTML]{0000FF} {\ul 0.238}}    & 0.211                                   & 0.251                                  & 0.221                                                & 0.254                  & 0.225                                  & 0.259                                 & 0.219                                  & 0.261                                 \\
                               & 336 & 0.258                                 & {\color[HTML]{0000FF} {\ul 0.280}}    & {\color[HTML]{0000FF} {\ul 0.251}}                    & 0.287                   & 0.265                                  & 0.290                                  & 0.292                                              & 0.307                 & 0.283                                                & 0.335                  & {\color[HTML]{FF0000} \textbf{0.237}}   & {\color[HTML]{FF0000} \textbf{0.265}}   & 0.262                                  & 0.291                               & 0.260                                 & 0.282                                 & 0.267                                   & 0.292                                  & 0.278                                                & 0.296                  & 0.278                                  & 0.297                                 & 0.280                                  & 0.306                                 \\
                               & 720 & {\color[HTML]{0000FF} {\ul 0.337}}    & {\color[HTML]{FF0000} \textbf{0.333}} & 0.339                                                 & 0.341                   & 0.342                                  & 0.340                                  & 0.365                                              & 0.354                 & 0.345                                                & 0.381                  & {\color[HTML]{FF0000} \textbf{0.312}}   & {\color[HTML]{0000FF} {\ul 0.334}}      & 0.343                                  & 0.343                               & 0.343                                 & 0.353                                 & 0.343                                   & 0.341                                  & 0.358                                                & 0.349                  & 0.354                                  & 0.348                                 & 0.365                                  & 0.359                                 \\ \cmidrule(l){2-26} 
      & Avg & {\color[HTML]{0000FF} {\ul 0.237}}    & {\color[HTML]{FF0000} \textbf{0.260}} & 0.240                                                 & 0.272                   & 0.245                                  & 0.272                                  & 0.273                                              & 0.292                 & 0.265                                                & 0.317                  & {\color[HTML]{FF0000} \textbf{0.226}}   & {\color[HTML]{0000FF} {\ul 0.262}}      & 0.242                                  & 0.272                               & 0.239                                 & 0.265                                 & 0.246                                   & 0.272                                  & 0.258                                                & 0.279                  & 0.259                                  & 0.281                                 & 0.259                                  & 0.287                                 \\ \midrule
\multirow{5}{*}{\rotatebox[origin=c]{90}{Solar-Energy}}      & 96  & {\color[HTML]{0000FF} {\ul 0.179}}    & {\color[HTML]{FF0000} \textbf{0.191}} & 0.189                                  & 0.259                                  & 0.205                                  & 0.242                                  & 0.319                                              & 0.353                 & 0.290                                                & 0.378                  & {\color[HTML]{FF0000} \textbf{0.171}}   & 0.231                                   & 0.197                                  & 0.241                               & 0.197                                 & {\color[HTML]{0000FF} {\ul 0.211}}    & 0.185                                   & 0.233                                  & 0.203                                                & 0.237                  & 0.234                                  & 0.286                                 & 0.250                                  & 0.292                                 \\
                               & 192 & {\color[HTML]{FF0000} \textbf{0.209}} & {\color[HTML]{FF0000} \textbf{0.213}} & 0.222                                  & 0.283                                  & 0.233                                  & 0.265                                  & 0.367                                              & 0.387                 & 0.320                                                & 0.398                  & {\color[HTML]{0000FF} {\ul 0.218}}      & 0.263                                   & 0.231                                  & 0.264                               & 0.234                                 & {\color[HTML]{0000FF} {\ul 0.234}}    & 0.227                                   & 0.253                                  & 0.233                                                & 0.261                  & 0.267                                  & 0.310                                 & 0.296                                  & 0.318                                 \\
                               & 336 & {\color[HTML]{0000FF} {\ul 0.231}}    & {\color[HTML]{FF0000} \textbf{0.229}} & {\color[HTML]{0000FF} {\ul 0.231}}     & 0.292                                  & 0.249                                  & 0.278                                  & 0.408                                              & 0.403                 & 0.353                                                & 0.415                  & {\color[HTML]{FF0000} \textbf{0.212}}   & 0.269                                   & 0.241                                  & 0.268                               & 0.256                                 & {\color[HTML]{0000FF} {\ul 0.250}}    & 0.246                                   & 0.284                                  & 0.248                                                & 0.273                  & 0.290                                  & 0.315                                 & 0.319                                  & 0.330                                 \\
                               & 720 & 0.241                                 & {\color[HTML]{FF0000} \textbf{0.236}} & {\color[HTML]{0000FF} {\ul 0.223}}     & 0.285                                  & 0.253                                  & 0.281                                  & 0.411                                              & 0.395                 & 0.356                                                & 0.413                  & {\color[HTML]{FF0000} \textbf{0.212}}   & 0.270                                   & 0.250                                  & 0.281                               & 0.260                                 & {\color[HTML]{0000FF} {\ul 0.254}}    & 0.247                                   & 0.276                                  & 0.249                                                & 0.275                  & 0.289                                  & 0.317                                 & 0.338                                  & 0.337                                 \\ \cmidrule(l){2-26} 
 & Avg & {\color[HTML]{0000FF} {\ul 0.215}}    & {\color[HTML]{FF0000} \textbf{0.217}} & 0.216                                  & 0.280                                  & 0.235                                  & 0.266                                  & 0.376                                              & 0.384                 & 0.330                                                & 0.401                  & {\color[HTML]{FF0000} \textbf{0.203}}   & 0.258                                   & 0.230                                  & 0.264                               & 0.237                                 & {\color[HTML]{0000FF} {\ul 0.237}}    & 0.226                                   & 0.262                                  & 0.233                                                & 0.262                  & 0.270                                  & 0.307                                 & 0.301                                  & 0.319                                 \\ \midrule
\multirow{5}{*}{\rotatebox[origin=c]{90}{PEMS03}}      & 12  & {\color[HTML]{FF0000} \textbf{0.060}} & {\color[HTML]{FF0000} \textbf{0.159}} & 0.076                                  & 0.188                                  & 0.071                                  & 0.177                                  & 0.117                                              & 0.226                 & 0.122                                                & 0.243                  & 0.097                                   & 0.208                                   & {\color[HTML]{0000FF} {\ul 0.063}}     & {\color[HTML]{0000FF} {\ul 0.164}}  & 0.072                                 & 0.177                                 & 0.068                                   & 0.174                                  & 0.071                                                & 0.174                  & 0.099                                  & 0.216                                 & 0.085                                  & 0.192                                 \\
                               & 24  & {\color[HTML]{FF0000} \textbf{0.078}} & {\color[HTML]{FF0000} \textbf{0.179}} & 0.113                                  & 0.226                                  & 0.102                                  & 0.213                                  & 0.235                                              & 0.324                 & 0.201                                                & 0.317                  & 0.120                                   & 0.230                                   & {\color[HTML]{0000FF} {\ul 0.080}}     & {\color[HTML]{0000FF} {\ul 0.185}}  & 0.107                                 & 0.217                                 & 0.093                                   & 0.202                                  & 0.093                                                & 0.201                  & 0.142                                  & 0.259                                 & 0.118                                  & 0.223                                 \\
                               & 48  & {\color[HTML]{FF0000} \textbf{0.104}} & {\color[HTML]{FF0000} \textbf{0.210}} & 0.191                                  & 0.292                                  & 0.162                                  & 0.272                                  & 0.541                                              & 0.521                 & 0.333                                                & 0.425                  & 0.170                                   & 0.272                                   & {\color[HTML]{0000FF} {\ul 0.124}}     & {\color[HTML]{0000FF} {\ul 0.226}}  & 0.194                                 & 0.302                                 & 0.146                                   & 0.258                                  & 0.125                                                & 0.236                  & 0.211                                  & 0.319                                 & 0.155                                  & 0.260                                 \\
                               & 96  & {\color[HTML]{FF0000} \textbf{0.140}} & {\color[HTML]{FF0000} \textbf{0.247}} & 0.288                                  & 0.363                                  & 0.244                                  & 0.340                                  & 1.062                                              & 0.790                 & 0.457                                                & 0.515                  & 0.274                                   & 0.342                                   & {\color[HTML]{0000FF} {\ul 0.160}}     & {\color[HTML]{0000FF} {\ul 0.266}}  & 0.323                                 & 0.402                                 & 0.228                                   & 0.330                                  & 0.164                                                & 0.275                  & 0.269                                  & 0.370                                 & 0.228                                  & 0.317                                 \\ \cmidrule(l){2-26} 
    & Avg & {\color[HTML]{FF0000} \textbf{0.095}} & {\color[HTML]{FF0000} \textbf{0.199}} & 0.167                                  & 0.267                                  & 0.145                                  & 0.251                                  & 0.489                                              & 0.465                 & 0.278                                                & 0.375                  & 0.165                                   & 0.263                                   & {\color[HTML]{0000FF} {\ul 0.107}}     & {\color[HTML]{0000FF} {\ul 0.210}}  & 0.174                                 & 0.275                                 & 0.135                                   & 0.243                                  & 0.113                                                & 0.221                  & 0.180                                  & 0.291                                 & 0.147                                  & 0.248                                 \\ \midrule
\multirow{5}{*}{\rotatebox[origin=c]{90}{PEMS04}}    & 12  & {\color[HTML]{FF0000} \textbf{0.068}} & {\color[HTML]{FF0000} \textbf{0.163}} & 0.092                                  & 0.204                                  & 0.082                                  & 0.190                                  & 0.129                                              & 0.239                 & 0.148                                                & 0.272                  & 0.099                                   & 0.214                                   & {\color[HTML]{0000FF} {\ul 0.071}}     & {\color[HTML]{0000FF} {\ul 0.172}}  & 0.089                                 & 0.194                                 & 0.085                                   & 0.189                                  & 0.078                                                & 0.183                  & 0.105                                  & 0.224                                 & 0.087                                  & 0.195                                 \\
                               & 24  & {\color[HTML]{FF0000} \textbf{0.079}} & {\color[HTML]{FF0000} \textbf{0.176}} & 0.128                                  & 0.243                                  & 0.110                                  & 0.224                                  & 0.246                                              & 0.337                 & 0.224                                                & 0.340                  & 0.115                                   & 0.231                                   & {\color[HTML]{0000FF} {\ul 0.087}}     & {\color[HTML]{0000FF} {\ul 0.193}}  & 0.128                                 & 0.234                                 & 0.117                                   & 0.224                                  & 0.095                                                & 0.205                  & 0.153                                  & 0.275                                 & 0.103                                  & 0.215                                 \\
                               & 48  & {\color[HTML]{FF0000} \textbf{0.095}} & {\color[HTML]{FF0000} \textbf{0.197}} & 0.213                                  & 0.315                                  & 0.160                                  & 0.276                                  & 0.568                                              & 0.539                 & 0.355                                                & 0.437                  & 0.144                                   & 0.261                                   & {\color[HTML]{0000FF} {\ul 0.113}}     & {\color[HTML]{0000FF} {\ul 0.222}}  & 0.224                                 & 0.321                                 & 0.174                                   & 0.276                                  & 0.120                                                & 0.233                  & 0.229                                  & 0.339                                 & 0.136                                  & 0.250                                 \\
                               & 96  & {\color[HTML]{FF0000} \textbf{0.122}} & {\color[HTML]{FF0000} \textbf{0.226}} & 0.307                                  & 0.384                                  & 0.234                                  & 0.343                                  & 1.181                                              & 0.843                 & 0.452                                                & 0.504                  & 0.185                                   & 0.297                                   & {\color[HTML]{0000FF} {\ul 0.141}}     & {\color[HTML]{0000FF} {\ul 0.252}}  & 0.382                                 & 0.445                                 & 0.273                                   & 0.354                                  & 0.150                                                & 0.262                  & 0.291                                  & 0.389                                 & 0.190                                  & 0.303                                 \\ \cmidrule(l){2-26} 
     & Avg & {\color[HTML]{FF0000} \textbf{0.091}} & {\color[HTML]{FF0000} \textbf{0.190}} & 0.185                                  & 0.287                                  & 0.146                                  & 0.258                                  & 0.531                                              & 0.489                 & 0.295                                                & 0.388                  & 0.136                                   & 0.251                                   & {\color[HTML]{0000FF} {\ul 0.103}}     & {\color[HTML]{0000FF} {\ul 0.210}}  & 0.206                                 & 0.299                                 & 0.162                                   & 0.261                                  & 0.111                                                & 0.221                  & 0.195                                  & 0.307                                 & 0.129                                  & 0.241                                 \\ \midrule
\multirow{5}{*}{\rotatebox[origin=c]{90}{PEMS07}}       & 12  & {\color[HTML]{FF0000} \textbf{0.052}} & {\color[HTML]{FF0000} \textbf{0.138}} & 0.073                                  & 0.184                                  & 0.064                                  & 0.163                                  & 0.109                                              & 0.222                 & 0.115                                                & 0.242                  & 0.090                                   & 0.197                                   & {\color[HTML]{0000FF} {\ul 0.055}}     & {\color[HTML]{0000FF} {\ul 0.145}}  & 0.068                                 & 0.166                                 & 0.063                                   & 0.158                                  & 0.067                                                & 0.165                  & 0.095                                  & 0.207                                 & 0.082                                  & 0.181                                 \\
                               & 24  & {\color[HTML]{FF0000} \textbf{0.065}} & {\color[HTML]{FF0000} \textbf{0.151}} & 0.111                                  & 0.219                                  & 0.093                                  & 0.200                                  & 0.230                                              & 0.327                 & 0.210                                                & 0.329                  & 0.110                                   & 0.219                                   & {\color[HTML]{0000FF} {\ul 0.070}}     & {\color[HTML]{0000FF} {\ul 0.164}}  & 0.103                                 & 0.206                                 & 0.089                                   & 0.192                                  & 0.088                                                & 0.190                  & 0.150                                  & 0.262                                 & 0.101                                  & 0.204                                 \\
                               & 48  & {\color[HTML]{FF0000} \textbf{0.084}} & {\color[HTML]{FF0000} \textbf{0.171}} & 0.237                                  & 0.328                                  & 0.137                                  & 0.248                                  & 0.551                                              & 0.531                 & 0.398                                                & 0.458                  & 0.149                                   & 0.256                                   & {\color[HTML]{0000FF} {\ul 0.094}}     & {\color[HTML]{0000FF} {\ul 0.192}}  & 0.165                                 & 0.268                                 & 0.136                                   & 0.241                                  & 0.110                                                & 0.215                  & 0.253                                  & 0.340                                 & 0.134                                  & 0.238                                 \\
                               & 96  & {\color[HTML]{FF0000} \textbf{0.108}} & {\color[HTML]{FF0000} \textbf{0.196}} & 0.303                                  & 0.354                                  & 0.198                                  & 0.306                                  & 1.112                                              & 0.809                 & 0.594                                                & 0.553                  & 0.258                                   & 0.359                                   & {\color[HTML]{0000FF} {\ul 0.117}}     & {\color[HTML]{0000FF} {\ul 0.217}}  & 0.258                                 & 0.346                                 & 0.197                                   & 0.298                                  & 0.139                                                & 0.245                  & 0.346                                  & 0.404                                 & 0.181                                  & 0.279                                 \\ \cmidrule(l){2-26} 
    & Avg & {\color[HTML]{FF0000} \textbf{0.077}} & {\color[HTML]{FF0000} \textbf{0.164}} & 0.181                                  & 0.271                                  & 0.123                                  & 0.229                                  & 0.500                                              & 0.472                 & 0.329                                                & 0.395                  & 0.152                                   & 0.258                                   & {\color[HTML]{0000FF} {\ul 0.084}}     & {\color[HTML]{0000FF} {\ul 0.180}}  & 0.149                                 & 0.247                                 & 0.121                                   & 0.222                                  & 0.101                                                & 0.204                  & 0.211                                  & 0.303                                 & 0.124                                  & 0.225                                 \\ \midrule
\multirow{5}{*}{\rotatebox[origin=c]{90}{PEMS08}}        & 12  & {\color[HTML]{FF0000} \textbf{0.068}} & {\color[HTML]{FF0000} \textbf{0.159}} & 0.091                                  & 0.201                                  & 0.080                                  & 0.182                                  & 0.122                                              & 0.233                 & 0.154                                                & 0.276                  & 0.119                                   & 0.222                                   & {\color[HTML]{0000FF} {\ul 0.071}}     & {\color[HTML]{0000FF} {\ul 0.171}}  & 0.080                                 & 0.181                                 & 0.081                                   & 0.185                                  & 0.079                                                & 0.182                  & 0.168                                  & 0.232                                 & 0.112                                  & 0.212                                 \\
                               & 24  & {\color[HTML]{FF0000} \textbf{0.089}} & {\color[HTML]{FF0000} \textbf{0.178}} & 0.137                                  & 0.246                                  & 0.114                                  & 0.219                                  & 0.236                                              & 0.330                 & 0.248                                                & 0.353                  & 0.149                                   & 0.249                                   & {\color[HTML]{0000FF} {\ul 0.091}}     & {\color[HTML]{0000FF} {\ul 0.189}}  & 0.118                                 & 0.220                                 & 0.112                                   & 0.214                                  & 0.115                                                & 0.219                  & 0.224                                  & 0.281                                 & 0.141                                  & 0.238                                 \\
                               & 48  & {\color[HTML]{FF0000} \textbf{0.123}} & {\color[HTML]{FF0000} \textbf{0.204}} & 0.265                                  & 0.343                                  & 0.184                                  & 0.284                                  & 0.562                                              & 0.540                 & 0.440                                                & 0.470                  & 0.206                                   & 0.292                                   & {\color[HTML]{0000FF} {\ul 0.128}}     & {\color[HTML]{0000FF} {\ul 0.219}}  & 0.199                                 & 0.289                                 & 0.174                                   & 0.267                                  & 0.186                                                & 0.235                  & 0.321                                  & 0.354                                 & 0.198                                  & 0.283                                 \\
                               & 96  & {\color[HTML]{FF0000} \textbf{0.173}} & {\color[HTML]{FF0000} \textbf{0.236}} & 0.410                                  & 0.407                                  & 0.309                                  & 0.356                                  & 1.216                                              & 0.846                 & 0.674                                                & 0.565                  & 0.329                                   & 0.355                                   & {\color[HTML]{0000FF} {\ul 0.198}}     & {\color[HTML]{0000FF} {\ul 0.266}}  & 0.405                                 & 0.431                                 & 0.277                                   & 0.335                                  & 0.221                                                & 0.267                  & 0.408                                  & 0.417                                 & 0.320                                  & 0.351                                 \\ \cmidrule(l){2-26} 
    & Avg & {\color[HTML]{FF0000} \textbf{0.113}} & {\color[HTML]{FF0000} \textbf{0.194}} & 0.226                                  & 0.299                                  & 0.172                                  & 0.260                                  & 0.534                                              & 0.487                 & 0.379                                                & 0.416                  & 0.200                                   & 0.279                                   & {\color[HTML]{0000FF} {\ul 0.122}}     & {\color[HTML]{0000FF} {\ul 0.211}}  & 0.201                                 & 0.280                                 & 0.161                                   & 0.250                                  & 0.150                                                & 0.226                  & 0.280                                  & 0.321                                 & 0.193                                  & 0.271                                 \\ \midrule
\multicolumn{2}{c}{1st Count}         & 33                                    & 49                                    & 0                                                     & 0                       & 0                                      & 0                                      & 0                                                  & 0                     & 0                                                    & 0                      & 29                                      & 14                                      & 1                                      & 0                                   & 2                                     & 1                                     & 2                                       & 2                                      & 0                                                    & 0                      & 1                                      & 1                                     & 0                                      & 0                                     \\ \bottomrule
\end{tabular}}
\end{table}

\begin{table}[p]
\caption{Full results for the short-term forecasting task (Part 1). For prediction lengths $\tau \in \left \{ 3,6,9,12 \right \}$, the lookback horizon is $T=12$; for $\tau \in \left \{ 24,36,48,60 \right \}$, it is $T=36$. The best results are highlighted in {\color[HTML]{FF0000} \textbf{bold}}, and the second-best results are {\color[HTML]{0000FF} {\ul underlined}}. This table presents the detailed version of Table~\ref{tab:short_term}.}
\label{tab:full_short_part1}
\centering
\setlength{\tabcolsep}{1.6pt}
\renewcommand{\arraystretch}{1.0} 
{\fontsize{5}{8}\selectfont
\begin{tabular}{@{}cccccccccccccccccccccccccc@{}}
\toprule
\multicolumn{2}{c}{Model}         & \multicolumn{2}{c}{\begin{tabular}[c]{@{}c@{}}OrthoLienar\\      (Ours)\end{tabular}} & 
\multicolumn{2}{c}{\begin{tabular}[c]{@{}c@{}}TimeMix.\\      \citeyear{timemixer} \end{tabular}} & 
\multicolumn{2}{c}{\begin{tabular}[c]{@{}c@{}}FilterNet\\      \citeyear{filternet} \end{tabular}} & 
\multicolumn{2}{c}{\begin{tabular}[c]{@{}c@{}}FITS\\      \citeyear{fits}  \end{tabular}} &
\multicolumn{2}{c}{\begin{tabular}[c]{@{}c@{}}DLinear\\      \citeyear{linear}  \end{tabular}} & 
\multicolumn{2}{c}{\begin{tabular}[c]{@{}c@{}}TimeMix.++\\      \citeyear{timemixer++} \end{tabular}} & 
\multicolumn{2}{c}{\begin{tabular}[c]{@{}c@{}}Leddam\\      \citeyear{Leddam_icml} \end{tabular}} & 
\multicolumn{2}{c}{\begin{tabular}[c]{@{}c@{}}CARD\\      \citeyear{card} \end{tabular}} & 
\multicolumn{2}{c}{\begin{tabular}[c]{@{}c@{}}Fredformer\\      \citeyear{fredformer} \end{tabular}} & 
\multicolumn{2}{c}{\begin{tabular}[c]{@{}c@{}}iTrans.\\      \citeyear{itransformer} \end{tabular}} & 
\multicolumn{2}{c}{\begin{tabular}[c]{@{}c@{}}PatchTST\\      \citeyear{patchtst} \end{tabular}} & 
\multicolumn{2}{c}{\begin{tabular}[c]{@{}c@{}}TimesNet\\      \citeyear{timesnet} \end{tabular}} \\ \midrule
\multicolumn{2}{c}{Metric}        & MSE                                       & MAE                                       & MSE                                    & MAE                                   & MSE                                     & MAE                                   & MSE                                  & MAE                                 & MSE                                                  & MAE                    & MSE                                      & MAE                                   & MSE                                    & MAE                                 & MSE                                  & MAE                                 & MSE                                     & MAE                                    & MSE                                   & MAE                                   & MSE                                    & MAE                                   & MSE                      & MAE                                                 \\ \midrule
\multirow{10}{*}{\rotatebox[origin=c]{90}{ILI}}     & 3   & {\color[HTML]{FF0000} \textbf{0.468}}     & {\color[HTML]{FF0000} \textbf{0.349}}     & 0.659                                  & 0.435                                 & 0.660                                   & 0.437                                 & 1.461                                & 0.743                               & 1.280                                                & 0.747                  & 0.658                                    & 0.430                                 & 0.551                                  & {\color[HTML]{0000FF} {\ul 0.388}}  & 0.597                                & 0.392                               & {\color[HTML]{0000FF} {\ul 0.528}}      & 0.403                                  & 0.555                                 & 0.395                                 & 0.646                                  & 0.417                                 & 0.627                    & 0.420                                               \\
                            & 6   & {\color[HTML]{FF0000} \textbf{0.923}}     & {\color[HTML]{FF0000} \textbf{0.516}}     & 1.306                                  & 0.643                                 & 1.140                                   & 0.606                                 & 2.337                                & 0.974                               & 2.054                                                & 0.967                  & 1.273                                    & 0.645                                 & {\color[HTML]{0000FF} {\ul 1.021}}     & {\color[HTML]{0000FF} {\ul 0.569}}  & 1.246                                & 0.610                               & 1.128                                   & 0.604                                  & 1.124                                 & 0.586                                 & 1.269                                  & 0.629                                 & 1.147                    & 0.610                                               \\
                            & 9   & {\color[HTML]{FF0000} \textbf{1.289}}     & {\color[HTML]{FF0000} \textbf{0.655}}     & 2.070                                  & 0.842                                 & 1.815                                   & 0.798                                 & 3.397                                & 1.197                               & 2.771                                                & 1.138                  & 2.009                                    & 0.840                                 & 1.881                                  & 0.795                               & 2.041                                & 0.829                               & 1.804                                   & 0.786                                  & {\color[HTML]{0000FF} {\ul 1.794}}    & {\color[HTML]{0000FF} {\ul 0.772}}    & 2.021                                  & 0.830                                 & 1.796                    & 0.785                                               \\
                            & 12  & {\color[HTML]{FF0000} \textbf{1.698}}     & {\color[HTML]{FF0000} \textbf{0.791}}     & 2.792                                  & 1.018                                 & 2.435                                   & 0.953                                 & 4.244                                & 1.361                               & 3.497                                                & 1.284                  & 2.683                                    & 1.004                                 & 2.421                                  & 0.964                               & 2.746                                & 0.996                               & 2.610                                   & 0.989                                  & {\color[HTML]{0000FF} {\ul 2.273}}    & {\color[HTML]{0000FF} {\ul 0.884}}    & 2.788                                  & 1.016                                 & 2.349                    & 0.920                                               \\ \cmidrule(l){2-26} 
                            & Avg & {\color[HTML]{FF0000} \textbf{1.094}}     & {\color[HTML]{FF0000} \textbf{0.578}}     & 1.707                                  & 0.734                                 & 1.512                                   & 0.698                                 & 2.860                                & 1.069                               & 2.400                                                & 1.034                  & 1.656                                    & 0.730                                 & 1.468                                  & 0.679                               & 1.658                                & 0.707                               & 1.518                                   & 0.696                                  & {\color[HTML]{0000FF} {\ul 1.437}}    & {\color[HTML]{0000FF} {\ul 0.659}}    & 1.681                                  & 0.723                                 & 1.480                    & 0.684                                               \\ \cmidrule(l){2-26} 
                            & 24  & {\color[HTML]{FF0000} \textbf{1.737}}     & {\color[HTML]{FF0000} \textbf{0.800}}     & 2.110                                  & 0.879                                 & 2.190                                   & 0.870                                 & 4.265                                & 1.523                               & 3.158                                                & 1.243                  & {\color[HTML]{0000FF} {\ul 1.877}}       & {\color[HTML]{0000FF} {\ul 0.826}}    & 2.085                                  & 0.883                               & 2.407                                & 0.970                               & 2.098                                   & 0.894                                  & 2.004                                 & 0.860                                 & 2.046                                  & 0.849                                 & 2.317                    & 0.934                                               \\
                            & 36  & {\color[HTML]{0000FF} {\ul 1.714}}        & {\color[HTML]{FF0000} \textbf{0.795}}     & 2.084                                  & 0.890                                 & 1.902                                   & {\color[HTML]{0000FF} {\ul 0.862}}    & 3.718                                & 1.363                               & 3.009                                                & 1.200                  & 2.276                                    & 0.912                                 & 2.017                                  & 0.892                               & 2.324                                & 0.948                               & {\color[HTML]{FF0000} \textbf{1.712}}   & 0.867                                  & 1.910                                 & 0.880                                 & 2.344                                  & 0.912                                 & 1.972                    & 0.920                                               \\
                            & 48  & {\color[HTML]{FF0000} \textbf{1.821}}     & {\color[HTML]{FF0000} \textbf{0.804}}     & 1.961                                  & 0.866                                 & 2.051                                   & 0.882                                 & 3.994                                & 1.422                               & 2.994                                                & 1.194                  & 1.921                                    & 0.850                                 & {\color[HTML]{0000FF} {\ul 1.860}}     & {\color[HTML]{0000FF} {\ul 0.847}}  & 2.133                                & 0.911                               & 2.054                                   & 0.922                                  & 2.036                                 & 0.891                                 & 2.123                                  & 0.883                                 & 2.238                    & 0.940                                               \\
                            & 60  & {\color[HTML]{0000FF} {\ul 1.785}}        & {\color[HTML]{FF0000} \textbf{0.810}}     & 1.926                                  & 0.878                                 & 2.151                                   & 0.925                                 & 4.543                                & 1.554                               & 3.172                                                & 1.232                  & {\color[HTML]{FF0000} \textbf{1.745}}    & {\color[HTML]{0000FF} {\ul 0.838}}    & 1.967                                  & 0.879                               & 2.177                                & 0.921                               & 1.925                                   & 0.913                                  & 2.022                                 & 0.919                                 & 2.001                                  & 0.895                                 & 2.027                    & 0.928                                               \\ \cmidrule(l){2-26} 
   & Avg & {\color[HTML]{FF0000} \textbf{1.764}}     & {\color[HTML]{FF0000} \textbf{0.802}}     & 2.020                                  & 0.878                                 & 2.073                                   & 0.885                                 & 4.130                                & 1.465                               & 3.083                                                & 1.217                  & 1.955                                    & {\color[HTML]{0000FF} {\ul 0.857}}    & 1.982                                  & 0.875                               & 2.260                                & 0.938                               & {\color[HTML]{0000FF} {\ul 1.947}}      & 0.899                                  & 1.993                                 & 0.887                                 & 2.128                                  & 0.885                                 & 2.139                    & 0.931                                               \\ \midrule
\multirow{10}{*}{\rotatebox[origin=c]{90}{COVID-19}}      & 3   & {\color[HTML]{FF0000} \textbf{1.100}}     & {\color[HTML]{FF0000} \textbf{0.487}}     & 1.237                                  & 0.547                                 & 1.195                                   & 0.555                                 & 2.039                                & 0.790                               & 2.386                                                & 0.909                  & 1.298                                    & 0.584                                 & 1.216                                  & 0.570                               & {\color[HTML]{0000FF} {\ul 1.103}}   & {\color[HTML]{0000FF} {\ul 0.521}}  & 1.165                                   & 0.548                                  & 1.193                                 & 0.561                                 & 1.220                                  & 0.573                                 & 2.021                    & 0.704                                               \\
                            & 6   & {\color[HTML]{0000FF} {\ul 1.750}}        & {\color[HTML]{FF0000} \textbf{0.619}}     & 2.003                                  & 0.739                                 & 1.839                                   & 0.711                                 & 2.683                                & 0.919                               & 3.220                                                & 1.053                  & 1.833                                    & {\color[HTML]{0000FF} {\ul 0.682}}    & 1.782                                  & 0.689                               & 1.919                                & 0.735                               & {\color[HTML]{FF0000} \textbf{1.465}}   & 0.685                                  & 1.933                                 & 0.755                                 & 1.982                                  & 0.762                                 & 2.405                    & 0.808                                               \\
                            & 9   & {\color[HTML]{0000FF} {\ul 2.239}}        & {\color[HTML]{FF0000} \textbf{0.734}}     & 2.594                                  & 0.860                                 & 2.537                                   & 0.897                                 & 3.147                                & 1.050                               & 3.803                                                & 1.160                  & 2.472                                    & {\color[HTML]{0000FF} {\ul 0.822}}    & 2.407                                  & 0.866                               & 2.358                                & 0.841                               & {\color[HTML]{FF0000} \textbf{2.145}}   & 0.845                                  & 2.441                                 & 0.879                                 & 2.633                                  & 0.916                                 & 2.858                    & 0.969                                               \\
                            & 12  & {\color[HTML]{FF0000} \textbf{2.538}}     & {\color[HTML]{FF0000} \textbf{0.831}}     & 3.103                                  & 0.981                                 & {\color[HTML]{0000FF} {\ul 2.782}}      & 0.956                                 & 3.630                                & 1.156                               & 4.524                                                & 1.288                  & 3.273                                    & 1.084                                 & 2.851                                  & 0.991                               & 2.857                                & 0.971                               & 2.833                                   & 0.984                                  & 2.819                                 & 0.984                                 & 3.050                                  & 1.030                                 & 2.993                    & {\color[HTML]{0000FF} {\ul 0.964}}                  \\ \cmidrule(l){2-26} 
                            & Avg & {\color[HTML]{0000FF} {\ul 1.907}}        & {\color[HTML]{FF0000} \textbf{0.668}}     & 2.234                                  & 0.782                                 & 2.088                                   & 0.780                                 & 2.875                                & 0.979                               & 3.483                                                & 1.102                  & 2.219                                    & 0.793                                 & 2.064                                  & 0.779                               & 2.059                                & 0.767                               & {\color[HTML]{FF0000} \textbf{1.902}}   & {\color[HTML]{0000FF} {\ul 0.765}}     & 2.096                                 & 0.795                                 & 2.221                                  & 0.820                                 & 2.569                    & 0.861                                               \\ \cmidrule(l){2-26} 
                            & 24  & {\color[HTML]{FF0000} \textbf{4.474}}     & {\color[HTML]{FF0000} \textbf{1.180}}     & 6.335                                  & 1.554                                 & 5.926                                   & 1.517                                 & 8.248                                & 1.829                               & 9.780                                                & 1.851                  & 6.539                                    & 1.618                                 & 4.860                                  & 1.342                               & 5.133                                & 1.394                               & 4.799                                   & 1.347                                  & {\color[HTML]{0000FF} {\ul 4.715}}    & {\color[HTML]{0000FF} {\ul 1.321}}    & 5.528                                  & 1.450                                 & 5.634                    & 1.442                                               \\
                            & 36  & {\color[HTML]{FF0000} \textbf{7.241}}     & {\color[HTML]{FF0000} \textbf{1.670}}     & 8.222                                  & 1.787                                 & 7.696                                   & 1.733                                 & 10.345                               & 2.050                               & 12.804                                               & 2.083                  & 7.986                                    & 1.770                                 & 7.378                                  & 1.708                               & 7.377                                & 1.725                               & 7.536                                   & 1.727                                  & {\color[HTML]{0000FF} {\ul 7.299}}    & {\color[HTML]{0000FF} {\ul 1.681}}    & 8.351                                  & 1.830                                 & 9.114                    & 1.848                                               \\
                            & 48  & 10.076                                    & {\color[HTML]{0000FF} {\ul 1.985}}        & 11.669                                 & 2.157                                 & 11.572                                  & 2.141                                 & 11.999                               & 2.197                               & 14.244                                               & 2.189                  & 11.655                                   & 2.156                                 & {\color[HTML]{0000FF} {\ul 10.051}}    & 1.999                               & 11.013                               & 2.103                               & {\color[HTML]{FF0000} \textbf{9.833}}   & {\color[HTML]{FF0000} \textbf{1.951}}  & 10.141                                & 2.012                                 & 11.259                                 & 2.114                                 & 10.940                   & 2.033                                               \\
                            & 60  & 12.079                                    & 2.182                                     & 12.188                                 & 2.173                                 & {\color[HTML]{FF0000} \textbf{11.311}}  & {\color[HTML]{FF0000} \textbf{2.066}} & 13.185                               & 2.282                               & 15.472                                               & 2.275                  & 12.734                                   & 2.235                                 & {\color[HTML]{0000FF} {\ul 11.467}}    & {\color[HTML]{0000FF} {\ul 2.119}}  & 12.528                               & 2.227                               & 12.455                                  & 2.209                                  & 11.871                                & 2.156                                 & 12.666                                 & 2.225                                 & 12.888                   & 2.186                                               \\ \cmidrule(l){2-26} 
 & Avg & {\color[HTML]{0000FF} {\ul 8.467}}        & {\color[HTML]{FF0000} \textbf{1.754}}     & 9.604                                  & 1.918                                 & 9.126                                   & 1.864                                 & 10.944                               & 2.089                               & 13.075                                               & 2.099                  & 9.728                                    & 1.945                                 & {\color[HTML]{FF0000} \textbf{8.439}}  & {\color[HTML]{0000FF} {\ul 1.792}}  & 9.013                                & 1.862                               & 8.656                                   & 1.808                                  & 8.506                                 & {\color[HTML]{0000FF} {\ul 1.792}}    & 9.451                                  & 1.905                                 & 9.644                    & 1.877                                               \\ \midrule
\multirow{10}{*}{\rotatebox[origin=c]{90}{METR-LA}}  & 3   & 0.207                                     & {\color[HTML]{FF0000} \textbf{0.171}}     & {\color[HTML]{0000FF} {\ul 0.205}}     & 0.192                                 & {\color[HTML]{FF0000} \textbf{0.204}}   & 0.189                                 & 0.211                                & 0.198                               & 0.218                                                & 0.231                  & {\color[HTML]{FF0000} \textbf{0.204}}    & 0.189                                 & {\color[HTML]{FF0000} \textbf{0.204}}  & 0.191                               & 0.210                                & {\color[HTML]{0000FF} {\ul 0.180}}  & {\color[HTML]{0000FF} {\ul 0.205}}      & 0.188                                  & {\color[HTML]{0000FF} {\ul 0.205}}    & 0.188                                 & {\color[HTML]{FF0000} \textbf{0.204}}  & 0.190                                 & 0.221                    & 0.204                                               \\
                            & 6   & 0.301                                     & {\color[HTML]{FF0000} \textbf{0.207}}     & 0.297                                  & 0.230                                 & {\color[HTML]{0000FF} {\ul 0.296}}      & 0.228                                 & 0.306                                & 0.237                               & 0.307                                                & 0.278                  & 0.298                                    & 0.230                                 & {\color[HTML]{FF0000} \textbf{0.293}}  & 0.227                               & 0.311                                & {\color[HTML]{0000FF} {\ul 0.219}}  & 0.298                                   & 0.227                                  & 0.300                                 & 0.229                                 & 0.298                                  & 0.227                                 & 0.308                    & 0.238                                               \\
                            & 9   & 0.382                                     & {\color[HTML]{FF0000} \textbf{0.238}}     & 0.381                                  & 0.264                                 & {\color[HTML]{0000FF} {\ul 0.377}}      & 0.261                                 & 0.392                                & 0.271                               & 0.386                                                & 0.316                  & 0.384                                    & 0.266                                 & {\color[HTML]{FF0000} \textbf{0.369}}  & 0.264                               & 0.401                                & {\color[HTML]{0000FF} {\ul 0.253}}  & 0.385                                   & 0.263                                  & 0.386                                 & 0.265                                 & 0.382                                  & 0.263                                 & 0.387                    & 0.273                                               \\
                            & 12  & 0.452                                     & {\color[HTML]{FF0000} \textbf{0.263}}     & 0.455                                  & 0.295                                 & {\color[HTML]{0000FF} {\ul 0.449}}      & 0.290                                 & 0.467                                & 0.300                               & 0.452                                                & 0.353                  & 0.456                                    & 0.294                                 & {\color[HTML]{FF0000} \textbf{0.442}}  & 0.292                               & 0.474                                & {\color[HTML]{0000FF} {\ul 0.281}}  & 0.457                                   & 0.292                                  & 0.460                                 & 0.295                                 & 0.456                                  & 0.292                                 & 0.462                    & 0.298                                               \\ \cmidrule(l){2-26} 
                            & Avg & 0.335                                     & {\color[HTML]{FF0000} \textbf{0.220}}     & 0.334                                  & 0.245                                 & {\color[HTML]{0000FF} {\ul 0.331}}      & 0.242                                 & 0.344                                & 0.251                               & 0.341                                                & 0.294                  & 0.335                                    & 0.245                                 & {\color[HTML]{FF0000} \textbf{0.327}}  & 0.243                               & 0.349                                & {\color[HTML]{0000FF} {\ul 0.233}}  & 0.336                                   & 0.242                                  & 0.338                                 & 0.244                                 & 0.335                                  & 0.243                                 & 0.344                    & 0.253                                               \\ \cmidrule(l){2-26} 
                            & 24  & 0.650                                     & {\color[HTML]{FF0000} \textbf{0.337}}     & 0.671                                  & 0.413                                 & 0.670                                   & 0.402                                 & 0.698                                & 0.416                               & {\color[HTML]{0000FF} {\ul 0.645}}                   & 0.458                  & {\color[HTML]{FF0000} \textbf{0.617}}    & 0.394                                 & 0.680                                  & 0.405                               & 0.700                                & {\color[HTML]{0000FF} {\ul 0.378}}  & 0.676                                   & 0.408                                  & 0.700                                 & 0.413                                 & 0.679                                  & 0.410                                 & 0.698                    & 0.415                                               \\
                            & 36  & 0.800                                     & {\color[HTML]{FF0000} \textbf{0.388}}     & 0.841                                  & 0.480                                 & 0.824                                   & 0.471                                 & 0.874                                & 0.490                               & {\color[HTML]{0000FF} {\ul 0.785}}                   & 0.533                  & {\color[HTML]{FF0000} \textbf{0.781}}    & 0.457                                 & 0.841                                  & 0.471                               & 0.874                                & {\color[HTML]{0000FF} {\ul 0.448}}  & 0.852                                   & 0.477                                  & 0.867                                 & 0.480                                 & 0.845                                  & 0.484                                 & 0.856                    & 0.475                                               \\
                            & 48  & 0.905                                     & {\color[HTML]{FF0000} \textbf{0.427}}     & 0.964                                  & 0.531                                 & 0.955                                   & 0.521                                 & 1.013                                & 0.546                               & {\color[HTML]{0000FF} {\ul 0.885}}                   & 0.585                  & {\color[HTML]{FF0000} \textbf{0.842}}    & 0.520                                 & 0.963                                  & 0.528                               & 1.017                                & {\color[HTML]{0000FF} {\ul 0.498}}  & 0.982                                   & 0.526                                  & 1.017                                 & 0.539                                 & 0.972                                  & 0.536                                 & 0.972                    & 0.518                                               \\
                            & 60  & 0.999                                     & {\color[HTML]{FF0000} \textbf{0.457}}     & 1.047                                  & 0.573                                 & 1.050                                   & 0.563                                 & 1.122                                & 0.589                               & {\color[HTML]{0000FF} {\ul 0.959}}                   & 0.623                  & {\color[HTML]{FF0000} \textbf{0.958}}    & 0.551                                 & 1.029                                  & 0.556                               & 1.126                                & {\color[HTML]{0000FF} {\ul 0.541}}  & 1.084                                   & 0.569                                  & 1.079                                 & 0.572                                 & 1.077                                  & 0.578                                 & 1.033                    & 0.543                                               \\ \cmidrule(l){2-26} 
  & Avg & 0.838                                     & {\color[HTML]{FF0000} \textbf{0.402}}     & 0.881                                  & 0.499                                 & 0.875                                   & 0.489                                 & 0.927                                & 0.510                               & {\color[HTML]{0000FF} {\ul 0.819}}                   & 0.550                  & {\color[HTML]{FF0000} \textbf{0.799}}    & 0.480                                 & 0.878                                  & 0.490                               & 0.929                                & {\color[HTML]{0000FF} {\ul 0.466}}  & 0.898                                   & 0.495                                  & 0.916                                 & 0.501                                 & 0.893                                  & 0.502                                 & 0.890                    & 0.488                                               \\ \midrule
\multirow{10}{*}{\rotatebox[origin=c]{90}{NASDAQ}}  & 3   & {\color[HTML]{0000FF} {\ul 0.036}}        & {\color[HTML]{FF0000} \textbf{0.092}}     & {\color[HTML]{FF0000} \textbf{0.035}}  & {\color[HTML]{0000FF} {\ul 0.093}}    & 0.038                                   & 0.100                                 & 0.042                                & 0.111                               & 0.044                                                & 0.123                  & 0.036                                    & 0.098                                 & 0.040                                  & 0.103                               & 0.037                                & 0.096                               & 0.039                                   & 0.102                                  & 0.040                                 & 0.105                                 & 0.038                                  & 0.099                                 & 0.049                    & 0.123                                               \\
                            & 6   & {\color[HTML]{FF0000} \textbf{0.049}}     & {\color[HTML]{FF0000} \textbf{0.117}}     & {\color[HTML]{FF0000} \textbf{0.049}}  & {\color[HTML]{0000FF} {\ul 0.118}}    & 0.052                                   & 0.126                                 & 0.056                                & 0.134                               & 0.062                                                & 0.155                  & 0.050                                    & 0.121                                 & 0.054                                  & 0.128                               & {\color[HTML]{0000FF} {\ul 0.052}}   & 0.123                               & 0.053                                   & 0.126                                  & 0.054                                 & 0.129                                 & 0.053                                  & 0.124                                 & 0.061                    & 0.142                                               \\
                            & 9   & {\color[HTML]{FF0000} \textbf{0.062}}     & {\color[HTML]{FF0000} \textbf{0.137}}     & {\color[HTML]{FF0000} \textbf{0.062}}  & {\color[HTML]{0000FF} {\ul 0.139}}    & 0.065                                   & 0.145                                 & 0.069                                & 0.154                               & 0.082                                                & 0.189                  & 0.063                                    & 0.144                                 & 0.066                                  & 0.147                               & {\color[HTML]{0000FF} {\ul 0.063}}   & 0.141                               & 0.067                                   & 0.147                                  & 0.068                                 & 0.150                                 & 0.065                                  & 0.145                                 & 0.073                    & 0.161                                               \\
                            & 12  & {\color[HTML]{FF0000} \textbf{0.073}}     & {\color[HTML]{FF0000} \textbf{0.154}}     & {\color[HTML]{FF0000} \textbf{0.073}}  & {\color[HTML]{0000FF} {\ul 0.156}}    & 0.076                                   & 0.162                                 & 0.081                                & 0.172                               & 0.100                                                & 0.215                  & 0.075                                    & 0.161                                 & 0.078                                  & 0.164                               & {\color[HTML]{0000FF} {\ul 0.075}}   & 0.158                               & 0.079                                   & 0.165                                  & 0.078                                 & 0.165                                 & 0.077                                  & 0.161                                 & 0.088                    & 0.179                                               \\ \cmidrule(l){2-26} 
                            & Avg & {\color[HTML]{FF0000} \textbf{0.055}}     & {\color[HTML]{FF0000} \textbf{0.125}}     & {\color[HTML]{FF0000} \textbf{0.055}}  & {\color[HTML]{0000FF} {\ul 0.126}}    & 0.058                                   & 0.133                                 & 0.062                                & 0.142                               & 0.072                                                & 0.170                  & 0.056                                    & 0.131                                 & 0.059                                  & 0.135                               & {\color[HTML]{0000FF} {\ul 0.057}}   & 0.130                               & 0.059                                   & 0.135                                  & 0.060                                 & 0.137                                 & 0.058                                  & 0.132                                 & 0.068                    & 0.151                                               \\ \cmidrule(l){2-26} 
                            & 24  & {\color[HTML]{FF0000} \textbf{0.121}}     & {\color[HTML]{FF0000} \textbf{0.216}}     & {\color[HTML]{0000FF} {\ul 0.122}}     & 0.221                                 & 0.130                                   & 0.230                                 & 0.140                                & 0.244                               & 0.155                                                & 0.274                  & 0.132                                    & 0.233                                 & 0.125                                  & 0.222                               & 0.124                                & {\color[HTML]{0000FF} {\ul 0.220}}  & 0.128                                   & 0.226                                  & 0.137                                 & 0.237                                 & 0.127                                  & 0.224                                 & 0.198                    & 0.299                                               \\
                            & 36  & {\color[HTML]{FF0000} \textbf{0.163}}     & {\color[HTML]{FF0000} \textbf{0.261}}     & 0.183                                  & 0.279                                 & 0.175                                   & 0.273                                 & 0.184                                & 0.284                               & 0.196                                                & 0.306                  & 0.177                                    & 0.278                                 & 0.174                                  & 0.271                               & {\color[HTML]{0000FF} {\ul 0.167}}   & {\color[HTML]{0000FF} {\ul 0.266}}  & 0.170                                   & 0.268                                  & 0.184                                 & 0.280                                 & 0.174                                  & 0.269                                 & 0.229                    & 0.326                                               \\
                            & 48  & {\color[HTML]{0000FF} {\ul 0.205}}        & {\color[HTML]{FF0000} \textbf{0.296}}     & {\color[HTML]{FF0000} \textbf{0.200}}  & {\color[HTML]{0000FF} {\ul 0.298}}    & 0.224                                   & 0.314                                 & 0.234                                & 0.324                               & 0.244                                                & 0.344                  & 0.216                                    & 0.311                                 & 0.222                                  & 0.312                               & 0.218                                & 0.307                               & 0.218                                   & 0.306                                  & 0.229                                 & 0.318                                 & 0.225                                  & 0.314                                 & 0.267                    & 0.352                                               \\
                            & 60  & 0.259                                     & {\color[HTML]{0000FF} {\ul 0.336}}        & {\color[HTML]{FF0000} \textbf{0.238}}  & {\color[HTML]{FF0000} \textbf{0.328}} & 0.259                                   & 0.340                                 & 0.282                                & 0.357                               & 0.318                                                & 0.401                  & {\color[HTML]{0000FF} {\ul 0.249}}       & 0.337                                 & 0.264                                  & 0.341                               & 0.264                                & 0.341                               & 0.262                                   & 0.339                                  & 0.279                                 & 0.352                                 & 0.265                                  & 0.339                                 & 0.327                    & 0.394                                               \\ \cmidrule(l){2-26} 
  & Avg & {\color[HTML]{0000FF} {\ul 0.187}}        & {\color[HTML]{FF0000} \textbf{0.277}}     & {\color[HTML]{FF0000} \textbf{0.186}}  & {\color[HTML]{0000FF} {\ul 0.281}}    & 0.197                                   & 0.289                                 & 0.210                                & 0.302                               & 0.228                                                & 0.331                  & 0.193                                    & 0.290                                 & 0.196                                  & 0.286                               & 0.193                                & 0.284                               & 0.194                                   & 0.285                                  & 0.207                                 & 0.297                                 & 0.198                                  & 0.286                                 & 0.255                    & 0.343                                               \\ \midrule
\multirow{10}{*}{\rotatebox[origin=c]{90}{Wiki}}    & 3   & 6.161                                     & {\color[HTML]{FF0000} \textbf{0.368}}     & 6.209                                  & 0.392                                 & 6.234                                   & 0.402                                 & 7.470                                & 0.496                               & 6.254                                                & 0.438                  & 6.149                                    & 0.389                                 & {\color[HTML]{0000FF} {\ul 6.148}}     & 0.383                               & 6.183                                & {\color[HTML]{0000FF} {\ul 0.378}}  & 6.190                                   & 0.387                                  & 6.237                                 & 0.393                                 & {\color[HTML]{FF0000} \textbf{6.112}}  & 0.380                                 & 7.597                    & 0.510                                               \\
                            & 6   & 6.453                                     & {\color[HTML]{FF0000} \textbf{0.385}}     & 6.475                                  & 0.402                                 & 6.460                                   & 0.401                                 & 8.326                                & 0.544                               & 6.579                                                & 0.467                  & {\color[HTML]{0000FF} {\ul 6.436}}       & 0.401                                 & 6.455                                  & 0.397                               & 6.465                                & {\color[HTML]{0000FF} {\ul 0.393}}  & 6.696                                   & 0.404                                  & 6.484                                 & 0.400                                 & {\color[HTML]{FF0000} \textbf{6.425}}  & 0.395                                 & 7.962                    & 0.515                                               \\
                            & 9   & {\color[HTML]{FF0000} \textbf{6.666}}     & {\color[HTML]{FF0000} \textbf{0.398}}     & 6.702                                  & 0.418                                 & 6.697                                   & 0.416                                 & 8.869                                & 0.564                               & 6.776                                                & 0.508                  & 6.714                                    & 0.420                                 & {\color[HTML]{0000FF} {\ul 6.687}}     & 0.412                               & 6.714                                & 0.415                               & 6.768                                   & {\color[HTML]{0000FF} {\ul 0.411}}     & 6.689                                 & {\color[HTML]{0000FF} {\ul 0.411}}    & 6.743                                  & 0.426                                 & 8.150                    & 0.524                                               \\
                            & 12  & {\color[HTML]{0000FF} {\ul 6.834}}        & {\color[HTML]{FF0000} \textbf{0.406}}     & 6.902                                  & 0.426                                 & 6.899                                   & 0.426                                 & 9.394                                & 0.608                               & 6.927                                                & 0.513                  & 6.852                                    & 0.421                                 & 6.899                                  & 0.424                               & 6.852                                & 0.415                               & 7.168                                   & 0.424                                  & 6.868                                 & 0.419                                 & {\color[HTML]{FF0000} \textbf{6.814}}  & {\color[HTML]{0000FF} {\ul 0.414}}    & 8.117                    & 0.533                                               \\ \cmidrule(l){2-26} 
                            & Avg & {\color[HTML]{0000FF} {\ul 6.528}}        & {\color[HTML]{FF0000} \textbf{0.389}}     & 6.572                                  & 0.409                                 & 6.572                                   & 0.411                                 & 8.515                                & 0.553                               & 6.634                                                & 0.481                  & 6.538                                    & 0.408                                 & 6.547                                  & 0.404                               & 6.553                                & {\color[HTML]{0000FF} {\ul 0.400}}  & 6.705                                   & 0.406                                  & 6.569                                 & 0.405                                 & {\color[HTML]{FF0000} \textbf{6.523}}  & 0.404                                 & 7.956                    & 0.520                                               \\ \cmidrule(l){2-26} 
                            & 24  & 6.894                                     & {\color[HTML]{FF0000} \textbf{0.423}}     & 6.900                                  & 0.446                                 & 6.907                                   & 0.443                                 & 7.887                                & 0.613                               & 6.883                                                & 0.520                  & 6.902                                    & 0.460                                 & 6.919                                  & 0.450                               & 6.925                                & 0.440                               & {\color[HTML]{FF0000} \textbf{6.531}}   & 0.432                                  & 6.886                                 & 0.437                                 & {\color[HTML]{0000FF} {\ul 6.858}}     & {\color[HTML]{0000FF} {\ul 0.430}}    & 8.023                    & 0.612                                               \\
                            & 36  & 6.446                                     & {\color[HTML]{FF0000} \textbf{0.439}}     & 6.520                                  & 0.467                                 & 6.514                                   & 0.467                                 & 7.774                                & 0.660                               & {\color[HTML]{0000FF} {\ul 6.393}}                   & 0.538                  & 6.539                                    & 0.473                                 & 6.456                                  & 0.457                               & 6.463                                & 0.451                               & {\color[HTML]{FF0000} \textbf{5.935}}   & 0.453                                  & 6.431                                 & 0.452                                 & 6.400                                  & {\color[HTML]{0000FF} {\ul 0.445}}    & 7.229                    & 0.595                                               \\
                            & 48  & 6.004                                     & {\color[HTML]{FF0000} \textbf{0.446}}     & 6.108                                  & 0.484                                 & 6.135                                   & 0.478                                 & 7.737                                & 0.706                               & {\color[HTML]{0000FF} {\ul 5.940}}                   & 0.547                  & 6.115                                    & 0.487                                 & 6.031                                  & 0.468                               & 6.031                                & 0.460                               & {\color[HTML]{FF0000} \textbf{5.871}}   & 0.464                                  & 6.101                                 & 0.483                                 & 5.959                                  & {\color[HTML]{0000FF} {\ul 0.449}}    & 7.184                    & 0.641                                               \\
                            & 60  & 5.705                                     & {\color[HTML]{0000FF} {\ul 0.454}}        & 5.732                                  & 0.476                                 & 5.811                                   & 0.482                                 & 7.855                                & 0.746                               & {\color[HTML]{0000FF} {\ul 5.605}}                   & 0.552                  & 5.736                                    & 0.497                                 & 5.740                                  & 0.478                               & 5.723                                & 0.463                               & {\color[HTML]{FF0000} \textbf{5.389}}   & 0.463                                  & 5.681                                 & 0.462                                 & 5.633                                  & {\color[HTML]{FF0000} \textbf{0.452}} & 6.805                    & 0.645                                               \\ \cmidrule(l){2-26} 
    & Avg & 6.262                                     & {\color[HTML]{FF0000} \textbf{0.440}}     & 6.315                                  & 0.468                                 & 6.342                                   & 0.467                                 & 7.813                                & 0.681                               & {\color[HTML]{0000FF} {\ul 6.205}}                   & 0.539                  & 6.323                                    & 0.479                                 & 6.286                                  & 0.463                               & 6.285                                & 0.453                               & {\color[HTML]{FF0000} \textbf{5.931}}   & 0.453                                  & 6.275                                 & 0.458                                 & 6.212                                  & {\color[HTML]{0000FF} {\ul 0.444}}    & 7.310                    & 0.623                                               \\ \midrule
\multicolumn{2}{c}{1st   Count}   & 19                                        & 47                                        & 8                                      & 1                                     & 2                                       & 1                                     & 0                                    & 0                                   & 0                                                    & 0                      & 7                                        & 0                                     & 6                                      & 0                                   & 0                                    & 0                                   & 10                                      & 1                                      & 0                                     & 0                                     & 5                                      & 1                                     & 0                        & 0                                                   \\ \bottomrule
\end{tabular}
}
\end{table}

\begin{table}[p]
\caption{Full results for the short-term forecasting task (Part 2). For prediction lengths $\tau \in \left \{ 3,6,9,12 \right \}$, the lookback horizon is $T=12$; for $\tau \in \left \{ 24,36,48,60 \right \}$, it is $T=36$. The best and second-best results are highlighted in {\color[HTML]{FF0000} \textbf{bold}}, and {\color[HTML]{0000FF} {\ul underlined}}, respectively. This table presents the detailed version of Table~\ref{tab:short_term}.}
\label{tab:full_short_part2}
\centering
\setlength{\tabcolsep}{1.6pt}
\renewcommand{\arraystretch}{1.0} 
{\fontsize{5}{7}\selectfont
\begin{tabular}{@{}cccccccccccccccccccccccccc@{}}
\toprule
\multicolumn{2}{c}{Model}         & \multicolumn{2}{c}{\begin{tabular}[c]{@{}c@{}}OrthoLienar\\      (Ours)\end{tabular}} & 
\multicolumn{2}{c}{\begin{tabular}[c]{@{}c@{}}TimeMix.\\      \citeyear{timemixer} \end{tabular}} & 
\multicolumn{2}{c}{\begin{tabular}[c]{@{}c@{}}FilterNet\\      \citeyear{filternet} \end{tabular}} & 
\multicolumn{2}{c}{\begin{tabular}[c]{@{}c@{}}FITS\\      \citeyear{fits}  \end{tabular}} &
\multicolumn{2}{c}{\begin{tabular}[c]{@{}c@{}}DLinear\\      \citeyear{linear}  \end{tabular}} & 
\multicolumn{2}{c}{\begin{tabular}[c]{@{}c@{}}TimeMix.++\\      \citeyear{timemixer++} \end{tabular}} & 
\multicolumn{2}{c}{\begin{tabular}[c]{@{}c@{}}Leddam\\      \citeyear{Leddam_icml} \end{tabular}} & 
\multicolumn{2}{c}{\begin{tabular}[c]{@{}c@{}}CARD\\      \citeyear{card} \end{tabular}} & 
\multicolumn{2}{c}{\begin{tabular}[c]{@{}c@{}}Fredformer\\      \citeyear{fredformer} \end{tabular}} & 
\multicolumn{2}{c}{\begin{tabular}[c]{@{}c@{}}iTrans.\\      \citeyear{itransformer} \end{tabular}} & 
\multicolumn{2}{c}{\begin{tabular}[c]{@{}c@{}}PatchTST\\      \citeyear{patchtst} \end{tabular}} & 
\multicolumn{2}{c}{\begin{tabular}[c]{@{}c@{}}TimesNet\\      \citeyear{timesnet} \end{tabular}} \\ \midrule
\multicolumn{2}{c}{Metric}        & MSE                                       & MAE                                       & MSE                                    & MAE                                   & MSE                                     & MAE                                   & MSE                                  & MAE                                 & MSE                                                  & MAE                    & MSE                                      & MAE                                   & MSE                                    & MAE                                 & MSE                                  & MAE                                 & MSE                                     & MAE                                    & MSE                                   & MAE                                   & MSE                                    & MAE                                   & MSE                      & MAE                                                 \\ \midrule
\multirow{10}{*}{\rotatebox[origin=c]{90}{SP500}}  & 3   & {\color[HTML]{FF0000} \textbf{0.035}}        & {\color[HTML]{FF0000} \textbf{0.126}}       & 0.038                                       & 0.137                                       & 0.046                                   & 0.154                                 & 0.046                                & 0.157                               & 0.047                                 & 0.152                                 & 0.040                                   & 0.141                                   & 0.047                                   & {\color[HTML]{0000FF} {\ul 0.155}} & {\color[HTML]{0000FF} {\ul 0.036}}     & {\color[HTML]{0000FF} {\ul 0.130}}          & {\color[HTML]{0000FF} {\ul 0.045}}      & 0.153                                  & 0.046                                       & 0.155                                 & 0.042                                       & 0.146                                 & 0.037                                  & 0.133                                 \\
                            & 6   & {\color[HTML]{FF0000} \textbf{0.053}}        & {\color[HTML]{FF0000} \textbf{0.158}}       & 0.056                                       & 0.167                                       & 0.071                                   & 0.193                                 & 0.067                                & 0.190                               & 0.070                                 & 0.197                                 & 0.057                                   & 0.168                                   & {\color[HTML]{0000FF} {\ul 0.064}}      & {\color[HTML]{0000FF} {\ul 0.181}} & {\color[HTML]{0000FF} {\ul 0.054}}     & {\color[HTML]{0000FF} {\ul 0.162}}          & 0.066                                   & 0.187                                  & 0.066                                       & 0.187                                 & 0.063                                       & 0.183                                 & 0.055                                  & 0.165                                 \\
                            & 9   & {\color[HTML]{FF0000} \textbf{0.070}}        & {\color[HTML]{FF0000} \textbf{0.181}}       & 0.074                                       & 0.192                                       & 0.080                                   & 0.203                                 & 0.087                                & 0.218                               & 0.091                                 & 0.220                                 & 0.076                                   & 0.196                                   & 0.081                                   & 0.206                              & 0.072                                  & {\color[HTML]{0000FF} {\ul 0.184}}          & 0.080                                   & 0.205                                  & {\color[HTML]{0000FF} {\ul 0.080}}          & {\color[HTML]{0000FF} {\ul 0.203}}    & 0.081                                       & 0.207                                 & {\color[HTML]{0000FF} {\ul 0.071}}     & 0.187                                 \\
                            & 12  & {\color[HTML]{FF0000} \textbf{0.088}}        & {\color[HTML]{FF0000} \textbf{0.204}}       & 0.092                                       & 0.213                                       & 0.097                                   & 0.224                                 & 0.106                                & 0.242                               & 0.113                                 & 0.251                                 & 0.093                                   & 0.213                                   & 0.104                                   & 0.236                              & {\color[HTML]{0000FF} {\ul 0.089}}     & {\color[HTML]{0000FF} {\ul 0.206}}          & 0.103                                   & 0.234                                  & {\color[HTML]{0000FF} {\ul 0.099}}          & {\color[HTML]{0000FF} {\ul 0.228}}    & 0.101                                       & 0.233                                 & {\color[HTML]{0000FF} {\ul 0.089}}     & 0.210                                 \\ \cmidrule(l){2-26} 
                            & Avg & {\color[HTML]{FF0000} \textbf{0.061}}        & {\color[HTML]{FF0000} \textbf{0.167}}       & 0.065                                       & 0.177                                       & 0.073                                   & 0.193                                 & 0.076                                & 0.202                               & 0.080                                 & 0.205                                 & 0.066                                   & 0.179                                   & 0.074                                   & 0.194                              & {\color[HTML]{0000FF} {\ul 0.063}}     & {\color[HTML]{0000FF} {\ul 0.170}}          & 0.073                                   & 0.195                                  & {\color[HTML]{0000FF} {\ul 0.073}}          & {\color[HTML]{0000FF} {\ul 0.193}}    & 0.072                                       & 0.192                                 & {\color[HTML]{0000FF} {\ul 0.063}}     & 0.174                                 \\ \cmidrule(l){2-26} 
                            & 24  & {\color[HTML]{FF0000} \textbf{0.155}}        & {\color[HTML]{FF0000} \textbf{0.271}}       & 0.159                                       & 0.288                                       & 0.181                                   & 0.317                                 & 0.193                                & 0.334                               & 0.189                                 & 0.330                                 & {\color[HTML]{0000FF} {\ul 0.172}}      & {\color[HTML]{0000FF} {\ul 0.305}}      & 0.175                                   & 0.308                              & {\color[HTML]{0000FF} {\ul 0.156}}     & {\color[HTML]{0000FF} {\ul 0.276}}          & 0.181                                   & 0.315                                  & 0.180                                       & 0.309                                 & 0.164                                       & 0.298                                 & 0.162                                  & 0.291                                 \\
                            & 36  & {\color[HTML]{0000FF} {\ul 0.209}}           & {\color[HTML]{FF0000} \textbf{0.317}}       & 0.218                                       & 0.343                                       & 0.224                                   & {\color[HTML]{0000FF} {\ul 0.341}}    & 0.259                                & 0.389                               & 0.250                                 & 0.363                                 & 0.227                                   & 0.344                                   & 0.232                                   & 0.358                              & {\color[HTML]{FF0000} \textbf{0.206}}  & {\color[HTML]{0000FF} {\ul 0.319}}          & {\color[HTML]{FF0000} \textbf{0.239}}   & 0.365                                  & 0.225                                       & 0.346                                 & 0.221                                       & 0.341                                 & 0.219                                  & 0.344                                 \\
                            & 48  & {\color[HTML]{FF0000} \textbf{0.258}}        & {\color[HTML]{0000FF} {\ul \textbf{0.358}}} & 0.264                                       & 0.367                                       & 0.280                                   & 0.384                                 & 0.324                                & 0.439                               & 0.291                                 & 0.398                                 & 0.272                                   & 0.383                                   & {\color[HTML]{0000FF} {\ul 0.276}}      & {\color[HTML]{0000FF} {\ul 0.388}} & {\color[HTML]{FF0000} \textbf{0.258}}  & {\color[HTML]{FF0000} \textbf{0.354}}       & 0.283                                   & 0.394                                  & 0.275                                       & 0.383                                 & 0.278                                       & 0.397                                 & {\color[HTML]{0000FF} {\ul 0.262}}     & 0.371                                 \\
                            & 60  & {\color[HTML]{0000FF} {\ul 0.305}}           & {\color[HTML]{0000FF} {\ul \textbf{0.387}}} & 0.322                                       & 0.416                                       & 0.332                                   & 0.416                                 & 0.391                                & 0.486                               & 0.377                                 & 0.475                                 & {\color[HTML]{FF0000} \textbf{0.319}}   & {\color[HTML]{0000FF} {\ul 0.413}}      & 0.325                                   & 0.423                              & {\color[HTML]{FF0000} \textbf{0.303}}  & {\color[HTML]{FF0000} \textbf{0.385}}       & 0.341                                   & 0.438                                  & 0.322                                       & 0.418                                 & 0.321                                       & 0.409                                 & {\color[HTML]{0000FF} {\ul 0.305}}     & 0.399                                 \\ \cmidrule(l){2-26} 
   & Avg & {\color[HTML]{FF0000} \textbf{0.231}}        & {\color[HTML]{FF0000} \textbf{0.333}}       & 0.241                                       & 0.353                                       & 0.254                                   & 0.365                                 & 0.291                                & 0.412                               & 0.277                                 & 0.391                                 & 0.247                                   & {\color[HTML]{0000FF} {\ul 0.361}}      & 0.252                                   & 0.369                              & {\color[HTML]{FF0000} \textbf{0.231}}  & {\color[HTML]{FF0000} \textbf{0.333}}       & {\color[HTML]{0000FF} {\ul 0.261}}      & 0.378                                  & 0.250                                       & 0.364                                 & 0.246                                       & 0.361                                 & {\color[HTML]{0000FF} {\ul 0.237}}     & {\color[HTML]{0000FF} {\ul 0.351}}    \\ \midrule
\multirow{10}{*}{\rotatebox[origin=c]{90}{DowJones}}   & 3   & {\color[HTML]{0000FF} {\ul \textbf{1.550}}}  & {\color[HTML]{FF0000} \textbf{0.276}}       & 1.568                                       & {\color[HTML]{FF0000} \textbf{0.276}}       & 1.587                                   & 0.278                                 & 1.611                                & 0.283                               & 1.574                                 & 0.280                                 & 1.555                                   & 0.278                                   & {\color[HTML]{FF0000} \textbf{1.538}}   & {\color[HTML]{0000FF} {\ul 0.277}} & {\color[HTML]{0000FF} {\ul 1.563}}     & {\color[HTML]{FF0000} {\ul \textbf{0.276}}} & 1.563                                   & {\color[HTML]{0000FF} {\ul 0.277}}     & 1.551                                       & {\color[HTML]{FF0000} \textbf{0.276}} & 1.556                                       & {\color[HTML]{FF0000} \textbf{0.276}} & 2.269                                  & 0.325                                 \\
                            & 6   & {\color[HTML]{FF0000} {\ul \textbf{2.569}}}  & {\color[HTML]{FF0000} \textbf{0.361}}       & 2.593                                       & {\color[HTML]{0000FF} {\ul 0.362}}          & 2.634                                   & 0.364                                 & 2.661                                & 0.370                               & 2.613                                 & 0.367                                 & 2.591                                   & {\color[HTML]{0000FF} {\ul 0.364}}      & 2.588                                   & 0.364                              & {\color[HTML]{0000FF} {\ul 2.579}}     & {\color[HTML]{0000FF} {\ul 0.362}}          & {\color[HTML]{FF0000} \textbf{2.619}}   & 0.364                                  & 2.589                                       & 0.363                                 & 2.588                                       & {\color[HTML]{0000FF} {\ul 0.362}}    & 3.934                                  & 0.430                                 \\
                            & 9   & {\color[HTML]{0000FF} {\ul 3.565}}           & {\color[HTML]{FF0000} \textbf{0.430}}       & {\color[HTML]{FF0000} \textbf{3.564}}       & {\color[HTML]{0000FF} {\ul 0.431}}          & 3.651                                   & 0.434                                 & 3.704                                & 0.443                               & 3.621                                 & 0.436                                 & 3.572                                   & {\color[HTML]{0000FF} {\ul 0.433}}      & 3.644                                   & 0.440                              & 3.569                                  & {\color[HTML]{FF0000} \textbf{0.430}}       & {\color[HTML]{FF0000} \textbf{3.575}}   & 0.432                                  & 3.599                                       & 0.433                                 & 3.585                                       & 0.432                                 & 4.715                                  & 0.492                                 \\
                            & 12  & {\color[HTML]{0000FF} {\ul \textbf{4.517}}}  & {\color[HTML]{FF0000} \textbf{0.490}}       & {\color[HTML]{FF0000} \textbf{4.472}}       & {\color[HTML]{FF0000} \textbf{0.490}}       & {\color[HTML]{0000FF} {\ul 4.634}}      & 0.496                                 & 4.666                                & 0.503                               & 4.586                                 & 0.497                                 & 4.630                                   & 0.496                                   & 4.622                                   & 0.498                              & 4.528                                  & {\color[HTML]{FF0000} \textbf{0.490}}       & 4.605                                   & 0.493                                  & 4.579                                       & 0.493                                 & 4.533                                       & {\color[HTML]{0000FF} {\ul 0.492}}    & 5.999                                  & {\color[HTML]{0000FF} {\ul 0.561}}    \\ \cmidrule(l){2-26} 
                            & Avg & {\color[HTML]{0000FF} {\ul 3.050}}           & {\color[HTML]{FF0000} \textbf{0.389}}       & {\color[HTML]{FF0000} \textbf{3.049}}       & {\color[HTML]{0000FF} {\ul 0.390}}          & 3.126                                   & 0.393                                 & 3.160                                & 0.400                               & 3.099                                 & 0.395                                 & 3.087                                   & 0.393                                   & 3.098                                   & 0.395                              & 3.060                                  & {\color[HTML]{FF0000} \textbf{0.389}}       & {\color[HTML]{FF0000} \textbf{3.090}}   & {\color[HTML]{0000FF} {\ul 0.391}}     & 3.079                                       & 0.391                                 & 3.065                                       & {\color[HTML]{0000FF} {\ul 0.390}}    & 4.229                                  & 0.452                                 \\ \cmidrule(l){2-26} 
                            & 24  & {\color[HTML]{0000FF} {\ul \textbf{7.432}}}  & {\color[HTML]{FF0000} \textbf{0.664}}       & 8.327                                       & 0.683                                       & 8.000                                   & 0.683                                 & 7.974                                & 0.690                               & 7.590                                 & 0.670                                 & 8.283                                   & 0.689                                   & 8.029                                   & 0.679                              & {\color[HTML]{FF0000} \textbf{7.416}}  & {\color[HTML]{0000FF} {\ul 0.665}}          & 7.758                                   & 0.672                                  & {\color[HTML]{0000FF} {\ul 7.925}}          & {\color[HTML]{0000FF} {\ul 0.677}}    & 7.641                                       & 0.670                                 & 11.535                                 & 0.834                                 \\
                            & 36  & {\color[HTML]{0000FF} {\ul \textbf{10.848}}} & {\color[HTML]{0000FF} {\ul \textbf{0.799}}} & 11.192                                      & 0.813                                       & 12.011                                  & 0.823                                 & 11.907                               & 0.837                               & 10.986                                & 0.803                                 & 14.754                                  & 0.856                                   & 11.962                                  & 0.828                              & {\color[HTML]{FF0000} \textbf{10.799}} & {\color[HTML]{FF0000} \textbf{0.798}}       & 11.456                                  & 0.808                                  & {\color[HTML]{0000FF} {\ul 12.087}}         & {\color[HTML]{0000FF} {\ul 0.827}}    & 11.210                                      & 0.807                                 & 16.922                                 & 0.982                                 \\
                            & 48  & {\color[HTML]{0000FF} {\ul 14.045}}          & {\color[HTML]{0000FF} {\ul 0.914}}          & 15.278                                      & 0.945                                       & 14.814                                  & 0.933                                 & 15.821                               & 0.969                               & 14.157                                & 0.922                                 & 16.893                                  & 0.970                                   & {\color[HTML]{0000FF} {\ul 15.266}}     & 0.937                              & {\color[HTML]{FF0000} \textbf{13.881}} & {\color[HTML]{FF0000} \textbf{0.912}}       & {\color[HTML]{FF0000} \textbf{14.696}}  & {\color[HTML]{FF0000} \textbf{0.921}}  & 14.787                                      & 0.930                                 & 14.866                                      & 0.935                                 & 19.501                                 & 1.093                                 \\
                            & 60  & {\color[HTML]{FF0000} \textbf{16.959}}       & {\color[HTML]{FF0000} \textbf{1.017}}       & 20.997                                      & 1.067                                       & {\color[HTML]{FF0000} \textbf{18.932}}  & {\color[HTML]{FF0000} \textbf{1.054}} & 19.320                               & 1.077                               & 18.018                                & 1.035                                 & 18.881                                  & 1.059                                   & {\color[HTML]{0000FF} {\ul 18.407}}     & {\color[HTML]{0000FF} {\ul 1.045}} & {\color[HTML]{0000FF} {\ul 17.257}}    & {\color[HTML]{0000FF} {\ul 1.021}}          & 18.058                                  & 1.032                                  & 18.298                                      & 1.041                                 & 17.947                                      & 1.036                                 & 22.804                                 & 1.177                                 \\ \cmidrule(l){2-26} 
 & Avg & {\color[HTML]{FF0000} {\ul \textbf{12.321}}} & {\color[HTML]{FF0000} \textbf{0.848}}       & 13.948                                      & 0.877                                       & 13.439                                  & 0.873                                 & 13.755                               & 0.893                               & 12.688                                & 0.857                                 & 14.703                                  & 0.893                                   & {\color[HTML]{FF0000} \textbf{13.416}}  & {\color[HTML]{0000FF} {\ul 0.872}} & {\color[HTML]{0000FF} {\ul 12.338}}    & {\color[HTML]{0000FF} {\ul 0.849}}          & 12.992                                  & 0.858                                  & 13.274                                      & {\color[HTML]{0000FF} {\ul 0.869}}    & 12.916                                      & 0.862                                 & 17.690                                 & 1.021                                 \\ \midrule
\multirow{10}{*}{\rotatebox[origin=c]{90}{CarSales}}   & 3   & 0.303                                        & {\color[HTML]{0000FF} {\ul \textbf{0.277}}} & {\color[HTML]{0000FF} {\ul 0.307}}          & 0.296                                       & {\color[HTML]{FF0000} \textbf{0.304}}   & 0.291                                 & 0.411                                & 0.375                               & 0.396                                 & 0.383                                 & {\color[HTML]{FF0000} \textbf{0.309}}   & 0.289                                   & {\color[HTML]{FF0000} \textbf{0.316}}   & 0.300                              & 0.318                                  & {\color[HTML]{0000FF} {\ul 0.295}}          & {\color[HTML]{0000FF} {\ul 0.311}}      & 0.300                                  & {\color[HTML]{FF0000} {\ul \textbf{0.281}}} & {\color[HTML]{FF0000} \textbf{0.276}} & {\color[HTML]{0000FF} {\ul \textbf{0.300}}} & 0.290                                 & 0.315                                  & 0.303                                 \\
                            & 6   & 0.315                                        & {\color[HTML]{FF0000} \textbf{0.285}}       & 0.327                                       & 0.307                                       & {\color[HTML]{0000FF} {\ul 0.317}}      & 0.300                                 & 0.401                                & 0.367                               & 0.390                                 & 0.376                                 & 0.324                                   & 0.300                                   & {\color[HTML]{FF0000} \textbf{0.320}}   & 0.303                              & 0.335                                  & {\color[HTML]{0000FF} {\ul 0.307}}          & 0.326                                   & 0.309                                  & {\color[HTML]{FF0000} \textbf{0.294}}       & {\color[HTML]{FF0000} \textbf{0.285}} & {\color[HTML]{0000FF} {\ul 0.311}}          & {\color[HTML]{0000FF} {\ul 0.295}}    & 0.319                                  & 0.306                                 \\
                            & 9   & 0.327                                        & {\color[HTML]{0000FF} {\ul \textbf{0.293}}} & 0.336                                       & 0.315                                       & {\color[HTML]{0000FF} {\ul 0.329}}      & 0.308                                 & 0.418                                & 0.380                               & 0.406                                 & 0.390                                 & 0.331                                   & 0.306                                   & {\color[HTML]{FF0000} \textbf{0.332}}   & 0.311                              & 0.351                                  & {\color[HTML]{0000FF} {\ul 0.319}}          & 0.339                                   & 0.318                                  & {\color[HTML]{FF0000} \textbf{0.305}}       & {\color[HTML]{FF0000} \textbf{0.292}} & {\color[HTML]{0000FF} {\ul 0.324}}          & 0.306                                 & 0.338                                  & 0.318                                 \\
                            & 12  & 0.336                                        & {\color[HTML]{0000FF} {\ul \textbf{0.301}}} & 0.343                                       & 0.319                                       & {\color[HTML]{0000FF} {\ul 0.335}}      & 0.312                                 & 0.424                                & 0.385                               & 0.410                                 & 0.390                                 & 0.343                                   & 0.314                                   & {\color[HTML]{FF0000} \textbf{0.336}}   & 0.313                              & 0.357                                  & {\color[HTML]{0000FF} {\ul 0.324}}          & 0.348                                   & 0.324                                  & {\color[HTML]{FF0000} \textbf{0.314}}       & {\color[HTML]{FF0000} \textbf{0.298}} & {\color[HTML]{0000FF} {\ul 0.331}}          & 0.310                                 & 0.338                                  & 0.320                                 \\ \cmidrule(l){2-26} 
                            & Avg & 0.320                                        & {\color[HTML]{0000FF} {\ul \textbf{0.289}}} & 0.328                                       & 0.309                                       & {\color[HTML]{0000FF} {\ul 0.321}}      & 0.303                                 & 0.413                                & 0.377                               & 0.400                                 & 0.385                                 & 0.327                                   & 0.302                                   & {\color[HTML]{FF0000} \textbf{0.326}}   & 0.306                              & 0.340                                  & {\color[HTML]{0000FF} {\ul 0.311}}          & 0.331                                   & 0.313                                  & {\color[HTML]{FF0000} \textbf{0.298}}       & {\color[HTML]{FF0000} \textbf{0.288}} & {\color[HTML]{0000FF} {\ul 0.316}}          & 0.300                                 & 0.327                                  & 0.312                                 \\ \cmidrule(l){2-26} 
                            & 24  & 0.320                                        & {\color[HTML]{FF0000} \textbf{0.302}}       & 0.320                                       & 0.318                                       & 0.318                                   & 0.319                                 & 0.359                                & 0.347                               & {\color[HTML]{0000FF} {\ul 0.354}}    & 0.350                                 & {\color[HTML]{FF0000} \textbf{0.323}}   & 0.320                                   & 0.325                                   & 0.322                              & 0.337                                  & {\color[HTML]{0000FF} {\ul 0.321}}          & 0.319                                   & 0.326                                  & {\color[HTML]{FF0000} \textbf{0.303}}       & {\color[HTML]{0000FF} {\ul 0.312}}    & 0.319                                       & 0.319                                 & {\color[HTML]{0000FF} {\ul 0.316}}     & 0.328                                 \\
                            & 36  & 0.334                                        & {\color[HTML]{FF0000} \textbf{0.315}}       & 0.332                                       & 0.331                                       & {\color[HTML]{0000FF} {\ul 0.331}}      & 0.330                                 & 0.373                                & 0.360                               & {\color[HTML]{0000FF} {\ul 0.368}}    & 0.365                                 & {\color[HTML]{FF0000} \textbf{0.351}}   & 0.348                                   & 0.337                                   & 0.333                              & 0.348                                  & {\color[HTML]{0000FF} {\ul 0.333}}          & 0.333                                   & 0.335                                  & {\color[HTML]{FF0000} \textbf{0.318}}       & {\color[HTML]{0000FF} {\ul 0.323}}    & 0.332                                       & 0.330                                 & 0.338                                  & 0.343                                 \\
                            & 48  & 0.347                                        & {\color[HTML]{FF0000} \textbf{0.327}}       & 0.345                                       & 0.343                                       & {\color[HTML]{0000FF} {\ul 0.342}}      & 0.341                                 & 0.385                                & 0.370                               & {\color[HTML]{0000FF} {\ul 0.382}}    & 0.379                                 & {\color[HTML]{FF0000} \textbf{0.351}}   & 0.342                                   & 0.351                                   & 0.346                              & 0.362                                  & {\color[HTML]{0000FF} {\ul 0.345}}          & 0.349                                   & 0.344                                  & {\color[HTML]{FF0000} \textbf{0.331}}       & {\color[HTML]{0000FF} {\ul 0.332}}    & 0.347                                       & 0.344                                 & 0.349                                  & 0.351                                 \\
                            & 60  & 0.358                                        & {\color[HTML]{FF0000} \textbf{0.337}}       & 0.355                                       & 0.351                                       & {\color[HTML]{0000FF} {\ul 0.352}}      & 0.349                                 & 0.399                                & 0.385                               & {\color[HTML]{0000FF} {\ul 0.388}}    & 0.380                                 & {\color[HTML]{FF0000} \textbf{0.363}}   & 0.352                                   & 0.361                                   & 0.353                              & 0.372                                  & {\color[HTML]{0000FF} {\ul 0.353}}          & 0.359                                   & 0.349                                  & {\color[HTML]{FF0000} \textbf{0.344}}       & {\color[HTML]{0000FF} {\ul 0.342}}    & 0.355                                       & 0.348                                 & 0.359                                  & 0.359                                 \\ \cmidrule(l){2-26} 
     & Avg & 0.340                                        & {\color[HTML]{FF0000} \textbf{0.320}}       & 0.338                                       & 0.336                                       & {\color[HTML]{0000FF} {\ul 0.336}}      & 0.335                                 & 0.379                                & 0.365                               & {\color[HTML]{0000FF} {\ul 0.373}}    & 0.368                                 & {\color[HTML]{FF0000} \textbf{0.347}}   & 0.340                                   & 0.343                                   & 0.338                              & 0.355                                  & {\color[HTML]{0000FF} {\ul 0.338}}          & 0.340                                   & 0.338                                  & {\color[HTML]{FF0000} \textbf{0.324}}       & {\color[HTML]{0000FF} {\ul 0.327}}    & 0.338                                       & 0.335                                 & 0.340                                  & 0.345                                 \\ \midrule
\multirow{10}{*}{\rotatebox[origin=c]{90}{Power}}    & 3   & {\color[HTML]{0000FF} {\ul 0.864}}           & {\color[HTML]{FF0000} \textbf{0.688}}       & {\color[HTML]{FF0000} \textbf{0.850}}       & {\color[HTML]{FF0000} {\ul \textbf{0.683}}} & {\color[HTML]{0000FF} {\ul 0.842}}      & {\color[HTML]{0000FF} {\ul 0.685}}    & 0.899                                & 0.712                               & 0.876                                 & 0.708                                 & 0.843                                   & {\color[HTML]{0000FF} {\ul 0.685}}      & 0.859                                   & 0.696                              & 0.896                                  & 0.691                                       & 0.865                                   & 0.697                                  & 0.863                                       & 0.696                                 & {\color[HTML]{FF0000} \textbf{0.830}}       & 0.687                                 & 0.929                                  & 0.716                                 \\
                            & 6   & {\color[HTML]{FF0000} \textbf{0.991}}        & {\color[HTML]{0000FF} {\ul \textbf{0.742}}} & {\color[HTML]{0000FF} {\ul \textbf{0.971}}} & {\color[HTML]{0000FF} {\ul 0.743}}          & 0.988                                   & 0.747                                 & 1.079                                & 0.785                               & 0.991                                 & 0.761                                 & 0.997                                   & 0.748                                   & 1.023                                   & 0.761                              & {\color[HTML]{0000FF} {\ul 1.013}}     & 0.750                                       & 0.994                                   & 0.751                                  & 0.990                                       & 0.748                                 & {\color[HTML]{FF0000} \textbf{0.948}}       & {\color[HTML]{FF0000} \textbf{0.741}} & 1.070                                  & 0.773                                 \\
                            & 9   & {\color[HTML]{FF0000} \textbf{1.062}}        & {\color[HTML]{FF0000} \textbf{0.770}}       & {\color[HTML]{FF0000} \textbf{1.024}}       & {\color[HTML]{FF0000} {\ul \textbf{0.763}}} & 1.063                                   & 0.778                                 & 1.150                                & 0.804                               & 1.051                                 & 0.790                                 & 1.050                                   & {\color[HTML]{0000FF} {\ul 0.767}}      & 1.071                                   & 0.783                              & {\color[HTML]{0000FF} {\ul 1.085}}     & 0.781                                       & 1.073                                   & 0.784                                  & 1.065                                       & 0.780                                 & {\color[HTML]{0000FF} {\ul 1.027}}          & 0.773                                 & 1.104                                  & 0.788                                 \\
                            & 12  & {\color[HTML]{FF0000} \textbf{1.119}}        & {\color[HTML]{0000FF} {\ul \textbf{0.789}}} & {\color[HTML]{0000FF} {\ul \textbf{1.087}}} & {\color[HTML]{FF0000} {\ul \textbf{0.788}}} & 1.125                                   & 0.803                                 & 1.266                                & 0.851                               & 1.110                                 & 0.814                                 & 1.107                                   & 0.792                                   & 1.130                                   & 0.804                              & {\color[HTML]{0000FF} {\ul 1.124}}     & 0.792                                       & 1.135                                   & 0.807                                  & 1.142                                       & 0.808                                 & {\color[HTML]{FF0000} \textbf{1.083}}       & 0.797                                 & 1.212                                  & 0.825                                 \\ \cmidrule(l){2-26} 
                            & Avg & {\color[HTML]{FF0000} \textbf{1.009}}        & {\color[HTML]{0000FF} {\ul \textbf{0.747}}} & {\color[HTML]{0000FF} {\ul \textbf{0.983}}} & {\color[HTML]{FF0000} {\ul \textbf{0.744}}} & 1.004                                   & 0.753                                 & 1.098                                & 0.788                               & 1.007                                 & 0.768                                 & 0.999                                   & 0.748                                   & 1.021                                   & 0.761                              & {\color[HTML]{0000FF} {\ul 1.029}}     & 0.753                                       & 1.017                                   & 0.760                                  & 1.015                                       & 0.758                                 & {\color[HTML]{FF0000} \textbf{0.972}}       & 0.749                                 & 1.079                                  & 0.775                                 \\ \cmidrule(l){2-26} 
                            & 24  & {\color[HTML]{FF0000} \textbf{1.343}}        & {\color[HTML]{FF0000} \textbf{0.870}}       & {\color[HTML]{0000FF} {\ul 1.341}}          & 0.881                                       & 1.410                                   & 0.916                                 & 1.491                                & 0.944                               & 1.390                                 & 0.916                                 & {\color[HTML]{FF0000} \textbf{1.340}}   & {\color[HTML]{0000FF} {\ul 0.877}}      & 1.397                                   & 0.909                              & 1.406                                  & {\color[HTML]{0000FF} {\ul 0.886}}          & 1.410                                   & 0.913                                  & 1.462                                       & 0.924                                 & 1.468                                       & 0.935                                 & 1.494                                  & 0.924                                 \\
                            & 36  & {\color[HTML]{0000FF} {\ul \textbf{1.445}}}  & {\color[HTML]{FF0000} \textbf{0.903}}       & {\color[HTML]{FF0000} \textbf{1.420}}       & {\color[HTML]{0000FF} {\ul 0.914}}          & 1.590                                   & 0.968                                 & 1.621                                & 0.994                               & 1.518                                 & 0.957                                 & 1.446                                   & 0.920                                   & 1.509                                   & 0.951                              & {\color[HTML]{0000FF} {\ul 1.506}}     & {\color[HTML]{0000FF} {\ul 0.921}}          & 1.538                                   & 0.953                                  & 1.582                                       & 0.964                                 & 1.593                                       & 0.972                                 & 1.526                                  & 0.950                                 \\
                            & 48  & {\color[HTML]{0000FF} {\ul 1.559}}           & {\color[HTML]{0000FF} {\ul \textbf{0.946}}} & {\color[HTML]{FF0000} \textbf{1.567}}       & {\color[HTML]{0000FF} {\ul 0.963}}          & 1.680                                   & 1.009                                 & 1.775                                & 1.052                               & 1.610                                 & 0.995                                 & {\color[HTML]{FF0000} \textbf{1.467}}   & {\color[HTML]{FF0000} \textbf{0.933}}   & 1.646                                   & 0.999                              & 1.583                                  & 0.957                                       & 1.652                                   & 1.008                                  & 1.696                                       & 1.011                                 & 1.710                                       & 1.020                                 & 1.581                                  & 0.981                                 \\
                            & 60  & {\color[HTML]{FF0000} \textbf{1.602}}        & {\color[HTML]{FF0000} {\ul \textbf{0.971}}} & {\color[HTML]{0000FF} {\ul \textbf{1.609}}} & {\color[HTML]{0000FF} {\ul \textbf{0.988}}} & 1.776                                   & 1.053                                 & 1.958                                & 1.122                               & 1.679                                 & 1.020                                 & {\color[HTML]{0000FF} {\ul 1.626}}      & 1.006                                   & 1.727                                   & 1.043                              & 1.693                                  & 1.003                                       & 1.752                                   & 1.049                                  & 1.796                                       & 1.061                                 & 1.829                                       & 1.064                                 & 1.625                                  & 1.010                                 \\ \cmidrule(l){2-26} 
   & Avg & {\color[HTML]{0000FF} {\ul 1.487}}           & {\color[HTML]{FF0000} \textbf{0.922}}       & {\color[HTML]{0000FF} {\ul \textbf{1.484}}} & {\color[HTML]{0000FF} {\ul 0.937}}          & 1.614                                   & 0.986                                 & 1.711                                & 1.028                               & 1.549                                 & 0.972                                 & {\color[HTML]{FF0000} \textbf{1.470}}   & {\color[HTML]{0000FF} {\ul 0.934}}      & 1.570                                   & 0.975                              & 1.547                                  & 0.942                                       & 1.588                                   & 0.981                                  & 1.634                                       & 0.990                                 & 1.650                                       & 0.998                                 & 1.556                                  & 0.966                                 \\ \midrule
\multirow{10}{*}{\rotatebox[origin=c]{90}{Website}}     & 3   & 0.077                                        & {\color[HTML]{0000FF} {\ul \textbf{0.199}}} & 0.086                                       & 0.215                                       & 0.084                                   & 0.213                                 & 0.191                                & 0.320                               & 0.159                                 & 0.288                                 & 0.083                                   & 0.208                                   & {\color[HTML]{0000FF} {\ul 0.082}}      & 0.210                              & 0.099                                  & {\color[HTML]{0000FF} {\ul 0.225}}          & 0.080                                   & 0.207                                  & {\color[HTML]{FF0000} \textbf{0.072}}       & {\color[HTML]{FF0000} \textbf{0.198}} & {\color[HTML]{FF0000} \textbf{0.089}}       & 0.217                                 & {\color[HTML]{0000FF} {\ul 0.075}}     & 0.200                                 \\
                            & 6   & {\color[HTML]{0000FF} {\ul 0.103}}           & {\color[HTML]{FF0000} \textbf{0.227}}       & 0.124                                       & 0.248                                       & 0.116                                   & 0.242                                 & 0.235                                & 0.356                               & 0.182                                 & 0.302                                 & {\color[HTML]{0000FF} {\ul 0.115}}      & 0.237                                   & 0.114                                   & 0.239                              & 0.135                                  & {\color[HTML]{0000FF} {\ul 0.255}}          & 0.116                                   & 0.241                                  & {\color[HTML]{FF0000} \textbf{0.098}}       & {\color[HTML]{FF0000} \textbf{0.227}} & {\color[HTML]{FF0000} \textbf{0.121}}       & 0.246                                 & 0.105                                  & {\color[HTML]{0000FF} {\ul 0.232}}    \\
                            & 9   & {\color[HTML]{0000FF} {\ul \textbf{0.135}}}  & {\color[HTML]{FF0000} \textbf{0.251}}       & 0.159                                       & 0.275                                       & 0.151                                   & 0.269                                 & 0.276                                & 0.372                               & 0.220                                 & 0.330                                 & 0.156                                   & 0.273                                   & {\color[HTML]{0000FF} {\ul 0.147}}      & 0.266                              & 0.171                                  & 0.282                                       & 0.150                                   & {\color[HTML]{0000FF} {\ul 0.268}}     & {\color[HTML]{FF0000} \textbf{0.124}}       & {\color[HTML]{0000FF} {\ul 0.252}}    & 0.157                                       & 0.273                                 & 0.138                                  & 0.262                                 \\
                            & 12  & {\color[HTML]{0000FF} {\ul 0.176}}           & {\color[HTML]{0000FF} {\ul \textbf{0.281}}} & 0.204                                       & 0.306                                       & 0.194                                   & 0.297                                 & 0.409                                & 0.484                               & 0.255                                 & 0.355                                 & 0.193                                   & 0.296                                   & 0.196                                   & 0.298                              & 0.213                                  & 0.307                                       & 0.196                                   & 0.301                                  & {\color[HTML]{FF0000} \textbf{0.144}}       & {\color[HTML]{FF0000} \textbf{0.269}} & {\color[HTML]{FF0000} \textbf{0.200}}       & {\color[HTML]{0000FF} {\ul 0.302}}    & {\color[HTML]{0000FF} {\ul 0.169}}     & 0.285                                 \\ \cmidrule(l){2-26} 
                            & Avg & {\color[HTML]{0000FF} {\ul 0.123}}           & {\color[HTML]{0000FF} {\ul \textbf{0.240}}} & 0.143                                       & 0.261                                       & 0.136                                   & 0.255                                 & 0.278                                & 0.383                               & 0.204                                 & 0.319                                 & 0.137                                   & 0.253                                   & 0.135                                   & 0.253                              & 0.154                                  & {\color[HTML]{0000FF} {\ul 0.267}}          & 0.135                                   & 0.254                                  & {\color[HTML]{FF0000} \textbf{0.109}}       & {\color[HTML]{FF0000} \textbf{0.237}} & {\color[HTML]{FF0000} \textbf{0.141}}       & 0.259                                 & {\color[HTML]{0000FF} {\ul 0.122}}     & 0.244                                 \\ \cmidrule(l){2-26} 
                            & 24  & {\color[HTML]{0000FF} {\ul 0.186}}           & {\color[HTML]{0000FF} {\ul \textbf{0.306}}} & 0.229                                       & 0.335                                       & 0.273                                   & 0.357                                 & 0.431                                & 0.469                               & 0.315                                 & 0.393                                 & 0.231                                   & 0.349                                   & 0.240                                   & 0.345                              & 0.325                                  & 0.370                                       & {\color[HTML]{FF0000} \textbf{0.216}}   & 0.335                                  & {\color[HTML]{FF0000} \textbf{0.181}}       & {\color[HTML]{FF0000} \textbf{0.305}} & {\color[HTML]{0000FF} {\ul 0.245}}          & {\color[HTML]{0000FF} {\ul 0.350}}    & 0.276                                  & 0.358                                 \\
                            & 36  & {\color[HTML]{0000FF} {\ul 0.272}}           & {\color[HTML]{0000FF} {\ul \textbf{0.356}}} & 0.361                                       & 0.420                                       & 0.401                                   & 0.441                                 & 0.554                                & 0.552                               & {\color[HTML]{0000FF} {\ul 0.385}}    & 0.447                                 & 0.328                                   & 0.397                                   & 0.327                                   & 0.405                              & 0.428                                  & 0.442                                       & {\color[HTML]{FF0000} \textbf{0.331}}   & 0.411                                  & {\color[HTML]{FF0000} \textbf{0.226}}       & {\color[HTML]{FF0000} \textbf{0.342}} & 0.370                                       & {\color[HTML]{0000FF} {\ul 0.429}}    & 0.377                                  & 0.436                                 \\
                            & 48  & {\color[HTML]{0000FF} {\ul 0.365}}           & {\color[HTML]{0000FF} {\ul \textbf{0.391}}} & 0.501                                       & 0.507                                       & 0.530                                   & 0.522                                 & 0.694                                & 0.647                               & {\color[HTML]{0000FF} {\ul 0.436}}    & 0.486                                 & 0.450                                   & 0.473                                   & 0.446                                   & 0.475                              & 0.457                                  & 0.478                                       & {\color[HTML]{FF0000} \textbf{0.483}}   & 0.496                                  & {\color[HTML]{FF0000} \textbf{0.263}}       & {\color[HTML]{FF0000} \textbf{0.370}} & 0.504                                       & {\color[HTML]{0000FF} {\ul 0.513}}    & 0.389                                  & 0.446                                 \\
                            & 60  & 0.486                                        & {\color[HTML]{0000FF} {\ul 0.481}}          & 0.571                                       & 0.562                                       & 0.630                                   & 0.592                                 & 0.736                                & 0.673                               & {\color[HTML]{0000FF} {\ul 0.468}}    & 0.510                                 & 0.525                                   & 0.517                                   & 0.561                                   & 0.549                              & 0.596                                  & 0.566                                       & {\color[HTML]{FF0000} \textbf{0.556}}   & 0.547                                  & {\color[HTML]{FF0000} \textbf{0.323}}       & {\color[HTML]{FF0000} \textbf{0.410}} & 0.587                                       & {\color[HTML]{FF0000} \textbf{0.565}} & {\color[HTML]{0000FF} {\ul 0.478}}     & 0.507                                 \\ \cmidrule(l){2-26} 
     & Avg & {\color[HTML]{0000FF} {\ul 0.327}}           & {\color[HTML]{0000FF} {\ul \textbf{0.383}}} & 0.415                                       & 0.456                                       & 0.458                                   & 0.478                                 & 0.604                                & 0.585                               & {\color[HTML]{0000FF} {\ul 0.401}}    & 0.459                                 & 0.384                                   & 0.434                                   & 0.393                                   & 0.443                              & 0.451                                  & 0.464                                       & {\color[HTML]{FF0000} \textbf{0.396}}   & 0.447                                  & {\color[HTML]{FF0000} \textbf{0.248}}       & {\color[HTML]{FF0000} \textbf{0.357}} & 0.426                                       & {\color[HTML]{0000FF} {\ul 0.464}}    & 0.380                                  & 0.437                                 \\ \midrule
\multirow{10}{*}{\rotatebox[origin=c]{90}{Unemp}}   & 3   & {\color[HTML]{FF0000} \textbf{0.012}}        & {\color[HTML]{FF0000} \textbf{0.047}}       & 0.015                                       & 0.074                                       & {\color[HTML]{FF0000} \textbf{0.012}}   & 0.062                                 & 0.161                                & 0.289                               & 0.072                                 & 0.200                                 & 0.014                                   & 0.069                                   & {\color[HTML]{0000FF} {\ul 0.013}}      & 0.061                              & 0.014                                  & 0.070                                       & {\color[HTML]{0000FF} {\ul 0.013}}      & 0.068                                  & {\color[HTML]{FF0000} \textbf{0.012}}       & 0.061                                 & {\color[HTML]{FF0000} \textbf{0.012}}       & {\color[HTML]{0000FF} {\ul 0.060}}    & 0.062                                  & 0.171                                 \\ \cmidrule(l){3-26} 
                            & 6   & {\color[HTML]{0000FF} {\ul 0.041}}           & {\color[HTML]{FF0000} \textbf{0.108}}       & 0.057                                       & 0.154                                       & 0.043                                   & 0.130                                 & 0.229                                & 0.345                               & 0.115                                 & 0.255                                 & 0.079                                   & 0.194                                   & {\color[HTML]{FF0000} \textbf{0.040}}   & {\color[HTML]{0000FF} {\ul 0.122}} & 0.042                                  & 0.127                                       & 0.046                                   & 0.139                                  & 0.047                                       & 0.136                                 & 0.043                                       & 0.127                                 & 0.111                                  & 0.227                                 \\
                            & 9   & {\color[HTML]{FF0000} \textbf{0.084}}        & {\color[HTML]{FF0000} \textbf{0.170}}       & 0.109                                       & 0.213                                       & 0.107                                   & 0.216                                 & 0.369                                & 0.443                               & 0.191                                 & 0.329                                 & 0.101                                   & 0.200                                   & 0.119                                   & 0.226                              & {\color[HTML]{0000FF} {\ul 0.093}}     & 0.194                                       & 0.095                                   & 0.198                                  & 0.098                                       & 0.199                                 & {\color[HTML]{0000FF} {\ul 0.093}}          & {\color[HTML]{0000FF} {\ul 0.190}}    & 0.176                                  & 0.281                                 \\
                            & 12  & {\color[HTML]{FF0000} \textbf{0.131}}        & {\color[HTML]{FF0000} \textbf{0.220}}       & 0.195                                       & 0.293                                       & 0.155                                   & 0.255                                 & 0.475                                & 0.500                               & 0.240                                 & 0.386                                 & 0.196                                   & 0.291                                   & {\color[HTML]{0000FF} {\ul 0.141}}      & {\color[HTML]{0000FF} {\ul 0.239}} & 0.155                                  & 0.256                                       & 0.148                                   & 0.250                                  & 0.158                                       & 0.256                                 & 0.164                                       & 0.261                                 & 0.240                                  & 0.326                                 \\ \cmidrule(l){2-26} 
                            & Avg & {\color[HTML]{FF0000} \textbf{0.067}}        & {\color[HTML]{FF0000} \textbf{0.136}}       & 0.094                                       & 0.183                                       & 0.079                                   & 0.166                                 & 0.308                                & 0.394                               & 0.154                                 & 0.292                                 & 0.097                                   & 0.188                                   & 0.078                                   & 0.162                              & 0.076                                  & 0.162                                       & {\color[HTML]{0000FF} {\ul 0.075}}      & 0.163                                  & 0.079                                       & 0.163                                 & 0.078                                       & {\color[HTML]{0000FF} {\ul 0.160}}    & 0.147                                  & 0.251                                 \\ \cmidrule(l){2-26} 
                            & 24  & {\color[HTML]{FF0000} \textbf{0.458}}        & {\color[HTML]{FF0000} \textbf{0.447}}       & 0.668                                       & 0.549                                       & 0.655                                   & 0.548                                 & 2.520                                & 1.255                               & 0.951                                 & 0.645                                 & 0.779                                   & 0.578                                   & 0.861                                   & 0.645                              & 0.583                                  & 0.522                                       & {\color[HTML]{0000FF} {\ul 0.545}}      & {\color[HTML]{0000FF} {\ul 0.496}}     & 0.719                                       & 0.595                                 & 0.620                                       & 0.536                                 & 1.798                                  & 0.967                                 \\
                            & 36  & {\color[HTML]{FF0000} \textbf{0.870}}        & {\color[HTML]{FF0000} \textbf{0.619}}       & 1.580                                       & 0.866                                       & 1.247                                   & 0.769                                 & 4.478                                & 1.703                               & {\color[HTML]{0000FF} {\ul 0.906}}    & {\color[HTML]{0000FF} {\ul 0.717}}    & 1.318                                   & 0.777                                   & 1.741                                   & 0.938                              & 1.205                                  & 0.762                                       & 1.107                                   & 0.725                                  & 2.212                                       & 1.103                                 & 2.245                                       & 1.120                                 & 2.820                                  & 1.171                                 \\
                            & 48  & {\color[HTML]{0000FF} {\ul 1.643}}           & {\color[HTML]{0000FF} {\ul 0.900}}          & 3.702                                       & 1.483                                       & 2.700                                   & 1.209                                 & 6.385                                & 2.052                               & {\color[HTML]{FF0000} \textbf{0.974}} & {\color[HTML]{FF0000} \textbf{0.747}} & 2.977                                   & 1.258                                   & 3.526                                   & 1.427                              & 2.737                                  & 1.218                                       & 6.192                                   & 2.055                                  & 2.577                                       & 1.178                                 & 2.644                                       & 1.191                                 & 5.909                                  & 1.620                                 \\
                            & 60  & {\color[HTML]{0000FF} {\ul 2.593}}           & {\color[HTML]{0000FF} {\ul 1.182}}          & 6.320                                       & 2.030                                       & 5.370                                   & 1.830                                 & 8.460                                & 2.392                               & {\color[HTML]{FF0000} \textbf{1.071}} & {\color[HTML]{FF0000} \textbf{0.798}} & 6.582                                   & 2.060                                   & 5.574                                   & 1.857                              & 4.475                                  & 1.624                                       & 8.236                                   & 2.384                                  & 5.441                                       & 1.799                                 & 4.074                                       & 1.511                                 & 7.511                                  & 2.053                                 \\ \cmidrule(l){2-26} 
       & Avg & {\color[HTML]{0000FF} {\ul 1.391}}           & {\color[HTML]{0000FF} {\ul 0.787}}          & 3.068                                       & 1.232                                       & 2.493                                   & 1.089                                 & 5.461                                & 1.850                               & {\color[HTML]{FF0000} \textbf{0.975}} & {\color[HTML]{FF0000} \textbf{0.727}} & 2.914                                   & 1.168                                   & 2.925                                   & 1.217                              & 2.250                                  & 1.031                                       & 4.020                                   & 1.415                                  & 2.737                                       & 1.169                                 & 2.396                                       & 1.089                                 & 4.510                                  & 1.452                                 \\ \midrule
\multicolumn{2}{c}{1st Count}     & 18                                           & 35                                          & 5                                           & 6                                           & 1                                       & 0                                     & 0                                    & 0                                   & 3                                     & 3                                     & 3                                       & 1                                       & 2                                       & 0                                  & 7                                      & 9                                           & 0                                       & 0                                      & 21                                          & 14                                    & 5                                           & 2                                     & 0                                      & 0                                     \\ \bottomrule
\end{tabular}
}
\end{table}


\begin{table}[ht]
\caption{Comparison of different transformation bases. For \textit{S1} and \textit{S2}, the prediction lengths are $\tau \in \{3,6,9,12\}$ and $\tau \in \{24,36,48,60\}$, respectively. \textit{Wavelet1} and \textit{Wavelet2} denote the classic Haar and discrete Meyer wavelets, respectively. \textit{Identity} means that no transformation is applied. This is a complete version of Table~\ref{tab:base}. }
\label{tab:base_full}
\centering
\setlength{\tabcolsep}{3.2pt}
\renewcommand{\arraystretch}{1.0} 
{\fontsize{8}{9}\selectfont
\begin{tabular}{@{}cccccccccccccccc@{}}
\toprule
\multicolumn{2}{c}{Base}          & \multicolumn{2}{c}{ECL}                                                       & \multicolumn{2}{c}{Solar-Energy}                                              & \multicolumn{2}{c}{PEMS03}                                                    & \multicolumn{2}{c}{Power (S2)}                                                & \multicolumn{2}{c}{ILI (S1)}                                                  & \multicolumn{2}{c}{COVID-19 (S2)}                                              & \multicolumn{2}{c}{METR-LA (S2)}                                              \\ \midrule
\multicolumn{2}{c}{Metric}        & MSE                                   & MAE                                   & MSE                                   & MAE                                   & MSE                                   & MAE                                   & MSE                                   & MAE                                   & MSE                                   & MAE                                   & MSE                                    & MAE                                   & MSE                                   & MAE                                   \\ \midrule
                            & H1  & {\color[HTML]{FF0000} \textbf{0.131}} & {\color[HTML]{FF0000} \textbf{0.221}} & {\color[HTML]{FF0000} \textbf{0.179}} & {\color[HTML]{FF0000} \textbf{0.191}} & {\color[HTML]{FF0000} \textbf{0.060}} & {\color[HTML]{FF0000} \textbf{0.159}} & {\color[HTML]{0000FF} {\ul 1.343}}    & {\color[HTML]{0000FF} {\ul 0.870}}    & 0.468                                 & {\color[HTML]{0000FF} {\ul 0.349}}    & {\color[HTML]{FF0000} \textbf{4.474}}  & {\color[HTML]{FF0000} \textbf{1.180}} & {\color[HTML]{0000FF} {\ul 0.650}}    & {\color[HTML]{0000FF} {\ul 0.337}}    \\
\bottomrule
\end{tabular}
}
\end{table}

\begin{table}[ht]
\caption{Full results of applying OrthoTrans to iTransformer, PatchTST and RLinear. `Imp.' denotes the improvement over the corresponding vanilla model. Note that RLinear consists of only a single linear layer that projects the lookback length $T$ to the prediction horizon $\tau$. Nevertheless, OrthoTrans consistently improves RLinear's performance, demonstrating its ability to enhance model capacity. This table is the complete version of Table~\ref{tab:base_iTrans}.}
\label{tab:base_iTrans_full}
\centering
\setlength{\tabcolsep}{3.3pt}
\renewcommand{\arraystretch}{1.0} 
{\fontsize{8}{9}\selectfont
\begin{tabular}{@{}cccccccccccccc@{}}
\toprule
\multicolumn{2}{c}{}                         & \multicolumn{4}{c}{\begin{tabular}[c]{@{}c@{}}iTransformer\\  \citeyear{itransformer} \end{tabular}}                                                                            & \multicolumn{4}{c}{\begin{tabular}[c]{@{}c@{}}PatchTST\\      \citeyear{patchtst} \end{tabular}}                                                                                & \multicolumn{4}{c}{\begin{tabular}[c]{@{}c@{}}RLinear\\     \citeyear{rlinear} \end{tabular}}                                                                                 \\ \cmidrule(l){3-14} 
\multicolumn{2}{c}{\multirow{-2}{*}{Model}} & \multicolumn{2}{c}{Vanilla}                                                   & \multicolumn{2}{c}{+OrthoTrans}                                               & \multicolumn{2}{c}{Vanilla}                                                   & \multicolumn{2}{c}{+OrthoTrans}                                               & \multicolumn{2}{c}{Vanilla}                                                   & \multicolumn{2}{c}{+OrthoTrans}                                               \\ \midrule
\multicolumn{2}{c}{Metric}                   & MSE                                   & MAE                                   & MSE                                   & MAE                                   & MSE                                   & MAE                                   & MSE                                   & MAE                                   & MSE                                   & MAE                                   & MSE                                   & MAE                                   \\ \midrule
                                   & 96      & {\color[HTML]{FF0000} \textbf{0.334}} & 0.368                                 & {\color[HTML]{FF0000} \textbf{0.334}} & {\color[HTML]{FF0000} \textbf{0.367}} & 0.329                                 & 0.367                                 & {\color[HTML]{FF0000} \textbf{0.322}} & {\color[HTML]{FF0000} \textbf{0.359}} & 0.355                                 & 0.376                                 & {\color[HTML]{FF0000} \textbf{0.348}} & {\color[HTML]{FF0000} \textbf{0.366}} \\
                                   & 192     & {\color[HTML]{FF0000} \textbf{0.377}} & 0.391                                 & {\color[HTML]{FF0000} \textbf{0.377}} & {\color[HTML]{FF0000} \textbf{0.389}} & 0.367                                 & {\color[HTML]{FF0000} \textbf{0.385}} & {\color[HTML]{FF0000} \textbf{0.363}} & {\color[HTML]{FF0000} \textbf{0.383}} & 0.391                                 & 0.392                                 & {\color[HTML]{FF0000} \textbf{0.386}} & {\color[HTML]{FF0000} \textbf{0.385}} \\
                                   & 336     & 0.426                                 & 0.420                                 & {\color[HTML]{FF0000} \textbf{0.414}} & {\color[HTML]{FF0000} \textbf{0.413}} & 0.399                                 & 0.410                                 & {\color[HTML]{FF0000} \textbf{0.394}} & {\color[HTML]{FF0000} \textbf{0.405}} & 0.424                                 & 0.415                                 & {\color[HTML]{FF0000} \textbf{0.419}} & {\color[HTML]{FF0000} \textbf{0.406}} \\
                                   & 720     & 0.491                                 & 0.459                                 & {\color[HTML]{FF0000} \textbf{0.490}} & {\color[HTML]{FF0000} \textbf{0.452}} & {\color[HTML]{FF0000} \textbf{0.454}} & {\color[HTML]{FF0000} \textbf{0.439}} & 0.457                                 & 0.445                                 & 0.487                                 & 0.450                                 & {\color[HTML]{FF0000} \textbf{0.478}} & {\color[HTML]{FF0000} \textbf{0.439}} \\ \cmidrule(l){2-14} 
                                   & Avg     & 0.407                                 & 0.410                                 & {\color[HTML]{FF0000} \textbf{0.404}} & {\color[HTML]{FF0000} \textbf{0.405}} & 0.387                                 & 0.400                                 & {\color[HTML]{FF0000} \textbf{0.384}} & {\color[HTML]{FF0000} \textbf{0.398}} & 0.414                                 & 0.407                                 & {\color[HTML]{FF0000} \textbf{0.408}} & {\color[HTML]{FF0000} \textbf{0.399}} \\
\multirow{-6}{*}{ETTm1}            & Imp.    & --                                    & --                                    & 0.80\%                                & 1.16\%                                & --                                    & --                                    & 0.82\%                                & 0.53\%                                & --                                    & --                                    & 1.51\%                                & 1.97\%                                \\ \midrule
                                   & 96      & 0.148                                 & 0.240                                 & {\color[HTML]{FF0000} \textbf{0.136}} & {\color[HTML]{FF0000} \textbf{0.233}} & 0.161                                 & 0.250                                 & {\color[HTML]{FF0000} \textbf{0.153}} & {\color[HTML]{FF0000} \textbf{0.243}} & 0.201                                 & 0.281                                 & {\color[HTML]{FF0000} \textbf{0.197}} & {\color[HTML]{FF0000} \textbf{0.273}} \\
                                   & 192     & 0.162                                 & 0.253                                 & {\color[HTML]{FF0000} \textbf{0.155}} & {\color[HTML]{FF0000} \textbf{0.250}} & 0.199                                 & 0.289                                 & {\color[HTML]{FF0000} \textbf{0.165}} & {\color[HTML]{FF0000} \textbf{0.254}} & 0.201                                 & 0.283                                 & {\color[HTML]{FF0000} \textbf{0.196}} & {\color[HTML]{FF0000} \textbf{0.276}} \\
                                   & 336     & 0.178                                 & 0.269                                 & {\color[HTML]{FF0000} \textbf{0.169}} & {\color[HTML]{FF0000} \textbf{0.265}} & 0.215                                 & 0.305                                 & {\color[HTML]{FF0000} \textbf{0.182}} & {\color[HTML]{FF0000} \textbf{0.272}} & 0.215                                 & 0.298                                 & {\color[HTML]{FF0000} \textbf{0.211}} & {\color[HTML]{FF0000} \textbf{0.291}} \\
                                   & 720     & {\color[HTML]{FF0000} \textbf{0.225}} & 0.317                                 & {\color[HTML]{FF0000} \textbf{0.225}} & {\color[HTML]{FF0000} \textbf{0.311}} & 0.256                                 & 0.337                                 & {\color[HTML]{FF0000} \textbf{0.222}} & {\color[HTML]{FF0000} \textbf{0.306}} & 0.257                                 & 0.331                                 & {\color[HTML]{FF0000} \textbf{0.253}} & {\color[HTML]{FF0000} \textbf{0.324}} \\ \cmidrule(l){2-14} 
                                   & Avg     & 0.178                                 & 0.270                                 & {\color[HTML]{FF0000} \textbf{0.171}} & {\color[HTML]{FF0000} \textbf{0.265}} & 0.208                                 & 0.295                                 & {\color[HTML]{FF0000} \textbf{0.181}} & {\color[HTML]{FF0000} \textbf{0.269}} & 0.219                                 & 0.298                                 & {\color[HTML]{FF0000} \textbf{0.214}} & {\color[HTML]{FF0000} \textbf{0.291}} \\
\multirow{-6}{*}{ECL}              & Imp.    & --                                    & --                                    & 3.97\%                                & 1.87\%                                & --                                    & --                                    & 13.12\%                               & 8.98\%                                & --                                    & --                                    & 2.17\%                                & 2.35\%                                \\ \midrule
                                   & 12      & 0.071                                 & 0.174                                 & {\color[HTML]{FF0000} \textbf{0.062}} & {\color[HTML]{FF0000} \textbf{0.164}} & 0.099                                 & 0.216                                 & {\color[HTML]{FF0000} \textbf{0.073}} & {\color[HTML]{FF0000} \textbf{0.176}} & 0.126                                 & 0.236                                 & {\color[HTML]{FF0000} \textbf{0.117}} & {\color[HTML]{FF0000} \textbf{0.226}} \\
                                   & 24      & 0.093                                 & 0.201                                 & {\color[HTML]{FF0000} \textbf{0.080}} & {\color[HTML]{FF0000} \textbf{0.185}} & 0.142                                 & 0.259                                 & {\color[HTML]{FF0000} \textbf{0.104}} & {\color[HTML]{FF0000} \textbf{0.210}} & 0.246                                 & 0.334                                 & {\color[HTML]{FF0000} \textbf{0.233}} & {\color[HTML]{FF0000} \textbf{0.322}} \\
                                   & 48      & 0.125                                 & 0.236                                 & {\color[HTML]{FF0000} \textbf{0.112}} & {\color[HTML]{FF0000} \textbf{0.222}} & 0.211                                 & 0.319                                 & {\color[HTML]{FF0000} \textbf{0.181}} & {\color[HTML]{FF0000} \textbf{0.268}} & 0.551                                 & 0.529                                 & {\color[HTML]{FF0000} \textbf{0.532}} & {\color[HTML]{FF0000} \textbf{0.514}} \\
                                   & 96      & 0.164                                 & 0.275                                 & {\color[HTML]{FF0000} \textbf{0.160}} & {\color[HTML]{FF0000} \textbf{0.269}} & 0.269                                 & 0.370                                 & {\color[HTML]{FF0000} \textbf{0.293}} & {\color[HTML]{FF0000} \textbf{0.338}} & 1.057                                 & 0.787                                 & {\color[HTML]{FF0000} \textbf{1.024}} & {\color[HTML]{FF0000} \textbf{0.760}} \\ \cmidrule(l){2-14} 
                                   & Avg     & 0.113                                 & 0.221                                 & {\color[HTML]{FF0000} \textbf{0.103}} & {\color[HTML]{FF0000} \textbf{0.210}} & 0.180                                 & 0.291                                 & {\color[HTML]{FF0000} \textbf{0.163}} & {\color[HTML]{FF0000} \textbf{0.248}} & 0.495                                 & 0.472                                 & {\color[HTML]{FF0000} \textbf{0.477}} & {\color[HTML]{FF0000} \textbf{0.456}} \\
\multirow{-6}{*}{PEMS03}           & Imp.    & --                                    & --                                    & 8.45\%                                & 5.02\%                                & --                                    & --                                    & 9.69\%                                & 14.81\%                               & --                                    & --                                    & 3.74\%                                & 3.50\%                                \\ \midrule
                                   & 12      & 0.067                                 & 0.165                                 & {\color[HTML]{FF0000} \textbf{0.055}} & {\color[HTML]{FF0000} \textbf{0.147}} & 0.095                                 & 0.207                                 & {\color[HTML]{FF0000} \textbf{0.065}} & {\color[HTML]{FF0000} \textbf{0.160}} & 0.118                                 & 0.235                                 & {\color[HTML]{FF0000} \textbf{0.109}} & {\color[HTML]{FF0000} \textbf{0.221}} \\
                                   & 24      & 0.088                                 & 0.190                                 & {\color[HTML]{FF0000} \textbf{0.071}} & {\color[HTML]{FF0000} \textbf{0.165}} & 0.150                                 & 0.262                                 & {\color[HTML]{FF0000} \textbf{0.097}} & {\color[HTML]{FF0000} \textbf{0.192}} & 0.242                                 & 0.341                                 & {\color[HTML]{FF0000} \textbf{0.228}} & {\color[HTML]{FF0000} \textbf{0.324}} \\
                                   & 48      & 0.110                                 & 0.215                                 & {\color[HTML]{FF0000} \textbf{0.093}} & {\color[HTML]{FF0000} \textbf{0.188}} & 0.253                                 & 0.340                                 & {\color[HTML]{FF0000} \textbf{0.164}} & {\color[HTML]{FF0000} \textbf{0.247}} & 0.562                                 & 0.541                                 & {\color[HTML]{FF0000} \textbf{0.539}} & {\color[HTML]{FF0000} \textbf{0.522}} \\
                                   & 96      & 0.139                                 & 0.245                                 & {\color[HTML]{FF0000} \textbf{0.122}} & {\color[HTML]{FF0000} \textbf{0.218}} & 0.346                                 & 0.404                                 & {\color[HTML]{FF0000} \textbf{0.264}} & {\color[HTML]{FF0000} \textbf{0.311}} & 1.096                                 & 0.795                                 & {\color[HTML]{FF0000} \textbf{1.065}} & {\color[HTML]{FF0000} \textbf{0.773}} \\ \cmidrule(l){2-14} 
                                   & Avg     & 0.101                                 & 0.204                                 & {\color[HTML]{FF0000} \textbf{0.085}} & {\color[HTML]{FF0000} \textbf{0.179}} & 0.211                                 & 0.303                                 & {\color[HTML]{FF0000} \textbf{0.147}} & {\color[HTML]{FF0000} \textbf{0.227}} & 0.504                                 & 0.478                                 & {\color[HTML]{FF0000} \textbf{0.485}} & {\color[HTML]{FF0000} \textbf{0.460}} \\
\multirow{-6}{*}{PEMS07}           & Imp.    & --                                    & --                                    & 15.59\%                               & 11.94\%                               & --                                    & --                                    & 30.17\%                               & 24.94\%                               & --                                    & --                                    & 3.76\%                                & 3.77\%                                \\ \midrule
                                   & 96      & 0.203                                 & {\color[HTML]{FF0000} \textbf{0.237}} & {\color[HTML]{FF0000} \textbf{0.194}} & {\color[HTML]{FF0000} \textbf{0.237}} & 0.234                                 & 0.286                                 & {\color[HTML]{FF0000} \textbf{0.208}} & {\color[HTML]{FF0000} \textbf{0.248}} & 0.322                                 & 0.339                                 & {\color[HTML]{FF0000} \textbf{0.302}} & {\color[HTML]{FF0000} \textbf{0.327}} \\
                                   & 192     & 0.233                                 & {\color[HTML]{FF0000} \textbf{0.261}} & {\color[HTML]{FF0000} \textbf{0.231}} & 0.265                                 & 0.267                                 & 0.310                                 & {\color[HTML]{FF0000} \textbf{0.237}} & {\color[HTML]{FF0000} \textbf{0.268}} & 0.359                                 & 0.356                                 & {\color[HTML]{FF0000} \textbf{0.342}} & {\color[HTML]{FF0000} \textbf{0.346}} \\
                                   & 336     & 0.248                                 & 0.273                                 & {\color[HTML]{FF0000} \textbf{0.242}} & {\color[HTML]{FF0000} \textbf{0.272}} & 0.290                                 & 0.315                                 & {\color[HTML]{FF0000} \textbf{0.255}} & {\color[HTML]{FF0000} \textbf{0.281}} & 0.397                                 & 0.369                                 & {\color[HTML]{FF0000} \textbf{0.384}} & {\color[HTML]{FF0000} \textbf{0.362}} \\
                                   & 720     & 0.249                                 & {\color[HTML]{FF0000} \textbf{0.275}} & {\color[HTML]{FF0000} \textbf{0.245}} & {\color[HTML]{FF0000} \textbf{0.275}} & 0.289                                 & 0.317                                 & {\color[HTML]{FF0000} \textbf{0.255}} & {\color[HTML]{FF0000} \textbf{0.281}} & 0.397                                 & {\color[HTML]{FF0000} \textbf{0.356}} & {\color[HTML]{FF0000} \textbf{0.387}} & {\color[HTML]{FF0000} \textbf{0.356}} \\
                                   & Avg     & 0.233                                 & {\color[HTML]{FF0000} \textbf{0.262}} & {\color[HTML]{FF0000} \textbf{0.228}} & {\color[HTML]{FF0000} \textbf{0.262}} & 0.270                                 & 0.307                                 & {\color[HTML]{FF0000} \textbf{0.239}} & {\color[HTML]{FF0000} \textbf{0.269}} & 0.369                                 & 0.356                                 & {\color[HTML]{FF0000} \textbf{0.354}} & {\color[HTML]{FF0000} \textbf{0.348}} \\ \cmidrule(l){2-14} 
\multirow{-6}{*}{Solar-Energy}     & Imp.    & --                                    & --                                    & 2.16\%                                & 0.00\%                                & --                                    & --                                    & 11.57\%                               & 12.22\%                               & --                                    & --                                    & 4.14\%                                & 2.37\%                                \\ \midrule
                                  & 96      & 0.174                                 & 0.214                                 & {\color[HTML]{FF0000} \textbf{0.164}} & {\color[HTML]{FF0000} \textbf{0.208}} & 0.177                                 & 0.218                                 & {\color[HTML]{FF0000} \textbf{0.164}} & {\color[HTML]{FF0000} \textbf{0.209}} & {\color[HTML]{FF0000} \textbf{0.192}} & {\color[HTML]{FF0000} \textbf{0.232}} & {\color[HTML]{FF0000} \textbf{0.192}} & {\color[HTML]{FF0000} \textbf{0.232}} \\
                                   & 192     & {\color[HTML]{FF0000} \textbf{0.221}} & 0.254                                 & {\color[HTML]{FF0000} \textbf{0.221}} & {\color[HTML]{FF0000} \textbf{0.259}} & 0.225                                 & 0.259                                 & {\color[HTML]{FF0000} \textbf{0.211}} & {\color[HTML]{FF0000} \textbf{0.251}} & 0.240                                 & 0.271                                 & {\color[HTML]{FF0000} \textbf{0.236}} & {\color[HTML]{FF0000} \textbf{0.268}} \\
                                   & 336     & 0.278                                 & 0.296                                 & {\color[HTML]{FF0000} \textbf{0.271}} & {\color[HTML]{FF0000} \textbf{0.295}} & 0.278                                 & 0.297                                 & {\color[HTML]{FF0000} \textbf{0.266}} & {\color[HTML]{FF0000} \textbf{0.292}} & 0.292                                 & 0.307                                 & {\color[HTML]{FF0000} \textbf{0.288}} & {\color[HTML]{FF0000} \textbf{0.304}} \\
                                   & 720     & 0.358                                 & 0.349                                 & {\color[HTML]{FF0000} \textbf{0.351}} & {\color[HTML]{FF0000} \textbf{0.347}} & 0.354                                 & 0.348                                 & {\color[HTML]{FF0000} \textbf{0.345}} & {\color[HTML]{FF0000} \textbf{0.343}} & 0.364                                 & 0.353                                 & {\color[HTML]{FF0000} \textbf{0.359}} & {\color[HTML]{FF0000} \textbf{0.350}} \\ \cmidrule(l){2-14} 
                                   & Avg     & 0.258                                 & 0.279                                 & {\color[HTML]{FF0000} \textbf{0.252}} & {\color[HTML]{FF0000} \textbf{0.277}} & 0.259                                 & 0.281                                 & {\color[HTML]{FF0000} \textbf{0.246}} & {\color[HTML]{FF0000} \textbf{0.274}} & 0.272                                 & 0.291                                 & {\color[HTML]{FF0000} \textbf{0.269}} & {\color[HTML]{FF0000} \textbf{0.289}} \\
\multirow{-6}{*}{Weather}          & Imp.    & --                                    & --                                    & 2.42\%                                & 0.63\%                                & --                                    & --                                    & 4.86\%                                & 2.59\%                                & --                                    & --                                    & 1.22\%                                & 0.86\%                                \\ \midrule
                                   & 3       & {\color[HTML]{FF0000} \textbf{0.205}} & {\color[HTML]{FF0000} \textbf{0.188}} & {\color[HTML]{FF0000} \textbf{0.205}} & 0.189                                 & 0.204                                 & 0.190                                 & {\color[HTML]{FF0000} \textbf{0.203}} & {\color[HTML]{FF0000} \textbf{0.187}} & 0.210                                 & 0.195                                 & {\color[HTML]{FF0000} \textbf{0.208}} & {\color[HTML]{FF0000} \textbf{0.194}} \\
                                   & 6       & 0.300                                 & 0.229                                 & {\color[HTML]{FF0000} \textbf{0.292}} & {\color[HTML]{FF0000} \textbf{0.227}} & 0.298                                 & {\color[HTML]{FF0000} \textbf{0.227}} & {\color[HTML]{FF0000} \textbf{0.296}} & 0.228                                 & 0.304                                 & {\color[HTML]{FF0000} \textbf{0.234}} & {\color[HTML]{FF0000} \textbf{0.303}} & 0.235                                 \\
                                   & 9       & 0.386                                 & 0.265                                 & {\color[HTML]{FF0000} \textbf{0.375}} & {\color[HTML]{FF0000} \textbf{0.260}} & 0.382                                 & {\color[HTML]{FF0000} \textbf{0.263}} & {\color[HTML]{FF0000} \textbf{0.380}} & {\color[HTML]{FF0000} \textbf{0.263}} & 0.389                                 & 0.268                                 & {\color[HTML]{FF0000} \textbf{0.387}} & {\color[HTML]{FF0000} \textbf{0.266}} \\
                                   & 12      & 0.460                                 & 0.295                                 & {\color[HTML]{FF0000} \textbf{0.444}} & {\color[HTML]{FF0000} \textbf{0.291}} & 0.456                                 & {\color[HTML]{FF0000} \textbf{0.292}} & {\color[HTML]{FF0000} \textbf{0.452}} & 0.294                                 & 0.465                                 & 0.298                                 & {\color[HTML]{FF0000} \textbf{0.464}} & {\color[HTML]{FF0000} \textbf{0.297}} \\ \cmidrule(l){2-14} 
                                   & Avg     & 0.338                                 & 0.244                                 & {\color[HTML]{FF0000} \textbf{0.329}} & {\color[HTML]{FF0000} \textbf{0.242}} & 0.335                                 & {\color[HTML]{FF0000} \textbf{0.243}} & {\color[HTML]{FF0000} \textbf{0.333}} & {\color[HTML]{FF0000} \textbf{0.243}} & 0.342                                 & 0.249                                 & {\color[HTML]{FF0000} \textbf{0.341}} & {\color[HTML]{FF0000} \textbf{0.248}} \\
\multirow{-6}{*}{METR-LA}          & Imp.    & --                                    & --                                    & 2.55\%                                & 1.02\%                                & --                                    & --                                    & 0.67\%                                & 0.00\%                                & --                                    & --                                    & 0.44\%                                & 0.28\%                                \\ \bottomrule
\end{tabular}
}
\end{table}

\begin{table}[h]
\caption{Ablation results for  replacing or removing components along the variate and temporal dimensions. `Hor.' denotes the prediction horizon. The simplified results are presented in Table~\ref{tab:var_temp}.}
\label{tab:var_temp_full}
\centering
\setlength{\tabcolsep}{2.7pt}
\renewcommand{\arraystretch}{1.1} 
{\fontsize{7}{9}\selectfont
\begin{tabular}{@{}ccccccccccccccccccc@{}}
\toprule
                          &                           &                           & \multicolumn{2}{c}{ECL}                                                       & \multicolumn{2}{c}{Traffic}                                                   & \multicolumn{2}{c}{Solar}                                                     & \multicolumn{2}{c}{PEMS03}                                                    & \multicolumn{2}{c}{Weather}                                                   & \multicolumn{2}{c}{ETTm1}                                                     & \multicolumn{2}{c}{NASDAQ (S1)}                                               & \multicolumn{2}{c}{ILI (S2)}                                                  \\ \cmidrule(l){4-19} 
\multirow{-2}{*}{Var.}    & \multirow{-2}{*}{Temp.}   & \multirow{-2}{*}{Hor.} & MSE                                   & MAE                                   & MSE                                   & MAE                                   & MSE                                   & MAE                                   & MSE                                   & MAE                                   & MSE                                   & MAE                                   & MSE                                   & MAE                                   & MSE                                   & MAE                                   & MSE                                   & MAE                                   \\
\bottomrule
\end{tabular}
}
\end{table}

\begin{table}[ht]
\caption{Comparison of NormLin with the attention variants using the OLinear architecture. This is a complete version of Table~\ref{tab:normlin_vs_attn}. }
\label{tab:normlin_vs_attn_full}
\centering
\setlength{\tabcolsep}{2.2pt}
\renewcommand{\arraystretch}{1.0} 
{\fontsize{7}{9}\selectfont
%
} \\ \midrule
\multicolumn{2}{c}{Metric}      & MSE                                       & MAE                                       & MSE                                       & MAE                                   & MSE                                    & MAE                                   & MSE                                     & MAE                                    & MSE                                     & MAE                                     & MSE                                     & MAE                                 & MSE                                   & MAE                                   & MSE                                      & MAE                                  & MSE                                    & MAE                                    \\ \midrule
\multirow{5}{*}{\rotatebox[origin=c]{90}{ECL}}  & 96  & {\color[HTML]{FF0000} \textbf{0.131}}     & {\color[HTML]{FF0000} \textbf{0.221}}     & 0.138                                     & 0.226                                 & {\color[HTML]{0000FF} {\ul 0.134}}     & 0.223                                 & 0.135                                   & 0.225                                  & 0.138                                   & 0.227                                   & 0.135                                   & 0.225                               & 0.142                                 & 0.231                                 & {\color[HTML]{0000FF} {\ul 0.134}}       & {\color[HTML]{0000FF} {\ul 0.222}}   & {\color[HTML]{FF0000} \textbf{0.131}}  & {\color[HTML]{FF0000} \textbf{0.221}}  \\
                          & 192 & {\color[HTML]{FF0000} \textbf{0.150}}     & {\color[HTML]{FF0000} \textbf{0.238}}     & 0.157                                     & 0.245                                 & {\color[HTML]{0000FF} {\ul 0.153}}     & {\color[HTML]{0000FF} {\ul 0.240}}    & 0.157                                   & 0.245                                  & 0.155                                   & 0.243                                   & 0.156                                   & 0.245                               & 0.164                                 & 0.250                                 & 0.155                                    & 0.243                                & 0.154                                  & 0.242                                  \\
                          & 336 & {\color[HTML]{FF0000} \textbf{0.165}}     & {\color[HTML]{FF0000} \textbf{0.254}}     & 0.180                                     & 0.267                                 & 0.169                                  & 0.257                                 & 0.171                                   & 0.261                                  & 0.177                                   & 0.266                                   & 0.172                                   & 0.262                               & 0.180                                 & 0.267                                 & 0.174                                    & 0.261                                & {\color[HTML]{0000FF} {\ul 0.166}}     & {\color[HTML]{0000FF} {\ul 0.255}}     \\
                          & 720 & 0.191                                     & {\color[HTML]{0000FF} {\ul 0.279}}        & {\color[HTML]{0000FF} {\ul 0.190}}        & 0.281                                 & 0.211                                  & 0.293                                 & 0.196                                   & 0.283                                  & 0.197                                   & 0.285                                   & 0.194                                   & 0.283                               & 0.219                                 & 0.302                                 & 0.190                                    & {\color[HTML]{0000FF} {\ul 0.279}}   & {\color[HTML]{FF0000} \textbf{0.187}}  & {\color[HTML]{FF0000} \textbf{0.277}}  \\ \cmidrule(l){2-20} 
    & Avg & {\color[HTML]{FF0000} \textbf{0.159}}     & {\color[HTML]{FF0000} \textbf{0.248}}     & 0.166                                     & 0.255                                 & 0.167                                  & 0.253                                 & 0.165                                   & 0.253                                  & 0.167                                   & 0.255                                   & 0.164                                   & 0.254                               & 0.176                                 & 0.262                                 & {\color[HTML]{0000FF} {\ul 0.163}}       & 0.251                                & {\color[HTML]{FF0000} \textbf{0.159}}  & {\color[HTML]{0000FF} {\ul 0.249}}     \\ \midrule
\multirow{5}{*}{\rotatebox[origin=c]{90}{Traffic}}    & 96  & {\color[HTML]{FF0000} \textbf{0.398}}     & {\color[HTML]{FF0000} \textbf{0.226}}     & 0.407                                     & {\color[HTML]{0000FF} {\ul 0.228}}    & 0.413                                  & {\color[HTML]{FF0000} \textbf{0.226}} & 0.411                                   & 0.229                                  & 0.407                                   & 0.229                                   & 0.417                                   & 0.229                               & 0.420                                 & 0.250                                 & 0.409                                    & {\color[HTML]{0000FF} {\ul 0.228}}   & {\color[HTML]{0000FF} {\ul 0.403}}     & 0.229                                  \\
                          & 192 & {\color[HTML]{0000FF} {\ul 0.439}}        & {\color[HTML]{FF0000} \textbf{0.241}}     & 0.445                                     & 0.243                                 & 0.440                                  & {\color[HTML]{FF0000} \textbf{0.241}} & 0.446                                   & 0.246                                  & 0.441                                   & 0.243                                   & 0.441                                   & {\color[HTML]{0000FF} {\ul 0.242}}  & 0.445                                 & 0.263                                 & 0.441                                    & 0.243                                & {\color[HTML]{FF0000} \textbf{0.424}}  & 0.243                                  \\
                          & 336 & 0.464                                     & {\color[HTML]{FF0000} \textbf{0.250}}     & {\color[HTML]{0000FF} {\ul 0.461}}        & {\color[HTML]{0000FF} {\ul 0.252}}    & 0.470                                  & {\color[HTML]{FF0000} \textbf{0.250}} & 0.468                                   & 0.255                                  & 0.464                                   & {\color[HTML]{FF0000} \textbf{0.250}}   & 0.474                                   & {\color[HTML]{0000FF} {\ul 0.252}}  & 0.463                                 & 0.271                                 & 0.469                                    & {\color[HTML]{0000FF} {\ul 0.252}}   & {\color[HTML]{FF0000} \textbf{0.447}}  & {\color[HTML]{0000FF} {\ul 0.252}}     \\
                          & 720 & 0.502                                     & {\color[HTML]{FF0000} \textbf{0.270}}     & 0.516                                     & 0.282                                 & 0.514                                  & {\color[HTML]{FF0000} \textbf{0.270}} & 0.516                                   & 0.275                                  & 0.528                                   & 0.283                                   & 0.526                                   & 0.273                               & {\color[HTML]{0000FF} {\ul 0.496}}    & 0.290                                 & 0.513                                    & 0.274                                & {\color[HTML]{FF0000} \textbf{0.482}}  & {\color[HTML]{0000FF} {\ul 0.271}}     \\ \cmidrule(l){2-20} 
 & Avg & {\color[HTML]{0000FF} {\ul 0.451}}        & {\color[HTML]{FF0000} \textbf{0.247}}     & 0.457                                     & 0.251                                 & 0.459                                  & {\color[HTML]{FF0000} \textbf{0.247}} & 0.460                                   & 0.251                                  & 0.460                                   & 0.251                                   & 0.464                                   & {\color[HTML]{0000FF} {\ul 0.249}}  & 0.456                                 & 0.269                                 & 0.458                                    & {\color[HTML]{0000FF} {\ul 0.249}}   & {\color[HTML]{FF0000} \textbf{0.439}}  & {\color[HTML]{0000FF} {\ul 0.249}}     \\ \midrule
\multirow{5}{*}{\rotatebox[origin=c]{90}{PEMS03}}     & 12  & {\color[HTML]{FF0000} \textbf{0.060}}     & {\color[HTML]{FF0000} \textbf{0.159}}     & {\color[HTML]{FF0000} \textbf{0.060}}     & {\color[HTML]{0000FF} {\ul 0.160}}    & {\color[HTML]{FF0000} \textbf{0.060}}  & {\color[HTML]{FF0000} \textbf{0.159}} & {\color[HTML]{FF0000} \textbf{0.060}}   & {\color[HTML]{0000FF} {\ul 0.160}}     & {\color[HTML]{0000FF} {\ul 0.061}}      & {\color[HTML]{0000FF} {\ul 0.160}}      & {\color[HTML]{FF0000} \textbf{0.060}}   & {\color[HTML]{0000FF} {\ul 0.160}}  & 0.062                                 & 0.161                                 & 0.062                                    & 0.162                                & {\color[HTML]{0000FF} {\ul 0.061}}     & {\color[HTML]{0000FF} {\ul 0.160}}     \\
                          & 24  & 0.078                                     & {\color[HTML]{0000FF} {\ul 0.179}}        & {\color[HTML]{FF0000} \textbf{0.075}}     & {\color[HTML]{0000FF} {\ul 0.179}}    & 0.077                                  & {\color[HTML]{FF0000} \textbf{0.178}} & {\color[HTML]{0000FF} {\ul 0.076}}      & {\color[HTML]{0000FF} {\ul 0.179}}     & {\color[HTML]{FF0000} \textbf{0.075}}   & {\color[HTML]{FF0000} \textbf{0.178}}   & 0.077                                   & 0.181                               & 0.078                                 & 0.181                                 & 0.078                                    & 0.182                                & 0.078                                  & 0.181                                  \\
                          & 48  & {\color[HTML]{0000FF} {\ul 0.104}}        & {\color[HTML]{0000FF} {\ul 0.210}}        & 0.105                                     & 0.212                                 & {\color[HTML]{FF0000} \textbf{0.103}}  & {\color[HTML]{FF0000} \textbf{0.208}} & 0.109                                   & 0.214                                  & {\color[HTML]{0000FF} {\ul 0.104}}      & {\color[HTML]{0000FF} {\ul 0.210}}      & 0.112                                   & 0.216                               & 0.115                                 & 0.220                                 & 0.111                                    & 0.215                                & 0.108                                  & 0.213                                  \\
                          & 96  & {\color[HTML]{FF0000} \textbf{0.140}}     & {\color[HTML]{FF0000} \textbf{0.247}}     & 0.149                                     & 0.256                                 & 0.144                                  & {\color[HTML]{0000FF} {\ul 0.248}}    & 0.149                                   & 0.255                                  & 0.147                                   & 0.255                                   & 0.155                                   & 0.259                               & 0.163                                 & 0.266                                 & 0.156                                    & 0.261                                & {\color[HTML]{0000FF} {\ul 0.142}}     & 0.251                                  \\ \cmidrule(l){2-20} 
  & Avg & {\color[HTML]{FF0000} \textbf{0.095}}     & {\color[HTML]{0000FF} {\ul 0.199}}        & 0.097                                     & 0.202                                 & {\color[HTML]{0000FF} {\ul 0.096}}     & {\color[HTML]{FF0000} \textbf{0.198}} & 0.099                                   & 0.202                                  & 0.097                                   & 0.201                                   & 0.101                                   & 0.204                               & 0.104                                 & 0.207                                 & 0.102                                    & 0.205                                & 0.097                                  & 0.201                                  \\ \midrule
\multirow{5}{*}{\rotatebox[origin=c]{90}{Weather}}    & 96  & {\color[HTML]{FF0000} \textbf{0.153}}     & {\color[HTML]{FF0000} \textbf{0.190}}     & {\color[HTML]{0000FF} {\ul 0.155}}        & 0.192                                 & {\color[HTML]{FF0000} \textbf{0.153}}  & {\color[HTML]{FF0000} \textbf{0.190}} & 0.156                                   & {\color[HTML]{0000FF} {\ul 0.191}}     & 0.157                                   & 0.194                                   & 0.156                                   & 0.193                               & {\color[HTML]{0000FF} {\ul 0.155}}    & 0.193                                 & 0.156                                    & 0.194                                & {\color[HTML]{0000FF} {\ul 0.155}}     & {\color[HTML]{0000FF} {\ul 0.191}}     \\
                          & 192 & {\color[HTML]{FF0000} \textbf{0.200}}     & {\color[HTML]{FF0000} \textbf{0.235}}     & 0.206                                     & 0.240                                 & 0.204                                  & {\color[HTML]{0000FF} {\ul 0.237}}    & 0.209                                   & 0.242                                  & 0.205                                   & 0.240                                   & 0.211                                   & 0.245                               & 0.205                                 & 0.241                                 & 0.211                                    & 0.245                                & {\color[HTML]{0000FF} {\ul 0.202}}     & 0.239                                  \\
                          & 336 & {\color[HTML]{FF0000} \textbf{0.258}}     & {\color[HTML]{FF0000} \textbf{0.280}}     & 0.267                                     & 0.288                                 & 0.267                                  & 0.285                                 & 0.264                                   & 0.284                                  & 0.264                                   & 0.285                                   & 0.268                                   & 0.288                               & {\color[HTML]{0000FF} {\ul 0.263}}    & 0.285                                 & 0.266                                    & 0.285                                & 0.267                                  & {\color[HTML]{0000FF} {\ul 0.282}}     \\
                          & 720 & {\color[HTML]{FF0000} \textbf{0.337}}     & {\color[HTML]{0000FF} {\ul 0.333}}        & 0.350                                     & 0.339                                 & {\color[HTML]{0000FF} {\ul 0.340}}     & {\color[HTML]{FF0000} \textbf{0.332}} & 0.341                                   & 0.336                                  & 0.341                                   & 0.337                                   & 0.349                                   & 0.340                               & 0.345                                 & 0.339                                 & 0.361                                    & 0.347                                & 0.341                                  & 0.336                                  \\ \cmidrule(l){2-20} 
 & Avg & {\color[HTML]{FF0000} \textbf{0.237}}     & {\color[HTML]{FF0000} \textbf{0.260}}     & 0.244                                     & 0.265                                 & 0.241                                  & {\color[HTML]{0000FF} {\ul 0.261}}    & 0.242                                   & 0.263                                  & 0.242                                   & 0.264                                   & 0.246                                   & 0.266                               & 0.242                                 & 0.264                                 & 0.248                                    & 0.267                                & {\color[HTML]{0000FF} {\ul 0.241}}     & 0.262    \\ \midrule
\multirow{5}{*}{\rotatebox[origin=c]{90}{PEMS08}}      & 12  & {\color[HTML]{FF0000} \textbf{0.068}}     & {\color[HTML]{0000FF} {\ul 0.159}}        & 0.070                                     & 0.164                                 & {\color[HTML]{0000FF} {\ul 0.069}}     & {\color[HTML]{FF0000} \textbf{0.157}} & {\color[HTML]{FF0000} \textbf{0.068}}   & 0.161                                  & 0.070                                   & 0.164                                   & 0.071                                   & 0.164                               & 0.072                                 & 0.167                                 & 0.070                                    & 0.164                                & {\color[HTML]{FF0000} \textbf{0.068}}  & {\color[HTML]{0000FF} {\ul 0.159}}     \\
                          & 24  & 0.089                                     & {\color[HTML]{0000FF} {\ul 0.178}}        & 0.095                                     & 0.187                                 & {\color[HTML]{0000FF} {\ul 0.087}}     & {\color[HTML]{FF0000} \textbf{0.175}} & 0.090                                   & 0.181                                  & 0.096                                   & 0.188                                   & 0.094                                   & 0.185                               & 0.095                                 & 0.190                                 & 0.094                                    & 0.184                                & {\color[HTML]{FF0000} \textbf{0.084}}  & {\color[HTML]{0000FF} {\ul 0.178}}     \\
                          & 48  & 0.123                                     & {\color[HTML]{0000FF} {\ul 0.204}}        & 0.130                                     & 0.215                                 & {\color[HTML]{0000FF} {\ul 0.118}}     & {\color[HTML]{FF0000} \textbf{0.201}} & 0.130                                   & 0.212                                  & 0.128                                   & 0.216                                   & 0.141                                   & 0.220                               & 0.139                                 & 0.228                                 & 0.133                                    & 0.217                                & {\color[HTML]{FF0000} \textbf{0.117}}  & {\color[HTML]{0000FF} {\ul 0.204}}     \\
                          & 96  & {\color[HTML]{FF0000} \textbf{0.173}}     & {\color[HTML]{0000FF} {\ul 0.236}}        & 0.207                                     & 0.265                                 & {\color[HTML]{0000FF} {\ul 0.174}}     & {\color[HTML]{FF0000} \textbf{0.234}} & 0.195                                   & 0.252                                  & 0.206                                   & 0.263                                   & 0.213                                   & 0.260                               & 0.226                                 & 0.276                                 & 0.192                                    & 0.255                                & 0.183                                  & 0.240                                  \\ \cmidrule(l){2-20} 
  & Avg & {\color[HTML]{0000FF} {\ul 0.113}}        & {\color[HTML]{0000FF} {\ul 0.194}}        & 0.125                                     & 0.208                                 & {\color[HTML]{FF0000} \textbf{0.112}}  & {\color[HTML]{FF0000} \textbf{0.192}} & 0.121                                   & 0.201                                  & 0.125                                   & 0.208                                   & 0.130                                   & 0.207                               & 0.133                                 & 0.215                                 & 0.122                                    & 0.205                                & {\color[HTML]{0000FF} {\ul 0.113}}     & 0.195                                  \\ \midrule
\multirow{5}{*}{\rotatebox[origin=c]{90}{Solar-Energy}}    & 96  & {\color[HTML]{FF0000} \textbf{0.179}}     & {\color[HTML]{0000FF} {\ul 0.191}}        & 0.192                                     & 0.198                                 & {\color[HTML]{FF0000} \textbf{0.179}}  & 0.192                                 & {\color[HTML]{0000FF} {\ul 0.180}}      & {\color[HTML]{FF0000} \textbf{0.190}}  & 0.193                                   & 0.198                                   & 0.200                                   & 0.198                               & 0.188                                 & 0.202                                 & 0.193                                    & 0.195                                & {\color[HTML]{0000FF} {\ul 0.180}}     & 0.192                                  \\
                          & 192 & {\color[HTML]{FF0000} \textbf{0.209}}     & {\color[HTML]{FF0000} \textbf{0.213}}     & 0.220                                     & 0.219                                 & {\color[HTML]{0000FF} {\ul 0.211}}     & {\color[HTML]{0000FF} {\ul 0.214}}    & 0.218                                   & 0.220                                  & 0.215                                   & 0.215                                   & 0.230                                   & 0.222                               & 0.221                                 & 0.224                                 & 0.230                                    & 0.223                                & {\color[HTML]{0000FF} {\ul 0.211}}     & {\color[HTML]{FF0000} \textbf{0.213}}  \\
                          & 336 & {\color[HTML]{FF0000} \textbf{0.231}}     & {\color[HTML]{FF0000} \textbf{0.229}}     & 0.237                                     & 0.236                                 & {\color[HTML]{0000FF} {\ul 0.232}}     & {\color[HTML]{0000FF} {\ul 0.230}}    & 0.237                                   & 0.234                                  & 0.240                                   & 0.237                                   & 0.246                                   & 0.238                               & 0.247                                 & 0.241                                 & 0.244                                    & 0.237                                & 0.234                                  & {\color[HTML]{0000FF} {\ul 0.230}}     \\
                          & 720 & {\color[HTML]{FF0000} \textbf{0.241}}     & {\color[HTML]{FF0000} \textbf{0.236}}     & {\color[HTML]{0000FF} {\ul 0.243}}        & {\color[HTML]{0000FF} {\ul 0.240}}    & {\color[HTML]{0000FF} {\ul 0.243}}     & {\color[HTML]{FF0000} \textbf{0.236}} & 0.252                                   & 0.242                                  & 0.246                                   & 0.241                                   & 0.251                                   & 0.242                               & 0.257                                 & 0.245                                 & 0.249                                    & 0.242                                & 0.244                                  & {\color[HTML]{FF0000} \textbf{0.236}}  \\ \cmidrule(l){2-20} 
  & Avg & {\color[HTML]{FF0000} \textbf{0.215}}     & {\color[HTML]{FF0000} \textbf{0.217}}     & 0.223                                     & 0.223                                 & {\color[HTML]{0000FF} {\ul 0.216}}     & {\color[HTML]{0000FF} {\ul 0.218}}    & 0.222                                   & 0.221                                  & 0.223                                   & 0.223                                   & 0.231                                   & 0.225                               & 0.228                                 & 0.228                                 & 0.229                                    & 0.224                                & 0.217                                  & {\color[HTML]{0000FF} {\ul 0.218}}     \\ \midrule
\multirow{5}{*}{\rotatebox[origin=c]{90}{ILI}}     & 24  & {\color[HTML]{FF0000} \textbf{1.737}}     & {\color[HTML]{0000FF} {\ul 0.800}}        & 2.404                                     & 0.888                                 & {\color[HTML]{0000FF} {\ul 1.776}}     & 0.805                                 & 2.015                                   & 0.842                                  & 2.565                                   & 0.895                                   & 2.723                                   & 0.909                               & 1.867                                 & {\color[HTML]{FF0000} \textbf{0.795}} & 2.580                                    & 0.903                                & 1.846                                  & 0.824                                  \\
                          & 36  & {\color[HTML]{FF0000} \textbf{1.714}}     & {\color[HTML]{FF0000} \textbf{0.795}}     & 1.979                                     & 0.857                                 & {\color[HTML]{0000FF} {\ul 1.769}}     & {\color[HTML]{0000FF} {\ul 0.806}}    & 1.820                                   & 0.832                                  & 1.973                                   & 0.856                                   & 1.944                                   & 0.864                               & 2.002                                 & 0.846                                 & 1.832                                    & 0.835                                & 1.794                                  & 0.829                                  \\
                          & 48  & {\color[HTML]{FF0000} \textbf{1.821}}     & {\color[HTML]{FF0000} \textbf{0.804}}     & 1.895                                     & 0.827                                 & 1.911                                  & {\color[HTML]{0000FF} {\ul 0.820}}    & 1.867                                   & 0.824                                  & 1.891                                   & 0.831                                   & 1.884                                   & 0.839                               & 1.996                                 & 0.837                                 & {\color[HTML]{0000FF} {\ul 1.854}}       & 0.829                                & 1.890                                  & 0.824                                  \\
                          & 60  & {\color[HTML]{FF0000} \textbf{1.785}}     & {\color[HTML]{FF0000} \textbf{0.810}}     & {\color[HTML]{0000FF} {\ul 1.811}}        & 0.818                                 & 1.827                                  & {\color[HTML]{0000FF} {\ul 0.814}}    & 1.823                                   & 0.815                                  & 1.837                                   & 0.816                                   & 1.987                                   & 0.872                               & 1.935                                 & 0.835                                 & 1.972                                    & 0.868                                & 1.982                                  & 0.868                                  \\ \cmidrule(l){2-20} 
    & Avg & {\color[HTML]{FF0000} \textbf{1.764}}     & {\color[HTML]{FF0000} \textbf{0.802}}     & 2.022                                     & 0.847                                 & {\color[HTML]{0000FF} {\ul 1.821}}     & {\color[HTML]{0000FF} {\ul 0.811}}    & 1.881                                   & 0.828                                  & 2.066                                   & 0.849                                   & 2.134                                   & 0.871                               & 1.950                                 & 0.828                                 & 2.059                                    & 0.859                                & 1.878                                  & 0.836                                  \\ \midrule
\multirow{5}{*}{\rotatebox[origin=c]{90}{NASDAQ}}   & 3   & {\color[HTML]{0000FF} {\ul 0.036}}        & {\color[HTML]{FF0000} \textbf{0.092}}     & {\color[HTML]{0000FF} {\ul 0.036}}        & 0.094                                 & {\color[HTML]{FF0000} \textbf{0.035}}  & {\color[HTML]{FF0000} \textbf{0.092}} & {\color[HTML]{0000FF} {\ul 0.036}}      & {\color[HTML]{0000FF} {\ul 0.093}}     & {\color[HTML]{0000FF} {\ul 0.036}}      & 0.094                                   & 0.037                                   & 0.096                               & {\color[HTML]{0000FF} {\ul 0.036}}    & {\color[HTML]{0000FF} {\ul 0.093}}    & {\color[HTML]{0000FF} {\ul 0.036}}       & 0.094                                & {\color[HTML]{0000FF} {\ul 0.036}}     & {\color[HTML]{0000FF} {\ul 0.093}}     \\
                          & 6   & {\color[HTML]{FF0000} \textbf{0.049}}     & {\color[HTML]{FF0000} \textbf{0.117}}     & {\color[HTML]{0000FF} {\ul 0.050}}        & {\color[HTML]{0000FF} {\ul 0.118}}    & {\color[HTML]{FF0000} \textbf{0.049}}  & {\color[HTML]{0000FF} {\ul 0.118}}    & {\color[HTML]{FF0000} \textbf{0.049}}   & {\color[HTML]{FF0000} \textbf{0.117}}  & {\color[HTML]{0000FF} {\ul 0.050}}      & 0.119                                   & 0.052                                   & 0.122                               & {\color[HTML]{0000FF} {\ul 0.050}}    & {\color[HTML]{0000FF} {\ul 0.118}}    & 0.051                                    & 0.119                                & {\color[HTML]{FF0000} \textbf{0.049}}  & {\color[HTML]{0000FF} {\ul 0.118}}     \\
                          & 9   & {\color[HTML]{FF0000} \textbf{0.062}}     & {\color[HTML]{FF0000} \textbf{0.137}}     & {\color[HTML]{0000FF} {\ul 0.063}}        & {\color[HTML]{0000FF} {\ul 0.138}}    & {\color[HTML]{FF0000} \textbf{0.062}}  & {\color[HTML]{FF0000} \textbf{0.137}} & {\color[HTML]{FF0000} \textbf{0.062}}   & {\color[HTML]{FF0000} \textbf{0.137}}  & {\color[HTML]{0000FF} {\ul 0.063}}      & {\color[HTML]{0000FF} {\ul 0.138}}      & 0.064                                   & 0.140                               & {\color[HTML]{FF0000} \textbf{0.062}} & {\color[HTML]{0000FF} {\ul 0.138}}    & {\color[HTML]{FF0000} \textbf{0.062}}    & 0.139                                & {\color[HTML]{FF0000} \textbf{0.062}}  & {\color[HTML]{0000FF} {\ul 0.138}}     \\
                          & 12  & {\color[HTML]{FF0000} \textbf{0.073}}     & {\color[HTML]{FF0000} \textbf{0.154}}     & {\color[HTML]{0000FF} {\ul 0.074}}        & {\color[HTML]{0000FF} {\ul 0.155}}    & {\color[HTML]{0000FF} {\ul 0.074}}     & {\color[HTML]{0000FF} {\ul 0.155}}    & {\color[HTML]{FF0000} \textbf{0.073}}   & {\color[HTML]{FF0000} \textbf{0.154}}  & {\color[HTML]{0000FF} {\ul 0.074}}      & {\color[HTML]{0000FF} {\ul 0.155}}      & 0.075                                   & 0.157                               & {\color[HTML]{0000FF} {\ul 0.074}}    & {\color[HTML]{0000FF} {\ul 0.155}}    & {\color[HTML]{0000FF} {\ul 0.074}}       & 0.156                                & {\color[HTML]{0000FF} {\ul 0.074}}     & {\color[HTML]{0000FF} {\ul 0.155}}     \\ \cmidrule(l){2-20} 
 & Avg & {\color[HTML]{FF0000} \textbf{0.055}}     & {\color[HTML]{FF0000} \textbf{0.125}}     & {\color[HTML]{0000FF} {\ul 0.056}}        & {\color[HTML]{0000FF} {\ul 0.126}}    & {\color[HTML]{FF0000} \textbf{0.055}}  & {\color[HTML]{FF0000} \textbf{0.125}} & {\color[HTML]{FF0000} \textbf{0.055}}   & {\color[HTML]{FF0000} \textbf{0.125}}  & {\color[HTML]{0000FF} {\ul 0.056}}      & {\color[HTML]{0000FF} {\ul 0.126}}      & 0.057                                   & 0.129                               & {\color[HTML]{FF0000} \textbf{0.055}} & {\color[HTML]{0000FF} {\ul 0.126}}    & {\color[HTML]{0000FF} {\ul 0.056}}       & 0.127                                & {\color[HTML]{FF0000} \textbf{0.055}}  & {\color[HTML]{0000FF} {\ul 0.126}}     \\ \midrule
\multicolumn{2}{c}{1st   Count} & 27                                        & 26                                        & 3                                         & 0                                     & 10                                     & 21                                    & 6                                       & 5                                      & 1                                       & 2                                       & 1                                       & 0                                   & 2                                     & 1                                     & 1                                        & 0                                    & 15                                     & 5                                      \\ \bottomrule
\end{tabular}
}
\end{table}

\begin{table}[ht]
\caption{Applying the NormLin module to Transformer-based forecasters: iTransformer, PatchTST, Leddam, and Fredformer. For a fair comparison, model hyperparameters, loss functions, and training strategies are kept unchanged. In the case of Leddam, only the \emph{cross-channel attention} module is updated. This table presents the complete results corresponding to Table~\ref{tab:normlin_itrans} in the main text.}

\label{tab:normlin_itrans_full}
\centering
\setlength{\tabcolsep}{2.3pt}
\renewcommand{\arraystretch}{1.0} 
{\fontsize{8}{9}\selectfont
\begin{tabular}{@{}cccccccccccccccccc@{}}
\toprule
\multicolumn{2}{c}{}                        & \multicolumn{4}{c}{\begin{tabular}[c]{@{}c@{}}iTransformer\\   \citeyear{itransformer} \end{tabular}}                                            & \multicolumn{4}{c}{\begin{tabular}[c]{@{}c@{}}PatchTST\\      \citeyear{patchtst} \end{tabular}}                                                                                & \multicolumn{4}{c}{\begin{tabular}[c]{@{}c@{}}Leddam\\   \citeyear{Leddam_icml}  \end{tabular}}                                                                                  & \multicolumn{4}{c}{\begin{tabular}[c]{@{}c@{}}Fredformer\\  \citeyear{fredformer}   \end{tabular}}                                                                              \\ \cmidrule(l){3-18} 
\multicolumn{2}{c}{\multirow{-2}{*}{Model}} & \multicolumn{2}{c}{Vanilla}                   & \multicolumn{2}{c}{+ NormLin}                                                 & \multicolumn{2}{c}{Vanilla}                                                   & \multicolumn{2}{c}{+ NormLin}                                                 & \multicolumn{2}{c}{Vanilla}                                                   & \multicolumn{2}{c}{+ NormLin}                                                 & \multicolumn{2}{c}{Vanilla}                                                   & \multicolumn{2}{c}{+ NormLin}                                                 \\ \midrule
\multicolumn{2}{c}{Metric}                  & MSE   & MAE                                   & MSE                                   & MAE                                   & MSE                                   & MAE                                   & MSE                                   & MAE                                   & MSE                                   & MAE                                   & MSE                                   & MAE                                   & MSE                                   & MAE                                   & MSE                                   & MAE                                   \\ \midrule
\multirow{5}{*}{\rotatebox[origin=c]{90}{ETTm1}}    & 96        & 0.334 & 0.368                                 & {\color[HTML]{FF0000} \textbf{0.320}} & {\color[HTML]{FF0000} \textbf{0.359}} & 0.329                                 & 0.367                                 & {\color[HTML]{FF0000} \textbf{0.323}} & {\color[HTML]{FF0000} \textbf{0.361}} & {\color[HTML]{FF0000} \textbf{0.319}} & 0.359                                 & {\color[HTML]{FF0000} \textbf{0.319}} & {\color[HTML]{FF0000} \textbf{0.357}} & 0.326                                 & 0.361                                 & {\color[HTML]{FF0000} \textbf{0.318}} & {\color[HTML]{FF0000} \textbf{0.356}} \\
                                & 192       & 0.377 & 0.391                                 & {\color[HTML]{FF0000} \textbf{0.365}} & {\color[HTML]{FF0000} \textbf{0.382}} & 0.367                                 & 0.385                                 & {\color[HTML]{FF0000} \textbf{0.357}} & {\color[HTML]{FF0000} \textbf{0.384}} & 0.369                                 & 0.383                                 & {\color[HTML]{FF0000} \textbf{0.360}} & {\color[HTML]{FF0000} \textbf{0.380}} & 0.363                                 & {\color[HTML]{FF0000} \textbf{0.380}} & {\color[HTML]{FF0000} \textbf{0.362}} & 0.383                                 \\
                                & 336       & 0.426 & 0.420                                 & {\color[HTML]{FF0000} \textbf{0.400}} & {\color[HTML]{FF0000} \textbf{0.405}} & 0.399                                 & 0.410                                 & {\color[HTML]{FF0000} \textbf{0.390}} & {\color[HTML]{FF0000} \textbf{0.408}} & 0.394                                 & 0.402                                 & {\color[HTML]{FF0000} \textbf{0.391}} & {\color[HTML]{FF0000} \textbf{0.401}} & 0.395                                 & {\color[HTML]{FF0000} \textbf{0.403}} & {\color[HTML]{FF0000} \textbf{0.391}} & 0.404                                 \\
                                & 720       & 0.491 & 0.459                                 & {\color[HTML]{FF0000} \textbf{0.467}} & {\color[HTML]{FF0000} \textbf{0.443}} & 0.454                                 & {\color[HTML]{FF0000} \textbf{0.439}} & {\color[HTML]{FF0000} \textbf{0.448}} & 0.443                                 & 0.460                                 & 0.442                                 & {\color[HTML]{FF0000} \textbf{0.455}} & {\color[HTML]{FF0000} \textbf{0.439}} & {\color[HTML]{FF0000} \textbf{0.453}} & {\color[HTML]{FF0000} \textbf{0.438}} & 0.454                                 & 0.440                                 \\ \cmidrule(l){2-18} 
         & Avg       & 0.407 & 0.410                                 & {\color[HTML]{FF0000} \textbf{0.388}} & {\color[HTML]{FF0000} \textbf{0.397}} & 0.387                                 & 0.400                                 & {\color[HTML]{FF0000} \textbf{0.379}} & {\color[HTML]{FF0000} \textbf{0.399}} & 0.386                                 & 0.397                                 & {\color[HTML]{FF0000} \textbf{0.381}} & {\color[HTML]{FF0000} \textbf{0.394}} & 0.384                                 & {\color[HTML]{FF0000} \textbf{0.396}} & {\color[HTML]{FF0000} \textbf{0.381}} & {\color[HTML]{FF0000} \textbf{0.396}} \\ \midrule
\multirow{5}{*}{\rotatebox[origin=c]{90}{ECL}}     & 96        & 0.148 & 0.240                                 & {\color[HTML]{FF0000} \textbf{0.138}} & {\color[HTML]{FF0000} \textbf{0.232}} & 0.161                                 & 0.250                                 & {\color[HTML]{FF0000} \textbf{0.155}} & {\color[HTML]{FF0000} \textbf{0.246}} & 0.141                                 & 0.235                                 & {\color[HTML]{FF0000} \textbf{0.139}} & {\color[HTML]{FF0000} \textbf{0.234}} & 0.147                                 & 0.241                                 & {\color[HTML]{FF0000} \textbf{0.138}} & {\color[HTML]{FF0000} \textbf{0.235}} \\
                                & 192       & 0.162 & 0.253                                 & {\color[HTML]{FF0000} \textbf{0.154}} & {\color[HTML]{FF0000} \textbf{0.248}} & 0.199                                 & 0.289                                 & {\color[HTML]{FF0000} \textbf{0.167}} & {\color[HTML]{FF0000} \textbf{0.256}} & 0.159                                 & 0.252                                 & {\color[HTML]{FF0000} \textbf{0.155}} & {\color[HTML]{FF0000} \textbf{0.248}} & 0.165                                 & 0.258                                 & {\color[HTML]{FF0000} \textbf{0.155}} & {\color[HTML]{FF0000} \textbf{0.251}} \\
                                & 336       & 0.178 & 0.269                                 & {\color[HTML]{FF0000} \textbf{0.171}} & {\color[HTML]{FF0000} \textbf{0.265}} & 0.215                                 & 0.305                                 & {\color[HTML]{FF0000} \textbf{0.183}} & {\color[HTML]{FF0000} \textbf{0.273}} & 0.173                                 & 0.268                                 & {\color[HTML]{FF0000} \textbf{0.171}} & {\color[HTML]{FF0000} \textbf{0.265}} & 0.177                                 & 0.273                                 & {\color[HTML]{FF0000} \textbf{0.172}} & {\color[HTML]{FF0000} \textbf{0.271}} \\
                                & 720       & 0.225 & 0.317                                 & {\color[HTML]{FF0000} \textbf{0.201}} & {\color[HTML]{FF0000} \textbf{0.297}} & 0.256                                 & 0.337                                 & {\color[HTML]{FF0000} \textbf{0.221}} & {\color[HTML]{FF0000} \textbf{0.307}} & 0.201                                 & 0.295                                 & {\color[HTML]{FF0000} \textbf{0.195}} & {\color[HTML]{FF0000} \textbf{0.293}} & 0.213                                 & {\color[HTML]{FF0000} \textbf{0.304}} & {\color[HTML]{FF0000} \textbf{0.210}} & {\color[HTML]{FF0000} \textbf{0.304}} \\ \cmidrule(l){2-18} 
                                & Avg       & 0.178 & 0.270                                 & {\color[HTML]{FF0000} \textbf{0.166}} & {\color[HTML]{FF0000} \textbf{0.260}} & 0.208                                 & 0.295                                 & {\color[HTML]{FF0000} \textbf{0.181}} & {\color[HTML]{FF0000} \textbf{0.270}} & 0.169                                 & 0.263                                 & {\color[HTML]{FF0000} \textbf{0.165}} & {\color[HTML]{FF0000} \textbf{0.260}} & 0.176                                 & 0.269                                 & {\color[HTML]{FF0000} \textbf{0.169}} & {\color[HTML]{FF0000} \textbf{0.265}} \\ \midrule
\multirow{5}{*}{\rotatebox[origin=c]{90}{PEMS03}}    & 12        & 0.071 & 0.174                                 & {\color[HTML]{FF0000} \textbf{0.062}} & {\color[HTML]{FF0000} \textbf{0.164}} & 0.099                                 & 0.216                                 & {\color[HTML]{FF0000} \textbf{0.071}} & {\color[HTML]{FF0000} \textbf{0.176}} & {\color[HTML]{FF0000} \textbf{0.063}} & {\color[HTML]{FF0000} \textbf{0.164}} & {\color[HTML]{FF0000} \textbf{0.063}} & {\color[HTML]{FF0000} \textbf{0.164}} & 0.068                                 & 0.174                                 & {\color[HTML]{FF0000} \textbf{0.064}} & {\color[HTML]{FF0000} \textbf{0.168}} \\
                                & 24        & 0.093 & 0.201                                 & {\color[HTML]{FF0000} \textbf{0.078}} & {\color[HTML]{FF0000} \textbf{0.184}} & 0.142                                 & 0.259                                 & {\color[HTML]{FF0000} \textbf{0.103}} & {\color[HTML]{FF0000} \textbf{0.211}} & 0.080                                 & 0.185                                 & {\color[HTML]{FF0000} \textbf{0.079}} & {\color[HTML]{FF0000} \textbf{0.182}} & 0.093                                 & 0.202                                 & {\color[HTML]{FF0000} \textbf{0.081}} & {\color[HTML]{FF0000} \textbf{0.188}} \\
                                & 48        & 0.125 & 0.236                                 & {\color[HTML]{FF0000} \textbf{0.110}} & {\color[HTML]{FF0000} \textbf{0.219}} & 0.211                                 & 0.319                                 & {\color[HTML]{FF0000} \textbf{0.160}} & {\color[HTML]{FF0000} \textbf{0.248}} & 0.124                                 & 0.226                                 & {\color[HTML]{FF0000} \textbf{0.110}} & {\color[HTML]{FF0000} \textbf{0.217}} & 0.146                                 & 0.258                                 & {\color[HTML]{FF0000} \textbf{0.119}} & {\color[HTML]{FF0000} \textbf{0.228}} \\
                                & 96        & 0.164 & 0.275                                 & {\color[HTML]{FF0000} \textbf{0.159}} & {\color[HTML]{FF0000} \textbf{0.269}} & 0.269                                 & 0.370                                 & {\color[HTML]{FF0000} \textbf{0.251}} & {\color[HTML]{FF0000} \textbf{0.331}} & 0.160                                 & {\color[HTML]{FF0000} \textbf{0.266}} & {\color[HTML]{FF0000} \textbf{0.159}} & {\color[HTML]{FF0000} \textbf{0.266}} & 0.228                                 & 0.330                                 & {\color[HTML]{FF0000} \textbf{0.170}} & {\color[HTML]{FF0000} \textbf{0.284}} \\ \cmidrule(l){2-18} 
        & Avg       & 0.113 & 0.221                                 & {\color[HTML]{FF0000} \textbf{0.102}} & {\color[HTML]{FF0000} \textbf{0.209}} & 0.180                                 & 0.291                                 & {\color[HTML]{FF0000} \textbf{0.146}} & {\color[HTML]{FF0000} \textbf{0.241}} & 0.107                                 & 0.210                                 & {\color[HTML]{FF0000} \textbf{0.103}} & {\color[HTML]{FF0000} \textbf{0.207}} & 0.134                                 & 0.241                                 & {\color[HTML]{FF0000} \textbf{0.108}} & {\color[HTML]{FF0000} \textbf{0.217}} \\ \midrule
\multirow{5}{*}{\rotatebox[origin=c]{90}{PEMS07}}    & 12        & 0.067 & 0.165                                 & {\color[HTML]{FF0000} \textbf{0.055}} & {\color[HTML]{FF0000} \textbf{0.145}} & 0.095                                 & 0.207                                 & {\color[HTML]{FF0000} \textbf{0.084}} & {\color[HTML]{FF0000} \textbf{0.199}} & 0.055                                 & {\color[HTML]{FF0000} \textbf{0.145}} & {\color[HTML]{FF0000} \textbf{0.054}} & {\color[HTML]{FF0000} \textbf{0.145}} & 0.063                                 & 0.158                                 & {\color[HTML]{FF0000} \textbf{0.057}} & {\color[HTML]{FF0000} \textbf{0.150}} \\
                                & 24        & 0.088 & 0.190                                 & {\color[HTML]{FF0000} \textbf{0.070}} & {\color[HTML]{FF0000} \textbf{0.162}} & 0.150                                 & 0.262                                 & {\color[HTML]{FF0000} \textbf{0.098}} & {\color[HTML]{FF0000} \textbf{0.198}} & 0.070                                 & 0.164                                 & {\color[HTML]{FF0000} \textbf{0.067}} & {\color[HTML]{FF0000} \textbf{0.160}} & 0.089                                 & 0.190                                 & {\color[HTML]{FF0000} \textbf{0.073}} & {\color[HTML]{FF0000} \textbf{0.168}} \\
                                & 48        & 0.110 & 0.215                                 & {\color[HTML]{FF0000} \textbf{0.091}} & {\color[HTML]{FF0000} \textbf{0.187}} & 0.253                                 & {\color[HTML]{FF0000} \textbf{0.340}} & {\color[HTML]{FF0000} \textbf{0.243}} & 0.341                                 & 0.094                                 & 0.192                                 & {\color[HTML]{FF0000} \textbf{0.087}} & {\color[HTML]{FF0000} \textbf{0.184}} & 0.135                                 & 0.239                                 & {\color[HTML]{FF0000} \textbf{0.099}} & {\color[HTML]{FF0000} \textbf{0.201}} \\
                                & 96        & 0.139 & 0.245                                 & {\color[HTML]{FF0000} \textbf{0.128}} & {\color[HTML]{FF0000} \textbf{0.220}} & 0.346                                 & 0.404                                 & {\color[HTML]{FF0000} \textbf{0.249}} & {\color[HTML]{FF0000} \textbf{0.310}} & {\color[HTML]{FF0000} \textbf{0.117}} & {\color[HTML]{FF0000} \textbf{0.217}} & 0.119                                 & 0.222                                 & 0.196                                 & 0.294                                 & {\color[HTML]{FF0000} \textbf{0.157}} & {\color[HTML]{FF0000} \textbf{0.267}} \\ \cmidrule(l){2-18} 
       & Avg       & 0.101 & 0.204                                 & {\color[HTML]{FF0000} \textbf{0.086}} & {\color[HTML]{FF0000} \textbf{0.178}} & 0.211                                 & 0.303                                 & {\color[HTML]{FF0000} \textbf{0.168}} & {\color[HTML]{FF0000} \textbf{0.262}} & 0.084                                 & 0.180                                 & {\color[HTML]{FF0000} \textbf{0.082}} & {\color[HTML]{FF0000} \textbf{0.178}} & 0.121                                 & 0.220                                 & {\color[HTML]{FF0000} \textbf{0.096}} & {\color[HTML]{FF0000} \textbf{0.196}} \\ \midrule
\multirow{5}{*}{\rotatebox[origin=c]{90}{Solar-Energy}}  & 96        & 0.203 & 0.237                                 & {\color[HTML]{FF0000} \textbf{0.194}} & {\color[HTML]{FF0000} \textbf{0.233}} & 0.234                                 & 0.286                                 & {\color[HTML]{FF0000} \textbf{0.208}} & {\color[HTML]{FF0000} \textbf{0.249}} & 0.197                                 & 0.241                                 & {\color[HTML]{FF0000} \textbf{0.186}} & {\color[HTML]{FF0000} \textbf{0.227}} & {\color[HTML]{FF0000} \textbf{0.185}} & 0.233                                 & 0.190                                 & {\color[HTML]{FF0000} \textbf{0.231}} \\
                                & 192       & 0.233 & 0.261                                 & {\color[HTML]{FF0000} \textbf{0.225}} & {\color[HTML]{FF0000} \textbf{0.258}} & 0.267                                 & 0.310                                 & {\color[HTML]{FF0000} \textbf{0.234}} & {\color[HTML]{FF0000} \textbf{0.266}} & 0.231                                 & 0.264                                 & {\color[HTML]{FF0000} \textbf{0.221}} & {\color[HTML]{FF0000} \textbf{0.258}} & 0.227                                 & {\color[HTML]{FF0000} \textbf{0.253}} & {\color[HTML]{FF0000} \textbf{0.222}} & 0.254                                 \\
                                & 336       & 0.248 & {\color[HTML]{FF0000} \textbf{0.273}} & {\color[HTML]{FF0000} \textbf{0.241}} & {\color[HTML]{FF0000} \textbf{0.273}} & 0.290                                 & 0.315                                 & {\color[HTML]{FF0000} \textbf{0.252}} & {\color[HTML]{FF0000} \textbf{0.278}} & {\color[HTML]{FF0000} \textbf{0.241}} & {\color[HTML]{FF0000} \textbf{0.268}} & 0.242                                 & 0.270                                 & 0.246                                 & 0.284                                 & {\color[HTML]{FF0000} \textbf{0.245}} & {\color[HTML]{FF0000} \textbf{0.275}} \\
                                & 720       & 0.249 & {\color[HTML]{FF0000} \textbf{0.275}} & {\color[HTML]{FF0000} \textbf{0.244}} & 0.277                                 & 0.289                                 & 0.317                                 & {\color[HTML]{FF0000} \textbf{0.253}} & {\color[HTML]{FF0000} \textbf{0.277}} & 0.250                                 & 0.281                                 & {\color[HTML]{FF0000} \textbf{0.239}} & {\color[HTML]{FF0000} \textbf{0.271}} & {\color[HTML]{FF0000} \textbf{0.247}} & {\color[HTML]{FF0000} \textbf{0.276}} & 0.249                                 & 0.281                                 \\ \cmidrule(l){2-18} 
        & Avg       & 0.233 & 0.262                                 & {\color[HTML]{FF0000} \textbf{0.226}} & {\color[HTML]{FF0000} \textbf{0.260}} & 0.270                                 & 0.307                                 & {\color[HTML]{FF0000} \textbf{0.237}} & {\color[HTML]{FF0000} \textbf{0.267}} & 0.230                                 & 0.264                                 & {\color[HTML]{FF0000} \textbf{0.222}} & {\color[HTML]{FF0000} \textbf{0.257}} & {\color[HTML]{FF0000} \textbf{0.226}} & 0.262                                 &  {\color[HTML]{FF0000} \textbf{0.226}}                                 & {\color[HTML]{FF0000} \textbf{0.260}} \\ \midrule
\multirow{5}{*}{\rotatebox[origin=c]{90}{Weather}}    & 96        & 0.174 & 0.214                                 & {\color[HTML]{FF0000} \textbf{0.160}} & {\color[HTML]{FF0000} \textbf{0.205}} & 0.177                                 & 0.218                                 & {\color[HTML]{FF0000} \textbf{0.161}} & {\color[HTML]{FF0000} \textbf{0.206}} & {\color[HTML]{FF0000} \textbf{0.156}} & 0.202                                 & {\color[HTML]{FF0000} \textbf{0.156}} & {\color[HTML]{FF0000} \textbf{0.201}} & 0.163                                 & 0.207                                 & {\color[HTML]{FF0000} \textbf{0.156}} & {\color[HTML]{FF0000} \textbf{0.202}} \\
                                & 192       & 0.221 & 0.254                                 & {\color[HTML]{FF0000} \textbf{0.209}} & {\color[HTML]{FF0000} \textbf{0.250}} & 0.225                                 & 0.259                                 & {\color[HTML]{FF0000} \textbf{0.207}} & {\color[HTML]{FF0000} \textbf{0.249}} & 0.207                                 & 0.250                                 & {\color[HTML]{FF0000} \textbf{0.206}} & {\color[HTML]{FF0000} \textbf{0.249}} & 0.211                                 & 0.251                                 & {\color[HTML]{FF0000} \textbf{0.205}} & {\color[HTML]{FF0000} \textbf{0.248}} \\
                            & 336       & 0.278 & 0.296 & {\color[HTML]{FF0000} \textbf{0.268}} & {\color[HTML]{FF0000} \textbf{0.294}}                                 & 0.278                                 & 0.297                                 & {\color[HTML]{FF0000} \textbf{0.266}} & {\color[HTML]{FF0000} \textbf{0.291}} & {\color[HTML]{FF0000} \textbf{0.262}} & {\color[HTML]{FF0000} \textbf{0.291}} & {\color[HTML]{FF0000} \textbf{0.262}} & {\color[HTML]{FF0000} \textbf{0.291}} & 0.267                                 & 0.292                                 & {\color[HTML]{FF0000} \textbf{0.261}} & {\color[HTML]{FF0000} \textbf{0.289}} \\
                        & 720       & 0.358 & 0.349 & {\color[HTML]{FF0000} \textbf{0.344}} & {\color[HTML]{FF0000} \textbf{0.344}} & 0.354                                 & 0.348                                 & {\color[HTML]{FF0000} \textbf{0.345}} & {\color[HTML]{FF0000} \textbf{0.344}} & {\color[HTML]{FF0000} \textbf{0.343}} & {\color[HTML]{FF0000} \textbf{0.343}} & 0.345                                 & 0.344                                 & 0.343                                 & {\color[HTML]{FF0000} \textbf{0.341}} & {\color[HTML]{FF0000} \textbf{0.340}} & {\color[HTML]{FF0000} \textbf{0.341}} \\ \cmidrule(l){2-18} 
      & Avg       & 0.258 & 0.279                                 & {\color[HTML]{FF0000} \textbf{0.245}} & {\color[HTML]{FF0000} \textbf{0.273}} & 0.259                                 & 0.281                                 & {\color[HTML]{FF0000} \textbf{0.245}} & {\color[HTML]{FF0000} \textbf{0.272}} & {\color[HTML]{FF0000} \textbf{0.242}} & 0.272                                 & {\color[HTML]{FF0000} \textbf{0.242}} & {\color[HTML]{FF0000} \textbf{0.271}} & 0.246                                 & 0.273                                 & {\color[HTML]{FF0000} \textbf{0.240}} & {\color[HTML]{FF0000} \textbf{0.270}} \\ \midrule
\multirow{5}{*}{\rotatebox[origin=c]{90}{METR-LA}}    & 3         & 0.205 & {\color[HTML]{FF0000} \textbf{0.188}} & {\color[HTML]{FF0000} \textbf{0.202}} & 0.189                                 & {\color[HTML]{FF0000} \textbf{0.204}} & 0.190                                 & 0.205                                 & {\color[HTML]{FF0000} \textbf{0.188}} & 0.204                                 & 0.191                                 & {\color[HTML]{FF0000} \textbf{0.202}} & {\color[HTML]{FF0000} \textbf{0.189}} & 0.205                                 & 0.188                                 & {\color[HTML]{FF0000} \textbf{0.200}} & {\color[HTML]{FF0000} \textbf{0.186}} \\
                                & 6         & 0.300 & 0.229                                 & {\color[HTML]{FF0000} \textbf{0.293}} & {\color[HTML]{FF0000} \textbf{0.226}} & 0.298                                 & {\color[HTML]{FF0000} \textbf{0.227}} & {\color[HTML]{FF0000} \textbf{0.297}} & 0.229                                 & 0.293                                 & 0.227                                 & {\color[HTML]{FF0000} \textbf{0.289}} & {\color[HTML]{FF0000} \textbf{0.226}} & 0.298                                 & {\color[HTML]{FF0000} \textbf{0.227}} & {\color[HTML]{FF0000} \textbf{0.296}} & {\color[HTML]{FF0000} \textbf{0.227}} \\
                                & 9         & 0.386 & 0.265                                 & {\color[HTML]{FF0000} \textbf{0.375}} & {\color[HTML]{FF0000} \textbf{0.259}} & 0.382                                 & 0.263                                 & {\color[HTML]{FF0000} \textbf{0.379}} & {\color[HTML]{FF0000} \textbf{0.262}} & 0.369                                 & 0.264                                 & {\color[HTML]{FF0000} \textbf{0.366}} & {\color[HTML]{FF0000} \textbf{0.259}} & 0.385                                 & {\color[HTML]{FF0000} \textbf{0.263}} & {\color[HTML]{FF0000} \textbf{0.369}} & {\color[HTML]{FF0000} \textbf{0.263}} \\
                                & 12        & 0.460 & 0.295                                 & {\color[HTML]{FF0000} \textbf{0.442}} & {\color[HTML]{FF0000} \textbf{0.289}} & 0.456                                 & {\color[HTML]{FF0000} \textbf{0.292}} & {\color[HTML]{FF0000} \textbf{0.453}} & 0.293                                 & 0.442                                 & 0.292                                 & {\color[HTML]{FF0000} \textbf{0.423}} & {\color[HTML]{FF0000} \textbf{0.287}} & 0.457                                 & 0.292                                 & {\color[HTML]{FF0000} \textbf{0.451}} & {\color[HTML]{FF0000} \textbf{0.291}} \\ \cmidrule(l){2-18} 
          & Avg       & 0.338 & 0.244                                 & {\color[HTML]{FF0000} \textbf{0.328}} & {\color[HTML]{FF0000} \textbf{0.241}} & 0.335                                 & {\color[HTML]{FF0000} \textbf{0.243}} & {\color[HTML]{FF0000} \textbf{0.333}} & {\color[HTML]{FF0000} \textbf{0.243}} & 0.327                                 & 0.243                                 & {\color[HTML]{FF0000} \textbf{0.320}} & {\color[HTML]{FF0000} \textbf{0.240}} & 0.336                                 & {\color[HTML]{FF0000} \textbf{0.242}} & {\color[HTML]{FF0000} \textbf{0.329}} & {\color[HTML]{FF0000} \textbf{0.242}} \\ \bottomrule
\end{tabular}
}
\end{table}

\section{OLinear-C} \label{append_orthoc}

\subsection{Forecasting performance}

Table~\ref{tab:ortho-c-full} compares OLinear and OLinear-C on both short- and long-term forecasting tasks. OLinear-C achieves performance comparable to OLinear while using fewer learnable parameters and offering improved efficiency (see Table~\ref{tab:GPU}). However, due to its fixed weight matrix, $\mathrm{NormLin}_c$ lacks the flexibility required to serve as a plug-and-play module for other forecasters. Therefore, we focus on OLinear in this work.

\begin{table}[t]
\caption{Full evaluation of OLinear-C. \textit{S1} and \textit{S2} correspond to  `Input-12, Predict-$\left \{ 3,6,9,12 \right \}$' and `Input-36, Predict-$\left \{ 24,36,48,60 \right \}$', respectively. This table presents the full results of Table~\ref{tab:OLinear-c_short}.}
\label{tab:ortho-c-full}
\centering
\setlength{\tabcolsep}{3.5pt}
\renewcommand{\arraystretch}{0.9} 
{\fontsize{7}{8}\selectfont
\begin{tabular}{@{}cccccccccccccccc@{}}
\toprule
\multicolumn{2}{c}{Dataset}           & \multicolumn{2}{c}{ETTm1}                                                     & \multicolumn{2}{c}{ETTm2}                                                     & \multicolumn{2}{c}{ETTh1}                                                     & \multicolumn{2}{c}{ETTh2}                                                     & \multicolumn{2}{c}{ECL}                                                       & \multicolumn{2}{c}{Exchange}                                                   & \multicolumn{2}{c}{Traffic}                                                   \\ \midrule
\multicolumn{2}{c}{Metric}            & MSE                                   & MAE                                   & MSE                                   & MAE                                   & MSE                                   & MAE                                   & MSE                                   & MAE                                   & MSE                                   & MAE                                   & MSE                                    & MAE                                   & MSE                                   & MAE                                   \\ \midrule
\multirow{5}{*}{\rotatebox[origin=c]{90}{OLinear}}    & 96  & 0.302                                 & 0.334                                 & 0.169                                 & 0.249                                 & 0.360                                 & 0.382                                 & 0.284                                 & 0.329                                 & 0.131                                 & 0.221                                 & 0.082                                  & 0.200                                 & 0.398                                 & 0.226                                 \\
\bottomrule
\end{tabular}
}
\end{table}

\subsection{Robustness} \label{robust_subsec_c}

Table~\ref{tab:robust_C} reports the standard deviation of OLinear-C across seven random seeds, demonstrating its robustness to independent runs. For a broader robustness comparison, Table~\ref{tab:robust_compare} presents 99\% confidence intervals for OLinear, OLinear-C, TimeMixer++, and iTransformer.

\begin{table}[t]
\caption{Robustness of OLinear-C performance. Standard deviations are calculated over seven random seeds.}
\label{tab:robust_C}
\centering
\setlength{\tabcolsep}{3pt}
\renewcommand{\arraystretch}{1.0} 
{\fontsize{9}{10}\selectfont

\begin{tabular}{@{}cccccccccc@{}}
\toprule
\multicolumn{2}{c}{Dataset}    & \multicolumn{2}{c}{ECL}           & \multicolumn{2}{c}{Traffic}    & \multicolumn{2}{c}{ETTm1}         & \multicolumn{2}{c}{Solar-Energy} \\ \midrule
\multicolumn{2}{c}{Metric}     & MSE              & MAE            & MSE            & MAE           & MSE              & MAE            & MSE             & MAE            \\ \midrule
\multirow{4}{*}{\rotatebox[origin=c]{90}{Horizon}} & 96  & 0.130±3e-4       & 0.220±4e-4     & 0.404±4e-3     & 0.227±7e-4    & 0.303±7e-4       & 0.335±7e-4     & 0.178±1e-3      & 0.191±7e-4     \\
                         & 192 & 0.156±1e-3       & 0.243±9e-4     & 0.435±2e-3     & 0.241±4e-4    & 0.357±2e-4       & 0.364±2e-4     & 0.209±8e-4      & 0.213±3e-4     \\
                         & 336 & 0.167±1e-3       & 0.256±8e-4     & 0.465±5e-3     & 0.250±2e-4    & 0.389±9e-4       & 0.386±5e-4     & 0.231±5e-4      & 0.229±4e-5     \\
                         & 720 & 0.192±9e-3       & 0.279±6e-3     & 0.502±5e-3     & 0.271±6e-4    & 0.452±6e-4       & 0.426±4e-4     & 0.241±4e-4      & 0.236±4e-4     \\ \midrule
\multicolumn{2}{c}{Dataset}    & \multicolumn{2}{c}{Weather}       & \multicolumn{2}{c}{PEMS03}     & \multicolumn{2}{c}{NASDAQ (S1)}   & \multicolumn{2}{c}{Wiki (S1)}    \\ \midrule
\multicolumn{2}{c}{Metric}     & MSE              & MAE            & MSE            & MAE           & MSE              & MAE            & MSE             & MAE            \\ \midrule
\multirow{4}{*}{\rotatebox[origin=c]{90}{Horizon}} & H1  & 0.154±1e-3       & 0.191±8e-4     & 0.061±4e-4     & 0.160±4e-4    & 0.035±2e-4       & 0.092±5e-4     & 6.187±9e-3      & 0.368±7e-4     \\
                         & H2  & 0.204±3-3        & 0.238±2e-3     & 0.076±8e-4     & 0.178±8e-4    & 0.049±8e-5       & 0.117±4e-4     & 6.457±7e-3      & 0.384±6e-4     \\
                         & H3  & 0.259±4e-3       & 0.280±4e-3     & 0.105±9e-4     & 0.211±6e-4    & 0.061±0.000      & 0.137±3e-4     & 6.652±4e-3      & 0.396±6e-4     \\
                         & H4  & 0.343±6e-3       & 0.334±3e-3     & 0.142±2e-3     & 0.248±1e-3    & 0.073±5e-5       & 0.153±2e-4     & 6.835±3e-3      & 0.406±2e-4     \\ \midrule
\multicolumn{2}{c}{Dataset}    & \multicolumn{2}{c}{DowJones (S2)} & \multicolumn{2}{c}{SP500 (S2)} & \multicolumn{2}{c}{CarSales (S1)} & \multicolumn{2}{c}{Power (S2)}   \\ \midrule
\multicolumn{2}{c}{Metric}     & MSE              & MAE            & MSE            & MAE           & MSE              & MAE            & MSE             & MAE            \\ \midrule
\multirow{4}{*}{\rotatebox[origin=c]{90}{Horizon}} & H1  & 7.496±4e-2       & 0.665±9e-4     & 0.156±2e-3     & 0.271±2e-3    & 0.303±1e-3       & 0.276±1e-3     & 1.460±4e-2      & 0.922±1e-2     \\
                         & H2  & 10.965±7e-2      & 0.802±1e-3     & 0.210±1e-3     & 0.319±1e-3    & 0.317±7e-4       & 0.287±6e-4     & 1.527±3e-2      & 0.939±2e-2     \\
                         & H3  & 14.161±5e-2      & 0.915±7e-4     & 0.262±2e-3     & 0.358±2e-3    & 0.328±7e-4       & 0.294±6e-4     & 1.721±7e-2      & 1.016±3e-2     \\
                         & H4  & 17.084±7e-2      & 1.018±1e-3     & 0.296±3e-3     & 0.378±3e-3    & 0.337±4e-4       & 0.300±4e-4     & 1.869±1e-2      & 1.077±5e-3     \\ \bottomrule
\end{tabular}
}
\end{table}

\subsection{Ablation studies}


We conduct ablation studies on different transformations of the Pearson correlation matrix $\mathrm{CorrMat}_v$, including $\mathrm{Softplus}$, $\mathrm{Sigmoid}$, $\mathrm{ReLU}$, and the identity function (i.e., no transformation). All variants are followed by row-wise L1 normalization. As shown in Table~\ref{tab:OL-c-abl}, $\mathrm{Softmax}(\mathrm{CorrMat}_v)$ consistently outperforms the other options and is adopted in the main experiments.

\begin{table}[th]
\caption{Ablation study on different transformations of $\mathrm{CorrMat}_v$}
\label{tab:OL-c-abl}
\centering
\setlength{\tabcolsep}{2pt}
\renewcommand{\arraystretch}{1.0} 
{\fontsize{7}{8}\selectfont
\begin{tabular}{@{}cccccccccccccccccccc@{}}
\toprule
\multicolumn{2}{c}{Dataset}      & \multicolumn{2}{c}{ECL}                                                       & \multicolumn{2}{c}{Traffic}                                                   & \multicolumn{2}{c}{Solar}                                                     & \multicolumn{2}{c}{PEMS03}                                                    & \multicolumn{2}{c}{Weather}                                                   & \multicolumn{2}{c}{NASDAQ S1}                                                 & \multicolumn{2}{c}{ILI S2}                                                    & \multicolumn{2}{c}{Web\_S1}                                                   & \multicolumn{2}{c}{Cars\_S2}                                                  \\ \midrule
\multicolumn{2}{c}{Metric}       & MSE                                   & MAE                                   & MSE                                   & MAE                                   & MSE                                   & MAE                                   & MSE                                   & MAE                                   & MSE                                   & MAE                                   & MSE                                   & MAE                                   & MSE                                   & MAE                                   & MSE                                   & MAE                                   & MSE                                   & MAE                                   \\
\bottomrule
\end{tabular}
}
\end{table}

\section{More experiments}

\subsection{Performance under varying lookback horizons}

We evaluate the forecasting performance of OLinear and its variant, OLinear-C, under varying lookback horizons. Instead of using a fixed lookback length, we search for the optimal horizon within the range of 96 to 720. As shown in Table~\ref{tab:best_var_lookback}, both models consistently achieve state-of-the-art results across different forecasting lengths and datasets, demonstrating their robustness and strong performance.

\begin{table}[ht]
\caption{Performance with varying lookback horizons. The best and second-best results are highlighted in {\color[HTML]{FF0000} \textbf{bold}} and {\color[HTML]{0000FF} {\ul underlined}}, respectively.  \textit{OLinear-C} refers to the OLinear variant where the weight matrix in NormLin is replaced by $\mathrm{Softmax} \left ( \mathrm{CorrMat} _v \right )$, with $\mathrm{CorrMat} _v$ being the cross-variate correlation matrix. }
\label{tab:best_var_lookback}
\centering
\setlength{\tabcolsep}{1.1pt}
\renewcommand{\arraystretch}{1.0} 
{\fontsize{7}{9}\selectfont
\begin{tabular}{@{}cccccccccccccccccccccc@{}}
\toprule
\multicolumn{2}{c}{Model}       & \multicolumn{2}{c}{\begin{tabular}[c]{@{}c@{}}OLinear\\      (Ours)\end{tabular}} & \multicolumn{2}{c}{\begin{tabular}[c]{@{}c@{}}OLinear-C\\      (Ours)\end{tabular}} & \multicolumn{2}{c}{\begin{tabular}[c]{@{}c@{}}Leddam\\      ()\end{tabular}} & \multicolumn{2}{c}{\begin{tabular}[c]{@{}c@{}}CARD\\      ()\end{tabular}}    & \multicolumn{2}{c}{\begin{tabular}[c]{@{}c@{}}Fredformer\\      ()\end{tabular}} & \multicolumn{2}{c}{\begin{tabular}[c]{@{}c@{}}iTrans.\\      ()\end{tabular}} & \multicolumn{2}{c}{\begin{tabular}[c]{@{}c@{}}TimeMixer\\      ()\end{tabular}} & \multicolumn{2}{c}{\begin{tabular}[c]{@{}c@{}}PatchTST\\      ()\end{tabular}} & \multicolumn{2}{c}{\begin{tabular}[c]{@{}c@{}}TimesNet\\      ()\end{tabular}} & \multicolumn{2}{c}{\begin{tabular}[c]{@{}c@{}}DLinear\\      ()\end{tabular}} \\ \midrule
\multicolumn{2}{c}{Metric}      & MSE                                       & MAE                                       & MSE                                        & MAE                                        & MSE                                   & MAE                                  & MSE                                   & MAE                                   & MSE                                                     & MAE                    & MSE                                   & MAE                                   & MSE                                    & MAE                                    & MSE                                    & MAE                                   & MSE                                    & MAE                                   & MSE                                   & MAE                                   \\ \midrule
\multirow{5}{*}{\rotatebox[origin=c]{90}{ECL}}    & 96  & {\color[HTML]{FF0000} \textbf{0.123}}     & {\color[HTML]{FF0000} \textbf{0.212}}     & {\color[HTML]{FF0000} \textbf{0.123}}      & {\color[HTML]{FF0000} \textbf{0.212}}      & 0.134                                 & 0.227                                & {\color[HTML]{0000FF} {\ul 0.129}}    & 0.223                                 & {\color[HTML]{0000FF} {\ul 0.129}}                      & 0.226                  & 0.132                                 & 0.227                                 & {\color[HTML]{0000FF} {\ul 0.129}}     & 0.224                                  & {\color[HTML]{0000FF} {\ul 0.129}}     & {\color[HTML]{0000FF} {\ul 0.222}}    & 0.168                                  & 0.272                                 & 0.140                                 & 0.237                                 \\
                          & 192 & {\color[HTML]{0000FF} {\ul 0.143}}        & {\color[HTML]{0000FF} {\ul 0.232}}        & 0.144                                      & 0.233                                      & 0.156                                 & 0.248                                & 0.154                                 & 0.245                                 & 0.148                                                   & 0.244                  & 0.154                                 & 0.251                                 & {\color[HTML]{FF0000} \textbf{0.140}}  & {\color[HTML]{FF0000} \textbf{0.220}}  & 0.147                                  & 0.240                                 & 0.184                                  & 0.322                                 & 0.153                                 & 0.249                                 \\
                          & 336 & {\color[HTML]{0000FF} {\ul 0.157}}        & {\color[HTML]{FF0000} \textbf{0.247}}     & {\color[HTML]{FF0000} \textbf{0.156}}      & {\color[HTML]{FF0000} \textbf{0.247}}      & 0.166                                 & 0.264                                & 0.161                                 & 0.257                                 & 0.165                                                   & 0.262                  & 0.170                                 & 0.268                                 & 0.161                                  & {\color[HTML]{0000FF} {\ul 0.255}}     & 0.163                                  & 0.259                                 & 0.198                                  & 0.300                                 & 0.169                                 & 0.267                                 \\
                          & 720 & {\color[HTML]{0000FF} {\ul 0.181}}        & {\color[HTML]{FF0000} \textbf{0.270}}     & {\color[HTML]{FF0000} \textbf{0.180}}      & {\color[HTML]{FF0000} \textbf{0.270}}      & 0.195                                 & 0.291                                & 0.185                                 & {\color[HTML]{0000FF} {\ul 0.278}}    & 0.193                                                   & 0.286                  & 0.193                                 & 0.288                                 & 0.194                                  & 0.287                                  & 0.197                                  & 0.290                                 & 0.220                                  & 0.320                                 & 0.203                                 & 0.301                                 \\ \cmidrule(l){2-22} 
 & Avg & {\color[HTML]{FF0000} \textbf{0.151}}     & {\color[HTML]{FF0000} \textbf{0.240}}     & {\color[HTML]{FF0000} \textbf{0.151}}      & {\color[HTML]{FF0000} \textbf{0.240}}      & 0.163                                 & 0.257                                & 0.157                                 & 0.251                                 & 0.159                                                   & 0.254                  & 0.162                                 & 0.258                                 & {\color[HTML]{0000FF} {\ul 0.156}}     & {\color[HTML]{0000FF} {\ul 0.247}}     & 0.159                                  & 0.253                                 & 0.193                                  & 0.304                                 & 0.166                                 & 0.264                                 \\ \midrule
 
\multirow{5}{*}{\rotatebox[origin=c]{90}{Traffic}}     & 96  & {\color[HTML]{FF0000} \textbf{0.338}}     & {\color[HTML]{FF0000} \textbf{0.221}}     & {\color[HTML]{0000FF} {\ul 0.340}}         & {\color[HTML]{FF0000} \textbf{0.221}}      & 0.366                                 & 0.260                                & 0.341                                 & {\color[HTML]{0000FF} {\ul 0.229}}    & 0.358                                                   & 0.257                  & 0.359                                 & 0.262                                 & 0.360                                  & 0.249                                  & 0.360                                  & 0.249                                 & 0.593                                  & 0.321                                 & 0.410                                 & 0.282                                 \\
                          & 192 & {\color[HTML]{FF0000} \textbf{0.361}}     & {\color[HTML]{FF0000} \textbf{0.233}}     & 0.369                                      & {\color[HTML]{0000FF} {\ul 0.235}}         & 0.394                                 & 0.270                                & {\color[HTML]{0000FF} {\ul 0.367}}    & 0.243                                 & 0.381                                                   & 0.272                  & 0.376                                 & 0.270                                 & 0.375                                  & 0.250                                  & 0.379                                  & 0.256                                 & 0.617                                  & 0.336                                 & 0.423                                 & 0.287                                 \\
                          & 336 & {\color[HTML]{0000FF} {\ul 0.386}}        & {\color[HTML]{FF0000} \textbf{0.241}}     & 0.390                                      & {\color[HTML]{FF0000} \textbf{0.241}}      & 0.400                                 & 0.283                                & 0.388                                 & {\color[HTML]{0000FF} {\ul 0.254}}    & 0.396                                                   & 0.277                  & 0.393                                 & 0.279                                 & {\color[HTML]{FF0000} \textbf{0.385}}  & 0.270                                  & 0.392                                  & 0.264                                 & 0.629                                  & 0.336                                 & 0.436                                 & 0.296                                 \\
                          & 720 & 0.444                                     & {\color[HTML]{FF0000} \textbf{0.264}}     & 0.451                                      & {\color[HTML]{0000FF} {\ul 0.266}}         & 0.442                                 & 0.297                                & {\color[HTML]{0000FF} {\ul 0.427}}    & 0.276                                 & {\color[HTML]{FF0000} \textbf{0.424}}                   & 0.296                  & 0.434                                 & 0.293                                 & 0.430                                  & 0.281                                  & 0.432                                  & 0.286                                 & 0.640                                  & 0.350                                 & 0.466                                 & 0.315                                 \\ \cmidrule(l){2-22} 
 & Avg & {\color[HTML]{0000FF} {\ul 0.382}}        & {\color[HTML]{FF0000} \textbf{0.240}}     & 0.387                                      & {\color[HTML]{0000FF} {\ul 0.241}}         & 0.400                                 & 0.278                                & {\color[HTML]{FF0000} \textbf{0.381}} & 0.251                                 & 0.390                                                   & 0.275                  & 0.390                                 & 0.276                                 & 0.388                                  & 0.263                                  & 0.391                                  & 0.264                                 & 0.620                                  & 0.336                                 & 0.434                                 & 0.295                                 \\ \midrule
\multirow{5}{*}{\rotatebox[origin=c]{90}{Weather}}       & 96  & {\color[HTML]{FF0000} \textbf{0.144}}     & {\color[HTML]{0000FF} {\ul 0.184}}        & {\color[HTML]{FF0000} \textbf{0.144}}      & {\color[HTML]{FF0000} \textbf{0.183}}      & 0.149                                 & 0.199                                & {\color[HTML]{0000FF} {\ul 0.145}}    & 0.186                                 & 0.150                                                   & 0.203                  & 0.165                                 & 0.214                                 & 0.147                                  & 0.197                                  & 0.149                                  & 0.198                                 & 0.172                                  & 0.220                                 & 0.176                                 & 0.237                                 \\
                          & 192 & 0.190                                     & 0.230                                     & {\color[HTML]{0000FF} {\ul 0.189}}         & {\color[HTML]{0000FF} {\ul 0.228}}         & 0.196                                 & 0.243                                & {\color[HTML]{FF0000} \textbf{0.187}} & {\color[HTML]{FF0000} \textbf{0.227}} & 0.194                                                   & 0.246                  & 0.208                                 & 0.253                                 & {\color[HTML]{0000FF} {\ul 0.189}}     & 0.239                                  & 0.194                                  & 0.241                                 & 0.219                                  & 0.261                                 & 0.220                                 & 0.282                                 \\
                          & 336 & {\color[HTML]{FF0000} \textbf{0.235}}     & {\color[HTML]{0000FF} {\ul 0.268}}        & {\color[HTML]{0000FF} {\ul 0.236}}         & {\color[HTML]{0000FF} {\ul 0.268}}         & 0.243                                 & 0.280                                & 0.238                                 & {\color[HTML]{FF0000} \textbf{0.258}} & 0.243                                                   & 0.284                  & 0.257                                 & 0.292                                 & 0.241                                  & 0.280                                  & 0.306                                  & 0.282                                 & 0.246                                  & 0.337                                 & 0.265                                 & 0.319                                 \\
                          & 720 & 0.316                                     & {\color[HTML]{0000FF} {\ul 0.323}}        & 0.316                                      & {\color[HTML]{0000FF} {\ul 0.323}}         & 0.321                                 & 0.334                                & {\color[HTML]{FF0000} \textbf{0.308}} & {\color[HTML]{FF0000} \textbf{0.321}} & {\color[HTML]{FF0000} \textbf{0.308}}                   & 0.333                  & 0.331                                 & 0.343                                 & {\color[HTML]{0000FF} {\ul 0.310}}     & 0.330                                  & 0.314                                  & 0.334                                 & 0.365                                  & 0.359                                 & 0.323                                 & 0.362                                 \\ \cmidrule(l){2-22} 
 & Avg & {\color[HTML]{0000FF} {\ul 0.221}}        & 0.251                                     & {\color[HTML]{0000FF} {\ul 0.221}}         & {\color[HTML]{0000FF} {\ul 0.250}}         & 0.227                                 & 0.264                                & {\color[HTML]{FF0000} \textbf{0.220}} & {\color[HTML]{FF0000} \textbf{0.248}} & 0.224                                                   & 0.266                  & 0.240                                 & 0.275                                 & 0.222                                  & 0.262                                  & 0.241                                  & 0.264                                 & 0.251                                  & 0.294                                 & 0.246                                 & 0.300                                 \\ \midrule
\multirow{5}{*}{\rotatebox[origin=c]{90}{Solar-Energy}}      & 96  & {\color[HTML]{0000FF} {\ul 0.169}}        & {\color[HTML]{FF0000} \textbf{0.194}}     & 0.195                                      & {\color[HTML]{0000FF} {\ul 0.207}}         & 0.186                                 & 0.242                                & 0.170                                 & {\color[HTML]{0000FF} {\ul 0.207}}    & 0.187                                                   & 0.236                  & 0.190                                 & 0.241                                 & {\color[HTML]{FF0000} \textbf{0.167}}  & 0.220                                  & 0.224                                  & 0.278                                 & 0.219                                  & 0.314                                 & 0.289                                 & 0.377                                 \\
                          & 192 & 0.190                                     & {\color[HTML]{FF0000} \textbf{0.212}}     & {\color[HTML]{0000FF} {\ul 0.189}}         & {\color[HTML]{FF0000} \textbf{0.212}}      & 0.208                                 & 0.262                                & 0.192                                 & {\color[HTML]{0000FF} {\ul 0.219}}    & 0.196                                                   & 0.251                  & 0.233                                 & 0.261                                 & {\color[HTML]{FF0000} \textbf{0.187}}  & 0.249                                  & 0.253                                  & 0.298                                 & 0.231                                  & 0.322                                 & 0.319                                 & 0.397                                 \\
                          & 336 & {\color[HTML]{FF0000} \textbf{0.192}}     & {\color[HTML]{FF0000} \textbf{0.215}}     & {\color[HTML]{0000FF} {\ul 0.194}}         & {\color[HTML]{0000FF} {\ul 0.219}}         & 0.218                                 & 0.265                                & 0.226                                 & 0.233                                 & 0.208                                                   & 0.265                  & 0.226                                 & 0.275                                 & 0.200                                  & 0.258                                  & 0.273                                  & 0.306                                 & 0.246                                  & 0.337                                 & 0.352                                 & 0.415                                 \\
                          & 720 & {\color[HTML]{0000FF} {\ul 0.204}}        & {\color[HTML]{FF0000} \textbf{0.225}}     & {\color[HTML]{FF0000} \textbf{0.203}}      & {\color[HTML]{0000FF} {\ul 0.226}}         & 0.208                                 & 0.273                                & 0.217                                 & 0.243                                 & 0.209                                                   & 0.272                  & 0.220                                 & 0.282                                 & 0.215                                  & 0.250                                  & 0.272                                  & 0.308                                 & 0.280                                  & 0.363                                 & 0.356                                 & 0.412                                 \\ \cmidrule(l){2-22} 
  & Avg & {\color[HTML]{FF0000} \textbf{0.189}}     & {\color[HTML]{FF0000} \textbf{0.211}}     & 0.195                                      & {\color[HTML]{0000FF} {\ul 0.216}}         & 0.205                                 & 0.261                                & 0.201                                 & 0.225                                 & 0.200                                                   & 0.256                  & 0.217                                 & 0.265                                 & {\color[HTML]{0000FF} {\ul 0.192}}     & 0.244                                  & 0.256                                  & 0.298                                 & 0.244                                  & 0.334                                 & 0.329                                 & 0.400                                 \\ \midrule
  
\multirow{5}{*}{\rotatebox[origin=c]{90}{ETTm1}}      & 96  & {\color[HTML]{FF0000} \textbf{0.275}}     & {\color[HTML]{FF0000} \textbf{0.326}}     & {\color[HTML]{FF0000} \textbf{0.275}}      & {\color[HTML]{FF0000} \textbf{0.326}}      & 0.294                                 & 0.347                                & 0.288                                 & {\color[HTML]{0000FF} {\ul 0.332}}    & {\color[HTML]{0000FF} {\ul 0.284}}                      & 0.338                  & 0.309                                 & 0.357                                 & 0.291                                  & 0.340                                  & 0.293                                  & 0.346                                 & 0.338                                  & 0.375                                 & 0.299                                 & 0.343                                 \\
                          & 192 & {\color[HTML]{0000FF} {\ul 0.318}}        & {\color[HTML]{0000FF} {\ul 0.352}}        & {\color[HTML]{FF0000} \textbf{0.317}}      & {\color[HTML]{FF0000} \textbf{0.351}}      & 0.336                                 & 0.369                                & 0.332                                 & 0.357                                 & 0.323                                                   & 0.364                  & 0.346                                 & 0.383                                 & 0.327                                  & 0.365                                  & 0.333                                  & 0.370                                 & 0.374                                  & 0.387                                 & 0.335                                 & 0.365                                 \\
                          & 336 & {\color[HTML]{FF0000} \textbf{0.352}}     & {\color[HTML]{FF0000} \textbf{0.373}}     & {\color[HTML]{FF0000} \textbf{0.352}}      & {\color[HTML]{FF0000} \textbf{0.373}}      & 0.364                                 & 0.389                                & 0.364                                 & {\color[HTML]{0000FF} {\ul 0.376}}    & {\color[HTML]{0000FF} {\ul 0.358}}                      & 0.387                  & 0.385                                 & 0.410                                 & 0.360                                  & 0.381                                  & 0.369                                  & 0.392                                 & 0.410                                  & 0.411                                 & 0.369                                 & 0.386                                 \\
                          & 720 & {\color[HTML]{FF0000} \textbf{0.406}}     & {\color[HTML]{0000FF} {\ul 0.408}}        & {\color[HTML]{FF0000} \textbf{0.406}}      & {\color[HTML]{0000FF} {\ul 0.408}}         & 0.421                                 & 0.419                                & {\color[HTML]{0000FF} {\ul 0.414}}    & {\color[HTML]{FF0000} \textbf{0.407}} & 0.420                                                   & 0.417                  & 0.440                                 & 0.442                                 & 0.415                                  & 0.417                                  & 0.416                                  & 0.420                                 & 0.478                                  & 0.450                                 & 0.425                                 & 0.421                                 \\ \cmidrule(l){2-22} 
  & Avg & {\color[HTML]{FF0000} \textbf{0.338}}     & {\color[HTML]{0000FF} {\ul 0.365}}        & {\color[HTML]{FF0000} \textbf{0.338}}      & {\color[HTML]{FF0000} \textbf{0.364}}      & 0.354                                 & 0.381                                & 0.350                                 & 0.368                                 & {\color[HTML]{0000FF} {\ul 0.346}}                      & 0.376                  & 0.370                                 & 0.398                                 & 0.348                                  & 0.376                                  & 0.353                                  & 0.382                                 & 0.400                                  & 0.406                                 & 0.357                                 & 0.379                                 \\ \midrule
\multicolumn{2}{c}{1st   Count} & 12                                        & 16                                        & 11                                         & 12                                         & 0                                     & 0                                    & 4                                     & 4                                     & 2                                                       & 0                      & 0                                     & 0                                     & 4                                      & 1                                      & 0                                      & 0                                     & 0                                      & 0                                     & 0                                     & 0                                     \\ \bottomrule
\end{tabular}
}
\end{table}

\subsection{Few-shot and zero-shot forecasting}

To evaluate OLinear's adaptability to sparse data and its transferability to unseen datasets, we conduct few-shot and zero-shot learning experiments, with the results shown in Tables~\ref{tab:few_shot} and \ref{tab:zero_shot}, respectively. As a linear-based model, OLinear exhibits surprisingly strong generalization capabilities, an attribute more commonly associated with Transformer-based models \citep{gpt3_2020}. Notably, OLinear achieves this with high computational efficiency (see Table~\ref{tab:GPU}). Furthermore, OLinear delivers performance on par with or superior to TimeMixer++ \citep{timemixer++}, despite the latter’s architectural complexity, which includes multi-scale design, series imaging, and dual-axis attention mechanisms.

This performance advantage may be attributed to the high-rank property of the NormLin weight matrix (see Appendix~\ref{low_rank}) and its enhanced gradient flow (see Appendix~\ref{jacob}). In contrast, attention matrices derived from the classic query-key mechanism typically exhibit low-rank characteristics (see Figures~\ref{fig:orthotrans_improve_rank}- \ref{fig:orthotrans_improve_rank_patchtst}), leading to a compressed representation space \citep{flattentrans}.

\vspace{10pt}
\begin{table}[h]
\caption{Few-shot forecasting performance with 10\% training data. Results are averaged over four prediction lengths $\left \{ 96,192,336,720 \right \}$.}
\label{tab:few_shot}
\centering
\setlength{\tabcolsep}{1.3pt}
\renewcommand{\arraystretch}{1.0} 
{\fontsize{7}{9}\selectfont
\begin{tabular}{@{}ccccccccccccccccccccc@{}}
\toprule
Model   & \multicolumn{2}{c}{\begin{tabular}[c]{@{}c@{}}OLinear\\      (Ours)\end{tabular}} & 
\multicolumn{2}{c}{\begin{tabular}[c]{@{}c@{}}TimeMixer++\\  \citeyear{timemixer++} \end{tabular}} & 
\multicolumn{2}{c}{\begin{tabular}[c]{@{}c@{}}TimeMixer\\   \citeyear{timemixer} \end{tabular}} & 
\multicolumn{2}{c}{\begin{tabular}[c]{@{}c@{}}iTrans.\\      \citeyear{itransformer} \end{tabular}} & 
\multicolumn{2}{c}{\begin{tabular}[c]{@{}c@{}}PatchTST\\      \citeyear{patchtst} \end{tabular}} & 
\multicolumn{2}{c}{\begin{tabular}[c]{@{}c@{}}TimesNet\\      \citeyear{timesnet} \end{tabular}} &
\multicolumn{2}{c}{\begin{tabular}[c]{@{}c@{}}DLinear\\      \citeyear{linear} \end{tabular}} & 
\multicolumn{2}{c}{\begin{tabular}[c]{@{}c@{}}TiDE\\      \citeyear{tide} \end{tabular}} & 
\multicolumn{2}{c}{\begin{tabular}[c]{@{}c@{}}Crossfm.\\      \citeyear{crossformer} \end{tabular}} & 
\multicolumn{2}{c}{\begin{tabular}[c]{@{}c@{}}FEDformer\\      \citeyear{fedformer} \end{tabular}} \\ \midrule
Metric  & MSE                                       & MAE                                       & MSE                                     & MAE                                     & MSE                                                   & MAE                     & MSE                                   & MAE                                   & MSE                                    & MAE                                   & MSE                                    & MAE                                   & MSE                                   & MAE                                   & MSE                                  & MAE                                 & MSE                                    & MAE                                   & MSE                                    & MAE                                    \\ \midrule
ETTh1   & {\color[HTML]{FF0000} \textbf{0.478}}     & {\color[HTML]{FF0000} \textbf{0.456}}     & {\color[HTML]{0000FF} {\ul 0.517}}      & {\color[HTML]{0000FF} {\ul 0.512}}      & 0.613                                                 & 0.520                   & 0.510                                 & 0.597                                 & 0.633                                  & 0.542                                 & 0.869                                  & 0.628                                 & 0.691                                 & 0.600                                 & 0.589                                & 0.535                               & 0.645                                  & 0.558                                 & 0.639                                  & 0.561                                  \\
ETTh2   & {\color[HTML]{0000FF} {\ul 0.390}}        & {\color[HTML]{0000FF} {\ul 0.407}}        & {\color[HTML]{FF0000} \textbf{0.379}}   & {\color[HTML]{FF0000} \textbf{0.391}}   & 0.402                                                 & 0.433                   & 0.455                                 & 0.461                                 & 0.415                                  & 0.431                                 & 0.479                                  & 0.465                                 & 0.605                                 & 0.538                                 & 0.395                                & 0.412                               & 0.428                                  & 0.447                                 & 0.466                                  & 0.475                                  \\
ETTm1   & {\color[HTML]{0000FF} {\ul 0.405}}        & {\color[HTML]{FF0000} \textbf{0.400}}     & {\color[HTML]{FF0000} \textbf{0.398}}   & {\color[HTML]{0000FF} {\ul 0.431}}      & 0.487                                                 & 0.461                   & 0.491                                 & 0.516                                 & 0.501                                  & 0.466                                 & 0.677                                  & 0.537                                 & 0.411                                 & 0.429                                 & 0.425                                & 0.458                               & 0.462                                  & 0.489                                 & 0.722                                  & 0.605                                  \\
ETTm2   & {\color[HTML]{FF0000} \textbf{0.287}}     & {\color[HTML]{FF0000} \textbf{0.330}}     & {\color[HTML]{0000FF} {\ul 0.291}}      & {\color[HTML]{0000FF} {\ul 0.351}}      & 0.311                                                 & 0.367                   & 0.375                                 & 0.412                                 & 0.296                                  & 0.343                                 & 0.320                                  & 0.353                                 & 0.316                                 & 0.368                                 & 0.317                                & 0.371                               & 0.343                                  & 0.389                                 & 0.463                                  & 0.488                                  \\
Weather & 0.245                                     & {\color[HTML]{FF0000} \textbf{0.265}}     & {\color[HTML]{FF0000} \textbf{0.241}}   & {\color[HTML]{0000FF} {\ul 0.271}}      & {\color[HTML]{0000FF} {\ul 0.242}}                    & 0.281                   & 0.291                                 & 0.331                                 & 0.242                                  & 0.279                                 & 0.279                                  & 0.301                                 & 0.241                                 & 0.283                                 & 0.249                                & 0.291                               & 0.267                                  & 0.306                                 & 0.284                                  & 0.324                                  \\ \bottomrule
\end{tabular}
}
\end{table}
\vspace{10pt}

\vspace{5pt}
\begin{table}[h]
\caption{Zero-shot forecasting performance. Results are averaged over four prediction lengths $\left \{ 96,192,336,720 \right \}$. $\mathcal{A} \rightarrow   \mathcal{B}$ denotes that the model is trained on dataset $\mathcal{A}$ but evaluated on  $\mathcal{B}$. }
\label{tab:zero_shot}
\centering
\setlength{\tabcolsep}{1.3pt}
\renewcommand{\arraystretch}{1.0} 
{\fontsize{7}{9}\selectfont
\begin{tabular}{@{}ccccccccccccccccccccc@{}}
\toprule
Model       & \multicolumn{2}{c}{\begin{tabular}[c]{@{}c@{}}OLinear\\      (Ours)\end{tabular}}     & \multicolumn{2}{c}{\begin{tabular}[c]{@{}c@{}}TimeMixer++\\  \citeyear{timemixer++} \end{tabular}}         & 
\multicolumn{2}{c}{\begin{tabular}[c]{@{}c@{}}TimeMixer\\    \citeyear{timemixer} \end{tabular}} & 
\multicolumn{2}{c}{\begin{tabular}[c]{@{}c@{}}LLMTime\\      \citeyear{LLMtime} \end{tabular}} & 
\multicolumn{2}{c}{\begin{tabular}[c]{@{}c@{}}DLinear\\      \citeyear{linear} \end{tabular}} & 
\multicolumn{2}{c}{\begin{tabular}[c]{@{}c@{}}PatchTST\\      \citeyear{patchtst} \end{tabular}} & 
\multicolumn{2}{c}{\begin{tabular}[c]{@{}c@{}}TimesNet\\      \citeyear{timesnet} \end{tabular}} & 
\multicolumn{2}{c}{\begin{tabular}[c]{@{}c@{}}iTrans.\\      \citeyear{itransformer} \end{tabular}} & 
\multicolumn{2}{c}{\begin{tabular}[c]{@{}c@{}}Crossfm.\\      \citeyear{crossformer} \end{tabular}} & 
\multicolumn{2}{c}{\begin{tabular}[c]{@{}c@{}}FEDformer\\      \citeyear{fedformer} \end{tabular}} \\ \midrule
Metric      & MSE                                         & MAE                                         & MSE                                         & MAE                                         & MSE                                                   & MAE                     & MSE                                   & MAE                                   & MSE                                   & MAE                                   & MSE                                    & MAE                                   & MSE                                    & MAE                                   & MSE                                   & MAE                                   & MSE                                    & MAE                                   & MSE                                    & MAE                                    \\ \midrule
ETTh1→ETTh2 & {\color[HTML]{FF0000} \textbf{0.367}}       & {\color[HTML]{FF0000} \textbf{0.391}}       & {\color[HTML]{FF0000} {\ul \textbf{0.367}}} & {\color[HTML]{FF0000} {\ul \textbf{0.391}}} & 0.427                                                 & 0.424                   & 0.992                                 & 0.708                                 & 0.493                                 & 0.488                                 & {\color[HTML]{0000FF} {\ul 0.380}}     & {\color[HTML]{0000FF} {\ul 0.405}}    & 0.421                                  & 0.431                                 & 0.481                                 & 0.474                                 & 0.555                                  & 0.574                                 & 0.712                                  & 0.693                                  \\
ETTh1→ETTm2 & {\color[HTML]{0000FF} {\ul 0.311}}          & {\color[HTML]{FF0000} {\ul \textbf{0.352}}} & {\color[HTML]{FF0000} \textbf{0.301}}       & {\color[HTML]{0000FF} {\ul \textbf{0.357}}} & 0.361                                                 & 0.397                   & 1.867                                 & 0.869                                 & 0.415                                 & 0.452                                 & 0.314                                  & 0.360                                 & 0.327                                  & 0.361                                 & {\color[HTML]{0000FF} {\ul 0.311}}    & 0.361                                 & 0.613                                  & 0.629                                 & 0.681                                  & 0.588                                  \\
ETTh2→ETTh1 & {\color[HTML]{FF0000} {\ul \textbf{0.507}}} & {\color[HTML]{FF0000} \textbf{0.479}}       & {\color[HTML]{0000FF} {\ul \textbf{0.511}}} & {\color[HTML]{0000FF} {\ul 0.498}}          & 0.679                                                 & 0.577                   & 1.961                                 & 0.981                                 & 0.703                                 & 0.574                                 & 0.565                                  & 0.513                                 & 0.865                                  & 0.621                                 & 0.552                                 & 0.511                                 & 0.587                                  & 0.518                                 & 0.612                                  & 0.624                                  \\
ETTm1→ETTh2 & {\color[HTML]{FF0000} \textbf{0.413}}       & {\color[HTML]{0000FF} {\ul \textbf{0.425}}} & {\color[HTML]{0000FF} {\ul 0.417}}          & {\color[HTML]{FF0000} {\ul \textbf{0.422}}} & 0.452                                                 & 0.441                   & 0.992                                 & 0.708                                 & 0.464                                 & 0.475                                 & 0.439                                  & 0.438                                 & 0.457                                  & 0.454                                 & 0.434                                 & 0.438                                 & 0.624                                  & 0.541                                 & 0.533                                  & 0.594                                  \\
ETTm1→ETTm2 & {\color[HTML]{FF0000} \textbf{0.291}}       & {\color[HTML]{FF0000} \textbf{0.327}}       & {\color[HTML]{FF0000} \textbf{0.291}}       & {\color[HTML]{0000FF} {\ul 0.331}}          & {\color[HTML]{0000FF} {\ul 0.329}}                    & 0.357                   & 1.867                                 & 0.869                                 & 0.335                                 & 0.389                                 & {\color[HTML]{0000FF} {\ul 0.296}}     & 0.334                                 & 0.322                                  & 0.354                                 & 0.324                                 & {\color[HTML]{0000FF} {\ul 0.331}}    & 0.595                                  & 0.572                                 & 0.612                                  & 0.611                                  \\
ETTm2→ETTm1 & {\color[HTML]{0000FF} {\ul 0.480}}          & {\color[HTML]{FF0000} \textbf{0.445}}       & {\color[HTML]{FF0000} \textbf{0.427}}       & {\color[HTML]{0000FF} {\ul 0.448}}          & 0.554                                                 & 0.478                   & 1.933                                 & 0.984                                 & 0.649                                 & 0.537                                 & 0.568                                  & 0.492                                 & 0.769                                  & 0.567                                 & 0.559                                 & 0.491                                 & 0.611                                  & 0.593                                 & 0.577                                  & 0.601                                  \\ \bottomrule
\end{tabular}
}
\end{table}
\vspace{10pt}

\subsection{Increasing lookback lengths}

\begin{figure}[h]
   \centering
   \includegraphics[width=1.0\linewidth]{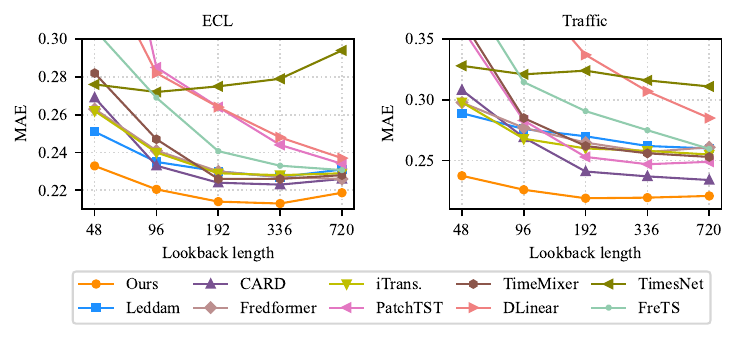}
   \caption{Performance change with increasing lookback lengths. The prediction length is $\tau=96$.}
   \label{fig_lookback}
\end{figure}

Performance under increasing lookback lengths reflects a model’s ability to effectively utilize historical information \citep{itransformer}. Linear-based forecasters have strong potential to benefit from longer lookback windows \citep{linear}. As shown in Figure~\ref{fig_lookback}, OLinear demonstrates consistent improvements as the lookback horizon increases from 48 to 720, and consistently outperforms state-of-the-art forecasters.

\vspace{10pt}

\subsection{Training with less data}

\begin{figure}[h]
   \centering
   \includegraphics[width=1.0\linewidth]{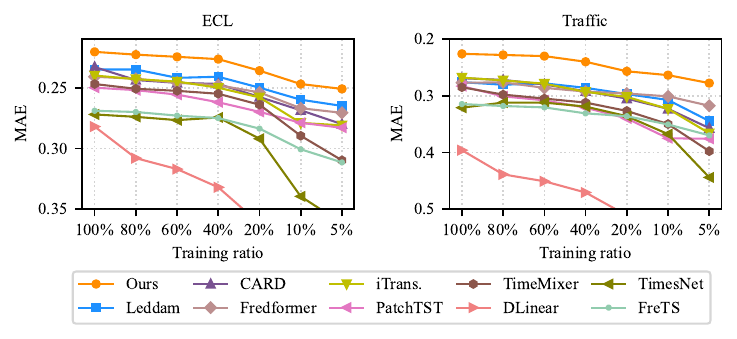}
   \caption{Performance change with less training data. The lookback and prediction lengths are set to 96. The Y-axis is inverted for clarity.}
   \label{fig_part}
\end{figure}

Training with less data reflects a model's adaptability and learning efficacy. As shown in Figure~\ref{fig_part}, OLinear adapts well to decreasing training set sizes, from 100\% down to 5\%. Our model consistently outperforms others across all training ratios, highlighting its robustness to data-sparse conditions.

\vspace{10pt}

\subsection{OLinear versus Timer}

We further compare OLinear with the large pre-trained time series model Timer \citep{timer} in terms of few-shot and zero-shot capabilities. As shown in Table~\ref{tab:timer}, OLinear surprisingly achieves comparable performance to Timer \citep{timer}, despite not being pre-trained on large-scale datasets. When the full training set is used, OLinear outperforms the fine-tuned Timer on the ECL, Traffic, Weather, and PEMS datasets. Even when only 5\% of training data are available, OLinear still performs better on the ETTm1 and ETTm2 datasets. For zero-shot learning, our model—trained on ETTh1/ETTm1/ETTm2—outperforms Timer on ETTh2/ETTm2/ETTm1, respectively. This phenomenon not only demonstrates OLinear’s strong generalization ability, but also suggests that large time series models are still in the early stages of development.

\begin{table}[ht]
\caption{Performance comparison with Timer on fine-tuning and zero-shot scenarios. MSEs are reported, with a prediction length of 96. Better results are highlighted in {\color[HTML]{FF0000} \textbf{bold}}. For OLinear's zero-shot performance, we adopt the $\mathrm{ETTh1} \leftrightarrow \mathrm{ETTh2}$ and $\mathrm{ETTm1} \leftrightarrow \mathrm{ETTm2}$ scheme (e.g., training on ETTh1 and evaluating on ETTh2, or vice versa).}
\label{tab:timer}
\centering
\setlength{\tabcolsep}{4.2pt}
\renewcommand{\arraystretch}{1.0} 
{\fontsize{9}{10}\selectfont
\begin{tabular}{@{}ccccccccc@{}}
\toprule
Train.   Ratio & \multicolumn{2}{c}{100\%}                                                     & \multicolumn{2}{c}{20\%}                                                      & \multicolumn{2}{c}{5\%}                                                       & \multicolumn{2}{c}{0\% (Zero-shot)}                                           \\ \midrule
Model          & Ours                                  & Timer                                 & Ours                                  & Timer                                 & Ours                                  & Timer                                 & Ours                                  & Timer-28B                             \\ \midrule
pre-trained    & No                                    & UTSD-12G                              & No                                    & UTSD-12G                              & No                                    & UTSD-12G                              & ETT                                   & \multicolumn{1}{l}{UTSD+LOTSA}        \\ \midrule
ETTh1          & 0.363                                 & {\color[HTML]{FF0000} \textbf{0.358}} & 0.396                                 & {\color[HTML]{FF0000} \textbf{0.359}} & 0.413                                 & {\color[HTML]{FF0000} \textbf{0.362}} & 0.416                                 & {\color[HTML]{FF0000} \textbf{0.393}} \\
ETTh2          & 0.276                                 & --                                    & 0.289                                 & {\color[HTML]{FF0000} \textbf{0.284}} & 0.303                                 & {\color[HTML]{FF0000} \textbf{0.280}} & {\color[HTML]{FF0000} \textbf{0.287}} & 0.308                                 \\
ETTm1          & 0.275                                 & --                                    & {\color[HTML]{FF0000} \textbf{0.300}} & 0.321                                 & {\color[HTML]{FF0000} \textbf{0.321}} & {\color[HTML]{FF0000} \textbf{0.321}} & {\color[HTML]{FF0000} \textbf{0.355}} & 0.420                                 \\
ETTm2          & 0.161                                 & --                                    & {\color[HTML]{FF0000} \textbf{0.167}} & 0.187                                 & {\color[HTML]{FF0000} \textbf{0.170}} & 0.176                                 & {\color[HTML]{FF0000} \textbf{0.176}} & 0.247                                 \\
ECL            & {\color[HTML]{FF0000} \textbf{0.123}} & 0.136                                 & {\color[HTML]{FF0000} \textbf{0.131}} & 0.134                                 & 0.153                                 & {\color[HTML]{FF0000} \textbf{0.132}} & -                                     & 0.147                                 \\
Traffic        & {\color[HTML]{FF0000} \textbf{0.338}} & 0.351                                 & 0.371                                 & {\color[HTML]{FF0000} \textbf{0.352}} & 0.419                                 & {\color[HTML]{FF0000} \textbf{0.361}} & -                                     & 0.414                                 \\
Weather        & {\color[HTML]{FF0000} \textbf{0.144}} & 0.154                                 & 0.156                                 & {\color[HTML]{FF0000} \textbf{0.151}} & 0.168                                 & {\color[HTML]{FF0000} \textbf{0.151}} & -                                     & 0.243                                 \\
PEMS03         & {\color[HTML]{FF0000} \textbf{0.103}} & 0.118                                 & {\color[HTML]{FF0000} \textbf{0.116}} & {\color[HTML]{FF0000} \textbf{0.116}} & 0.226                                 & {\color[HTML]{FF0000} \textbf{0.125}} & -                                     & -                                     \\
PEMS04         & {\color[HTML]{FF0000} \textbf{0.086}} & 0.107                                 & {\color[HTML]{FF0000} \textbf{0.195}} & 0.120                                 & 0.542                                 & {\color[HTML]{FF0000} \textbf{0.135}} & -                                     & -                                     \\ \bottomrule
\end{tabular}
}
\end{table}

\subsection{Hyperparameter sensitivity} \label{hyperpara}

\begin{figure}[h]
   \centering
   \includegraphics[width=1.0\linewidth]{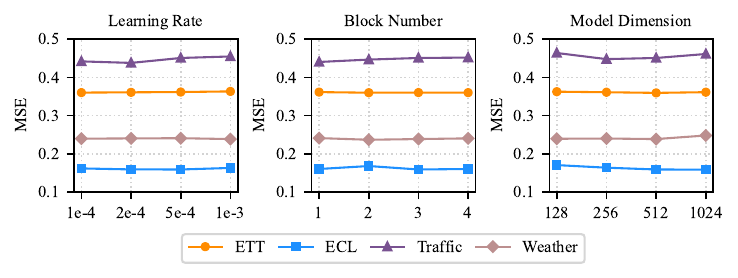}
   \caption{Hyperparameter sensitivity with respect to the learning rate $lr$, the number of blocks $L$, and the model dimension $D$. Average MSEs are reported across four prediction lengths $\tau \in \left \{ 96,192,336,720 \right \}$. ETT denotes the average performance over four subsets:  $\mathrm{ETT} \left \{ \mathrm{h} 1,\mathrm{h} 2,\mathrm{m} 1,\mathrm{m} 2 \right \} $. The lookback length $T$ is uniformly set as 96.}
   \label{fig_hyper}
\end{figure}

We evaluate the hyperparameter sensitivity of OLinear with respect to four key factors: the learning rate ($lr$), the number of blocks ($L$), the model dimension ($D$), and the embedding size ($d$). As shown in Figure~\ref{fig_hyper}, OLinear generally maintains stable performance across different choices of $lr$, $L$, and $D$, although each dataset exhibits its own preferences.  Regarding the embedding size $d$, Table~\ref{tab:embed_size} indicates that $d = 16$ represents a \textit{sweet spot}, which we adopt as the default setting in our experiments.

\begin{table}[ht]
\caption{Performance with various embedding size $d$ settings. In our main experiments, we set $d=16$ without specific statements.}
\label{tab:embed_size}
\centering
\setlength{\tabcolsep}{6pt}
\renewcommand{\arraystretch}{1.0} 
{\fontsize{8}{10}\selectfont
\begin{tabular}{@{}cccccccccccccc@{}}
\toprule
\multicolumn{2}{c}{Emb.   size}      & \multicolumn{2}{c}{1}                                                         & \multicolumn{2}{c}{4}                                                         & \multicolumn{2}{c}{8}                                                         & \multicolumn{2}{c}{16}                                                        & \multicolumn{2}{c}{32}                                                        & \multicolumn{2}{c}{64}                                                        \\ \midrule
\multicolumn{2}{c}{Metric}           & MSE                                   & MAE                                   & MSE                                   & MAE                                   & MSE                                   & MAE                                   & MSE                                   & MAE                                   & MSE                                   & MAE                                   & MSE                                   & MAE                                   \\ \midrule
\multirow{5}{*}{\rotatebox[origin=c]{90}{ECL}}     & 96  & 0.133                                 & {\color[HTML]{0000FF} {\ul 0.223}}    & {\color[HTML]{0000FF} {\ul 0.132}}    & {\color[HTML]{FF0000} \textbf{0.221}} & {\color[HTML]{0000FF} {\ul 0.132}}    & {\color[HTML]{FF0000} \textbf{0.221}} & {\color[HTML]{FF0000} \textbf{0.131}} & {\color[HTML]{FF0000} \textbf{0.221}} & {\color[HTML]{FF0000} \textbf{0.131}} & {\color[HTML]{FF0000} \textbf{0.221}} & {\color[HTML]{FF0000} \textbf{0.131}} & {\color[HTML]{FF0000} \textbf{0.221}} \\
                               & 192 & {\color[HTML]{0000FF} {\ul 0.152}}    & {\color[HTML]{0000FF} {\ul 0.240}}    & 0.153                                 & 0.241                                 & {\color[HTML]{0000FF} {\ul 0.152}}    & {\color[HTML]{0000FF} {\ul 0.240}}    & {\color[HTML]{FF0000} \textbf{0.150}} & {\color[HTML]{FF0000} \textbf{0.238}} & 0.153                                 & 0.241                                 & {\color[HTML]{0000FF} {\ul 0.152}}    & {\color[HTML]{0000FF} {\ul 0.240}}    \\
                               & 336 & {\color[HTML]{0000FF} {\ul 0.166}}    & 0.256                                 & 0.169                                 & 0.257                                 & {\color[HTML]{FF0000} \textbf{0.165}} & {\color[HTML]{0000FF} {\ul 0.255}}    & {\color[HTML]{FF0000} \textbf{0.165}} & {\color[HTML]{FF0000} \textbf{0.254}} & {\color[HTML]{FF0000} \textbf{0.165}} & {\color[HTML]{0000FF} {\ul 0.255}}    & 0.168                                 & 0.258                                 \\
                               & 720 & {\color[HTML]{FF0000} \textbf{0.188}} & {\color[HTML]{FF0000} \textbf{0.277}} & 0.198                                 & 0.282                                 & 0.207                                 & 0.290                                 & {\color[HTML]{0000FF} {\ul 0.191}}    & {\color[HTML]{0000FF} {\ul 0.279}}    & 0.193                                 & 0.280                                 & 0.200                                 & 0.286                                 \\ \cmidrule(l){2-14} 
         & Avg & {\color[HTML]{FF0000} \textbf{0.159}} & {\color[HTML]{0000FF} {\ul 0.249}}    & 0.163                                 & 0.250                                 & 0.164                                 & 0.252                                 & {\color[HTML]{FF0000} \textbf{0.159}} & {\color[HTML]{FF0000} \textbf{0.248}} & 0.161                                 & {\color[HTML]{0000FF} {\ul 0.249}}    & 0.163                                 & 0.251                                 \\ \midrule
\multirow{5}{*}{\rotatebox[origin=c]{90}{Traffic}}    & 96  & {\color[HTML]{FF0000} \textbf{0.398}} & 0.230                                 & 0.411                                 & {\color[HTML]{0000FF} {\ul 0.227}}    & 0.403                                 & {\color[HTML]{FF0000} \textbf{0.226}} & {\color[HTML]{FF0000} \textbf{0.398}} & {\color[HTML]{FF0000} \textbf{0.226}} & {\color[HTML]{0000FF} {\ul 0.402}}    & {\color[HTML]{FF0000} \textbf{0.226}} & 0.404                                 & {\color[HTML]{FF0000} \textbf{0.226}} \\
                               & 192 & {\color[HTML]{FF0000} \textbf{0.425}} & 0.243                                 & {\color[HTML]{0000FF} {\ul 0.433}}    & {\color[HTML]{FF0000} \textbf{0.241}} & 0.443                                 & {\color[HTML]{FF0000} \textbf{0.241}} & 0.439                                 & {\color[HTML]{FF0000} \textbf{0.241}} & 0.436                                 & {\color[HTML]{FF0000} \textbf{0.241}} & 0.435                                 & {\color[HTML]{0000FF} {\ul 0.242}}    \\
                               & 336 & {\color[HTML]{FF0000} \textbf{0.450}} & 0.251                                 & 0.462                                 & {\color[HTML]{0000FF} {\ul 0.250}}    & 0.465                                 & {\color[HTML]{FF0000} \textbf{0.249}} & 0.464                                 & {\color[HTML]{0000FF} {\ul 0.250}}    & 0.460                                 & 0.251                                 & {\color[HTML]{0000FF} {\ul 0.459}}    & 0.254                                 \\
                               & 720 & {\color[HTML]{FF0000} \textbf{0.496}} & {\color[HTML]{FF0000} \textbf{0.269}} & 0.518                                 & {\color[HTML]{FF0000} \textbf{0.269}} & 0.520                                 & {\color[HTML]{0000FF} {\ul 0.270}}    & {\color[HTML]{0000FF} {\ul 0.502}}    & {\color[HTML]{0000FF} {\ul 0.270}}    & 0.507                                 & 0.274                                 & 0.515                                 & 0.275                                 \\ \cmidrule(l){2-14} 
     & Avg & {\color[HTML]{FF0000} \textbf{0.442}} & 0.248                                 & 0.456                                 & {\color[HTML]{0000FF} {\ul 0.247}}    & 0.458                                 & {\color[HTML]{FF0000} \textbf{0.246}} & {\color[HTML]{0000FF} {\ul 0.451}}    & {\color[HTML]{0000FF} {\ul 0.247}}    & {\color[HTML]{0000FF} {\ul 0.451}}    & 0.248                                 & 0.453                                 & 0.249                                 \\ \midrule
\multirow{5}{*}{\rotatebox[origin=c]{90}{Solar-Energy}}    & 96  & 0.180                                 & 0.194                                 & {\color[HTML]{0000FF} {\ul 0.179}}    & 0.193                                 & 0.180                                 & {\color[HTML]{0000FF} {\ul 0.191}}    & {\color[HTML]{0000FF} {\ul 0.179}}    & {\color[HTML]{0000FF} {\ul 0.191}}    & {\color[HTML]{FF0000} \textbf{0.177}} & {\color[HTML]{FF0000} \textbf{0.190}} & 0.182                                 & 0.192                                 \\
                               & 192 & 0.212                                 & 0.216                                 & {\color[HTML]{0000FF} {\ul 0.211}}    & {\color[HTML]{0000FF} {\ul 0.214}}    & {\color[HTML]{FF0000} \textbf{0.209}} & {\color[HTML]{FF0000} \textbf{0.213}} & {\color[HTML]{FF0000} \textbf{0.209}} & {\color[HTML]{FF0000} \textbf{0.213}} & 0.212                                 & {\color[HTML]{FF0000} \textbf{0.213}} & {\color[HTML]{0000FF} {\ul 0.211}}    & {\color[HTML]{0000FF} {\ul 0.214}}    \\
                               & 336 & 0.234                                 & 0.233                                 & 0.232                                 & {\color[HTML]{0000FF} {\ul 0.230}}    & {\color[HTML]{FF0000} \textbf{0.230}} & {\color[HTML]{0000FF} {\ul 0.230}}    & {\color[HTML]{0000FF} {\ul 0.231}}    & {\color[HTML]{FF0000} \textbf{0.229}} & {\color[HTML]{0000FF} {\ul 0.231}}    & {\color[HTML]{0000FF} {\ul 0.230}}    & 0.233                                 & 0.231                                 \\
                               & 720 & 0.244                                 & 0.238                                 & {\color[HTML]{FF0000} \textbf{0.241}} & {\color[HTML]{0000FF} {\ul 0.236}}    & {\color[HTML]{FF0000} \textbf{0.241}} & {\color[HTML]{FF0000} \textbf{0.235}} & {\color[HTML]{FF0000} \textbf{0.241}} & {\color[HTML]{0000FF} {\ul 0.236}}    & {\color[HTML]{0000FF} {\ul 0.242}}    & {\color[HTML]{0000FF} {\ul 0.236}}    & 0.243                                 & 0.238                                 \\ \cmidrule(l){2-14} 
& Avg & 0.217                                 & 0.220                                 & {\color[HTML]{0000FF} {\ul 0.216}}    & {\color[HTML]{0000FF} {\ul 0.218}}    & {\color[HTML]{FF0000} \textbf{0.215}} & {\color[HTML]{FF0000} \textbf{0.217}} & {\color[HTML]{FF0000} \textbf{0.215}} & {\color[HTML]{FF0000} \textbf{0.217}} & {\color[HTML]{FF0000} \textbf{0.215}} & {\color[HTML]{FF0000} \textbf{0.217}} & 0.217                                 & 0.219                                 \\ \midrule
\multirow{5}{*}{\rotatebox[origin=c]{90}{Weather}}      & 96  & {\color[HTML]{0000FF} {\ul 0.152}}    & {\color[HTML]{0000FF} {\ul 0.189}}    & 0.158                                 & 0.192                                 & {\color[HTML]{FF0000} \textbf{0.150}} & {\color[HTML]{FF0000} \textbf{0.187}} & {\color[HTML]{0000FF} {\ul 0.152}}    & {\color[HTML]{0000FF} {\ul 0.189}}    & {\color[HTML]{0000FF} {\ul 0.152}}    & 0.190                                 & 0.157                                 & 0.194                                 \\
                               & 192 & 0.203                                 & {\color[HTML]{0000FF} {\ul 0.238}}    & {\color[HTML]{FF0000} \textbf{0.200}} & {\color[HTML]{FF0000} \textbf{0.234}} & 0.207                                 & 0.242                                 & 0.207                                 & 0.243                                 & {\color[HTML]{0000FF} {\ul 0.202}}    & {\color[HTML]{0000FF} {\ul 0.238}}    & 0.205                                 & 0.239                                 \\
                               & 336 & {\color[HTML]{0000FF} {\ul 0.261}}    & {\color[HTML]{0000FF} {\ul 0.282}}    & {\color[HTML]{FF0000} \textbf{0.259}} & {\color[HTML]{FF0000} \textbf{0.280}} & {\color[HTML]{FF0000} \textbf{0.259}} & {\color[HTML]{FF0000} \textbf{0.280}} & {\color[HTML]{0000FF} {\ul 0.261}}    & {\color[HTML]{0000FF} {\ul 0.282}}    & 0.263                                 & {\color[HTML]{0000FF} {\ul 0.282}}    & 0.269                                 & 0.287                                 \\
                               & 720 & {\color[HTML]{FF0000} \textbf{0.342}} & {\color[HTML]{0000FF} {\ul 0.337}}    & {\color[HTML]{0000FF} {\ul 0.343}}    & 0.338                                 & {\color[HTML]{0000FF} {\ul 0.343}}    & {\color[HTML]{0000FF} {\ul 0.337}}    & 0.344                                 & {\color[HTML]{0000FF} {\ul 0.337}}    & 0.350                                 & 0.339                                 & 0.344                                 & {\color[HTML]{FF0000} \textbf{0.336}} \\ \cmidrule(l){2-14} 
      & Avg & {\color[HTML]{FF0000} \textbf{0.240}} & {\color[HTML]{FF0000} \textbf{0.261}} & {\color[HTML]{FF0000} \textbf{0.240}} & {\color[HTML]{FF0000} \textbf{0.261}} & {\color[HTML]{FF0000} \textbf{0.240}} & {\color[HTML]{FF0000} \textbf{0.261}} & {\color[HTML]{0000FF} {\ul 0.241}}    & {\color[HTML]{0000FF} {\ul 0.262}}    & 0.242                                 & {\color[HTML]{0000FF} {\ul 0.262}}    & 0.244                                 & 0.264                                 \\ \midrule
\multirow{5}{*}{\rotatebox[origin=c]{90}{ILI}}     & 3   & 0.487                                 & 0.355                                 & 0.545                                 & 0.380                                 & 0.500                                 & 0.353                                 & {\color[HTML]{FF0000} \textbf{0.468}} & {\color[HTML]{FF0000} \textbf{0.349}} & 0.515                                 & 0.362                                 & {\color[HTML]{0000FF} {\ul 0.486}}    & {\color[HTML]{0000FF} {\ul 0.352}}    \\
                               & 6   & 0.915                                 & {\color[HTML]{0000FF} {\ul 0.515}}    & 1.021                                 & 0.533                                 & {\color[HTML]{FF0000} \textbf{0.890}} & {\color[HTML]{FF0000} \textbf{0.506}} & 0.923                                 & 0.516                                 & {\color[HTML]{0000FF} {\ul 0.914}}    & {\color[HTML]{0000FF} {\ul 0.515}}    & 1.059                                 & 0.558                                 \\
                               & 9   & 1.301                                 & 0.662                                 & 1.363                                 & 0.666                                 & 1.376                                 & 0.667                                 & 1.289                                 & 0.655                                 & {\color[HTML]{0000FF} {\ul 1.265}}    & {\color[HTML]{FF0000} \textbf{0.644}} & {\color[HTML]{FF0000} \textbf{1.264}} & {\color[HTML]{0000FF} {\ul 0.645}}    \\
                               & 12  & {\color[HTML]{0000FF} {\ul 1.638}}    & 0.783                                 & 1.693                                 & 0.788                                 & 1.810                                 & 0.799                                 & 1.698                                 & 0.791                                 & {\color[HTML]{FF0000} \textbf{1.619}} & {\color[HTML]{FF0000} \textbf{0.769}} & 1.747                                 & {\color[HTML]{0000FF} {\ul 0.773}}    \\ \cmidrule(l){2-14} 
         & Avg & {\color[HTML]{0000FF} {\ul 1.085}}    & 0.579                                 & 1.155                                 & 0.592                                 & 1.144                                 & 0.581                                 & 1.094                                 & {\color[HTML]{0000FF} {\ul 0.578}}    & {\color[HTML]{FF0000} \textbf{1.078}} & {\color[HTML]{FF0000} \textbf{0.572}} & 1.139                                 & 0.582                                 \\ \midrule
\multirow{5}{*}{\rotatebox[origin=c]{90}{PEMS03}}       & 12  & {\color[HTML]{0000FF} {\ul 0.060}}    & {\color[HTML]{0000FF} {\ul 0.159}}    & {\color[HTML]{0000FF} {\ul 0.060}}    & {\color[HTML]{0000FF} {\ul 0.159}}    & {\color[HTML]{0000FF} {\ul 0.060}}    & {\color[HTML]{0000FF} {\ul 0.159}}    & {\color[HTML]{0000FF} {\ul 0.060}}    & {\color[HTML]{0000FF} {\ul 0.159}}    & 0.061                                 & 0.160                                 & {\color[HTML]{FF0000} \textbf{0.059}} & {\color[HTML]{FF0000} \textbf{0.158}} \\
                               & 24  & {\color[HTML]{0000FF} {\ul 0.077}}    & {\color[HTML]{FF0000} \textbf{0.179}} & {\color[HTML]{0000FF} {\ul 0.078}}    & {\color[HTML]{FF0000} \textbf{0.179}} & {\color[HTML]{0000FF} {\ul 0.078}}    & {\color[HTML]{0000FF} {\ul 0.180}}    & {\color[HTML]{0000FF} {\ul 0.078}}    & {\color[HTML]{FF0000} \textbf{0.179}} & {\color[HTML]{FF0000} \textbf{0.077}} & {\color[HTML]{FF0000} \textbf{0.179}} & {\color[HTML]{FF0000} \textbf{0.077}} & {\color[HTML]{FF0000} \textbf{0.179}} \\
                               & 48  & 0.107                                 & 0.213                                 & 0.106                                 & {\color[HTML]{0000FF} {\ul 0.211}}    & 0.106                                 & {\color[HTML]{0000FF} {\ul 0.211}}    & {\color[HTML]{FF0000} \textbf{0.104}} & {\color[HTML]{FF0000} \textbf{0.210}} & 0.106                                 & {\color[HTML]{0000FF} {\ul 0.211}}    & {\color[HTML]{0000FF} {\ul 0.105}}    & {\color[HTML]{FF0000} \textbf{0.210}} \\
                               & 96  & 0.146                                 & 0.253                                 & 0.143                                 & {\color[HTML]{0000FF} {\ul 0.248}}    & {\color[HTML]{0000FF} {\ul 0.141}}    & {\color[HTML]{0000FF} {\ul 0.248}}    & {\color[HTML]{FF0000} \textbf{0.140}} & {\color[HTML]{FF0000} \textbf{0.247}} & {\color[HTML]{FF0000} \textbf{0.140}} & {\color[HTML]{FF0000} \textbf{0.247}} & 0.143                                 & {\color[HTML]{0000FF} {\ul 0.248}}    \\ \cmidrule(l){2-14} 
       & Avg & 0.097                                 & 0.201                                 & 0.097                                 & {\color[HTML]{FF0000} \textbf{0.199}} & {\color[HTML]{0000FF} {\ul 0.096}}    & {\color[HTML]{0000FF} {\ul 0.200}}    & {\color[HTML]{FF0000} \textbf{0.095}} & {\color[HTML]{FF0000} \textbf{0.199}} & {\color[HTML]{0000FF} {\ul 0.096}}    & {\color[HTML]{FF0000} \textbf{0.199}} & {\color[HTML]{0000FF} {\ul 0.096}}    & {\color[HTML]{FF0000} \textbf{0.199}} \\ \midrule
\multicolumn{2}{c}{1st Count}        & 9                                     & 4                                     & 4                                     & 8                                     & 9                                     & 12                                    & 12                                    & 14                                    & 8                                     & 12                                    & 4                                     & 7                                     \\ \bottomrule
\end{tabular}
}
\end{table}

\subsection{More metrics}

We further conduct a comprehensive evaluation of forecasting performance using three scale-free metrics: the Coefficient of Determination ($R^2$), the Pearson Correlation Coefficient ($r$), and the Mean Absolute Scaled Error (MASE), which are defined as follows:

\begin{equation}
\begin{aligned}
  R^2&=\frac{1}{N}\sum_{n=1}^{N}\left ( 1-\frac{\left | \hat{\mathbf{Y}} _{n:}-\mathbf{Y}_{n:} \right | ^2_2 }{\left | \mathbf{Y} _{n:}-\overline{\mathbf{Y}_{n:}}  \right | ^2_2}  \right )  , \\
  r &=  \frac{1}{N}\sum_{n=1}^{N}\frac{ {\textstyle \sum_{t=1}^{\tau}}  \left ( \hat{\mathbf{Y}} _{n,t}-\overline{\hat{\mathbf{Y}} _{n:}}  \right )\left ( \mathbf{Y} _{n,t}-\overline{\mathbf{Y} _{n:}}  \right )  }
{  \sqrt{ {\textstyle \sum_{t=1}^{\tau}} \left ( \hat{\mathbf{Y}} _{n,t}-\overline{\hat{\mathbf{Y}} _{n:}}  \right )^2} \sqrt{ {\textstyle \sum_{t=1}^{\tau}} \left ( \mathbf{Y} _{n,t}-\overline{\mathbf{Y} _{n:}}  \right )^2}     } ,   \\
  \mathrm{MASE} &= \frac{1}{N}\sum_{n=1}^{N}\frac{\frac{1}{\tau} \left \| \hat{\mathbf{Y}} _{n:}-\mathbf{Y}_{n:} \right \|_1  }{\frac{1}{\tau-1}  {\textstyle \sum_{t=1}^{\tau-1} \left | \mathbf{Y} _{n,t}-\mathbf{Y}_{t-1,n} \right | } }.
\end{aligned}
\label{more_metrics}
\end{equation} 

Here $\mathbf{Y} _{n:}$ and $\hat{\mathbf{Y}} _{n:}$ denote the $n$-th variate of the ground truth $\mathbf{Y}$ and the prediction $\hat{\mathbf{Y}}$, respectively. $\mathbf{Y} _{n,t}$ refers to the $t$-th timestep of the $n$-th variate, and $\overline{(\cdot)}$ represents the mean.

As shown in Table~\ref{tab:more_metrics}, OLinear consistently achieves state-of-the-art performance on these metrics. For example, it an average Pearson correlation coefficient of $r = 0.903$ on the ECL dataset,  indicating strong alignment between predictions and ground truth. On the Traffic dataset, OLinear achieves a MASE of 0.764, outperforming the second-best model, CARD \citep{card}, by a notable margin of 11.4\%.

\begin{table}[ht]
 \caption{Forecasting performance on the scale-free metrics: Coefficient of Determination ($R^2$), Pearson Correlation Coefficient ($r$), and MASE. The symbols $\downarrow$ and $\uparrow$ indicate that lower and higher values are better, respectively. The best and second-best results are highlighted in {\color[HTML]{FF0000} \textbf{bold}} and {\color[HTML]{0000FF} {\ul underlined}}, respectively. The lookback length $T$ is uniformly set as 96.}
\label{tab:more_metrics}
\centering
\setlength{\tabcolsep}{1.5pt}
\renewcommand{\arraystretch}{1.0} 
{\fontsize{7}{9}\selectfont
\begin{tabular}{@{}ccccccccccccccccc@{}}
\toprule
\multicolumn{2}{c}{Model}       & \multicolumn{3}{c}{\begin{tabular}[c]{@{}c@{}}OLinear\\      (Ours)\end{tabular}}                                 & 
\multicolumn{3}{c}{\begin{tabular}[c]{@{}c@{}}Leddam\\  \citeyear{Leddam_icml} 
 \end{tabular}}                                 & 
\multicolumn{3}{c}{\begin{tabular}[c]{@{}c@{}}CARD\\      \citeyear{card}  \end{tabular}}      & 
\multicolumn{3}{c}{\begin{tabular}[c]{@{}c@{}}iTrans.\\      \citeyear{itransformer}  \end{tabular}}   & 
\multicolumn{3}{c}{\begin{tabular}[c]{@{}c@{}}TimeMixer\\      \citeyear{timemixer}  \end{tabular}} \\ \midrule
\multicolumn{2}{c}{Metric}      & $R^2$ ($\uparrow$)                                   & $r$ ($\uparrow$)                                   & MASE ($\downarrow$)                                 & $R^2$ ($\uparrow$)                                & $r$ ($\uparrow$)                                & MASE ($\downarrow$)                              & $R^2$ ($\uparrow$)                                & $r$ ($\uparrow$)   & MASE ($\downarrow$)                              & $R^2$ ($\uparrow$)                                & $r$ ($\uparrow$)                                & MASE ($\downarrow$) & $R^2$ ($\uparrow$)                       & $r$ ($\uparrow$)                      & MASE ($\downarrow$)                    \\ \midrule
\multirow{5}{*}{\rotatebox[origin=c]{90}{ECL}}  & 96  & {\color[HTML]{FF0000} \textbf{0.640}} & {\color[HTML]{FF0000} \textbf{0.916}} & {\color[HTML]{FF0000} \textbf{0.902}} & 0.527                              & {\color[HTML]{0000FF} {\ul 0.911}} & {\color[HTML]{0000FF} {\ul 0.964}} & 0.533                              & 0.905 & 0.975                              & {\color[HTML]{0000FF} {\ul 0.550}} & 0.907                              & 0.990 & 0.456                     & 0.897                    & 1.022                    \\
                          & 192 & {\color[HTML]{FF0000} \textbf{0.638}} & {\color[HTML]{FF0000} \textbf{0.908}} & {\color[HTML]{FF0000} \textbf{0.983}} & {\color[HTML]{0000FF} {\ul 0.593}} & {\color[HTML]{0000FF} {\ul 0.902}} & 1.041                              & 0.547                              & 0.899 & {\color[HTML]{0000FF} {\ul 1.028}} & 0.590                              & 0.900                              & 1.057 & 0.575                     & 0.893                    & 1.076                    \\
                          & 336 & {\color[HTML]{FF0000} \textbf{0.710}} & {\color[HTML]{FF0000} \textbf{0.901}} & {\color[HTML]{FF0000} \textbf{1.061}} & 0.670                              & {\color[HTML]{0000FF} {\ul 0.893}} & 1.134                              & {\color[HTML]{0000FF} {\ul 0.678}} & 0.891 & {\color[HTML]{0000FF} {\ul 1.113}} & {\color[HTML]{0000FF} {\ul 0.678}} & {\color[HTML]{0000FF} {\ul 0.893}} & 1.133 & 0.640                     & 0.886                    & 1.155                    \\
                          & 720 & {\color[HTML]{FF0000} \textbf{0.701}} & {\color[HTML]{FF0000} \textbf{0.889}} & {\color[HTML]{FF0000} \textbf{1.174}} & {\color[HTML]{0000FF} {\ul 0.661}} & {\color[HTML]{0000FF} {\ul 0.882}} & {\color[HTML]{0000FF} {\ul 1.251}} & 0.653                              & 0.877 & 1.273                              & 0.638                              & 0.872                              & 1.315 & 0.634                     & 0.871                    & 1.314                    \\ \cmidrule(l){2-17} 
   & Avg & {\color[HTML]{FF0000} \textbf{0.672}} & {\color[HTML]{FF0000} \textbf{0.903}} & {\color[HTML]{FF0000} \textbf{1.030}} & 0.613                              & {\color[HTML]{0000FF} {\ul 0.897}} & 1.098                              & 0.603                              & 0.893 & {\color[HTML]{0000FF} {\ul 1.097}} & {\color[HTML]{0000FF} {\ul 0.614}} & 0.893                              & 1.124 & 0.576                     & 0.887                    & 1.142                    \\ \midrule
\multirow{5}{*}{\rotatebox[origin=c]{90}{Traffic}}   & 96  & {\color[HTML]{FF0000} \textbf{0.737}} & {\color[HTML]{FF0000} \textbf{0.899}} & {\color[HTML]{FF0000} \textbf{0.729}} & 0.681                              & 0.885                              & 0.909                              & {\color[HTML]{0000FF} {\ul 0.690}} & 0.882 & {\color[HTML]{0000FF} {\ul 0.855}} & {\color[HTML]{0000FF} {\ul 0.690}} & {\color[HTML]{0000FF} {\ul 0.889}} & 0.860 & 0.640                     & 0.888                    & 0.979                    \\
                          & 192 & {\color[HTML]{FF0000} \textbf{0.719}} & {\color[HTML]{FF0000} \textbf{0.884}} & {\color[HTML]{FF0000} \textbf{0.756}} & 0.666                              & 0.865                              & 0.949                              & 0.683                              & 0.869 & {\color[HTML]{0000FF} {\ul 0.856}} & {\color[HTML]{0000FF} {\ul 0.689}} & {\color[HTML]{0000FF} {\ul 0.877}} & 0.879 & 0.660                     & 0.862                    & 0.959                    \\
                          & 336 & {\color[HTML]{FF0000} \textbf{0.730}} & {\color[HTML]{FF0000} \textbf{0.875}} & {\color[HTML]{FF0000} \textbf{0.758}} & 0.707                              & {\color[HTML]{0000FF} {\ul 0.867}} & 0.897                              & 0.702                              & 0.862 & {\color[HTML]{0000FF} {\ul 0.843}} & {\color[HTML]{0000FF} {\ul 0.713}} & {\color[HTML]{0000FF} {\ul 0.867}} & 0.883 & 0.685                     & 0.852                    & 0.960                    \\
                          & 720 & {\color[HTML]{FF0000} \textbf{0.703}} & {\color[HTML]{FF0000} \textbf{0.860}} & {\color[HTML]{FF0000} \textbf{0.815}} & 0.687                              & 0.853                              & 0.947                              & 0.679                              & 0.848 & {\color[HTML]{0000FF} {\ul 0.894}} & {\color[HTML]{0000FF} {\ul 0.693}} & {\color[HTML]{0000FF} {\ul 0.857}} & 0.913 & 0.672                     & 0.842                    & 0.986                    \\ \cmidrule(l){2-17} 
 & Avg & {\color[HTML]{FF0000} \textbf{0.722}} & {\color[HTML]{FF0000} \textbf{0.879}} & {\color[HTML]{FF0000} \textbf{0.764}} & 0.685                              & 0.868                              & 0.926                              & 0.689                              & 0.865 & {\color[HTML]{0000FF} {\ul 0.862}} & {\color[HTML]{0000FF} {\ul 0.696}} & {\color[HTML]{0000FF} {\ul 0.873}} & 0.884 & 0.664                     & 0.861                    & 0.971                    \\ \midrule
\multicolumn{2}{c}{Model}       & 
\multicolumn{3}{c}{\begin{tabular}[c]{@{}c@{}}FilterNet\\      \citeyear{filternet}  \end{tabular}}                                       & 
\multicolumn{3}{c}{\begin{tabular}[c]{@{}c@{}}FITS\\      \citeyear{fits}  \end{tabular}}                                   & \multicolumn{3}{c}{\begin{tabular}[c]{@{}c@{}}DLinear\\      \citeyear{linear}  \end{tabular}}   & 
\multicolumn{3}{c}{\begin{tabular}[c]{@{}c@{}}PatchTST\\      \citeyear{patchtst}  \end{tabular}}  & 
\multicolumn{3}{c}{\begin{tabular}[c]{@{}c@{}}TimesNet\\      \citeyear{timesnet}  \end{tabular}}  \\ \midrule
\multicolumn{2}{c}{Metric}      & $R^2$ ($\uparrow$)                                   & $r$ ($\uparrow$)                                   & MASE ($\downarrow$)                                 & $R^2$ ($\uparrow$)                                & $r$ ($\uparrow$)                                & MASE ($\downarrow$)                              & $R^2$ ($\uparrow$)                                & $r$ ($\uparrow$)   & MASE ($\downarrow$)                              & $R^2$ ($\uparrow$)                                & $r$ ($\uparrow$)                                & MASE ($\downarrow$) & $R^2$ ($\uparrow$)                       & $r$ ($\uparrow$)                      & MASE ($\downarrow$)                    \\ \midrule
\multirow{5}{*}{\rotatebox[origin=c]{90}{ECL}}    & 96  & 0.537                                 & 0.905                                 & 0.996                                 & 0.352                              & 0.883                              & 1.144                              & 0.120                              & 0.870 & 1.300                              & 0.426                              & 0.890                              & 1.119 & 0.405                     & 0.892                    & 1.118                    \\
                          & 192 & 0.576                                 & 0.898                                 & 1.052                                 & 0.446                              & 0.880                              & 1.153                              & 0.215                              & 0.867 & 1.312                              & 0.509                              & 0.887                              & 1.140 & 0.503                     & 0.885                    & 1.195                    \\
                          & 336 & 0.670                                 & 0.891                                 & 1.139                                 & 0.588                              & 0.873                              & 1.229                              & 0.247                              & 0.858 & 1.387                              & 0.604                              & 0.880                              & 1.216 & 0.592                     & 0.875                    & 1.262                    \\
                          & 720 & 0.628                                 & 0.875                                 & 1.328                                 & 0.593                              & 0.858                              & 1.376                              & 0.148                              & 0.844 & 1.536                              & 0.605                              & 0.864                              & 1.371 & 0.603                     & 0.864                    & 1.388                    \\ \cmidrule(l){2-17} 
     & Avg & 0.603                                 & 0.892                                 & 1.129                                 & 0.495                              & 0.873                              & 1.225                              & 0.182                              & 0.860 & 1.384                              & 0.536                              & 0.880                              & 1.211 & 0.526                     & 0.879                    & 1.241                    \\ \midrule
\multirow{5}{*}{\rotatebox[origin=c]{90}{Traffic}}   & 96  & 0.637                                 & 0.876                                 & 0.987                                 & 0.458                              & 0.811                              & 1.284                              & 0.441                              & 0.811 & 1.346                              & 0.668                              & 0.880                              & 0.903 & 0.625                     & 0.877                    & 1.023                    \\
                          & 192 & 0.665                                 & 0.867                                 & 0.950                                 & 0.542                              & 0.814                              & 1.151                              & 0.551                              & 0.815 & 1.189                              & 0.677                              & 0.871                              & 0.887 & 0.635                     & 0.867                    & 1.009                    \\
                          & 336 & 0.696                                 & 0.861                                 & 0.930                                 & 0.582                              & 0.810                              & 1.119                              & 0.582                              & 0.810 & 1.167                              & 0.700                              & 0.863                              & 0.874 & 0.654                     & 0.852                    & 1.079                    \\
                          & 720 & 0.678                                 & 0.847                                 & 0.982                                 & 0.568                              & 0.794                              & 1.169                              & 0.564                              & 0.794 & 1.224                              & 0.681                              & 0.850                              & 0.925 & 0.661                     & 0.850                    & 1.047                    \\ \cmidrule(l){2-17} 
 & Avg & 0.669                                 & 0.862                                 & 0.962                                 & 0.537                              & 0.807                              & 1.181                              & 0.535                              & 0.808 & 1.232                              & 0.682                              & 0.866                              & 0.897 & 0.644                     & 0.862                    & 1.040                    \\ \bottomrule
\end{tabular}
}
\end{table}

\subsection{Less training data to compute Q matrices} \label{Q_mat_robust}

To evaluate the amount of training data required for computing the orthogonal transformation matrices $\mathbf{Q}_i$ and $\mathbf{Q}_o$, we conduct experiments using only partial data of the training set. As shown in Table~\ref{tab:part_Q}, OLinear exhibits strong robustness, with minimal performance variation  across different data proportions. This property not only enables more efficient preprocessing before training, but also ensures robust performance in few-shot learning scenarios (see Appendix~\ref{few-shot}).

\begin{table}[ht]
\caption{Forecasting performance with orthogonal matrices computed from varying proportions of the training data, denoted as `Q\_ratio'. All experiments are trained on the full training set.}
\label{tab:part_Q}
\centering
\setlength{\tabcolsep}{5pt}
\renewcommand{\arraystretch}{1.0} 
{\fontsize{8}{9}\selectfont
\begin{tabular}{@{}cccccccccccccc@{}}
\toprule
\multicolumn{2}{c}{Q\_ratio}     & \multicolumn{2}{c}{100\%}                                                     & \multicolumn{2}{c}{80\%}                                                      & \multicolumn{2}{c}{60\%}                                                      & \multicolumn{2}{c}{40\%}                                                      & \multicolumn{2}{c}{20\%}                                                      & \multicolumn{2}{c}{10\%}                                                      \\ \midrule
\multicolumn{2}{c}{Metric}       & MSE                                   & MAE                                   & MSE                                   & MAE                                   & MSE                                   & MAE                                   & MSE                                   & MAE                                   & MSE                                   & MAE                                   & MSE                                   & MAE                                   \\ \midrule
                           & 96  & 0.131                                 & 0.221                                 & 0.131                                 & 0.221                                 & 0.131                                 & 0.221                                 & 0.131                                 & 0.221                                 & 0.130                                 & 0.221                                 & 0.131                                 & 0.221                                 \\
                           & 192 & 0.150                                 & 0.238                                 & 0.155                                 & 0.243                                 & 0.152                                 & 0.239                                 & 0.153                                 & 0.240                                 & 0.151                                 & 0.240                                 & 0.152                                 & 0.240                                 \\
                           & 336 & 0.165                                 & 0.254                                 & 0.169                                 & 0.257                                 & 0.164                                 & 0.253                                 & 0.166                                 & 0.255                                 & 0.165                                 & 0.254                                 & 0.166                                 & 0.255                                 \\
                           & 720 & 0.191                                 & 0.279                                 & 0.193                                 & 0.284                                 & 0.188                                 & 0.277                                 & 0.193                                 & 0.284                                 & 0.193                                 & 0.280                                 & 0.189                                 & 0.278                                 \\ \cmidrule(l){2-14} 
\multirow{-5}{*}{ECL}      & Avg & {\color[HTML]{FF0000} \textbf{0.159}} & {\color[HTML]{0000FF} {\ul 0.248}}    & 0.162                                 & 0.251                                 & {\color[HTML]{FF0000} \textbf{0.159}} & {\color[HTML]{FF0000} \textbf{0.247}} & 0.161                                 & 0.250                                 & {\color[HTML]{0000FF} {\ul 0.160}}    & {\color[HTML]{0000FF} {\ul 0.248}}    & {\color[HTML]{FF0000} \textbf{0.159}} & {\color[HTML]{0000FF} {\ul 0.248}}    \\ \midrule
                           & 96  & 0.398                                 & 0.226                                 & 0.403                                 & 0.226                                 & 0.403                                 & 0.227                                 & 0.401                                 & 0.226                                 & 0.404                                 & 0.226                                 & 0.402                                 & 0.226                                 \\
                           & 192 & 0.439                                 & 0.241                                 & 0.436                                 & 0.240                                 & 0.434                                 & 0.240                                 & 0.432                                 & 0.240                                 & 0.434                                 & 0.240                                 & 0.431                                 & 0.240                                 \\
                           & 336 & 0.464                                 & 0.250                                 & 0.461                                 & 0.250                                 & 0.462                                 & 0.250                                 & 0.463                                 & 0.250                                 & 0.459                                 & 0.250                                 & 0.465                                 & 0.250                                 \\
                           & 720 & 0.502                                 & 0.270                                 & 0.499                                 & 0.271                                 & 0.499                                 & 0.271                                 & 0.501                                 & 0.271                                 & 0.502                                 & 0.271                                 & 0.512                                 & 0.271                                 \\ \cmidrule(l){2-14} 
\multirow{-5}{*}{Traffic}  & Avg & 0.451                                 & {\color[HTML]{FF0000} \textbf{0.247}} & {\color[HTML]{0000FF} {\ul 0.450}}    & {\color[HTML]{FF0000} \textbf{0.247}} & {\color[HTML]{FF0000} \textbf{0.449}} & {\color[HTML]{FF0000} \textbf{0.247}} & {\color[HTML]{FF0000} \textbf{0.449}} & {\color[HTML]{FF0000} \textbf{0.247}} & {\color[HTML]{0000FF} {\ul 0.450}}    & {\color[HTML]{FF0000} \textbf{0.247}} & 0.453                                 & {\color[HTML]{FF0000} \textbf{0.247}} \\ \midrule
                           & 96  & 0.153                                 & 0.190                                 & 0.150                                 & 0.188                                 & 0.153                                 & 0.191                                 & 0.153                                 & 0.190                                 & 0.154                                 & 0.191                                 & 0.149                                 & 0.187                                 \\
                           & 192 & 0.200                                 & 0.235                                 & 0.202                                 & 0.237                                 & 0.204                                 & 0.239                                 & 0.203                                 & 0.238                                 & 0.202                                 & 0.237                                 & 0.203                                 & 0.239                                 \\
                           & 336 & 0.258                                 & 0.280                                 & 0.260                                 & 0.280                                 & 0.260                                 & 0.281                                 & 0.264                                 & 0.284                                 & 0.260                                 & 0.281                                 & 0.259                                 & 0.281                                 \\
                           & 720 & 0.337                                 & 0.333                                 & 0.339                                 & 0.333                                 & 0.338                                 & 0.332                                 & 0.337                                 & 0.332                                 & 0.335                                 & 0.330                                 & 0.339                                 & 0.332                                 \\ \cmidrule(l){2-14} 
\multirow{-5}{*}{Weather}  & Avg & {\color[HTML]{FF0000} \textbf{0.237}} & {\color[HTML]{0000FF} {\ul 0.260}}    & {\color[HTML]{0000FF} {\ul 0.238}}    & {\color[HTML]{FF0000} \textbf{0.259}} & 0.239                                 & 0.261                                 & 0.239                                 & 0.261                                 & {\color[HTML]{0000FF} {\ul 0.238}}    & {\color[HTML]{0000FF} {\ul 0.260}}    & {\color[HTML]{FF0000} \textbf{0.237}} & {\color[HTML]{0000FF} {\ul 0.260}}    \\ \midrule
                           & 12  & 0.060                                 & 0.159                                 & 0.060                                 & 0.159                                 & 0.060                                 & 0.159                                 & 0.060                                 & 0.159                                 & 0.061                                 & 0.160                                 & 0.060                                 & 0.159                                 \\
                           & 24  & 0.078                                 & 0.179                                 & 0.077                                 & 0.179                                 & 0.077                                 & 0.180                                 & 0.077                                 & 0.179                                 & 0.077                                 & 0.180                                 & 0.077                                 & 0.180                                 \\
                           & 48  & 0.104                                 & 0.210                                 & 0.105                                 & 0.210                                 & 0.104                                 & 0.210                                 & 0.104                                 & 0.210                                 & 0.104                                 & 0.210                                 & 0.105                                 & 0.210                                 \\
                           & 96  & 0.140                                 & 0.247                                 & 0.141                                 & 0.248                                 & 0.142                                 & 0.248                                 & 0.141                                 & 0.248                                 & 0.143                                 & 0.249                                 & 0.140                                 & 0.249                                 \\ \cmidrule(l){2-14} 
\multirow{-5}{*}{PEMS03}   & Avg & {\color[HTML]{FF0000} \textbf{0.095}} & {\color[HTML]{FF0000} \textbf{0.199}} & {\color[HTML]{0000FF} {\ul 0.096}}    & {\color[HTML]{FF0000} \textbf{0.199}} & {\color[HTML]{0000FF} {\ul 0.096}}    & {\color[HTML]{FF0000} \textbf{0.199}} & {\color[HTML]{FF0000} \textbf{0.095}} & {\color[HTML]{FF0000} \textbf{0.199}} & {\color[HTML]{0000FF} {\ul 0.096}}    & {\color[HTML]{FF0000} \textbf{0.199}} & {\color[HTML]{FF0000} \textbf{0.095}} & {\color[HTML]{FF0000} \textbf{0.199}} \\ \midrule
                           & 3   & 0.036                                 & 0.092                                 & 0.036                                 & 0.092                                 & 0.036                                 & 0.092                                 & 0.036                                 & 0.093                                 & 0.036                                 & 0.093                                 & 0.036                                 & 0.093                                 \\
                           & 6   & 0.049                                 & 0.117                                 & 0.049                                 & 0.117                                 & 0.049                                 & 0.117                                 & 0.049                                 & 0.117                                 & 0.049                                 & 0.117                                 & 0.049                                 & 0.118                                 \\
                           & 9   & 0.062                                 & 0.137                                 & 0.062                                 & 0.137                                 & 0.062                                 & 0.137                                 & 0.062                                 & 0.137                                 & 0.062                                 & 0.137                                 & 0.062                                 & 0.137                                 \\
                           & 12  & 0.073                                 & 0.154                                 & 0.073                                 & 0.154                                 & 0.073                                 & 0.154                                 & 0.073                                 & 0.154                                 & 0.073                                 & 0.154                                 & 0.073                                 & 0.154                                 \\ \cmidrule(l){2-14} 
\multirow{-5}{*}{NASDAQ}   & Avg & {\color[HTML]{FF0000} \textbf{0.055}} & {\color[HTML]{FF0000} \textbf{0.125}} & {\color[HTML]{FF0000} \textbf{0.055}} & {\color[HTML]{FF0000} \textbf{0.125}} & {\color[HTML]{FF0000} \textbf{0.055}} & {\color[HTML]{FF0000} \textbf{0.125}} & {\color[HTML]{FF0000} \textbf{0.055}} & {\color[HTML]{FF0000} \textbf{0.125}} & {\color[HTML]{FF0000} \textbf{0.055}} & {\color[HTML]{FF0000} \textbf{0.125}} & {\color[HTML]{FF0000} \textbf{0.055}} & {\color[HTML]{FF0000} \textbf{0.125}} \\ \midrule
                           & 24  & 0.320                                 & 0.302                                 & 0.319                                 & 0.302                                 & 0.319                                 & 0.301                                 & 0.319                                 & 0.302                                 & 0.320                                 & 0.302                                 & 0.320                                 & 0.302                                 \\
                           & 36  & 0.334                                 & 0.315                                 & 0.334                                 & 0.315                                 & 0.334                                 & 0.315                                 & 0.334                                 & 0.315                                 & 0.334                                 & 0.315                                 & 0.334                                 & 0.316                                 \\
                           & 48  & 0.347                                 & 0.327                                 & 0.347                                 & 0.327                                 & 0.347                                 & 0.327                                 & 0.347                                 & 0.327                                 & 0.347                                 & 0.327                                 & 0.347                                 & 0.328                                 \\
                           & 60  & 0.358                                 & 0.337                                 & 0.358                                 & 0.337                                 & 0.358                                 & 0.337                                 & 0.358                                 & 0.337                                 & 0.358                                 & 0.337                                 & 0.358                                 & 0.337                                 \\ \cmidrule(l){2-14} 
\multirow{-5}{*}{CarSales} & Avg & {\color[HTML]{0000FF} {\ul 0.340}}    & {\color[HTML]{FF0000} \textbf{0.320}} & {\color[HTML]{FF0000} \textbf{0.339}} & {\color[HTML]{FF0000} \textbf{0.320}} & {\color[HTML]{FF0000} \textbf{0.339}} & {\color[HTML]{FF0000} \textbf{0.320}} & {\color[HTML]{0000FF} {\ul 0.340}}    & {\color[HTML]{FF0000} \textbf{0.320}} & {\color[HTML]{0000FF} {\ul 0.340}}    & {\color[HTML]{FF0000} \textbf{0.320}} & {\color[HTML]{0000FF} {\ul 0.340}}    & {\color[HTML]{FF0000} \textbf{0.320}} \\ \bottomrule
\end{tabular}
}
\end{table}

\section{More ablation studies}

\subsection{Ablations on NormLin design} \label{normlin_abl}

In the NormLin layer, the $\mathrm{Softplus}$ function followed by L1 row-wise normalization is applied to the learnable weight matrix $\mathbf{W}$ to produce attention-like matrices, i.e., $\mathrm{RowNorm}_{\mathrm{L1}}(\mathrm{Softplus}(\mathbf{W}))$. To assess the effects of different transformations and normalization strategies, we consider transformation functions from the set $\{\mathrm{Softplus}, \mathrm{Identity}, \mathrm{Softmax}, \mathrm{Sigmoid}, \mathrm{ReLU}\}$ and and normalization methods from ${\mathrm{L1}, \mathrm{L2}}$ (all applied row-wise). Here, $\mathrm{Identity}$ denotes no transformation.

As shown in Table~\ref{tab:abl_weight_norm}, the combination `$\mathrm{Softplus}$ + $\mathrm{L1}$' yields the best performance among the 10 candidates. On average, L1 normalization leads to a 10\% performance improvement over L2, while $\mathrm{Softplus}$ slightly outperforms $\mathrm{Softmax}$ and other transformation functions.

\begin{table}[ht]
\caption{Ablation study on transformation functions and normalization methods in NormLin. For \textit{S1} and \textit{S2}, the prediction lengths are $\left \{ 3,6,9,12 \right \} $ and $\left \{ 24,36,48,60 \right \} $, respectively. }
\label{tab:abl_weight_norm}
\centering
\setlength{\tabcolsep}{3.5pt}
\renewcommand{\arraystretch}{1.2} 
{\fontsize{7}{8}\selectfont
\begin{tabular}{@{}ccccccccccccccccc@{}}
\toprule
Fun.                       & Norm                 & Hor. & \multicolumn{2}{c}{ECL}                                                       & \multicolumn{2}{c}{Traffic}                                                   & \multicolumn{2}{c}{Solar}                                                     & \multicolumn{2}{c}{PEMS03}                                                    & \multicolumn{2}{c}{Weather}                                                   & \multicolumn{2}{c}{NASDAQ (S1)}                                               & \multicolumn{2}{c}{ILI (S2)}                                                  \\ \midrule
                           &                      & H1   & {\color[HTML]{FF0000} \textbf{0.131}} & {\color[HTML]{FF0000} \textbf{0.221}} & {\color[HTML]{FF0000} \textbf{0.398}} & {\color[HTML]{FF0000} \textbf{0.226}} & {\color[HTML]{0000FF} {\ul 0.179}}    & {\color[HTML]{FF0000} \textbf{0.191}} & {\color[HTML]{FF0000} \textbf{0.060}} & {\color[HTML]{FF0000} \textbf{0.159}} & 0.153                                 & 0.190                                 & {\color[HTML]{0000FF} {\ul 0.036}}    & {\color[HTML]{FF0000} \textbf{0.092}} & 1.737                                 & 0.800                                 \\
                           &                      & H2   & {\color[HTML]{FF0000} \textbf{0.150}} & {\color[HTML]{FF0000} \textbf{0.238}} & {\color[HTML]{FF0000} \textbf{0.439}} & {\color[HTML]{FF0000} \textbf{0.241}} & {\color[HTML]{FF0000} \textbf{0.209}} & {\color[HTML]{FF0000} \textbf{0.213}} & {\color[HTML]{0000FF} {\ul 0.078}}    & {\color[HTML]{FF0000} \textbf{0.179}} & {\color[HTML]{0000FF} {\ul 0.200}}    & {\color[HTML]{FF0000} \textbf{0.235}} & {\color[HTML]{FF0000} \textbf{0.049}} & {\color[HTML]{FF0000} \textbf{0.117}} & {\color[HTML]{FF0000} \textbf{1.714}} & {\color[HTML]{FF0000} \textbf{0.795}} \\
\bottomrule
\end{tabular}
}
\end{table}

\subsection{Pre- and post-linear layers in NormLin module} \label{pre_post_lin}

In CSL, the pre- and post-linear layers, named based on their positions in the execution sequence, are applied around the NormLin layer (see Equation~\ref{eq:csl}): $\mathrm{Post\_Lin} \left ( \mathrm{NormLin} \left ( \mathrm{Pre\_Lin} ( \cdot ) \right ) \right )$.

We conduct ablation studies to evaluate the contributions of these two layers. As shown in Table~\ref{tab:abl_pre_post_lin}, incorporating both pre- and post-linear layers yields an average performance gain of 6\% compared to the variant without them, highlighting their effectiveness in refining inputs for multivariate correlation modeling and downstream series representation learning.

\begin{table}[ht]
\caption{Ablation study on pre- and post-linear layers around the NormLin layer. `Hor.' denotes prediction horizons. For \textit{S1} and \textit{S2}, the prediction lengths are $\left \{ 3,6,9,12 \right \} $ and $\left \{ 24,36,48,60 \right \} $, respectively. }
\label{tab:abl_pre_post_lin}
\centering
\setlength{\tabcolsep}{1.2pt}
\renewcommand{\arraystretch}{1.2} 
{\fontsize{7}{8}\selectfont
\begin{tabular}{@{}ccccccccccccccccccccc@{}}
\toprule
                           &                             &                        & \multicolumn{2}{c}{ECL}                                                       & \multicolumn{2}{c}{Traffic}                                                   & \multicolumn{2}{c}{Solar}                                                     & \multicolumn{2}{c}{PEMS03}                                                    & \multicolumn{2}{c}{Weather}                                                   & \multicolumn{2}{c}{NASDAQ (S1)}                                                 & \multicolumn{2}{c}{ILI (S2)}                                                    & \multicolumn{2}{c}{Wiki (S1)}                                                  & \multicolumn{2}{c}{ETTm1}                                                     \\ \cmidrule(l){4-21} 
\multirow{-2}{*}{Pre\_Lin} & \multirow{-2}{*}{Post\_Lin} & \multirow{-2}{*}{Hor.} & MSE                                   & MAE                                   & MSE                                   & MAE                                   & MSE                                   & MAE                                   & MSE                                   & MAE                                   & MSE                                   & MAE                                   & MSE                                   & MAE                                   & MSE                                   & MAE                                   & MSE                                   & MAE                                   & MSE                                   & MAE                                   \\ \midrule
                           &                             & H1                     & 0.135                                 & 0.229                                 & 0.474                                 & 0.253                                 & 0.192                                 & 0.206                                 & 0.065                                 & 0.167                                 & 0.159                                 & 0.195                                 & 0.040                                 & 0.104                                 & 1.744                                 & 0.861                                 & 6.184                                 & 0.372                                 & 0.320                                 & 0.352                                 \\
                           &                             & H2                     & 0.153                                 & 0.246                                 & 0.492                                 & 0.264                                 & 0.225                                 & 0.228                                 & 0.085                                 & 0.191                                 & 0.207                                 & 0.241                                 & 0.053                                 & 0.126                                 & 1.970                                 & 0.891                                 & 6.471                                 & 0.387                                 & 0.373                                 & 0.384                                 \\
                           &                             & H3                     & 0.168                                 & 0.264                                 & 0.511                                 & 0.273                                 & 0.247                                 & 0.244                                 & 0.125                                 & 0.233                                 & 0.264                                 & 0.285                                 & 0.065                                 & 0.146                                 & 2.153                                 & 0.892                                 & 6.673                                 & 0.403                                 & 0.401                                 & 0.400                                 \\
                           &                             & H4                     & 0.204                                 & 0.297                                 & 0.555                                 & 0.294                                 & 0.255                                 & 0.247                                 & 0.178                                 & 0.281                                 & 0.342                                 & 0.336                                 & 0.077                                 & 0.162                                 & 2.091                                 & 0.890                                 & 6.846                                 & 0.410                                 & 0.461                                 & 0.435                                 \\ \cmidrule(l){3-21} 
\multirow{-5}{*}{\ding{55}}        & \multirow{-5}{*}{\ding{55}}         & Avg                    & {\color[HTML]{0000FF} {\ul 0.165}}    & 0.259                                 & 0.508                                 & 0.271                                 & 0.230                                 & 0.231                                 & 0.113                                 & 0.218                                 & {\color[HTML]{0000FF} {\ul 0.243}}    & {\color[HTML]{0000FF} {\ul 0.264}}    & {\color[HTML]{0000FF} {\ul 0.058}}    & 0.134                                 & 1.989                                 & 0.883                                 & {\color[HTML]{0000FF} {\ul 6.543}}                                 & {\color[HTML]{0000FF} {\ul 0.393}}                                 & 0.389                                 & 0.393                                 \\ \midrule
                           &                             & H1                     & 0.131                                 & 0.222                                 & 0.403                                 & 0.228                                 & 0.179                                 & 0.192                                 & 0.061                                 & 0.161                                 & 0.155                                 & 0.192                                 & 0.036                                 & 0.094                                 & 1.926                                 & 0.858                                 & 6.176                                 & 0.368                                 & 0.303                                 & 0.334                                 \\
                           &                             & H2                     & 0.151                                 & 0.240                                 & 0.434                                 & 0.242                                 & 0.210                                 & 0.213                                 & 0.077                                 & 0.180                                 & 0.205                                 & 0.240                                 & 0.049                                 & 0.118                                 & 1.796                                 & 0.827                                 & 6.447                                 & 0.384                                 & 0.356                                 & 0.364                                 \\
                           &                             & H3                     & 0.166                                 & 0.256                                 & 0.457                                 & 0.251                                 & 0.231                                 & 0.230                                 & 0.107                                 & 0.212                                 & 0.262                                 & 0.282                                 & 0.062                                 & 0.137                                 & 1.960                                 & 0.833                                 & 6.650                                 & 0.399                                 & 0.389                                 & 0.387                                 \\
                           &                             & H4                     & 0.211                                 & 0.298                                 & 0.500                                 & 0.272                                 & 0.243                                 & 0.236                                 & 0.146                                 & 0.253                                 & 0.352                                 & 0.340                                 & 0.073                                 & 0.155                                 & 1.799                                 & 0.811                                 & 6.839                                 & 0.406                                 & 0.452                                 & 0.426                                 \\ \cmidrule(l){3-21} 
\multirow{-5}{*}{\ding{55}}        & \multirow{-5}{*}{\ding{52}}         & Avg                    & {\color[HTML]{0000FF} {\ul 0.165}}    & 0.254                                 & {\color[HTML]{FF0000} \textbf{0.448}} & {\color[HTML]{0000FF} {\ul 0.248}}    & {\color[HTML]{0000FF} {\ul 0.216}}    & {\color[HTML]{0000FF} {\ul 0.218}}    & 0.098                                 & {\color[HTML]{0000FF} {\ul 0.201}}    & {\color[HTML]{0000FF} {\ul 0.243}}    & {\color[HTML]{FF0000} \textbf{0.263}} & {\color[HTML]{FF0000} \textbf{0.055}} & {\color[HTML]{0000FF} {\ul 0.126}}    & 1.870                                 & {\color[HTML]{0000FF} {\ul 0.832}}    & {\color[HTML]{FF0000} \textbf{6.528}} & {\color[HTML]{FF0000} \textbf{0.389}} & {\color[HTML]{0000FF} {\ul 0.375}}    & {\color[HTML]{0000FF} {\ul 0.378}}    \\ \midrule
                           &                             & H1                     & 0.131                                 & 0.222                                 & 0.404                                 & 0.228                                 & 0.178                                 & 0.192                                 & 0.061                                 & 0.160                                 & 0.154                                 & 0.192                                 & 0.036                                 & 0.094                                 & 1.924                                 & 0.858                                 & 6.175                                 & 0.368                                 & 0.303                                 & 0.334                                 \\
                           &                             & H2                     & 0.155                                 & 0.243                                 & 0.434                                 & 0.242                                 & 0.211                                 & 0.214                                 & 0.077                                 & 0.180                                 & 0.204                                 & 0.239                                 & 0.049                                 & 0.118                                 & 1.796                                 & 0.828                                 & 6.448                                 & 0.385                                 & 0.356                                 & 0.364                                 \\
                           &                             & H3                     & 0.165                                 & 0.255                                 & 0.457                                 & 0.251                                 & 0.231                                 & 0.230                                 & 0.107                                 & 0.212                                 & 0.263                                 & 0.282                                 & 0.062                                 & 0.137                                 & 1.959                                 & 0.833                                 & 6.649                                 & 0.399                                 & 0.389                                 & 0.387                                 \\
                           &                             & H4                     & 0.209                                 & 0.293                                 & 0.499                                 & 0.272                                 & 0.242                                 & 0.236                                 & 0.144                                 & 0.252                                 & 0.344                                 & 0.339                                 & 0.073                                 & 0.155                                 & 1.799                                 & 0.811                                 & 6.839                                 & 0.406                                 & 0.452                                 & 0.426                                 \\ \cmidrule(l){3-21} 
\multirow{-5}{*}{\ding{52}}        & \multirow{-5}{*}{\ding{55}}         & Avg                    & {\color[HTML]{0000FF} {\ul 0.165}}    & {\color[HTML]{0000FF} {\ul 0.253}}    & {\color[HTML]{FF0000} \textbf{0.448}} & {\color[HTML]{0000FF} {\ul 0.248}}    & {\color[HTML]{FF0000} \textbf{0.215}} & {\color[HTML]{0000FF} {\ul 0.218}}    & {\color[HTML]{0000FF} {\ul 0.097}}    & {\color[HTML]{0000FF} {\ul 0.201}}    & {\color[HTML]{FF0000} \textbf{0.241}} & {\color[HTML]{FF0000} \textbf{0.263}} & {\color[HTML]{FF0000} \textbf{0.055}} & {\color[HTML]{0000FF} {\ul 0.126}}    & {\color[HTML]{0000FF} {\ul 1.869}}    & {\color[HTML]{0000FF} {\ul 0.832}}    & {\color[HTML]{FF0000} \textbf{6.528}} & {\color[HTML]{FF0000} \textbf{0.389}} & {\color[HTML]{0000FF} {\ul 0.375}}    & {\color[HTML]{0000FF} {\ul 0.378}}    \\ \midrule
                           &                             & H1                     & 0.131                                 & 0.221                                 & 0.398                                 & 0.226                                 & 0.179                                 & 0.191                                 & 0.060                                 & 0.159                                 & 0.152                                 & 0.189                                 & 0.036                                 & 0.092                                 & 1.737                                 & 0.800                                 & 6.161                                 & 0.368                                 & 0.302                                 & 0.334                                 \\
                           &                             & H2                     & 0.150                                 & 0.238                                 & 0.439                                 & 0.241                                 & 0.209                                 & 0.213                                 & 0.078                                 & 0.179                                 & 0.207                                 & 0.243                                 & 0.049                                 & 0.117                                 & 1.714                                 & 0.795                                 & 6.453                                 & 0.385                                 & 0.357                                 & 0.364                                 \\
                           &                             & H3                     & 0.165                                 & 0.254                                 & 0.464                                 & 0.250                                 & 0.231                                 & 0.229                                 & 0.104                                 & 0.210                                 & 0.261                                 & 0.282                                 & 0.062                                 & 0.137                                 & 1.821                                 & 0.804                                 & 6.666                                 & 0.398                                 & 0.387                                 & 0.385                                 \\
                           &                             & H4                     & 0.191                                 & 0.279                                 & 0.502                                 & 0.270                                 & 0.241                                 & 0.236                                 & 0.140                                 & 0.247                                 & 0.344                                 & 0.337                                 & 0.073                                 & 0.154                                 & 1.785                                 & 0.810                                 & 6.834                                 & 0.406                                 & 0.452                                 & 0.426                                 \\ \cmidrule(l){3-21} 
\multirow{-5}{*}{\ding{52}}        & \multirow{-5}{*}{\ding{52}}         & Avg                    & {\color[HTML]{FF0000} \textbf{0.159}} & {\color[HTML]{FF0000} \textbf{0.248}} & {\color[HTML]{0000FF} {\ul 0.451}}    & {\color[HTML]{FF0000} \textbf{0.247}} & {\color[HTML]{FF0000} \textbf{0.215}} & {\color[HTML]{FF0000} \textbf{0.217}} & {\color[HTML]{FF0000} \textbf{0.095}} & {\color[HTML]{FF0000} \textbf{0.199}} & {\color[HTML]{FF0000} \textbf{0.241}} & {\color[HTML]{FF0000} \textbf{0.263}} & {\color[HTML]{FF0000} \textbf{0.055}} & {\color[HTML]{FF0000} \textbf{0.125}} & {\color[HTML]{FF0000} \textbf{1.764}} & {\color[HTML]{FF0000} \textbf{0.802}} & {\color[HTML]{FF0000} \textbf{6.528}} & {\color[HTML]{FF0000} \textbf{0.389}} & {\color[HTML]{FF0000} \textbf{0.374}} & {\color[HTML]{FF0000} \textbf{0.377}} \\ \bottomrule
\end{tabular}
}
\end{table}

\subsection{Ablations on Q matrices} \label{abl_qi_qo}

The effectiveness of OrthoTrans has been demonstrated in Tables~\ref{tab:base} and~\ref{tab:base_iTrans}. To further analyze the individual contributions of $\mathbf{Q}_i$ and $\mathbf{Q}_o$, we conduct ablation studies on these transformation matrices.

As shown in Table~\ref{tab:abl_qi_qo}, using both $\mathbf{Q}_i$ and $\mathbf{Q}_o$ achieves the best performance—reducing the average MSE by 5.3\% compared to the baseline without transformations, by 4.4\% compared to using $\mathbf{Q}_o$ only, and by 0.8\% compared to using $\mathbf{Q}_i$ only. Notably, the variant using only $\mathbf{Q}_i$ outperforms the one using only $\mathbf{Q}_o$, highlighting the critical role of temporal decorrelation in effective encoding and representation learning.

\begin{table}[ht]
\caption{Ablation study on $\mathbf{Q}_i$ and $\mathbf{Q}_o$. `Hor.' denotes prediction horizons. For \textit{S1} and \textit{S2}, the prediction lengths are $ \{ 3, 6, 9, 12  \} $ and $\{ 24, 36, 48, 60 \} $, respectively. }
\label{tab:abl_qi_qo}
\centering
\setlength{\tabcolsep}{2.7pt}
\renewcommand{\arraystretch}{1.2} 
{\fontsize{7}{8}\selectfont
\begin{tabular}{@{}ccccccccccccccccccc@{}}
\toprule
                       &                        &                        & \multicolumn{2}{c}{ECL}                                                       & \multicolumn{2}{c}{Solar-Energy}                                              & \multicolumn{2}{c}{PEMS03}                                                    & \multicolumn{2}{c}{PEMS08}                                                    & \multicolumn{2}{c}{NASDAQ (S1)}                                               & \multicolumn{2}{c}{ILI (S2)}                                                  & \multicolumn{2}{c}{COVID-19 (S2)}                                              & \multicolumn{2}{c}{Exchange}                                                  \\ \cmidrule(l){4-19} 
\multirow{-2}{*}{$\mathbf{Q}_i$}   & \multirow{-2}{*}{$\mathbf{Q}_o$}   & \multirow{-2}{*}{Hor.} & MSE                                   & MAE                                   & MSE                                   & MAE                                   & MSE                                   & MAE                                   & MSE                                   & MAE                                   & MSE                                   & MAE                                   & MSE                                   & MAE                                   & MSE                                    & MAE                                   & MSE                                   & MAE                                   \\ \midrule
                       &                        & H1                     & 0.134                                 & 0.224                                 & {\color[HTML]{0000FF} {\ul 0.186}}    & 0.199                                 & 0.063                                 & 0.163                                 & 0.070                                 & {\color[HTML]{0000FF} {\ul 0.162}}    & {\color[HTML]{0000FF} {\ul 0.036}}    & {\color[HTML]{0000FF} {\ul 0.093}}    & 1.891                                 & 0.847                                 & 4.841                                  & 1.284                                 & {\color[HTML]{0000FF} {\ul 0.083}}    & {\color[HTML]{0000FF} {\ul 0.203}}    \\
                       &                        & H2                     & 0.156                                 & 0.245                                 & 0.221                                 & 0.220                                 & 0.080                                 & 0.184                                 & 0.091                                 & 0.183                                 & {\color[HTML]{0000FF} {\ul 0.050}}    & {\color[HTML]{0000FF} {\ul 0.119}}    & 1.889                                 & 0.844                                 & 7.828                                  & 1.757                                 & 0.173                                 & {\color[HTML]{0000FF} {\ul 0.295}}    \\
                       &                        & H3                     & {\color[HTML]{0000FF} {\ul 0.167}}    & {\color[HTML]{0000FF} {\ul 0.257}}    & 0.244                                 & {\color[HTML]{0000FF} {\ul 0.237}}    & 0.117                                 & 0.222                                 & 0.128                                 & {\color[HTML]{0000FF} {\ul 0.216}}    & {\color[HTML]{0000FF} {\ul 0.063}}    & {\color[HTML]{0000FF} {\ul 0.139}}    & 1.854                                 & 0.811                                 & 10.153                                 & 2.003                                 & 0.335                                 & {\color[HTML]{0000FF} {\ul 0.416}}    \\
                       &                        & H4                     & 0.194                                 & {\color[HTML]{0000FF} {\ul 0.282}}    & 0.256                                 & 0.244                                 & 0.165                                 & 0.266                                 & 0.231                                 & 0.262                                 & {\color[HTML]{0000FF} {\ul 0.074}}    & {\color[HTML]{0000FF} {\ul 0.156}}    & 1.802                                 & 0.816                                 & 12.603                                 & 2.232                                 & {\color[HTML]{0000FF} {\ul 0.846}}    & {\color[HTML]{0000FF} {\ul 0.692}}    \\ \cmidrule(l){3-19} 
\multirow{-5}{*}{\ding{55}}  & \multirow{-5}{*}{\ding{55}}  & Avg                    & 0.163                                 & 0.252                                 & 0.227                                 & {\color[HTML]{0000FF} {\ul 0.225}}    & 0.106                                 & 0.209                                 & 0.130                                 & {\color[HTML]{0000FF} {\ul 0.206}}    & {\color[HTML]{0000FF} {\ul 0.056}}    & {\color[HTML]{0000FF} {\ul 0.127}}    & 1.859                                 & 0.829                                 & 8.856                                  & 1.819                                 & {\color[HTML]{0000FF} {\ul 0.359}}    & {\color[HTML]{0000FF} {\ul 0.401}}    \\ \midrule
                       &                        & H1                     & {\color[HTML]{0000FF} {\ul 0.132}}    & {\color[HTML]{0000FF} {\ul 0.223}}    & {\color[HTML]{0000FF} {\ul 0.186}}    & {\color[HTML]{0000FF} {\ul 0.198}}    & 0.062                                 & {\color[HTML]{0000FF} {\ul 0.162}}    & {\color[HTML]{0000FF} {\ul 0.069}}    & {\color[HTML]{0000FF} {\ul 0.162}}    & {\color[HTML]{0000FF} {\ul 0.036}}    & {\color[HTML]{0000FF} {\ul 0.093}}    & {\color[HTML]{0000FF} {\ul 1.844}}    & {\color[HTML]{0000FF} {\ul 0.830}}    & {\color[HTML]{0000FF} {\ul 4.645}}     & {\color[HTML]{0000FF} {\ul 1.260}}    & {\color[HTML]{0000FF} {\ul 0.083}}    & {\color[HTML]{0000FF} {\ul 0.203}}    \\
                       &                        & H2                     & {\color[HTML]{0000FF} {\ul 0.153}}    & 0.243                                 & 0.219                                 & 0.219                                 & 0.079                                 & 0.183                                 & 0.092                                 & 0.183                                 & {\color[HTML]{0000FF} {\ul 0.050}}    & {\color[HTML]{0000FF} {\ul 0.119}}    & {\color[HTML]{0000FF} {\ul 1.779}}    & {\color[HTML]{0000FF} {\ul 0.811}}    & 8.137                                  & 1.798                                 & 0.173                                 & {\color[HTML]{0000FF} {\ul 0.295}}    \\
                       &                        & H3                     & {\color[HTML]{0000FF} {\ul 0.167}}    & 0.258                                 & 0.246                                 & 0.238                                 & 0.115                                 & 0.221                                 & 0.132                                 & 0.217                                 & {\color[HTML]{0000FF} {\ul 0.063}}    & {\color[HTML]{0000FF} {\ul 0.139}}    & 1.854                                 & 0.809                                 & {\color[HTML]{FF0000} \textbf{10.054}} & {\color[HTML]{0000FF} {\ul 1.990}}    & 0.334                                 & {\color[HTML]{0000FF} {\ul 0.416}}    \\
                       &                        & H4                     & 0.194                                 & 0.285                                 & {\color[HTML]{0000FF} {\ul 0.255}}    & 0.244                                 & 0.164                                 & 0.267                                 & 0.216                                 & 0.261                                 & {\color[HTML]{0000FF} {\ul 0.074}}    & {\color[HTML]{0000FF} {\ul 0.156}}    & 1.787                                 & 0.813                                 & {\color[HTML]{0000FF} {\ul 12.000}}    & 2.189                                 & 0.848                                 & 0.693                                 \\ \cmidrule(l){3-19} 
\multirow{-5}{*}{\ding{55}}  & \multirow{-5}{*}{\ding{52}} & Avg                    & {\color[HTML]{0000FF} {\ul 0.161}}    & 0.252                                 & {\color[HTML]{0000FF} {\ul 0.226}}    & {\color[HTML]{0000FF} {\ul 0.225}}    & {\color[HTML]{0000FF} {\ul 0.105}}    & {\color[HTML]{0000FF} {\ul 0.208}}    & 0.127                                 & {\color[HTML]{0000FF} {\ul 0.206}}    & {\color[HTML]{0000FF} {\ul 0.056}}    & {\color[HTML]{0000FF} {\ul 0.127}}    & {\color[HTML]{0000FF} {\ul 1.816}}    & {\color[HTML]{0000FF} {\ul 0.816}}    & 8.709                                  & 1.809                                 & {\color[HTML]{0000FF} {\ul 0.359}}    & {\color[HTML]{0000FF} {\ul 0.401}}    \\ \midrule
                       &                        & H1                     & {\color[HTML]{FF0000} \textbf{0.131}} & {\color[HTML]{FF0000} \textbf{0.221}} & {\color[HTML]{FF0000} \textbf{0.179}} & {\color[HTML]{FF0000} \textbf{0.191}} & {\color[HTML]{0000FF} {\ul 0.061}}    & {\color[HTML]{FF0000} \textbf{0.159}} & {\color[HTML]{FF0000} \textbf{0.068}} & {\color[HTML]{FF0000} \textbf{0.159}} & {\color[HTML]{FF0000} \textbf{0.036}} & {\color[HTML]{FF0000} \textbf{0.092}} & 1.897                                 & 0.847                                 & 4.957                                  & 1.330                                 & {\color[HTML]{FF0000} \textbf{0.082}} & {\color[HTML]{FF0000} \textbf{0.200}} \\
                       &                        & H2                     & 0.154                                 & {\color[HTML]{0000FF} {\ul 0.241}}    & {\color[HTML]{0000FF} {\ul 0.210}}    & {\color[HTML]{FF0000} \textbf{0.212}} & {\color[HTML]{FF0000} \textbf{0.076}} & {\color[HTML]{FF0000} \textbf{0.178}} & {\color[HTML]{FF0000} \textbf{0.087}} & {\color[HTML]{FF0000} \textbf{0.177}} & {\color[HTML]{FF0000} \textbf{0.049}} & {\color[HTML]{FF0000} \textbf{0.117}} & 1.886                                 & 0.836                                 & {\color[HTML]{FF0000} \textbf{7.100}}  & {\color[HTML]{FF0000} \textbf{1.654}} & {\color[HTML]{0000FF} {\ul 0.172}}    & {\color[HTML]{FF0000} \textbf{0.293}} \\
                       &                        & H3                     & {\color[HTML]{FF0000} \textbf{0.165}} & {\color[HTML]{FF0000} \textbf{0.254}} & {\color[HTML]{FF0000} \textbf{0.230}} & {\color[HTML]{FF0000} \textbf{0.229}} & {\color[HTML]{0000FF} {\ul 0.105}}    & {\color[HTML]{0000FF} {\ul 0.211}}    & {\color[HTML]{FF0000} \textbf{0.121}} & {\color[HTML]{FF0000} \textbf{0.204}} & {\color[HTML]{FF0000} \textbf{0.062}} & {\color[HTML]{FF0000} \textbf{0.137}} & {\color[HTML]{FF0000} \textbf{1.809}} & {\color[HTML]{FF0000} \textbf{0.802}} & 10.224                                 & 2.009                                 & {\color[HTML]{0000FF} {\ul 0.332}}    & {\color[HTML]{FF0000} \textbf{0.414}} \\
                       &                        & H4                     & {\color[HTML]{FF0000} \textbf{0.188}} & {\color[HTML]{FF0000} \textbf{0.279}} & {\color[HTML]{FF0000} \textbf{0.241}} & {\color[HTML]{FF0000} \textbf{0.235}} & {\color[HTML]{0000FF} {\ul 0.142}}    & {\color[HTML]{0000FF} {\ul 0.249}}    & {\color[HTML]{0000FF} {\ul 0.182}}    & {\color[HTML]{0000FF} {\ul 0.238}}    & {\color[HTML]{FF0000} \textbf{0.073}} & {\color[HTML]{FF0000} \textbf{0.154}} & {\color[HTML]{FF0000} \textbf{1.784}} & {\color[HTML]{0000FF} {\ul 0.811}}    & {\color[HTML]{FF0000} \textbf{11.779}} & {\color[HTML]{FF0000} \textbf{2.160}} & {\color[HTML]{FF0000} \textbf{0.837}} & {\color[HTML]{FF0000} \textbf{0.688}} \\ \cmidrule(l){3-19} 
\multirow{-5}{*}{\ding{52}} & \multirow{-5}{*}{\ding{55}}  & Avg                    & {\color[HTML]{FF0000} \textbf{0.159}} & {\color[HTML]{0000FF} {\ul 0.249}}    & {\color[HTML]{FF0000} \textbf{0.215}} & {\color[HTML]{FF0000} \textbf{0.217}} & 0.096                                 & {\color[HTML]{FF0000} \textbf{0.199}} & {\color[HTML]{0000FF} {\ul 0.114}}    & {\color[HTML]{FF0000} \textbf{0.194}} & {\color[HTML]{FF0000} \textbf{0.055}} & {\color[HTML]{FF0000} \textbf{0.125}} & {\color[HTML]{FF0000} \textbf{1.844}} & 0.824                                 & {\color[HTML]{0000FF} {\ul 8.515}}     & {\color[HTML]{0000FF} {\ul 1.788}}    & {\color[HTML]{FF0000} \textbf{0.355}} & {\color[HTML]{FF0000} \textbf{0.399}} \\ \midrule
                       &                        & H1                     & {\color[HTML]{FF0000} \textbf{0.131}} & {\color[HTML]{FF0000} \textbf{0.221}} & {\color[HTML]{FF0000} \textbf{0.179}} & {\color[HTML]{FF0000} \textbf{0.191}} & {\color[HTML]{FF0000} \textbf{0.060}} & {\color[HTML]{FF0000} \textbf{0.159}} & {\color[HTML]{FF0000} \textbf{0.068}} & {\color[HTML]{FF0000} \textbf{0.159}} & {\color[HTML]{FF0000} \textbf{0.036}} & {\color[HTML]{FF0000} \textbf{0.092}} & {\color[HTML]{FF0000} \textbf{1.737}} & {\color[HTML]{FF0000} \textbf{0.800}} & {\color[HTML]{FF0000} \textbf{4.474}}  & {\color[HTML]{FF0000} \textbf{1.180}} & {\color[HTML]{FF0000} \textbf{0.082}} & {\color[HTML]{FF0000} \textbf{0.200}} \\
                       &                        & H2                     & {\color[HTML]{FF0000} \textbf{0.150}} & {\color[HTML]{FF0000} \textbf{0.238}} & {\color[HTML]{FF0000} \textbf{0.209}} & {\color[HTML]{0000FF} {\ul 0.213}}    & {\color[HTML]{0000FF} {\ul 0.078}}    & {\color[HTML]{0000FF} {\ul 0.179}}    & {\color[HTML]{0000FF} {\ul 0.089}}    & {\color[HTML]{0000FF} {\ul 0.178}}    & {\color[HTML]{FF0000} \textbf{0.049}} & {\color[HTML]{FF0000} \textbf{0.117}} & {\color[HTML]{FF0000} \textbf{1.714}} & {\color[HTML]{FF0000} \textbf{0.795}} & {\color[HTML]{0000FF} {\ul 7.241}}     & {\color[HTML]{0000FF} {\ul 1.670}}    & {\color[HTML]{FF0000} \textbf{0.171}} & {\color[HTML]{FF0000} \textbf{0.293}} \\
                       &                        & H3                     & {\color[HTML]{FF0000} \textbf{0.165}} & {\color[HTML]{FF0000} \textbf{0.254}} & {\color[HTML]{0000FF} {\ul 0.231}}    & {\color[HTML]{FF0000} \textbf{0.229}} & {\color[HTML]{FF0000} \textbf{0.104}} & {\color[HTML]{FF0000} \textbf{0.210}} & {\color[HTML]{0000FF} {\ul 0.123}}    & {\color[HTML]{FF0000} \textbf{0.204}} & {\color[HTML]{FF0000} \textbf{0.062}} & {\color[HTML]{FF0000} \textbf{0.137}} & {\color[HTML]{0000FF} {\ul 1.821}}    & {\color[HTML]{0000FF} {\ul 0.804}}    & {\color[HTML]{0000FF} {\ul 10.076}}    & {\color[HTML]{FF0000} \textbf{1.985}} & {\color[HTML]{FF0000} \textbf{0.331}} & {\color[HTML]{FF0000} \textbf{0.414}} \\
                       &                        & H4                     & {\color[HTML]{0000FF} {\ul 0.191}}    & {\color[HTML]{FF0000} \textbf{0.279}} & {\color[HTML]{FF0000} \textbf{0.241}} & {\color[HTML]{0000FF} {\ul 0.236}}    & {\color[HTML]{FF0000} \textbf{0.140}} & {\color[HTML]{FF0000} \textbf{0.247}} & {\color[HTML]{FF0000} \textbf{0.173}} & {\color[HTML]{FF0000} \textbf{0.236}} & {\color[HTML]{FF0000} \textbf{0.073}} & {\color[HTML]{FF0000} \textbf{0.154}} & {\color[HTML]{0000FF} {\ul 1.785}}    & {\color[HTML]{FF0000} \textbf{0.810}} & 12.079                                 & {\color[HTML]{0000FF} {\ul 2.182}}    & {\color[HTML]{FF0000} \textbf{0.837}} & {\color[HTML]{FF0000} \textbf{0.688}} \\ \cmidrule(l){3-19} 
\multirow{-5}{*}{\ding{52}} & \multirow{-5}{*}{\ding{52}} & Avg                    & {\color[HTML]{FF0000} \textbf{0.159}} & {\color[HTML]{FF0000} \textbf{0.248}} & {\color[HTML]{FF0000} \textbf{0.215}} & {\color[HTML]{FF0000} \textbf{0.217}} & {\color[HTML]{FF0000} \textbf{0.095}} & {\color[HTML]{FF0000} \textbf{0.199}} & {\color[HTML]{FF0000} \textbf{0.113}} & {\color[HTML]{FF0000} \textbf{0.194}} & {\color[HTML]{FF0000} \textbf{0.055}} & {\color[HTML]{FF0000} \textbf{0.125}} & {\color[HTML]{FF0000} \textbf{1.764}} & {\color[HTML]{FF0000} \textbf{0.802}} & {\color[HTML]{FF0000} \textbf{8.467}}  & {\color[HTML]{FF0000} \textbf{1.754}} & {\color[HTML]{FF0000} \textbf{0.355}} & {\color[HTML]{FF0000} \textbf{0.399}} \\ \bottomrule
\end{tabular}
}
\end{table}

\subsection{Swapping Q matrices}

Table~\ref{tab:swap_q_mat} reports the forecasting performance when both training and testing are conducted using Q matrices ($\mathbf{Q}_i$ and $\mathbf{Q}_o$) derived from different datasets. As shown, using the Q matrices from the same dataset yields the best performance, highlighting the importance of incorporating dataset-specific temporal correlation information in time series forecasting.

\begin{table}[ht]
\caption{Forecasting performance when both training and testing use Q matrices $\mathbf{Q}_i$ and $\mathbf{Q}_o$ from different datasets. $\mathcal{A} \to \mathcal{B}$ indicates that the Q matrices computed from dataset $\mathcal{A}$ are applied in experiments on dataset $\mathcal{B}$.}
\label{tab:swap_q_mat}
\centering
\setlength{\tabcolsep}{20pt}
\renewcommand{\arraystretch}{1.2} 
{\fontsize{8}{9}\selectfont
\begin{tabular}{@{}ccccccc@{}}
\toprule
Setting & \multicolumn{2}{c}{ILI→ILI}                                                   & \multicolumn{2}{c}{Power→ILI}                                                 & \multicolumn{2}{c}{Web→ILI}                                                   \\ \midrule
Metric  & MSE                                   & MAE                                   & MSE                                   & MAE                                   & MSE                                   & MAE                                   \\ \midrule
24      & 1.737                                 & 0.800                                 & 1.846                                 & 0.830                                 & 1.887                                 & 0.842                                 \\
36      & 1.714                                 & 0.795                                 & 1.721                                 & 0.801                                 & 1.743                                 & 0.807                                 \\
48      & 1.821                                 & 0.804                                 & 1.849                                 & 0.810                                 & 1.875                                 & 0.817                                 \\
60      & 1.785                                 & 0.810                                 & 1.824                                 & 0.814                                 & 1.815                                 & 0.813                                 \\ \midrule
Avg     & {\color[HTML]{FF0000} \textbf{1.764}} & {\color[HTML]{FF0000} \textbf{0.802}} & 1.810                                 & 0.814                                 & 1.830                                 & 0.820                                 \\ \midrule
Setting & \multicolumn{2}{c}{ILI→Power}                                                 & \multicolumn{2}{c}{Power→Power}                                               & \multicolumn{2}{c}{Web→Power}                                                 \\ \midrule
Metric  & MSE                                   & MAE                                   & MSE                                   & MAE                                   & MSE                                   & MAE                                   \\ \midrule
24      & 1.391                                 & 0.895                                 & 1.343                                 & 0.870                                 & 1.393                                 & 0.889                                 \\
36      & 1.457                                 & 0.902                                 & 1.445                                 & 0.903                                 & 1.493                                 & 0.915                                 \\
48      & 1.612                                 & 0.972                                 & 1.559                                 & 0.946                                 & 1.680                                 & 0.997                                 \\
60      & 1.733                                 & 1.035                                 & 1.602                                 & 0.971                                 & 1.773                                 & 1.048                                 \\ \midrule
Avg     & 1.548                                 & 0.951                                 & {\color[HTML]{FF0000} \textbf{1.487}} & {\color[HTML]{FF0000} \textbf{0.922}} & 1.585                                 & 0.962                                 \\ \midrule
Setting & \multicolumn{2}{c}{ILI→Web}                                                   & \multicolumn{2}{c}{Power→Web}                                                 & \multicolumn{2}{c}{Web→Web}                                                   \\ \midrule
Metric  & MSE                                   & MAE                                   & MSE                                   & MAE                                   & MSE                                   & MAE                                   \\ \midrule
24      & 0.208                                 & 0.319                                 & 0.185                                 & 0.306                                 & 0.186                                 & 0.306                                 \\
36      & 0.285                                 & 0.369                                 & 0.289                                 & 0.364                                 & 0.272                                 & 0.356                                 \\
48      & 0.377                                 & 0.403                                 & 0.400                                 & 0.427                                 & 0.365                                 & 0.391                                 \\
60      & 0.480                                 & 0.476                                 & 0.467                                 & 0.466                                 & 0.486                                 & 0.481                                 \\ \midrule
Avg     & 0.337                                 & 0.391                                 & 0.335                                 & 0.391                                 & {\color[HTML]{FF0000} \textbf{0.327}} & {\color[HTML]{FF0000} \textbf{0.383}} \\ \bottomrule
\end{tabular}
}
\end{table}

\section{Model efficiency} \label{gpu_appendix}

As a linear model, OLinear achieves remarkable training and inference efficiency while delivering state-of-the-art forecasting performance.  Figure~\ref{fig:gpu_main_text} shows that OLinear achieves better efficiency than Transformer-based forecasters, benefiting from its relatively simple architecture. Detailed resource consumption for both training and inference is summarized in Table~\ref{tab:GPU}. Moreover, Table~\ref{tab:GPU_normlin_plugin_compare} presents the efficiency differences before and after applying the NormLin module to Transformer-based forecasters.


\begin{table}[ht]
\caption{Training and inference resource footprint for OLinear and baseline forecasters. `T.T.', `T.M.', `I.T.' and `I.M.' denote training time, training GPU memory usage, inference time, and inference GPU memory usage, respectively. All experiments use a lookback/prediction of 96 and a batch size of 16. To ensure a fair comparison, the number of layers (or blocks) is fixed at 2, except for DLinear, which employs only 1 linear layer. `OL' refers to OLinear. `OOM' indicates that the experiment runs out of memory on a 24 GB GPU.}
\label{tab:GPU}
\centering
\setlength{\tabcolsep}{2.5pt}
\renewcommand{\arraystretch}{0.9} 
{\fontsize{7}{9}\selectfont
\begin{tabular}{@{}cccccccccccccc@{}}
\toprule
\multicolumn{2}{c}{Model}                     & \begin{tabular}[c]{@{}c@{}}OL\\      (Ours)\end{tabular} & \begin{tabular}[c]{@{}c@{}}OL-C\\      (Ours)\end{tabular} & 
\begin{tabular}[c]{@{}c@{}}Leddam\\  \citeyear{Leddam_icml} \end{tabular} & \begin{tabular}[c]{@{}c@{}}CARD\\      \citeyear{card}\end{tabular} & 
\begin{tabular}[c]{@{}c@{}}Fredformer\\  \citeyear{fredformer} \end{tabular} & 
\begin{tabular}[c]{@{}c@{}}iTrans.\\  \citeyear{itransformer} \end{tabular} & \begin{tabular}[c]{@{}c@{}}TimeMix.++\\      \citeyear{timemixer++} \end{tabular} & \begin{tabular}[c]{@{}c@{}}TimeMix.\\  \citeyear{timemixer}   \end{tabular} & \begin{tabular}[c]{@{}c@{}}Dlinear\\     \citeyear{linear} \end{tabular} & \begin{tabular}[c]{@{}c@{}}TimesNet\\      \citeyear{timesnet} \end{tabular} & \begin{tabular}[c]{@{}c@{}}PatchTST\\   \citeyear{patchtst} \end{tabular} & \begin{tabular}[c]{@{}c@{}}FreTS\\     \citeyear{frets} \end{tabular} \\ \midrule
\multirow{6}{*}{\rotatebox[origin=c]{90}{ETT}}          & Params(M)      & 4.52                                                              & 4.52                                                                & 8.55                                                     & 0.03                                                   & 8.59                                                         & 4.83                                                      & 1.19                                                          & 0.08                                                       & 0.02                                                      & 299.94                                                     & 3.76                                                       & 0.42                                                    \\
                              & FLOPs(M)       & 33.74                                                             & 33.74                                                               & 111.54                                                   & 1.41                                                   & 135.01                                                       & 33.99                                                     & 1396.90                                                       & 10.72                                                      & 0.13                                                      & 289708                                                     & 272.20                                                     & 2.94                                                    \\
                              & T.T. (ms/iter) & 8.09                                                              & 7.26                                                                & 25.02                                                    & 29.10                                                  & 26.48                                                        & 10.11                                                     & 74.74                                                         & 27.99                                                      & 1.42                                                      & 501.98                                                     & 8.37                                                       & 3.63                                                    \\
                              & T.M.(GB)       & 0.20                                                              & 0.2                                                                 & 0.18                                                     & 0.03                                                   & 0.24                                                         & 0.21                                                      & 0.27                                                          & 0.05                                                       & 0.02                                                      & 5.80                                                       & 0.14                                                       & 0.03                                                    \\
                              & I.T.(ms/iter)  & 1.19                                                              & 1.09                                                                & 2.62                                                     & 4.05                                                   & 3.73                                                         & 1.72                                                      & 28.15                                                         & 3.86                                                       & 0.21                                                      & 155.97                                                     & 1.38                                                       & 0.49                                                    \\
                              & I.M. (MB)      & 156.41                                                            & 156.41                                                              & 51.04                                                    & 11.02                                                  & 78.04                                                        & 156.00                                                    & 149.58                                                        & 18.06                                                      & 8.48                                                      & 1396.24                                                    & 51.83                                                      & 14.47                                                   \\ \midrule
\multirow{6}{*}{\rotatebox[origin=c]{90}{ECL}}          & Params(M)      & 4.79                                                              & 4.58                                                                & 8.56                                                     & 1.39                                                   & 12.12                                                        & 4.83                                                      & 1.19                                                          & 0.12                                                       & 0.02                                                      & 300.58                                                     & 3.76                                                       & 0.42                                                    \\
                              & FLOPs(G)       & 1.65                                                              & 1.65                                                                & 5.32                                                     & 5.07                                                   & 5.55                                                         & 1.87                                                      & 64.02                                                         & 1.74                                                       & 0.01                                                      & 293.98                                                     & 12.48                                                      & 0.13                                                    \\
                              & T.T. (ms/iter) & 7.75                                                              & 6.94                                                                & 65.19                                                    & 78.25                                                  & 26.59                                                        & 10.66                                                     & 106.06                                                        & 57.09                                                      & 1.37                                                      & 509.17                                                     & 65.83                                                      & 3.78                                                    \\
                              & T.M.(GB)       & 0.45                                                              & 0.42                                                                & 1.44                                                     & 5.08                                                   & 1.24                                                         & 0.70                                                      & 1.02                                                          & 4.24                                                       & 0.04                                                      & 5.82                                                       & 3.02                                                       & 0.30                                                    \\
                              & I.T.(ms/iter)  & 2.11                                                              & 2.09                                                                & 14.78                                                    & 28.21                                                  & 7.08                                                         & 3.46                                                      & 34.10                                                         & 22.79                                                      & 0.20                                                      & 157.13                                                     & 20.52                                                      & 0.48                                                    \\
                              & I.M. (GB)      & 0.17                                                              & 0.17                                                                & 0.49                                                     & 0.63                                                   & 0.26                                                         & 0.23                                                      & 0.36                                                          & 0.82                                                       & 0.02                                                      & 1.41                                                       & 0.93                                                       & 0.23                                                    \\ \midrule
\multirow{6}{*}{\rotatebox[origin=c]{90}{Exchange}}     & Params(M)      & 1.74                                                              & 1.74                                                                & 8.56                                                     & 1.39                                                   & 8.59                                                         & 4.83                                                      & 1.19                                                          & 0.08                                                       & 0.02                                                      & 299.94                                                     & 3.76                                                       & 0.42                                                    \\
                              & FLOPs(M)       & 16.27                                                             & 16.27                                                               & 127.49                                                   & 114.83                                                 & 135.01                                                       & 38.88                                                     & 1595.72                                                       & 12.25                                                      & 0.15                                                      & 287909                                                     & 311.08                                                     & 3.36                                                    \\
                              & T.T. (ms/iter) & 7.33                                                              & 7.53                                                                & 24.81                                                    & 35.57                                                  & 28.56                                                        & 10.14                                                     & 97.05                                                         & 27.46                                                      & 1.47                                                      & 505.39                                                     & 9.75                                                       & 3.51                                                    \\
                              & T.M.(GB)       & 0.17                                                              & 0.17                                                                & 0.18                                                     & 0.17                                                   & 0.24                                                         & 0.21                                                      & 0.28                                                          & 0.06                                                       & 0.02                                                      & 5.80                                                       & 0.15                                                       & 0.03                                                    \\
                              & I.T.(ms/iter)  & 1.18                                                              & 1.89                                                                & 2.62                                                     & 6.02                                                   & 6.86                                                         & 1.70                                                      & 28.46                                                         & 3.81                                                       & 0.21                                                      & 156.33                                                     & 2.47                                                       & 0.48                                                    \\
                              & I.M. (GB)      & 0.02                                                              & 0.02                                                                & 0.05                                                     & 0.03                                                   & 0.08                                                         & 0.16                                                      & 0.15                                                          & 0.02                                                       & 0.01                                                      & 1.40                                                       & 0.05                                                       & 0.02                                                    \\ \midrule
\multirow{6}{*}{\rotatebox[origin=c]{90}{Traffic}}      & Params(M)      & 6.17                                                              & 4.69                                                                & 8.56                                                     & 0.98                                                   & 11.09                                                        & 4.83                                                      & 4.73                                                          & 0.12                                                       & 0.02                                                      & 301.69                                                     & 3.76                                                       & 0.42                                                    \\
                              & FLOPs(G)       & 4.91                                                              & 4.91                                                                & 15.24                                                    & 10.47                                                  & 14.78                                                        & 6.45                                                      & 588.47                                                        & 4.66                                                       & 0.02                                                      & 288.72                                                     & 33.52                                                      & 0.36                                                    \\
                              & T.T. (s/iter)  & 0.02                                                              & 0.02                                                                & 0.19                                                     & 0.18                                                   & 0.10                                                         & 0.04                                                      & OOM                                                           & 0.17                                                       & 0.001                                                     & 0.50                                                       & 0.17                                                       & 0.01                                                    \\
                              & T.M.(GB)       & 1.01                                                              & 0.95                                                                & 3.74                                                     & 9.50                                                   & 5.76                                                         & 2.95                                                      & OOM                                                           & 11.33                                                      & 0.07                                                      & 5.86                                                       & 8.01                                                       & 0.78                                                    \\
                              & I.T.(ms/iter)  & 5.71                                                              & 5.67                                                                & 45.68                                                    & 61.21                                                  & 31.60                                                        & 13.45                                                     & 1610.66                                                       & 67.21                                                      & 0.20                                                      & 15.64                                                      & 55.10                                                      & 3.48                                                    \\
                              & I.M. (GB)      & 0.43                                                              & 0.43                                                                & 1.25                                                     & 1.69                                                   & 1.84                                                         & 1.27                                                      & 9.35                                                          & 2.20                                                       & 0.04                                                      & 1.42                                                       & 2.44                                                       & 0.59                                                    \\ \midrule
\multirow{6}{*}{\rotatebox[origin=c]{90}{Weather}}      & Params(M)      & 4.52                                                              & 4.52                                                                & 8.56                                                     & 0.03                                                   & 0.50                                                         & 4.83                                                      & 2.37                                                          & 0.10                                                       & 0.02                                                      & 299.97                                                     & 3.76                                                       & 0.42                                                    \\
                              & FLOPs(G)       & 0.10                                                              & 0.10                                                                & 0.34                                                     & 0.004                                                  & 0.01                                                         & 0.10                                                      & 8.36                                                          & 0.05                                                       & 0.000                                                     & 291.21                                                     & 0.82                                                       & 0.01                                                    \\
                              & T.T. (ms/iter) & 7.47                                                              & 6.90                                                                & 27.55                                                    & 28.69                                                  & 37.85                                                        & 9.45                                                      & 102.04                                                        & 33.30                                                      & 1.38                                                      & 505.88                                                     & 9.89                                                       & 3.53                                                    \\
                              & T.M.(GB)       & 0.21                                                              & 0.21                                                                & 0.21                                                     & 0.05                                                   & 0.05                                                         & 0.22                                                      & 0.86                                                          & 0.16                                                       & 0.02                                                      & 5.80                                                       & 0.27                                                       & 0.04                                                    \\
                              & I.T.(ms/iter)  & 1.18                                                              & 1.05                                                                & 3.15                                                     & 4.11                                                   & 7.19                                                         & 1.66                                                      & 30.18                                                         & 5.10                                                       & 0.20                                                      & 156.27                                                     & 2.96                                                       & 0.48                                                    \\
                              & I.M. (MB)      & 39.26                                                             & 39.26                                                               & 70.06                                                    & 14.45                                                  & 18.28                                                        & 34.88                                                     & 316.49                                                        & 36.03                                                      & 9.06                                                      & 1397.18                                                    & 90.78                                                      & 24.37                                                   \\ \midrule
\multirow{6}{*}{\rotatebox[origin=c]{90}{Solar-Energy}} & Params(M)      & 4.58                                                              & 4.54                                                                & 8.56                                                     & 1.39                                                   & 4.61                                                         & 4.83                                                      & 2.37                                                          & 0.12                                                       & 0.02                                                      & 300.21                                                     & 3.76                                                       & 0.42                                                    \\
                              & FLOPs(G)       & 0.68                                                              & 0.68                                                                & 2.22                                                     & 2.05                                                   & 1.20                                                         & 0.72                                                      & 46.81                                                         & 0.74                                                       & 0.003                                                     & 290.64                                                     & 5.33                                                       & 0.06                                                    \\
                              & T.T. (ms/iter) & 9.12                                                              & 7.46                                                                & 48.64                                                    & 36.28                                                  & 31.77                                                        & 9.97                                                      & 515.44                                                        & 35.90                                                      & 1.44                                                      & 506.58                                                     & 28.89                                                      & 3.60                                                    \\
                              & T.M.(GB)       & 0.30                                                              & 0.30                                                                & 0.68                                                     & 2.22                                                   & 0.61                                                         & 0.37                                                      & 5.11                                                          & 1.83                                                       & 0.03                                                      & 5.81                                                       & 1.34                                                       & 0.15                                                    \\
                              & I.T.(ms/iter)  & 1.20                                                              & 1.08                                                                & 7.65                                                     & 10.44                                                  & 5.16                                                         & 1.96                                                      & 189.20                                                        & 8.54                                                       & 0.20                                                      & 156.66                                                     & 8.49                                                       & 0.48                                                    \\
                              & I.M. (GB)      & 0.21                                                              & 0.21                                                                & 0.23                                                     & 0.28                                                   & 0.16                                                         & 0.18                                                      & 1.94                                                          & 0.36                                                       & 0.01                                                      & 1.40                                                       & 0.42                                                       & 0.11                                                    \\ \midrule
\multirow{6}{*}{\rotatebox[origin=c]{90}{PEMS07}}       & Params(M)      & 4.75                                                              & 3.19                                                                & 8.56                                                     & 1.39                                                   & 13.44                                                        & 4.83                                                      & 2.38                                                          & 0.12                                                       & 0.02                                                      & 301.73                                                     & 3.76                                                       & 0.42                                                    \\
                              & FLOPs(G)       & 3.67                                                              & 3.67                                                                & 15.65                                                    & 16.10                                                  & 20.53                                                        & 6.66                                                      & 301.92                                                        & 4.77                                                       & 0.02                                                      & 290.97                                                     & 34.34                                                      & 0.37                                                    \\
                              & T.T. (s/iter)  & 0.01                                                              & 0.01                                                                & 0.19                                                     & 0.28                                                   & 0.11                                                         & 0.04                                                      & OOM                                                           & 0.17                                                       & 0.002                                                     & 0.51                                                       & 0.19                                                       & 0.01                                                    \\
                              & T.M.(GB)       & 0.83                                                              & 0.76                                                                & 3.85                                                     & 13.99                                                  & 6.42                                                         & 3.07                                                      & OOM                                                           & 11.62                                                      & 0.07                                                      & 5.86                                                       & 9.18                                                       & 0.79                                                    \\
                              & I.T.(ms/iter)  & 1.21                                                              & 1.10                                                                & 47.06                                                    & 94.01                                                  & 36.54                                                        & 13.92                                                     & 1370.91                                                       & 68.96                                                      & 0.20                                                      & 156.78                                                     & 56.49                                                       & 0.50                                                    \\
                              & I.M. (GB)      & 0.43                                                              & 0.43                                                                & 1.28                                                     & 1.74                                                   & 2.03                                                         & 1.32                                                      & 12.37                                                         & 2.25                                                       & 0.04                                                      & 1.43                                                       & 2.50                                                       & 0.61                                                    \\ \bottomrule
\end{tabular}
}
\end{table}

\begin{table}[ht]
\caption{Efficiency comparison of Leddam, iTransformer, PatchTST, and Fredformer before and after replacing multi-head self-attention with the NormLin module. In Leddam, only the \emph{cross-channel attention} module is updated. Metrics include number of learnable parameters (Params), FLOPs, training time (T.T.), training memory (T.M.), inference time (I.T.), and inference memory (I.M.). All experiments use a lookback/prediction length of 96 and a batch size of 16. Notably, in its official implementation, Fredformer employs Nystromformer \citep{nystromformer}—an approximate yet more efficient self-attention mechanism—for the ETTm1, ECL, and Weather datasets.}
\label{tab:GPU_normlin_plugin_compare}
\centering
\setlength{\tabcolsep}{5.5pt}
\renewcommand{\arraystretch}{1.1} 
{\fontsize{8}{9}\selectfont
\begin{tabular}{@{}cccccccccc@{}}
\toprule
\multicolumn{2}{c}{\multirow{2}{*}{Model}}     & \multicolumn{2}{c}{\begin{tabular}[c]{@{}c@{}}Leddam\\  \citeyear{Leddam_icml} \end{tabular}} & \multicolumn{2}{c}{\begin{tabular}[c]{@{}c@{}}iTrans.\\     \citeyear{itransformer} \end{tabular}} & \multicolumn{2}{c}{\begin{tabular}[c]{@{}c@{}}PatchTST\\      \citeyear{patchtst} \end{tabular}} & \multicolumn{2}{c}{\begin{tabular}[c]{@{}c@{}}Fredformer\\    \citeyear{fredformer} \end{tabular}} \\ \cmidrule(l){3-10} 
\multicolumn{2}{c}{}                           & Attn.                                & NormLin                               & Attn.                                 & NormLin                               & Attn.                                 & NormLin                                & Attn.                                  & NormLin                                 \\ \midrule
\multirow{6}{*}{ETTm1}        & Params(M)      & 8.55                                 & 7.51                                  & 4.83                                  & 2.21                                  & 3.76                                  & 2.71                                   & 8.59                                   & 8.39                                    \\
                              & FLOPs(M)       & 111.54                               & 104.15                                & 33.99                                 & 15.51                                 & 272.20                                & 182.20                                 & 135.01                                 & 79.10                                   \\
                              & T.T. (ms/iter) & 25.02                                & 23.67                                 & 10.11                                 & 7.10                                  & 8.37                                  & 6.82                                   & 26.48                                  & 10.06                                   \\
                              & T.M.(GB)       & 0.18                                 & 0.16                                  & 0.21                                  & 0.17                                  & 0.14                                  & 0.12                                   & 0.24                                   & 0.18                                    \\
                              & I.T.(ms/iter)  & 2.62                                 & 2.50                                  & 1.72                                  & 1.25                                  & 1.38                                  & 1.14                                   & 3.73                                   & 1.36                                    \\
                              & I.M. (MB)      & 51.04                                & 47.03                                 & 156.00                                & 146.20                                & 51.83                                 & 50.45                                  & 78.04                                  & 52.44                                   \\ \midrule
\multirow{6}{*}{ECL}          & Params(M)      & 8.56                                 & 7.71                                  & 4.83                                  & 2.41                                  & 3.76                                  & 2.71                                   & 12.12                                  & 12.92                                   \\
                              & FLOPs(G)       & 5.32                                 & 4.88                                  & 1.87                                  & 0.81                                  & 12.48                                 & 8.35                                   & 5.55                                   & 6.04                                    \\
                              & T.T. (ms/iter) & 65.19                                & 64.93                                 & 10.66                                 & 7.39                                  & 65.83                                 & 50.82                                  & 26.59                                  & 21.03                                   \\
                              & T.M.(GB)       & 1.44                                 & 1.41                                  & 0.70                                  & 0.41                                  & 3.02                                  & 2.78                                   & 1.24                                   & 1.27                                    \\
                              & I.T.(ms/iter)  & 14.78                                & 14.52                                 & 3.46                                  & 1.01                                  & 20.52                                 & 14.08                                  & 7.08                                   & 6.71                                    \\
                              & I.M. (GB)      & 0.49                                 & 0.48                                  & 0.23                                  & 0.10                                  & 0.93                                  & 0.99                                   & 0.26                                   & 0.26                                    \\ \midrule
\multirow{6}{*}{Weather}      & Params(M)      & 8.56                                 & 7.52                                  & 4.83                                  & 2.21                                  & 3.76                                  & 2.71                                   & 0.50                                   & 0.48                                    \\
                              & FLOPs(G)       & 0.34                                 & 0.31                                  & 0.10                                  & 0.05                                  & 0.82                                  & 0.55                                   & 0.01                                   & 0.003                                   \\
                              & T.T. (ms/iter) & 27.55                                & 27.21                                 & 9.45                                  & 7.49                                  & 9.89                                  & 7.47                                   & 37.85                                  & 28.19                                   \\
                              & T.M.(GB)       & 0.21                                 & 0.19                                  & 0.22                                  & 0.18                                  & 0.27                                  & 0.24                                   & 0.05                                   & 0.03                                    \\
                              & I.T.(ms/iter)  & 3.15                                 & 3.08                                  & 1.66                                  & 0.99                                  & 2.96                                  & 1.12                                   & 7.19                                   & 3.55                                    \\
                              & I.M. (MB)      & 70.06                                & 66.05                                 & 34.88                                 & 25.51                                 & 90.78                                 & 92.69                                  & 18.28                                  & 13.77                                   \\ \midrule
\multirow{6}{*}{Solar-Energy} & Params(M)      & 8.56                                 & 7.54                                  & 4.83                                  & 2.24                                  & 3.76                                  & 2.71                                   & 4.61                                   & 4.32                                    \\
                              & FLOPs(G)       & 2.22                                 & 2.06                                  & 0.72                                  & 0.32                                  & 5.33                                  & 3.56                                   & 1.20                                   & 0.91                                    \\
                              & T.T. (ms/iter) & 48.64                                & 48.12                                 & 9.97                                  & 7.40                                  & 28.89                                 & 22.31                                  & 31.77                                  & 12.32                                   \\
                              & T.M.(GB)       & 0.68                                 & 0.67                                  & 0.37                                  & 0.26                                  & 1.34                                  & 1.23                                   & 0.61                                   & 0.46                                    \\
                              & I.T.(ms/iter)  & 7.65                                 & 7.59                                  & 1.96                                  & 1.01                                  & 8.49                                  & 5.83                                   & 5.16                                   & 2.89                                    \\
                              & I.M. (GB)      & 0.23                                 & 0.22                                  & 0.18                                  & 0.18                                  & 0.42                                  & 0.44                                   & 0.16                                   & 0.09                                    \\ \midrule
\multirow{6}{*}{PEMS07}       & Params(M)      & 8.56                                 & 9.07                                  & 4.83                                  & 3.77                                  & 3.76                                  & 2.71                                   & 13.44                                  & 15.29                                   \\
                              & FLOPs(G)       & 15.65                                & 13.93                                 & 6.66                                  & 2.75                                  & 34.34                                 & 22.96                                  & 20.53                                  & 17.57                                   \\
                              & T.T. (s/iter)  & 0.19                                 & 0.18                                  & 0.04                                  & 0.01                                  & 0.19                                  & 0.14                                   & 0.11                                   & 0.06                                    \\
                              & T.M.(GB)       & 3.85                                 & 3.74                                  & 3.07                                  & 0.82                                  & 9.18                                  & 7.57                                   & 6.42                                   & 3.16                                    \\
                              & I.T.(ms/iter)  & 47.06                                & 45.16                                 & 13.92                                 & 3.02                                  & 56.49                                 & 38.63                                  & 36.54                                  & 19.16                                   \\
                              & I.M. (GB)      & 1.28                                 & 1.28                                  & 1.32                                  & 0.27                                  & 2.50                                  & 2.70                                   & 2.03                                   & 0.60                                    \\ \bottomrule
\end{tabular}
}
\end{table}

\section{Limitations} \label{limitations}

While this work includes extensive experiments, it primarily focuses on time series forecasting, and our claims are confined to this setting. Future work includes larger-scale pre-training and a wider range of time series analysis tasks (e.g., imputation and anomaly detection), in order to further validate the robustness of OLinear and its core components, OrthoTrans and NormLin.
In addition, although NormLin is linear-based, it still incurs $\mathcal{O}(N^2)$ computational and memory complexity with respect to the number of variates $N$. Developing simpler yet effective alternatives for modeling multivariate correlations—and more generally, for learning general-purpose token dependencies—remains a promising direction.

\section{Broader impacts} \label{broader_impacts}

Multivariate time series forecasting is a fundamental field with a wide range of real-world applications across domains such as energy, finance, healthcare, and transportation. While this work does not target a specific application domain, OLinear could have general real-world applicability (e.g., weather forecasting and traffic planning).
We contribute two key insights: (1) the OrthoTrans scheme, which reformulates temporally correlated forecasting as inter-independent feature prediction; and (2) the linear-based NormLin module, which effectively models multivariate correlations, enhances representation learning among temporal and frequency tokens, and remains compatible with decoder architectures and large-scale pre-training.
As a step toward simpler and more efficient token dependency learners, the NormLin module shows potential for broader use in deep learning and may inspire future research in this direction.


\end{document}